\documentclass{article}

\newif\ifarxiv
\newif\ifcameraready
\camerareadytrue
\ifdefined\arxivmode\arxivtrue\camerareadyfalse\fi
\newif\ifpublicbuild
\ifarxiv\publicbuildtrue\fi
\ifcameraready\publicbuildtrue\fi

\usepackage{graphicx}
\usepackage{subcaption}
\usepackage{booktabs}
\usepackage[pagebackref,breaklinks]{hyperref}
\PassOptionsToPackage{numbers,compress}{natbib}
\ifarxiv
  \usepackage[preprint]{neurips_2026}
\else\ifcameraready
  \usepackage[main,final]{neurips_2026}
\else
  \usepackage{neurips_2026}
\fi\fi
\makeatletter
\g@addto@macro\normalsize{%
  \setlength\abovedisplayskip{4pt}%
  \setlength\belowdisplayskip{4pt}%
  \setlength\abovedisplayshortskip{2pt}%
  \setlength\belowdisplayshortskip{2pt}%
}
\makeatother
\usepackage{titlesec}
\titlespacing*{\section}{0pt}{4pt}{2pt}
\titlespacing*{\subsection}{0pt}{3pt}{1pt}
\titlespacing*{\subsubsection}{0pt}{2pt}{0pt}
\usepackage{amsmath}
\usepackage{amssymb}
\usepackage{mathtools}
\usepackage{amsthm}
\usepackage[dvipsnames]{xcolor}
\usepackage[capitalize,noabbrev]{cleveref}
\usepackage{nicefrac}
\usepackage{siunitx}
\usepackage{tikz}
\usetikzlibrary{positioning,calc}
\usepackage{enumitem}
\usepackage{mwe} %
\usepackage{multirow}
\usepackage{wrapfig}
\usepackage{float}
\usepackage{placeins} %
\usepackage{afterpage} %
\usepackage{needspace} %
\usepackage{algorithm}
\usepackage{algpseudocode}

\def\eqref#1{equation~\ref{#1}}

\def\1{\bm{1}}

\DeclareMathAlphabet{\mathsfit}{\encodingdefault}{\sfdefault}{m}{sl}
\SetMathAlphabet{\mathsfit}{bold}{\encodingdefault}{\sfdefault}{bx}{n}

\newcommand{\E}{\mathbb{E}}

\newcommand{\KL}{D_{\mathrm{KL}}}

\makeatletter
\usepackage{xspace} 
\def\@onedot{\ifx\@let@token.\else.\null\fi\xspace}
\DeclareRobustCommand\onedot{\futurelet\@let@token\@onedot}

\usepackage{amsmath}
\usepackage{amsthm}

\newcommand{\fitlinewidth}[1]{\resizebox{\ifdim\width>\linewidth\linewidth\else\width\fi}{!}{#1}}

\definecolor{orange}{HTML}{ff7f0e}
\definecolor{blue}{HTML}{1f77b4}
\definecolor{burntorange}{rgb}{0.8, 0.33, 0.0}
\definecolor{NVgreen}{HTML}{76B900}

\newcommand{\colorE}[1]{\textcolor{RoyalPurple}{#1}}

\newcommand{\Fig}{Fig.}

\newcommand{\Sec}{Sec.}
\newcommand{\App}{App.}
\newcommand{\Eq}{Eq.}
\newcommand{\Tab}{Table}
\newcommand{\Tabs}{Tables}
\newcommand{\Algo}{Algorithm}

\providecommand{\E}{\mathbb{E}}
\DeclareMathOperator*{\Eop}{\E}
\providecommand{\KL}{\mathrm{KL}}
\providecommand{\Normal}{\mathcal{N}}
\providecommand{\Uniform}{\mathcal{U}}
\newcommand{\identity}{\mathbf{I}}
\newcommand{\uniformOnUnit}{\Uniform(0,1)}
\newcommand{\standardNormal}{\Normal(\mathbf{0},\identity)}
\newcommand{\diffd}{\mathrm{d}}                     %
\newcommand{\jacobian}{\mathbf{J}}
\newcommand{\timevar}{t}                           %
\newcommand{\maxTimevar}{T}                           %
\newcommand{\logsnr}{\lambda}                      %
\newcommand{\logsnrSchedule}{f_{\logsnr}}          %
\newcommand{\logsnrMin}{\logsnr_{\min}}            %
\newcommand{\logsnrMax}{\logsnr_{\max}}            %
\newcommand{\signalcoef}{\alpha_{\timevar}}
\newcommand{\noisecoeff}{\sigma_{\timevar}}
\newcommand{\dataSample}{\mathbf{x}}               %
\newcommand{\dataLabel}{\mathbf{y}}
\newcommand{\noiseVec}{\boldsymbol{\epsilon}}      %
\newcommand{\noisePred}{\hat{\noiseVec}_{\denParams}}           %
\newcommand{\genParams}{{\boldsymbol{\theta}}}       %
\newcommand{\denParams}{{\boldsymbol{\phi}}}         %
\newcommand{\generator}{G_{\genParams}}            %
\newcommand{\render}{g}                            %
\newcommand{\prerender}{\render'}
\newcommand{\textCond}{\mathbf{c}}                 %
\newcommand{\cfgScale}{\omega}                     %
\newcommand{\weight}{w}                            %
\newcommand{\importanceWeight}{\tilde{\weight}}
\newcommand{\sdsWeight}{\weight_{\textnormal{SDS}}}
\newcommand{\quantile}{u}
\newcommand{\dataset}{\mathcal{D}}
\newcommand{\loss}{\mathcal{L}}
\newcommand{\cost}{\ell}
\newcommand{\lossDiffusion}{\loss_{\mathrm{Diff}}} %
\newcommand{\lossWeighted}{\loss_{\mathrm{wDiff}}}              %
 
\newcommand{\costDiffusion}{\cost_{\mathrm{Diff}}} 
\newcommand{\lossSDS}{\loss_{\mathrm{SDS}}}        %
\newcommand{\lossDenoise}{\loss_{\mathrm{denoise}}}%
\newcommand{\lossReg}{\loss_{\mathrm{reg}}}        %

\newcommand{\labelBase}{\mathrm{base}}
\newcommand{\labelFake}{\mathrm{fake}}
\newcommand{\labelReal}{\mathrm{real}}
\newcommand{\mean}{\boldsymbol{\mu}}
\newcommand{\estimatedMean}{\hat{\mean}}
\newcommand{\meanBase}{\mean_{\labelBase}}
\newcommand{\meanFake}{\mean_{\labelFake}^{\denParams}}
\newcommand{\score}{\mathbf{s}}
\newcommand{\scoreReal}{\score_{\labelReal}}
\newcommand{\scoreFake}{\score_{\labelFake}}
\newcommand{\realDensity}{p_{\labelReal}}
\newcommand{\fakeDensity}{p_{\labelFake}}
\newcommand{\proposalDensity}{q}
\newcommand{\testFunc}{\mathbf{F}}
\newcommand{\numSamples}{N}
\newcommand{\sampleIndex}{n}
\newcommand{\numChannels}{C}
\newcommand{\numSpatial}{S}
\newcommand{\numStrata}{B}                         %
\newcommand{\stratum}{\mathcal{S}}                 %
\newcommand{\stratumIndex}{b}                      %
\newcommand{\samplePerStratum}{M_{\stratumIndex}}  %
\newcommand{\sampleIndexInStratum}{m}              %
\newcommand{\encoder}{\textnormal{Encode}}                                  %
\newcommand{\numRenders}{R}                               %
\newcommand{\numReNoises}{K}                              %
\newcommand{\generalCost}{c}
\newcommand{\costRender}{\generalCost_{\mathrm{render+encode}}}      %
\newcommand{\costDenoise}{\generalCost_{\mathrm{denoise}}}           %
\newcommand{\budget}{\mathcal{B}}                         %
\newcommand{\gterm}{\mathbf{f}}                         %
\newcommand{\renderIndex}{r}                                   %
\newcommand{\renoiseIndex}{k}                                   %
\newcommand{\encodedData}{\mathbf{z}} %
\newcommand{\noisedData}{\encodedData_{\timevar}}
\newcommand{\cameraSample}{\boldsymbol{\pi}}

\newcommand{\update}{\mathbf{u}}
\newcommand{\sdsupdate}{\update_{\textnormal{SDS}}}
\newcommand{\sdsupdatehat}{\hat{\update}_{\textnormal{SDS}}}
\newcommand{\residual}{\mathbf{r}}
\newcommand{\sampleStep}{k}
\newcommand{\numSampleStep}{K}
\newcommand{\randomness}{\boldsymbol{\xi}}
\newcommand{\hessian}{\mathbf{H}}
\newcommand{\influence}{I}
\newcommand{\motiveGrad}{\mathbf{g}}
\newcommand{\query}{\mathrm{query}}
\newcommand{\ntrain}{n}
\newcommand{\sampleSet}{\mathcal{T}}
\newcommand{\defeq}{:=}

\newcommand{\ourMethod}{CARV}
\newcommand{\ourTitle}{Variance Reduction for Gradients from Diffusion Teachers}

\makeatletter
\ifarxiv
  \renewcommand{\@maketitle}{%
    \vbox{%
      \hsize\textwidth
      \linewidth\hsize
      \vskip 0.1in
      \@toptitlebar
      \centering
      {\fontsize{15pt}{18pt}\selectfont\bfseries \@title\par}
      \@bottomtitlebar
      \begin{tabular}[t]{c}\rule{\z@}{16\p@}\bfseries\@author\end{tabular}%
      \vskip 0.3in \@minus 0.1in
    }
  }
\else
  \renewcommand{\@toptitlebar}{\hrule height 4\p@\vskip 0.25in\vskip -\parskip}
  \renewcommand{\@bottomtitlebar}{\vskip 0.25in\vskip -\parskip\hrule height 1\p@\vskip 0.07in}
  \renewcommand{\@maketitle}{%
    \vbox{%
      \hsize\textwidth
      \linewidth\hsize
      \vskip 0.05in
      \@toptitlebar
      \centering
      {\fontsize{14pt}{16pt}\selectfont\bfseries \@title\par}
      \@bottomtitlebar
      \if@anonymous
        \begin{tabular}[t]{c}\bfseries\rule{\z@}{12\p@}
          Anonymous Author(s) \\
          Affiliation \\
          Address \\
          \texttt{email} \\
        \end{tabular}%
      \else
        \def\And{\end{tabular}\hfil\linebreak[0]\hfil\begin{tabular}[t]{c}\bfseries\rule{\z@}{12\p@}\ignorespaces}%
        \def\AND{\end{tabular}\hfil\linebreak[4]\hfil\begin{tabular}[t]{c}\bfseries\rule{\z@}{12\p@}\ignorespaces}%
        \begin{tabular}[t]{c}\bfseries\rule{\z@}{12\p@}\@author\end{tabular}%
      \fi
      \vskip 0.3in \@minus 0.1in
    }
  }
\fi
\makeatother

\title{\ourTitle}

\author{%
  Jesse Bettencourt\textsuperscript{1\,2\,4}\quad
  Xindi Wu\textsuperscript{1\,3}\quad
  Matan Atzmon\textsuperscript{1}\quad
  James Lucas\textsuperscript{1}\quad
  Jonathan Lorraine\textsuperscript{1} \\[4pt]
  \textsuperscript{1}NVIDIA\quad
  \textsuperscript{2}University of Toronto\quad
  \textsuperscript{3}Princeton University\quad
  \textsuperscript{4}Vector Institute \\[4pt]
  Project page: \href{https://research.nvidia.com/labs/sil/projects/CARV/}{\color{NVgreen}\nolinkurl{research.nvidia.com/labs/sil/projects/CARV/}}
}

\ifpublicbuild
  \hypersetup{
    colorlinks=true,
    linkcolor=NavyBlue,
    citecolor=NavyBlue,
    urlcolor=NavyBlue,
    pdftitle={\ourTitle},
    pdfauthor={Jesse Bettencourt, Xindi Wu, Matan Atzmon, James Lucas, Jonathan Lorraine},
    pdfsubject={Variance reduction for Monte Carlo gradient estimators with diffusion teachers},
    pdfkeywords={diffusion models, variance reduction, Monte Carlo, score distillation sampling, importance sampling, stratified sampling, compute-aware estimator, hierarchical Monte Carlo}
  }
\fi

\begin{document}

\maketitle
\vspace{-1.5em}

\begin{abstract}
    Pretrained diffusion models serve as frozen teachers for pipelines such as text-to-3D, single-step distillation, and data attribution. The teacher gradients these pipelines consume are expectations over noise levels and Gaussian noise, estimated by Monte Carlo (MC). Each MC draw requires expensive upstream work (rendering, simulation, encoding), so estimator variance dominates compute cost. We introduce \ourMethod{}, a compute-aware variance-accounting framework that casts these gradients as a classical two-stage sampling problem with asymmetric per-level costs. This motivates a hierarchical MC estimator that amortizes each expensive upstream computation by reusing it across cheap diffusion draws, with within-unit variance further reduced by timestep importance sampling (IS) and stratified inverse-CDF sampling. \ourMethod{} delivers $2$ to $3\times$ effective compute multipliers on text-to-3D and data attribution without changing the objective. In text-to-3D, reuse provides most of the gain and IS with stratification adds about $25\%$. In single-step distillation, the same techniques cut gradient variance by an order of magnitude but do not improve downstream FID, marking the regime where MC variance is no longer the~bottleneck.
\end{abstract}

\vspace{-0.02\textheight}
\section{Introduction}
    Diffusion models generate state-of-the-art images, video, audio, and 3D/4D by repeatedly evaluating a denoiser on a noisy input and stepping along the predicted denoising direction. That direction can instead be used as a gradient by a downstream pipeline, with the model frozen as a teacher: in text-guided 3D the gradient from an image model steers a differentiable renderer, in one-step distillation it trains a student generator, and in data attribution it scores training examples. The signal in each case is the teacher gradient, this direction averaged over noise levels and Gaussian noise, and it must be estimated by Monte Carlo. Each draw requires expensive upstream work (rendering, simulation, encoding), so estimator variance dominates the compute cost.

    Most variance-reduction work targets teacher training via loss reweighting and schedule design. Downstream, practitioners inherit the teacher's timestep distribution, average draws ad hoc, sometimes with bias, and tune compute without knowing which source of randomness dominates variance or what each source costs. This leaves three open questions: which estimator components dominate variance, how to reduce variance without biasing the objective, and how to trade cheap operations (noising, denoising) against expensive ones (rendering, encoding) under a fixed~budget.

    We propose \ourMethod{}, which treats frozen-teacher gradient estimation as an instance of classical two-level sampling with asymmetric per-level costs: expensive upstream computations are the primary units and cheap diffusion draws are the subunits. Classical theory then supplies the variance decomposition, the cost-optimal allocation between levels, importance sampling, and stratification. \ourMethod{} makes them operational with an explicit cost model and parameter-gradient variance measurement. The resulting drop-ins are unbiased under the stated sampling construction and match baseline variance at a third to half the cost ($2$ to $3\times$ effective compute multipliers) in text-to-3D optimization and attribution, with no downstream FID gain in one-step distillation.
    Our contributions include:\vspace{-0.005\textheight}
    \setlist{nosep,leftmargin=*,topsep=2pt,itemsep=1pt,parsep=0pt,partopsep=0pt}
    \begin{enumerate}
        \item \textbf{Frozen-teacher gradients as two-stage sampling.} We identify the frozen-teacher gradient estimator as the classical two-stage sample mean with asymmetric per-level costs; classical theory then gives the variance decomposition, the allocation $\numReNoises^\star$, and the within-level variance reductions (importance sampling and stratification), which we operationalize with an explicit cost model and parameter-gradient variance measurement (\Sec~\ref{sec:method}).
        \item \textbf{Timestep importance sampling using the explicit teacher weight.} A drop-in proposal $\proposalDensity \!\propto\! p\,\sdsWeight$ giving ${\sim}1.2\times$ variance reduction over uniform at equal per-iteration cost (\Tab~\ref{tab:relative_improvement_by_m_param}; toy in \Fig~\ref{fig:toyIWVisualization}).
        \item \textbf{Stratified-inverse-CDF sampling over noise levels.} Adapts the standard equal-mass quantile construction to frozen-teacher timestep allocation, composing stratification with importance sampling per render (\App~\Fig~\ref{fig:stratifiedVisualization}, \Fig~\ref{fig:inverseTransformVisualization}); close to a Sinkhorn-optimized two-sample allocation (\App~\Sec~\ref{app:optimal_pair_distributions}).
        \item \textbf{Compute-aware variance-accounting framework (\ourMethod{}) with broad evaluation.} A measurement framework for diffusion gradient variance (\Sec~\ref{sec:method-variance-framework}, \Fig~\ref{fig:quantifying_variance_hierarchical_cost_aware_iw_strat_main}) applied to diffusion-teacher-guided optimization (\Sec~\ref{sec:experiments-sds}), one-step distillation (\Sec~\ref{sec:experiments-dmd}), and data attribution (\Sec~\ref{sec:experiments-motive}).
    \end{enumerate}

\section{Related Work}\label{sec:related}
    \textbf{Classical Monte Carlo variance reduction.}
        Importance sampling, stratification, control variates, and antithetic variates are standard \citep{rubinstein2016simulation, owen2013monte}, including for gradient estimation in machine learning \citep{mohamed2020monte}. Two-stage sampling with unequal per-stage costs, including the cost-optimal number of subunits per primary unit, is classical sampling theory \citep{hansen1943theory, cochran1977sampling, thompson2012sampling, lohr2021sampling}, as is stratified inverse-transform sampling \citep{glasserman2004monte}. Graphics amortizes expensive scene queries over reused samples in the same~spirit~\citep{bitterli2020spatiotemporal, lin2022generalized}.

    \textbf{Teacher-side schedule and weighting design.}
        Most diffusion variance reduction targets training the teacher: noise schedules \citep{ho2020denoising, nichol2021improved, hoogeboom2023simple}, learned signal-to-noise schedules that minimize the variance of the training objective \citep{kingma2023variational}, the noise schedule read as an importance-sampling density over noise levels \citep{kingma2023understanding}, and loss reweighting \citep{salimans2022progressive, karras2022elucidating, choi2022perception, hang2023minsnr}. These optimize gradients with respect to denoiser parameters, under a cost structure with no expensive upstream stage.

    \textbf{Downstream frozen-teacher methods.}
        SDS-style optimization \citep{poole2022dreamfusion, lin2023magic3d, wang2023prolificdreamer}, few-step distillation \citep{yin2024one, yin2024improved, salimans2024multistep}, and gradient-based attribution \citep{park2023trak, xie2024data, wu2026motion} consume teacher gradients as Monte Carlo estimates but treat the estimator monolithically: they inherit the teacher's timestep distribution or reshape it without a likelihood-ratio correction \citep{huang2024dreamtime}; SDS-style and distillation methods draw one diffusion sample per expensive upstream computation (RewardSDS \citep{chachy2025rewardsds} reweights several noise draws per render by a reward, for alignment rather than variance). VSD \citep{wang2023prolificdreamer} and SteinDreamer \citep{wang2025steindreamer} lower SDS variance by changing the objective or adding a control variate; our estimator changes compose with both (\App~\Sec~\ref{app:comparison-vsd}, \App~\Sec~\ref{app:comparison-stein}).

    Our central claim is the identification that frozen-teacher gradient estimation is an instance of the classical two-level sampling problem with asymmetric per-level costs, with renders as primary sampling units and diffusion draws as subunits. Extended related work is in \App~\Sec~\ref{app:related}.

\section{Background}\label{sec:background}
    Further background is in \App~\Sec~\ref{app:sec_background}.

    \subsection{Diffusion Models}\label{sec:background-diffusion-models}
        We use conditional latent diffusion via an autoencoder \cite{ho2020denoising, rombach2022high}. Let $(\dataSample,\textCond)$ be a data point and its conditioning (e.g., image+text), and $\encodedData \!=\! \encoder(\dataSample)$ its latent code. Forward noising at level $\timevar\in[0,1]$, from clean data at $\timevar\!=\!0$ to pure noise at $\timevar\!=\!1$, makes $\noisedData=\signalcoef \encodedData+\noisecoeff \noiseVec$ with $\noiseVec\sim\standardNormal$, where $\signalcoef,\noisecoeff$ are schedule-dependent coefficients (clamped away from $\timevar\!=\!0$). We use continuous notation; many systems use a discrete grid of $\maxTimevar$ timesteps (mapping in \App~\Sec~\ref{app:algorithm}), and some flow-matching codebases run $\timevar$ in the opposite direction.

        The teacher $\noisePred$, with parameters $\denParams$, is trained on $\E_{(\dataSample,\textCond),\timevar,\noiseVec}[\weight(\timevar)\costDiffusion]$, where $\weight(\timevar)$ is the timestep weight, $\costDiffusion\defeq\|\residual\|_2^2$, and $\residual\defeq\noisePred(\noisedData,\timevar,\textCond)-\noiseVec$ (\App~\Sec~\ref{app:sec_diffusion_models}).
        \App~\Sec~\ref{sec:background-diffusion-models-sampling-app} covers the essentials of sampling from diffusion models and classifier-free guidance. \Sec~\ref{sec:background-reducing-variance-importance_sampling} covers diffusion noise schedules and their connection to importance sampling.

    \subsection{Reducing Estimator Variance}\label{sec:background-reducing-variance}
        We consider Monte Carlo estimators for expectations that arise in diffusion training and downstream uses of frozen diffusion teachers. These are typically vector-valued (e.g., a parameter gradient or an update direction), so for an unbiased estimator $\hat\mean$ of a vector mean $\mean$, mean-squared error and variance coincide, and we measure dispersion by:
        \begin{equation}\label{eq:var}
            \mathrm{Var}(\hat\mean)\defeq\E[\|\hat\mean-\mean\|_2^2]=\mathrm{tr}(\mathrm{Cov}(\hat\mean)).
        \end{equation}

        \begin{figure*}[t!]
            \centering
            \vspace{-0.005\textheight}
            \scalebox{1.04}{
            \begin{tikzpicture}
            \centering
                \node (img1){\includegraphics[trim={0.9cm 1.0cm 1.0cm 1.55cm}, clip, width=.5\linewidth]{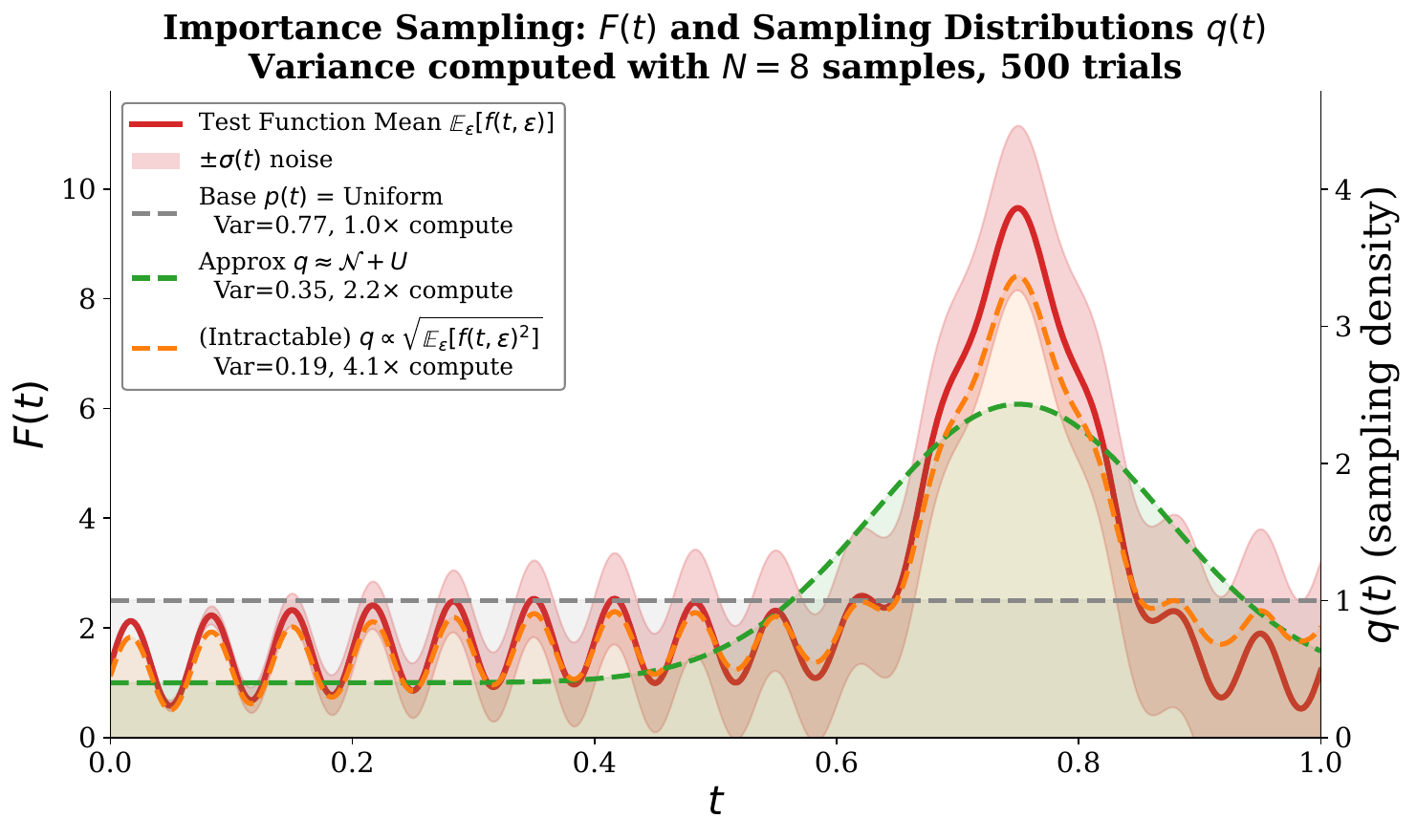}};
                \node[right=of img1, node distance=0cm, xshift=-.95cm, yshift=1.75cm, rotate=270, font=\color{black}]{\scriptsize{Proposal Density $\proposalDensity(\timevar)$}};
                \node[left=of img1, node distance=0cm, rotate=90, xshift=1.5cm, yshift=-.95cm,  font=\color{black}]{\scriptsize{Test Function $f(\timevar, \noiseVec)$}};
                \node[below=of img1, node distance=0cm, xshift=0.0cm, yshift=1.25cm,  font=\color{black}]{\scriptsize{Timestep $\timevar$}};
                
                \node [right=of img1, xshift=0.5cm] (img2){\includegraphics[trim={0.9cm 0.9cm 0cm 0.1cm}, clip, width=.27\linewidth]{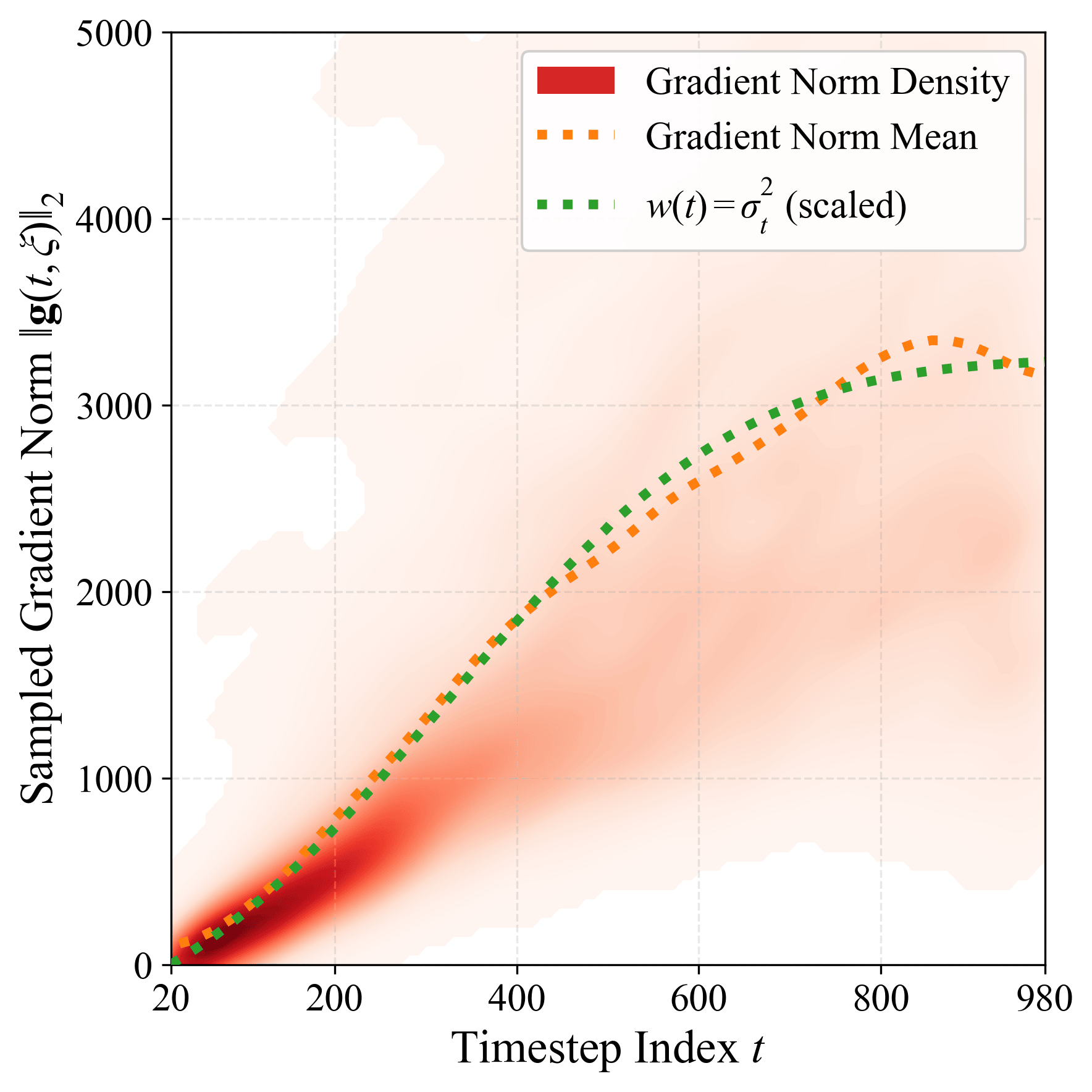}};
                \node[left=of img2, node distance=0cm, rotate=90, xshift=2.0cm, yshift=-.8cm,  font=\color{black}]{\scriptsize{Parameter Gradient Norm $\|\gterm(\timevar,\randomness)\|$}};
                \node[below=of img2, node distance=0cm, xshift=0.0cm, yshift=1.25cm,  font=\color{black}]{\scriptsize{Timestep $\timevar$}};
            \end{tikzpicture}
            }
            \vspace{-0.00\textheight}
            \caption{
                \textbf{Importance sampling for timestep allocation:}
                \emph{Left:} Toy example showing a test function $F$, uniform proposal, oracle optimal proposal from \Eq~\ref{eq:opt_proposal}, and a practical approximation adding a Gaussian at the peak. The oracle variance equals that of spending $\sim\!3$--$4\times$ the compute of uniform, while the approximation equals $\sim\!2\times$.
                \emph{Right:} Real parameter gradient norms from text-to-3D optimization. Our importance sampling weight function $\sdsWeight(\timevar)$ closely tracks the mean gradient norm, making it an effective sampling proxy: it reduces variance as much as ${\sim}20\%$ more compute would (\Tab~\ref{tab:relative_improvement_by_m_param}). Shading is the density of sampled norms over prompts, cameras, and noise. Additional analysis is in \App~\Fig~\ref{fig:sds_optimal_importance}. 
            }\label{fig:toyIWVisualization}
            \vspace{-0.00\textheight}
        \end{figure*}

        \subsubsection{Importance Sampling \& Noise Schedules}\label{sec:background-reducing-variance-importance_sampling}
            Importance sampling reduces Monte Carlo variance by sampling from important regions while preserving the expectation via a likelihood ratio. For noise level $\timevar\!\in\![0,1]$ with base density $p(\timevar)$ and randomness $\randomness$, we estimate the mean:\vspace{-0.005\textheight}
            \begin{equation}
                \smash{
                \mean=\E_{\timevar\sim p,\randomness\sim p(\cdot\mid\timevar)}[\gterm(\timevar,\randomness)]
                }
            \end{equation}
            where $\gterm(\timevar,\randomness)$ is a vector-valued contribution such as a gradient. Sampling $\timevar^{(\sampleIndex)}\sim\proposalDensity$ with importance weight $\importanceWeight(\timevar)=p(\timevar)/\proposalDensity(\timevar)$ gives the unbiased estimator:\vspace{-0.005\textheight}
            \begin{equation}
                \hat\mean_{\proposalDensity}=\nicefrac{1}{\numSamples}\sum\nolimits_{\sampleIndex=1}^{\numSamples}\importanceWeight(\timevar^{(\sampleIndex)})\gterm(\timevar^{(\sampleIndex)},\randomness^{(\sampleIndex)}).
            \end{equation}
            The variance-minimizing proposal is $\proposalDensity^\star(\timevar)\!\propto\!p(\timevar)\sqrt{\E[\|\gterm(\timevar,\randomness)\|_2^2\mid\timevar]}$ \cite{kahn1953methods, rubinstein2016simulation}. This oracle is impractical, so practitioners use cheap surrogates such as the per-timestep loss \cite{nichol2021improved}, fixed non-uniform samplers \cite{zheng2024non}, or learned schedules \cite{kingma2023variational}. The right surrogate depends on the form of $\gterm$, which is task-specific; we instantiate $\proposalDensity^\star$ and pick a proxy for the SDS gradient in \Sec~\ref{sec:method-noise-schedules}. As shown in \Fig~\ref{fig:toyIWVisualization} (left), the oracle proposal is variance-equivalent to $\sim\!3$--$4\times$ uniform compute and a Gaussian-at-peak approximation to $\sim\!2\times$; full derivation in \App~\Sec~\ref{sec:background-reducing-estimator-variance-app}.
            
            \textbf{Noise Schedules as Importance Sampling:}
            A noise schedule defines how signal and noise coefficients $\signalcoef,\noisecoeff$ vary with $\timevar\in[0,1]$. With uniform sampling $\timevar\sim\uniformOnUnit$, a schedule induces a distribution over noise levels. Schedule design can thus be viewed as choosing an importance distribution for the diffusion training objective; \citet{kingma2023variational} parametrize this in terms of signal-to-noise ratio and learn a schedule that minimizes the variance of the training-objective estimator. In downstream tasks, the schedule is inherited from the pretrained teacher, so we augment the timestep distribution via explicit importance sampling (\Sec~\ref{sec:method-noise-schedules}) rather than retraining. See \App~\Sec~\ref{sec:background-diffusion-models-noise-schedule-app} for details.

        \subsubsection{Stratified Sampling}\label{sec:background-reducing-variance-stratified}
            Stratified sampling (visualized in \App~\Fig~\ref{fig:stratifiedVisualization}) reduces variance by partitioning the domain into $\numStrata$ disjoint strata and averaging estimates within strata before combining with probabilities. Partition the domain into strata $\{\stratum_{\stratumIndex}\}_{\stratumIndex=1}^{\numStrata}$ with probabilities $p_{\stratumIndex}$ and $\samplePerStratum$ samples per stratum. The estimator is:\vspace{-0.005\textheight}
            \begin{equation}
                \!\!\estimatedMean_{\mathrm{strat}} \!\!=\!\! \sum\nolimits_{\stratumIndex=1}^{\numStrata} \!p_{\stratumIndex}\! \frac{1}{\samplePerStratum} \!\!\sum\nolimits_{\sampleIndexInStratum=1}^{\samplePerStratum}\!\! \testFunc(\timevar_{\stratumIndex,\sampleIndexInStratum}), \,\timevar_{\stratumIndex,\sampleIndexInStratum} \!\!\sim p(\timevar \!\mid\! \stratum_{\stratumIndex})\vspace{-0.003\textheight}
            \end{equation}
            This estimator is unbiased for $\E_{\timevar \sim p(\timevar)}[\testFunc(\timevar)]$ and, with proportional allocation (e.g., equal-probability strata with equal $\samplePerStratum$), has variance no greater than standard IID sampling from $p(\timevar)$, with strict reduction whenever the stratum means of $\testFunc$ differ \citep{thompson2012sampling}. As a simple example with a continuous domain $\timevar \in [0,1]$, one can use $\numStrata$ equal-width bins. With batch size $\numStrata$ and $\samplePerStratum=1$, each bin contributes one draw; \Sec~\ref{sec:method-stratified-sampling} applies this to diffusion timesteps.

    \subsection{Diffusion Model Applications}\label{sec:background-diffusion-applications}
        \vspace{-0.0\textheight}
        \subsubsection{Diffusion Priors for Optimization}\label{sec:background-sds}
            Score Distillation Sampling (SDS) uses a frozen pretrained diffusion model to supply gradients to a differentiable generator, renderer, or simulator, and underlies text-guided 3D/4D, material, and audio optimization \cite{poole2022dreamfusion, bahmani20244d, liu2024physics3d, deng2024flashtex, richter2025audiosds}. Given generator parameters $\genParams$ and rendering condition $\cameraSample$ (e.g., camera pose), we form an encoded observation:
            \begin{equation}
                \smash{
                \encodedData=\render(\genParams,\cameraSample)=\encoder(\prerender(\genParams,\cameraSample))
                }
            \end{equation}
            where $\prerender$ is the differentiable renderer (or generator or simulator) and $\encoder$ maps its output into the teacher's latent space, and we use the teacher to update $\genParams$. With $\noisedData=\signalcoef \encodedData+\noisecoeff \noiseVec$ and residual $\residual=\noisePred(\noisedData,\timevar,\textCond;\cfgScale)-\noiseVec$, SDS uses proxy score-direction $\weight(\timevar)\residual\,\diffd\noisedData/\diffd\encodedData$ and applies the chain rule:\vspace{-0.005\textheight}
            \begin{equation}\label{eq:sds_update}
                \smash{\sdsupdate(\genParams) \!=\! \Eop_{\cameraSample,\timevar,\noiseVec}\!\big[\sdsWeight(\timevar)\,\residual\,\nicefrac{\diffd \encodedData}{\diffd \genParams}\big],\ \text{where } \sdsWeight(\timevar) \!=\! \weight(\timevar)\signalcoef \text{ (using } \nicefrac{\diffd \noisedData}{\diffd \encodedData}\!=\!\signalcoef\identity\text{)}}
            \end{equation}
            so $\sdsWeight(\timevar)$ absorbs schedule factors, simplifying backprop and enabling importance sampling. The teacher is frozen; the stop-gradient surrogate is in \App~\Sec~\ref{sec:background-sds-app}.

        \subsubsection{Single-Step Diffusion Distillation}\label{sec:background-single-step-distillation}
            Distribution Matching Distillation (DMD) \cite{yin2024one} trains a one-step generator by averaging $\weight(\timevar)\signalcoef(\scoreFake-\scoreReal)\,\partial\encodedData/\partial\genParams$ over standard Gaussian generator inputs, $\timevar\sim p$, and independent standard Gaussian forward noise. Here $\encodedData$ is the generator output and $\scoreReal,\scoreFake$ are scores of the noised data and generator distributions at $\noisedData$, from the frozen teacher and a fake-score model trained on generator outputs (\Eq~\ref{eq:dmd_mc_estimator}; \App~\Sec~\ref{sec:background-single-step-distillation-app}).

        \subsubsection{Data Attribution for Video Generation}\label{sec:background-motive}
            Gradient-based attribution ranks training examples by averaging inner products of unit $\ell_2$-normalized per-draw loss gradients $\nabla_{\denParams}\costDiffusion$ from a query and training example, using shared $(\timevar,\noiseVec)$ draws (\Eq~\ref{eq:diffusion_attrib_cosine}). For video, we use MOTIVE's motion-weighted loss and projected gradients \cite{wu2026motion}, reducing variance in the underlying gradient estimator (\Sec~\ref{sec:experiments-motive}; \App~\Sec~\ref{app:background-motive}).

\section{Our Method}\label{sec:method}
    We present \ourMethod{}: frozen-teacher gradient estimation as two-level sampling with asymmetric per-level costs (\Sec~\ref{sec:method-variance-reduction-strategies}), the allocation between levels and the two within-level variance reductions this view prescribes (Secs.~\ref{sec:method-compute-reuse}--\ref{sec:method-stratified-sampling}), and a compute-aware procedure for measuring variance and setting the allocation (\Sec~\ref{sec:method-variance-framework}). Implementation details are in \App~\Sec~\ref{app:sec_method}.

    \subsection{Frozen-Teacher Gradients as Two-Level Sampling}\label{sec:method-variance-reduction-strategies}
        \textbf{The estimand.}
            Every gradient we estimate has the form $\E_{\dataSample}\big[\E_{(\timevar,\noiseVec)\mid\dataSample}[\mathbf{h}(\dataSample,\timevar,\noiseVec)]\big]$, an average over an \emph{expensive upstream draw} $\dataSample$ and over \emph{cheap diffusion randomness} $(\timevar,\noiseVec)$ drawn given $\dataSample$. The upstream draw is a render at a sampled camera pose (SDS), a generator output for a sampled input noise (DMD), or a fixed training example (data attribution, where only the inner expectation is estimated). In SDS and DMD, $\mathbf{h}=\gterm(\dataSample,\timevar,\noiseVec)\,\nicefrac{\partial\encodedData}{\partial\genParams}$, where $\gterm$ is the vector multiplying the renderer or generator Jacobian: $\gterm=\sdsWeight(\timevar)\residual$ for SDS (\Sec~\ref{sec:background-sds}), and a weighted score difference for DMD (\Sec~\ref{sec:background-single-step-distillation}). The Jacobian is shared by every $(\timevar,\noiseVec)$ drawn at the same $\dataSample$, so the backward pass through the renderer or generator runs once per upstream draw, on the averaged inner vector; this is why inner draws are cheap.

        \textbf{The identification and the estimator.}
            This is the classical two-level, or two-stage, sampling problem \citep{hansen1943theory, cochran1977sampling, lohr2021sampling}: upstream draws are the primary sampling units and diffusion draws are the subunits, with very different per-unit costs. The standard estimator draws one subunit per primary unit, using a fresh render $\smash{\dataSample^{(\renderIndex)}=\prerender(\genParams,\cameraSample^{(\renderIndex)})}$ per sample:
            \begin{equation}\label{eq:naive-per-sample}
                    \smash{
                    \hat{\nabla}_{\genParams}^{\mathrm{naive}} =\nicefrac{1}{\numRenders}\sum\nolimits_{\renderIndex=1}^{\numRenders} \gterm(\dataSample^{(\renderIndex)},\timevar^{(\renderIndex)},\noiseVec^{(\renderIndex)}) \nicefrac{\partial \encodedData^{(\renderIndex)}}{\partial \genParams}
                    }
                \end{equation}
            where $(\cameraSample^{(\renderIndex)}, \timevar^{(\renderIndex)},\noiseVec^{(\renderIndex)})$ are IID, with cost $\numRenders(\costRender+\costDenoise)$. The two-stage estimator draws $\numRenders$ primary units and $\numReNoises$ subunits $(\timevar,\noiseVec)$ for each, and averages within units and then across them:
            \begin{equation}\label{eq:reuse-estimator}
                    \!\hat{\nabla}_{\genParams}^{\mathrm{reuse}}\!=\!\nicefrac{1}{\numRenders}\!\sum\nolimits_{\renderIndex=1}^{\numRenders}\!\Big(\nicefrac{1}{\numReNoises}\!\!\sum\nolimits_{\renoiseIndex=1}^{\numReNoises}\!\gterm(\dataSample^{(\renderIndex)},\timevar^{(\renderIndex,\renoiseIndex)},\noiseVec^{(\renderIndex,\renoiseIndex)})\!\Big)\nicefrac{\partial \encodedData^{(\renderIndex)}}{\partial \genParams}
                \end{equation}
            It is unbiased for every $\numReNoises\!\ge\!1$ because $(\timevar,\noiseVec)$ are drawn independently given $\dataSample^{(\renderIndex)}$, and it reduces to \Eq~\ref{eq:naive-per-sample} at $\numReNoises\!=\!1$. Our hierarchical MC estimator is exactly this classical estimator; caching each upstream draw and re-noising it (compute reuse) is how it is implemented.

        \textbf{The cost model.}
            Per optimization step, the two-stage estimator costs
            \begin{equation}\label{eq:cost-model}
                \mathrm{cost}\approx\numRenders\,(\costRender+\numReNoises\,\costDenoise).
            \end{equation}
            The upstream cost $\costRender$ is paid once per primary unit: a forward and backward pass through the renderer or generator and the encoder in SDS and DMD, and encoding the training example in data attribution. The denoising cost $\costDenoise$ is paid once per subunit. In SDS and DMD it covers forward passes of the frozen teacher (and, in DMD, of the fake-score model) with no backward pass through them, since the gradients omit their Jacobians (\Sec~\ref{sec:background-diffusion-applications}); in data attribution it is the forward and backward pass that produces a per-example gradient. Both costs are wall-clock times measured on the hardware in use, including parallelization (\App~\Sec~\ref{sec:method-variance-framework-app}), so $\costRender/\costDenoise$ must be measured (\Sec~\ref{sec:method-variance-framework}).

        \textbf{Variance decomposition.}
            Define the across-unit variance of the conditional mean and the mean within-unit variance, $\sigma_A^2=\mathrm{Var}_{\dataSample}\big(\E[\mathbf{h}\mid\dataSample]\big)$ and $\sigma_W^2=\E_{\dataSample}\big[\mathrm{Var}(\mathbf{h}\mid\dataSample)\big]$, with trace covariances as in \Eq~\ref{eq:var}. Writing $\hat{\mathbf{h}}^{(\renderIndex)}$ for the average of $\mathbf{h}$ over the $\numReNoises$ subunits of unit $\renderIndex$, \Eq~\ref{eq:reuse-estimator} is $\hat{\nabla}_{\genParams}=\nicefrac{1}{\numRenders}\sum_{\renderIndex}\hat{\mathbf{h}}^{(\renderIndex)}$. The $\hat{\mathbf{h}}^{(\renderIndex)}$ are IID across $\renderIndex$, and given $\dataSample^{(\renderIndex)}$ the subunit draws are IID, so the law of total variance gives $\mathrm{Var}(\hat{\mathbf{h}}^{(\renderIndex)})=\sigma_A^2+\sigma_W^2/\numReNoises$ and the classical decomposition \citep{cochran1977sampling} (\App~\Sec~\ref{app:two-stage})
            \begin{equation}\label{eq:two-level}
                \mathrm{Var}(\hat{\nabla}_{\genParams}) = \frac{\sigma_A^2}{\numRenders} + \frac{\sigma_W^2}{\numRenders\numReNoises}.
            \end{equation}
            The two terms respond differently: $\sigma_A^2/\numRenders$ falls only with more primary units, while $\sigma_W^2/(\numRenders\numReNoises)$ falls with more subunits or with a smaller $\sigma_W^2$.

        \textbf{Efficiency and the three strategies.}
            We compare estimators by efficiency $\propto 1/(\mathrm{Var}\cdot\mathrm{cost})$ \citep{hammersley1964monte, glynn1992asymptotic}; \Sec~\ref{sec:method-variance-framework} defines the two metrics we report. For this structure, classical theory offers three strategies. Allocation between levels keeps the estimator of \Eq~\ref{eq:reuse-estimator} and chooses only $(\numRenders,\numReNoises)$, the split of a budget between primary units and subunits (\Sec~\ref{sec:method-compute-reuse}). Importance sampling of the subunit draws (\Sec~\ref{sec:method-noise-schedules}) and their stratification within each primary unit (\Sec~\ref{sec:method-stratified-sampling}) change how $\timevar$ is drawn to lower $\sigma_W^2$, leaving $\sigma_A^2$ untouched.
        
        \subsubsection{Allocation Between Levels and Compute Reuse}\label{sec:method-compute-reuse}
            \textbf{The allocation problem.}
                At a fixed budget $\budget$, \Eq~\ref{eq:cost-model} gives $\numRenders=\budget/(\costRender+\numReNoises\costDenoise)$ (treating $\numRenders$ as continuous), and substituting into \Eq~\ref{eq:two-level} leaves the variance as a function of $\numReNoises$ alone:
                \begin{equation}\label{eq:budget-variance}
                    \mathrm{Var}(\hat{\nabla}_{\genParams})=\frac{1}{\budget}\Big(\sigma_A^2\costRender+\sigma_W^2\costDenoise+\sigma_A^2\costDenoise\,\numReNoises+\frac{\sigma_W^2\costRender}{\numReNoises}\Big).
                \end{equation}
                The first two terms do not depend on $\numReNoises$. The third grows with $\numReNoises$, because more subunits per unit leave fewer units; the fourth shrinks with $\numReNoises$, because the upstream cost is amortized over more subunits.

            \textbf{The optimum under the cost model.}
                \Eq~\ref{eq:budget-variance} is convex in $\numReNoises>0$ and minimized where $\sigma_A^2\costDenoise=\sigma_W^2\costRender/\numReNoises^2$, at $\numReNoises^\star=\sqrt{(\sigma_W^2/\sigma_A^2)(\costRender/\costDenoise)}$ \citep{cochran1977sampling}; by convexity the best integer is the better of $\lfloor\numReNoises^\star\rfloor$ and $\lceil\numReNoises^\star\rceil$ (at least $1$). The optimum exceeds one exactly when $\sigma_W^2\costRender>\sigma_A^2\costDenoise$, and $\numReNoises=2$ beats $\numReNoises=1$ when $\sigma_W^2\costRender>2\sigma_A^2\costDenoise$ ($\numReNoises^\star>\sqrt{2}$), so re-noising helps when upstream cost dominates, unless the within-unit variance is small relative to the across-unit variance. When upstream cost vanishes, $\numReNoises^\star\to0$, so $\numReNoises=1$ is best for IID subunits and re-noising alone cannot help; only the $\sigma_W^2$-lowering strategies of Secs.~\ref{sec:method-noise-schedules} and~\ref{sec:method-stratified-sampling} remain. All four quantities in $\numReNoises^\star$ can be estimated from a short pilot before training (\Sec~\ref{sec:method-variance-framework}). The optimum minimizes estimator variance at fixed cost; downstream metrics such as FID can respond differently (\Sec~\ref{sec:experiments-dmd}).

\begin{figure}[t]
                \centering
                \begin{minipage}[c]{0.44\textwidth}
                \centering
                \fitlinewidth{%
                \begin{tikzpicture}
                \centering
                    \node (img11){\includegraphics[trim={0.0cm 1.0cm 0.0cm 0.0cm}, clip, width=.95\linewidth]{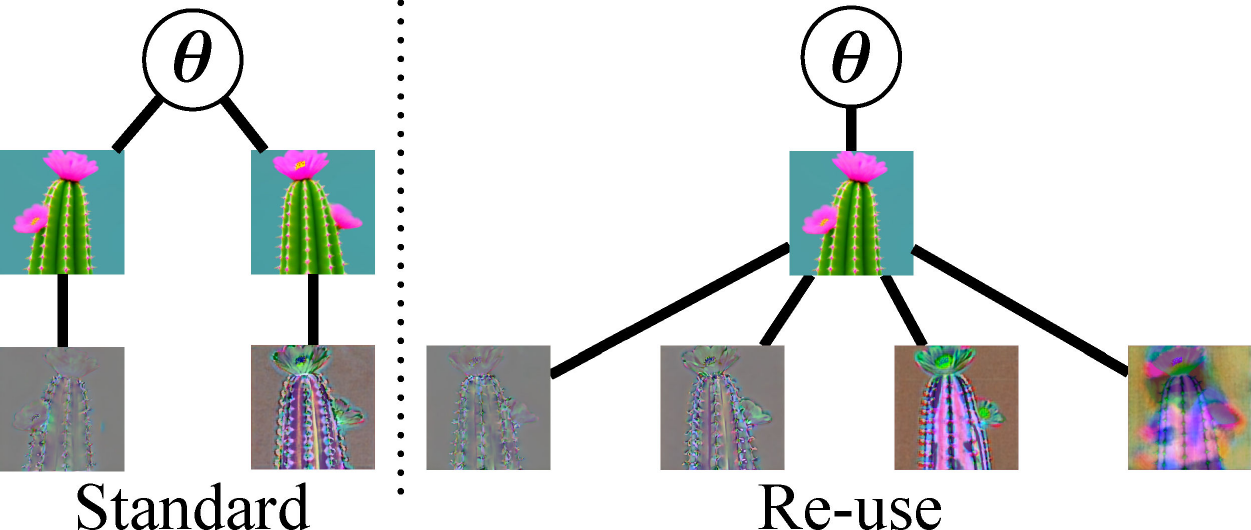}};
                    \node[below=of img11, node distance=0cm, xshift=-0.1cm, yshift=1.15cm,  font=\color{black}]{\scriptsize{$\numRenders \times \numReNoises$ sampled vectors $\gterm$ (\Eq~\ref{eq:reuse-estimator})}};
                    \node[above=of img11, node distance=0cm, xshift=-2.4cm, yshift=-1.15cm,  font=\color{black}]{\scriptsize{Baseline $(\numRenders\!=\!2, \numReNoises\!=\!1)$}};
                    \node[above=of img11, node distance=0cm, xshift=1.4cm, yshift=-1.15cm,  font=\color{black}]{\scriptsize{Reuse $(\numRenders\!=\!1, \numReNoises\!=\!4)$}};
                \end{tikzpicture}
                }
                \end{minipage}\hfill
                \begin{minipage}[c]{0.54\textwidth}
                \captionof{figure}{
                    \textbf{Compute Reuse Visualization:}
                    Baseline (left, $\numReNoises\!=\!1$) and re-noising (right, $\numReNoises\!>\!1$) use $\numRenders$ sampled renders and $\numReNoises$ noisings per render. Both take $\genParams$ (e.g., NeRF weights or generator), render, encode, noise, denoise, combine into a residual, and backpropagate. Re-noising helps when (a) $(\timevar, \noiseVec)$ drives variance and (b) denoising is cheaper than rendering. \Fig~\ref{fig:quantifying_variance_hierarchical_cost_aware_iw_strat} quantifies the trade-off. $\gterm$ is in the diffusion latent space but visualized in pixel space.
                }\label{fig:computeReuseVisualization}
                \end{minipage}
                \vspace{-0.012\textheight}
            \end{figure}

            \textbf{Compute reuse implements \boldmath$\numReNoises>1$.}
                We cache each upstream draw and re-noise it with $\numReNoises$ independent $(\timevar,\noiseVec)$ draws (\Fig~\ref{fig:computeReuseVisualization}). Since $\costRender$ includes backpropagation through the renderer or generator while $\costDenoise$ needs only forward passes of the score models, $\costRender\!\gg\!\costDenoise$ often holds in SDS and DMD. With latent diffusion, each render is encoded once and its $\numReNoises$ noised latents are formed from the cached encoding. Per-task instantiations and the combined pseudocode (\Algo~\ref{alg:combined}) are in \App~\Sec~\ref{app:algorithm}.

        \subsubsection{Importance Sampling the Timestep}\label{sec:method-noise-schedules}
            The within-unit variance $\sigma_W^2$ is an expectation over $\timevar\!\sim\!p$. Drawing $\timevar\!\sim\!\proposalDensity$ with weight $p/\proposalDensity$ leaves $\sigma_A^2$ and the form of \Eq~\ref{eq:two-level} unchanged and replaces $\sigma_W^2$ with the within-unit variance under $\proposalDensity$, which $\proposalDensity^\star\!\propto\!p\sqrt{\E[\|\mathbf{h}\|_2^2\mid\timevar]}$ minimizes \citep{kahn1953methods, rubinstein2016simulation}.

            For SDS, the per-sample contribution multiplying the renderer Jacobian is $\gterm(\dataSample,\timevar,\noiseVec)\!=\!\sdsWeight(\timevar)\,\residual$, giving parameter-gradient form $\sdsWeight(\timevar)\,\jacobian_{\genParams}^\top\residual$ at fixed render. The variance-minimizing proposal from \Sec~\ref{sec:background-reducing-variance-importance_sampling} is then $\proposalDensity^\star(\timevar)\!\propto\!p(\timevar)\sqrt{\E[\|\sdsWeight\jacobian_{\genParams}^\top\residual\|_2^2\!\mid\!\timevar]}$, which requires per-timestep gradient norms over renders and noise and is impractical. Loss-based proxies use only $\|\residual\|_2^2$ and ignore $\sdsWeight^2$ and $\|\jacobian_{\genParams}\|_2^2$; since $\|\sdsWeight\jacobian_{\genParams}^\top\residual\|_2^2 \!\leq\! \sdsWeight^2\|\jacobian_{\genParams}\|_2^2\|\residual\|_2^2$, they can misrank timesteps when $\sdsWeight$ varies with $\timevar$ or $\|\jacobian_{\genParams}^\top\residual\|_2$ correlates poorly with $\|\residual\|_2$, and ill-conditioned Jacobian chains through encoders and differentiable generators can amplify this.

            Because $\sdsWeight$ is deterministic given $\timevar$, $\proposalDensity^\star\!\propto\!p\,\sdsWeight\sqrt{\E[\|\jacobian_{\genParams}^\top\residual\|_2^2\!\mid\!\timevar]}$, so $\proposalDensity\!\propto\!p\,\sdsWeight$ is exact when this conditional second moment is flat in $\timevar$; the estimator keeps the Jacobian, and only the proposal omits it. Empirically, $\sdsWeight$ dominates the timestep dependence of the gradient norm (\App~\Fig~\ref{fig:iwvis_latent_vs_param}), so we use the negligible-cost proposal $\proposalDensity(\timevar)\!\propto\!p(\timevar)\sdsWeight(\timevar)$ with likelihood-ratio correction.

            For data attribution (\Sec~\ref{sec:experiments-motive}), gradient norms are approximately constant across timesteps (\App~\Fig~\ref{fig:optimal_importance_motive}), so we use uniform sampling. For DMD (\Sec~\ref{sec:experiments-dmd}), the weighting function is non-monotonic with data-dependent normalization, so we focus on stratification and compute reuse.

        \subsubsection{Stratified Sampling of the Re-noisings}\label{sec:method-stratified-sampling}
            Taking the $\numReNoises$ subunit draws of each primary unit one per equal-mass stratum of $\proposalDensity$ removes the between-stratum part of $\sigma_W^2$ and leaves $\sigma_A^2$ unchanged \citep{cochran1977sampling, owen2013monte}; the inverse-CDF quantile construction \citep{glasserman2004monte} composes this with any proposal. Timestep discretization details are in \App~\Sec~\ref{app:algorithm}.

            \textbf{Per-render stratification.}
                With re-noising, set the number of strata equal to the number of re-noisings, $\numStrata \defeq \numReNoises$. For a uniform $p$, partition $[0,1]$ into equal-width bins $\smash{\stratum_{\stratumIndex}=[(\stratumIndex-1)/\numStrata,\stratumIndex/\numStrata]}$ for $\stratumIndex=1,\dots,\numStrata$, each with probability $\nicefrac{1}{\numStrata}$. For each render $\dataSample^{(\renderIndex)}$, draw one timestep per bin and independent noise $\smash{\noiseVec^{(\renderIndex)}_{\stratumIndex}}$. The per-render contribution is:
                \begin{equation}\label{eq:per-render-strat}
                    \smash{
                    \bar{\gterm}^{(\renderIndex)}_{\mathrm{strat}} = \nicefrac{1}{\numStrata}\sum\nolimits_{\stratumIndex=1}^{\numStrata} \gterm(\dataSample^{(\renderIndex)},\timevar^{(\renderIndex)}_{\stratumIndex},\noiseVec^{(\renderIndex)}_{\stratumIndex})
                    }
                \end{equation}
                where $\smash{\timevar^{(\renderIndex)}_{\stratumIndex}}$ is uniform on $\stratum_{\stratumIndex}$, and the gradient estimate is $\smash{\tfrac{1}{\numRenders}\sum_{\renderIndex}\bar{\gterm}^{(\renderIndex)}_{\mathrm{strat}}\tfrac{\partial \encodedData^{(\renderIndex)}}{\partial \genParams}}$ (see \Eq~\ref{eq:reuse-estimator}). This stratifies the subunits within each primary unit.
            \begin{figure}[t]
                \centering
                \begin{minipage}[c]{0.45\textwidth}
                \centering
                \fitlinewidth{%
                    \begin{tikzpicture}
                    \centering
                        \node (img21){\includegraphics[trim={0.8cm 0.8cm 0.8cm 0.9cm}, clip, width=.75\linewidth]{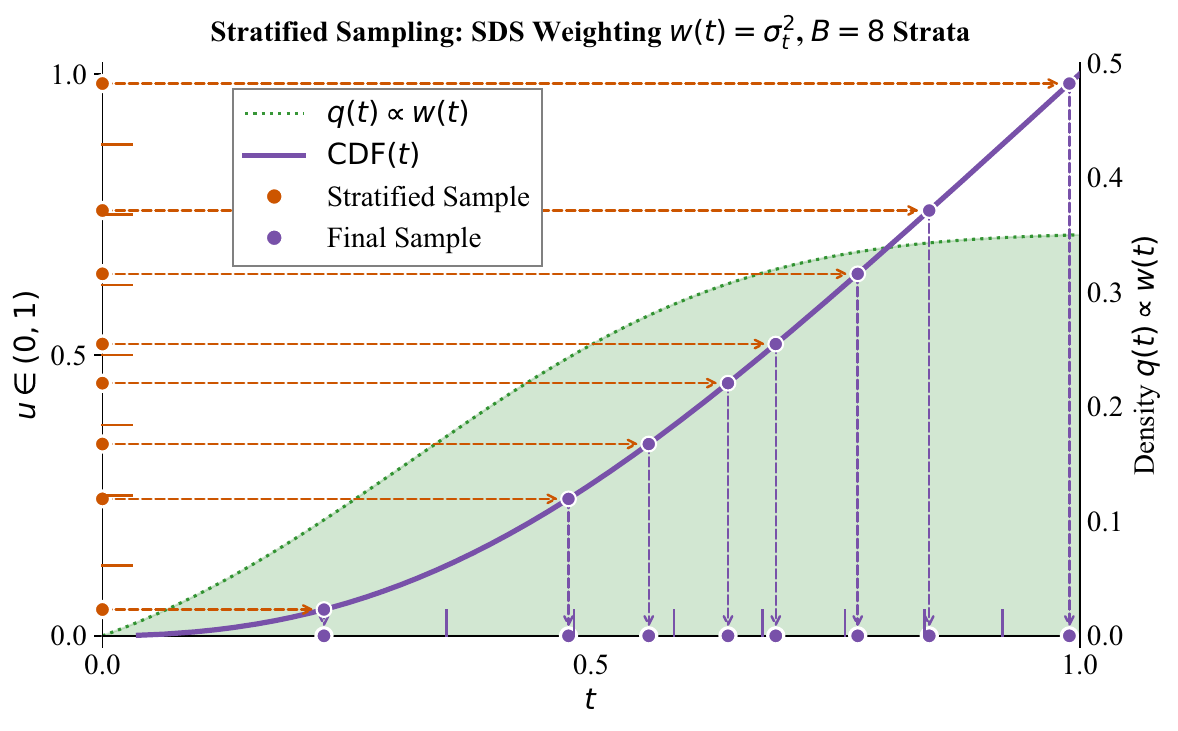}};
                        \node[right=of img21, node distance=0cm, xshift=-.95cm, yshift=1.0cm, rotate=270, font=\scriptsize\color{black}]{Density $\proposalDensity(\timevar)$};
                        \node[left=of img21, node distance=0cm, rotate=90, xshift=1.4cm, yshift=-.95cm, font=\scriptsize\color{black}]{Quantile Level $\quantile$};
                        \node[below=of img21, node distance=0cm, xshift=0.0cm, yshift=1.21cm, font=\scriptsize\color{black}]{Timestep $\timevar = \textnormal{CDF}_{\proposalDensity}^{-1}(\quantile)$, with stratified $\quantile$};
                    \end{tikzpicture}
                    }
                \end{minipage}\hfill
                \begin{minipage}[c]{0.53\textwidth}
                \captionof{figure}{
                        \textbf{Combining Stratified Sampling with Importance Weighting:}
                        We illustrate how to use inverse-transform sampling to map a stratified sample uniformly in $[0, 1]$ (see \App~\Fig~\ref{fig:stratifiedVisualization}) into a ``stratified'' sample for our target importance distribution, using the inverse-CDF, with the now non-uniform bins shown via \colorE{purple} ticks. This allows us to combine the benefits of both strategies, forming better estimators (see \Fig~\ref{fig:quantifying_variance_hierarchical_cost_aware_iw_strat}).
                    }\label{fig:inverseTransformVisualization}
                \end{minipage}
                \vspace{-0.0159\textheight}
            \end{figure}

            \textbf{Stratified importance sampling.}
                To combine IS and stratification for a non-uniform proposal, stratify in proposal-quantile space \citep{glasserman2004monte} rather than $\timevar$ (\Fig~\ref{fig:inverseTransformVisualization}). For each render $\dataSample^{(\renderIndex)}$ and stratum $\stratumIndex=1\dots\numStrata$, with proposal $\proposalDensity(\timevar)$ and IID jitters $\smash{\randomness^{(\renderIndex)}_{\stratumIndex}\sim\Uniform(0,1)}$, define:
                \begin{equation}
                    \smash{
                    \!\!\!\!\quantile^{(\renderIndex)}_{\stratumIndex}=(\stratumIndex-1+\randomness^{(\renderIndex)}_{\stratumIndex})/\numStrata,
                    \quad
                    \timevar^{(\renderIndex)}_{\stratumIndex}=\textnormal{CDF}_{\proposalDensity}^{-1}(\quantile^{(\renderIndex)}_{\stratumIndex})
                    }
                \end{equation}
                so that $\smash{\{\timevar^{(\renderIndex)}_{\stratumIndex}\}_{\stratumIndex=1}^{\numStrata}}$ contains one draw from each equal-mass stratum of $\proposalDensity$, or equivalently, non-uniform bins in $\timevar$ whose $\proposalDensity$-probabilities are all $1/\numStrata$. With independent $\smash{\noiseVec^{(\renderIndex)}_{\stratumIndex}\sim\standardNormal}$ and importance weights $\importanceWeight(\timevar)=\nicefrac{p(\timevar)}{\proposalDensity(\timevar)}$, the per-render stratified-IS contribution is:
                \begin{equation}\label{eq:stratified-IS}
                    \smash{
                    \!\!\!\!\bar{\gterm}^{(\renderIndex)}_{\mathrm{strat\mbox{-}IS}}
                    \!=\!\nicefrac{1}{\numStrata}\!\!\sum\nolimits_{\stratumIndex=1}^{\numStrata}\!\!\!
                    \importanceWeight(\timevar^{(\renderIndex)}_{\stratumIndex}\!)
                    \gterm(\dataSample^{(\renderIndex)},\timevar^{(\renderIndex)}_{\stratumIndex},\noiseVec^{(\renderIndex)}_{\stratumIndex})
                    }
                \end{equation}
                This remains unbiased for sampling $\timevar\sim p$ while reallocating draws toward timesteps with larger contributions via $\proposalDensity$ and reducing variance by enforcing balanced coverage across the $\proposalDensity$-quantiles even when $\proposalDensity$ is highly~\mbox{non-uniform}.

            \textbf{Design choice.}
                Both schemes add negligible compute. Per-render stratification (\Eq~\ref{eq:per-render-strat}) is preferred when $\numReNoises\!>\!1$, since it stratifies the subunits and composes with the allocation of \Sec~\ref{sec:method-compute-reuse}; when $\numReNoises\!=\!1$ it reduces to IID sampling, so the batch instead uses $\numStrata=\numRenders\numReNoises$ equal-mass strata, with one timestep drawn per stratum (global stratification, \Eq~\ref{eq:global-strat} in \App~\Sec~\ref{app:algorithm}). \App~\Fig~\ref{fig:variance_ablation_stratified} confirms the $\numReNoises\!>\!1$ preference, and we use per-render stratification for SDS and DMD, where $\numReNoises\!>\!1$ is efficient; for data attribution, global stratification is most effective (\Sec~\ref{sec:experiments-motive}).

            \textbf{Why the gains compose.}
                Allocation trades the two terms of \Eq~\ref{eq:two-level} against each other. Importance sampling with a good proposal and stratification both lower $\sigma_W^2$, by changing the measure and by removing between-stratum variance under it, and neither touches $\sigma_A^2$, so the three constructions compose and, empirically, their gains approximately multiply (\Tab~\ref{tab:relative_improvement_by_m_param}; \App~\Sec~\ref{app:ecm-worked-example}); the across-unit term eventually limits all three.

    \subsection{Variance Measurement and Allocation Procedure}\label{sec:method-variance-framework}
        \textbf{Variance estimation.}
        We estimate the variance of each estimator (\Eq~\ref{eq:var}) with Welford's online algorithm~\citep{welford1962note}, without storing samples, until the estimate converges (stopping rule in \App~\Sec~\ref{sec:experiments-sds-app-details}). For some experiments, a high-sample reference validates convergence (the MSE to it agrees with the online estimate) and gives cosine-similarity measures of directional accuracy.

        \textbf{Pilot estimation.}
            The same machinery estimates both terms of \Eq~\ref{eq:two-level} from a short pilot (\App~\Sec~\ref{app:practitioner-guide}): the Welford variance over re-noisings of each cached render, averaged over renders, estimates $\sigma_W^2$; the variance of the per-render means minus $\sigma_W^2/\numReNoises$ estimates $\sigma_A^2$ (extend the pilot if this estimate is not positive); with the timed costs, these give $\numReNoises^\star$ (\Sec~\ref{sec:method-compute-reuse}).

        \textbf{Allocation procedure.}
            In practice: (i) time $\costRender$ and $\costDenoise$; (ii) choose the proposal: if a known timestep weight dominates how gradient norms vary with $\timevar$ ($\sdsWeight$ in SDS), sample from $p$ times that weight with the likelihood-ratio correction, and otherwise keep $p$ (\Sec~\ref{sec:method-noise-schedules}); (iii) run a short pilot under that proposal at $\numReNoises\!>\!1$ to estimate $\sigma_A^2$ and $\sigma_W^2$ as above; (iv) set $\numReNoises$ to $\lfloor\numReNoises^\star\rfloor$ or $\lceil\numReNoises^\star\rceil$, whichever has the lower variance in \Eq~\ref{eq:budget-variance} (at least $1$); (v) stratify per render when $\numReNoises\!>\!1$ and globally when $\numReNoises\!=\!1$ (\Sec~\ref{sec:method-stratified-sampling}). Stratification lowers $\sigma_W^2$ further, so check the final configuration by its measured ECM. \App~\Sec~\ref{app:practitioner-guide} gives the full decision tree with thresholds.

        \textbf{Efficiency metrics.}
            To compare estimators differing in both variance and cost (wall-clock), we use efficiency $\propto 1/(\mathrm{Var}\cdot\mathrm{cost})$ and report two metrics ($>\!1$ better; see \App~\Fig~\ref{fig:compute_mult_explanation} for intuition):
            \begin{itemize}[nosep]
                \item \emph{Effective compute multiplier} (ECM) compares to a baseline (uniform-IID with $\numReNoises\!=\!1$) at iso-variance: $\mathrm{ECM} = \mathrm{cost}_{\mathrm{baseline}} / \mathrm{cost}_{\mathrm{method}}$. Computing ECM requires estimating $\mathrm{cost}_{\mathrm{baseline}}$ at the method's variance; we interpolate in log-log space along the baseline Pareto curve, exploiting the standard Monte Carlo variance rate $\mathrm{Var} \propto 1/\numRenders$ to extrapolate when needed.
                \item \emph{Relative efficiency} (RE) compares to uniform-IID at identical $(\numRenders, \numReNoises)$: $\mathrm{RE}\!=\!\mathrm{Var}_\mathrm{u}/\mathrm{Var}_\mathrm{m}$, isolating importance weighting (IW) and stratification from batch-size effects.
            \end{itemize}

        \textbf{Task-specific quantities.}
            For SDS (\Sec~\ref{sec:experiments-sds}), we measure variance of $\sdsupdatehat(\genParams)$ estimating $\sdsupdate(\genParams)$ (\Eq~\ref{eq:sds_update}). For DMD (\Sec~\ref{sec:experiments-dmd}), we measure the variance of the generator gradient from the score difference, with a positive control (\App~\Sec~\ref{sec:experiments-dmd-disconnect}). For data attribution (\Sec~\ref{sec:experiments-motive}), we measure the variance of per-example gradients used for influence scores. Details in \App~\Sec~\ref{sec:method-variance-framework-app}.

        \textbf{What is classical and what is new.}
            The decomposition in \Eq~\ref{eq:two-level}, the two-level allocation, importance sampling, stratification, and their composition through equal-mass quantiles are classical \citep{cochran1977sampling, owen2013monte}. New to this setting are: the identification of frozen-teacher gradient estimation as a two-level problem with asymmetric per-level costs; the teacher's own weight $\sdsWeight$ as a zero-cost proxy for the variance-optimal proposal; stratified-IS timestep allocation composed with per-render stratification, an adaptation of the standard construction; a cost model separating $\costRender$ from $\costDenoise$ (ECM, RE); and parameter-gradient variance measurement across three applications, which exposes the DMD boundary (\Sec~\ref{sec:experiments-dmd}). Details in \App~\Sec~\ref{app:novelty-discussion}.

\section{Experiments}\label{sec:experiments}
    We apply our methods to three diffusion-teacher tasks: optimization with diffusion priors (\Sec~\ref{sec:experiments-sds}), single-step distillation (\Sec~\ref{sec:experiments-dmd}), and data attribution (\Sec~\ref{sec:experiments-motive}). Across tasks, our framework reveals lower-variance setups per compute budget; details in \App~\Sec~\ref{app:sec_experiments}.

    \begin{figure}[t!]
        \vspace{-0.0\textheight}
        \centering
        \begin{minipage}[b]{0.48\textwidth}
            \centering
            \resizebox{\linewidth}{!}{%
            \scalebox{1}[1.0]{%
            \begin{tikzpicture}
            \centering
                \node (img11){\includegraphics[trim={1.35cm 1.9cm 0cm 0.8cm}, clip, width=.96\linewidth, height=.65\linewidth]{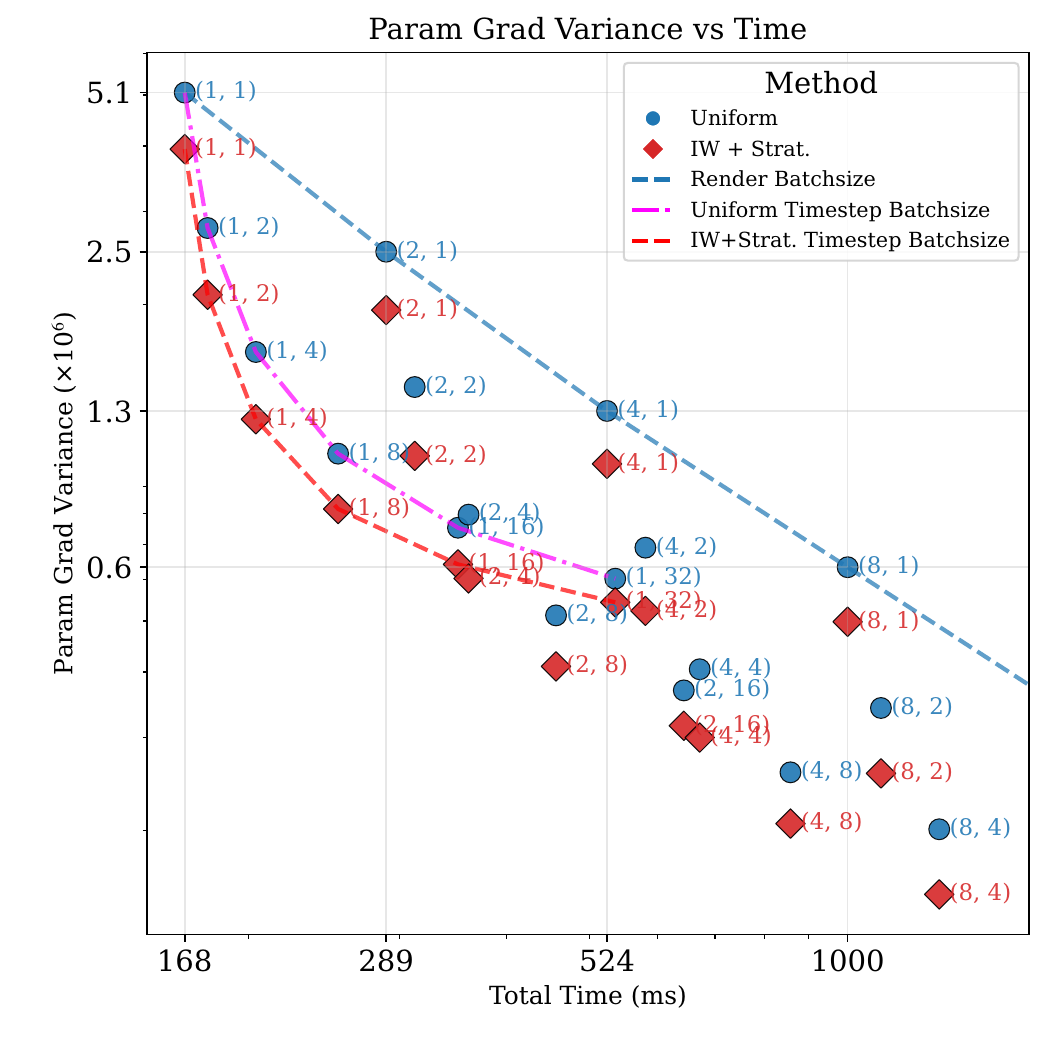}};
                \node[left=of img11, node distance=0cm, rotate=90, xshift=1.0cm, yshift=-1.0cm,  font=\color{black}]{\scriptsize{Variance $(\times10^6)$}};

                \node [below=of img11, node distance=0cm, yshift=1.1cm](img21){\includegraphics[trim={1.5cm 1.25cm 0cm 0.8cm}, clip, width=.96\linewidth, height=.65\linewidth]{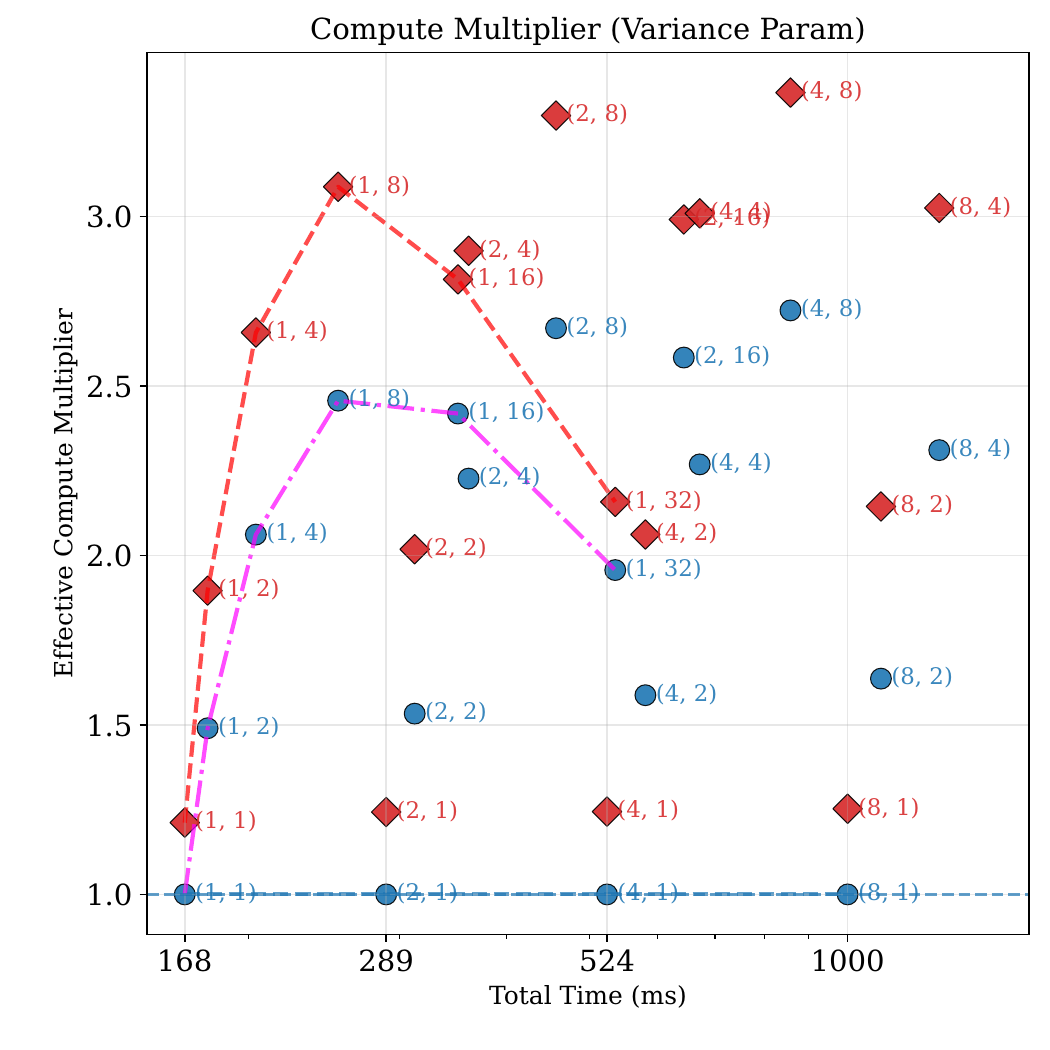}};
                \node[left=of img21, node distance=0cm, rotate=90, xshift=2.25cm, yshift=-1.0cm,  font=\color{black}]{\scriptsize{Effective Compute Mult. to Baseline}};
                \node[below=of img21, node distance=0cm, xshift=0.0cm, yshift=1.15cm,  font=\color{black}]{\scriptsize{Per-Iteration Compute (ms)}};
            \end{tikzpicture}
            }
            }
            \captionof{figure}{
                \textbf{Quantifying variance reduction from IW and stratification (SDS).}
                \emph{Top:} Variance of the parameter-gradient estimator (\Eq~\ref{eq:two-level}) late in training vs.\ compute. Colors: uniform baseline and IW+Strat. Points annotated by $(\numRenders,\numReNoises)$.
                    \emph{Bottom:} Effective compute multiplier vs.\ uniform baseline. Lines trace $(\numRenders\!=\!1,\numReNoises)$ and peak at $\numReNoises\!=\!8$. Averaged over $\numRenders$, $\numReNoises\!=\!8$ gives $\sim\!2.6\times$ (uniform), $\sim\!3.3\times$ (IW+Strat; \Tab~\ref{tab:absolute_ecm_by_m_param}). Ablations in \App~\Fig~\ref{fig:quantifying_variance_hierarchical_cost_aware_iw_strat}; RE in \Tab~\ref{tab:relative_improvement_by_m_param}.
            }\label{fig:quantifying_variance_hierarchical_cost_aware_iw_strat_main}
        \end{minipage}\hfill
        \begin{minipage}[b]{0.48\textwidth}
            \centering
            \resizebox{\linewidth}{!}{%
            \begin{tikzpicture}
            \centering
                \node (img11b){\includegraphics[trim={0.3cm 0.0cm 0.0cm 0.25cm}, clip, width=.9\linewidth]{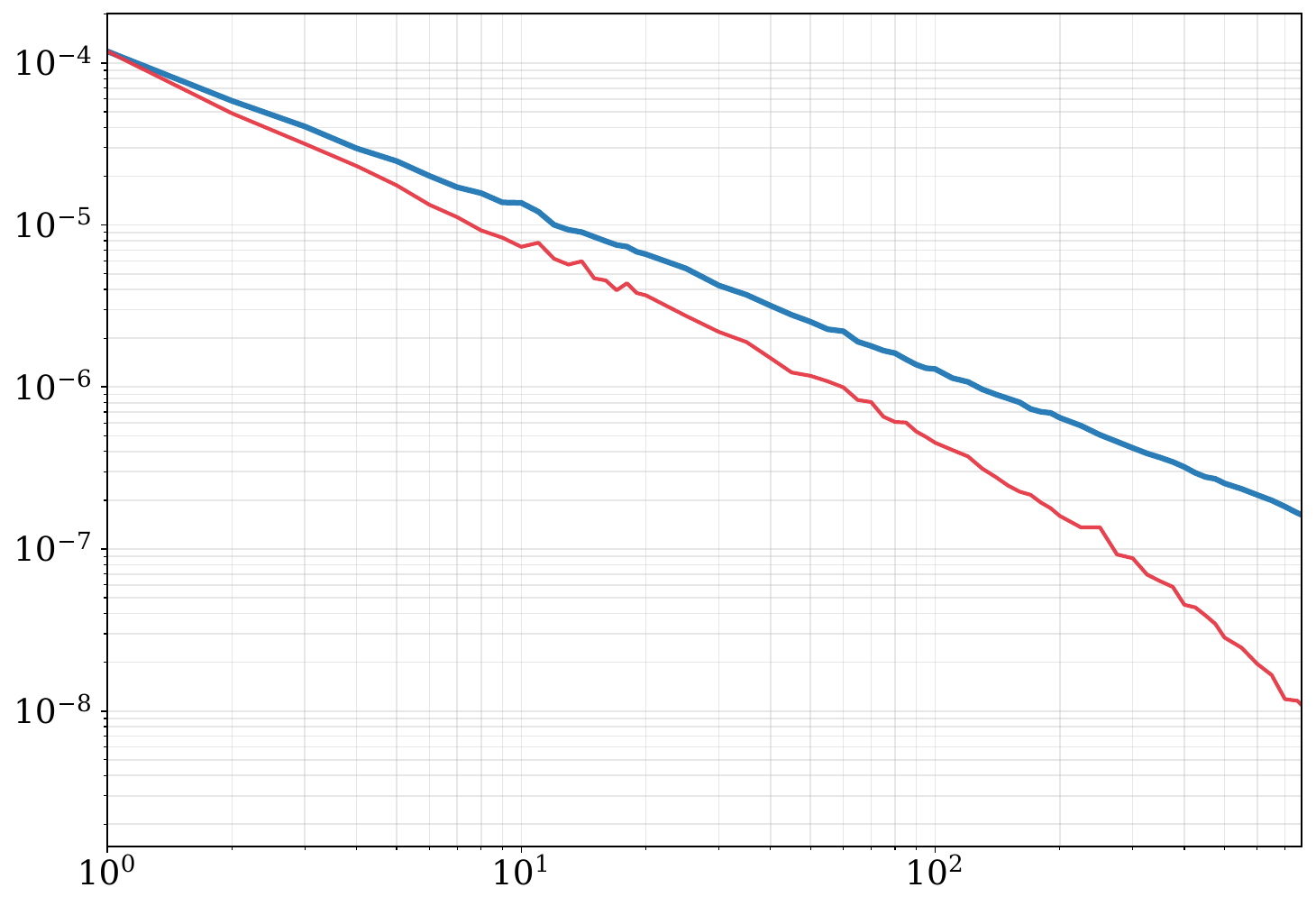}};
                \node[left=of img11b, node distance=0cm, rotate=90, xshift=1.5cm, yshift=-.8cm,  font=\color{black}]{\normalsize{Gradient Variance}};

                \node [below=of img11b, yshift=1.25cm](img21b){\includegraphics[trim={0.0cm 0.0cm 0.0cm 0.25cm}, clip, width=.9\linewidth]{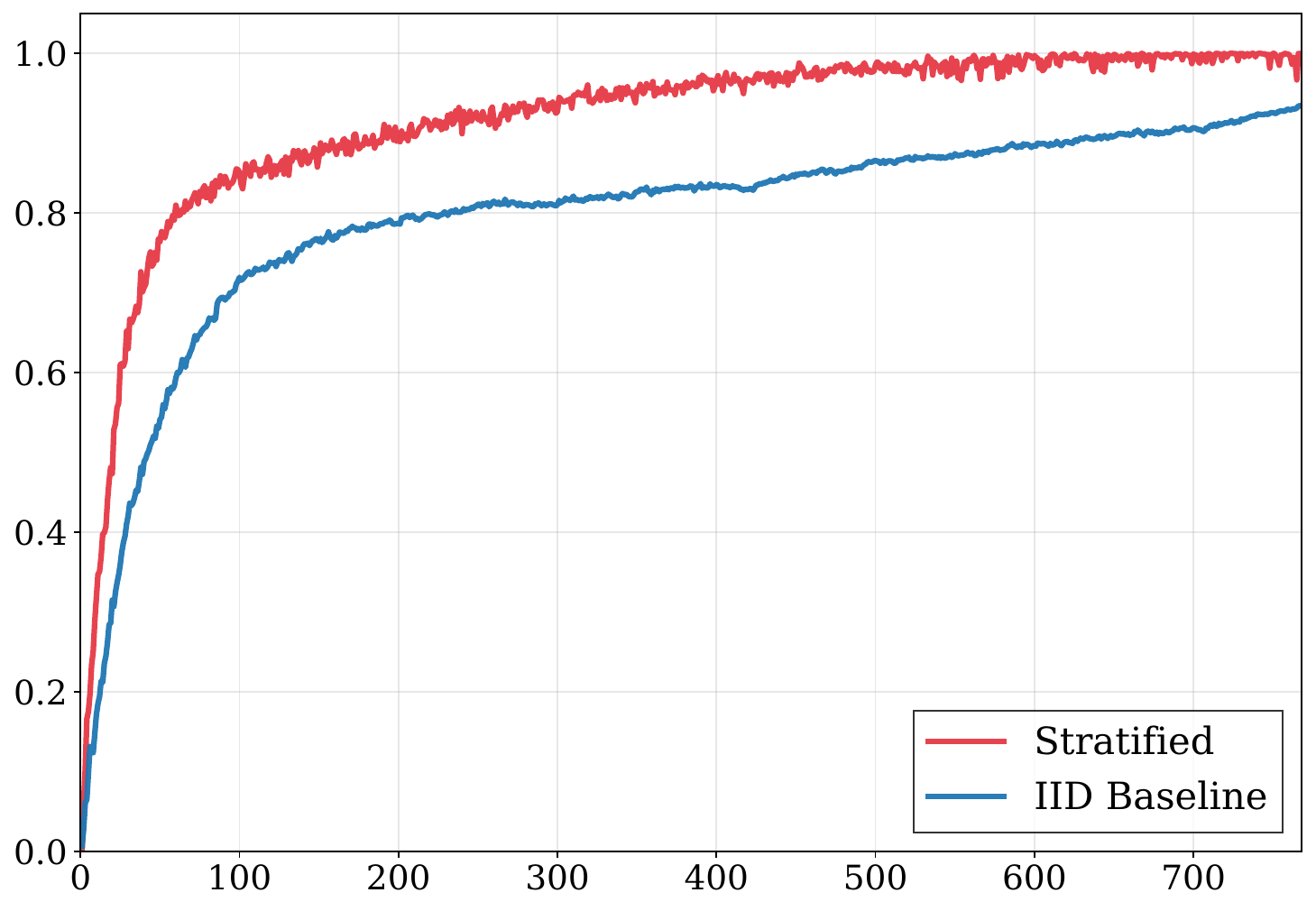}};
                \node[left=of img21b, node distance=0cm, rotate=90, xshift=1.5cm, yshift=-.8cm,  font=\color{black}]{\normalsize{Mean Correlation}};
                \node[below=of img21b, node distance=0cm, xshift=0.0cm, yshift=1.15cm,  font=\color{black}]{\normalsize{Gradient Samples per Data Point}};
            \end{tikzpicture}
            \vspace{-0.01\textheight}
            }
            \captionof{figure}{
                \textbf{Quantifying Changes in Data Attribution.}
                \emph{Top:} Gradient variance vs.\ evaluations per data point. Stratified sampling beats uniform sampling at an equal budget.
                \emph{Bottom:} Mean correlation of limited-evaluation rankings with ground-truth rankings (${\sim}1.3$--$3.8\times$ compute multiplier at the tabulated budgets, $>\!2\times$ at typical practical budgets; see \Tab~\ref{tab:motive_gradient_estimation_summary} and \App~\Fig~\ref{fig:quantifying_attribution_full}). Qualitative examples in \App~\Fig~\ref{fig:motive_qualitative}.
            }\label{fig:quantifying_attribution}
        \end{minipage}
        \vspace{-0.005\textheight}
    \end{figure}

    \subsection{Diffusion Priors for Optimization}\label{sec:experiments-sds}

\begin{figure}[t!]
                \centering
                \vspace{-0.0\textheight}
                \begin{minipage}[b]{0.40\textwidth}
                    \centering
                    \vspace*{-0.0em}
                    \resizebox{\linewidth}{!}{%
                    \scalebox{1}[0.9]{%
                    \begin{tikzpicture}
                    \centering
                        \node (img11){\includegraphics[trim={1.1cm 1.25cm 0cm 0cm}, clip, width=.77\linewidth]{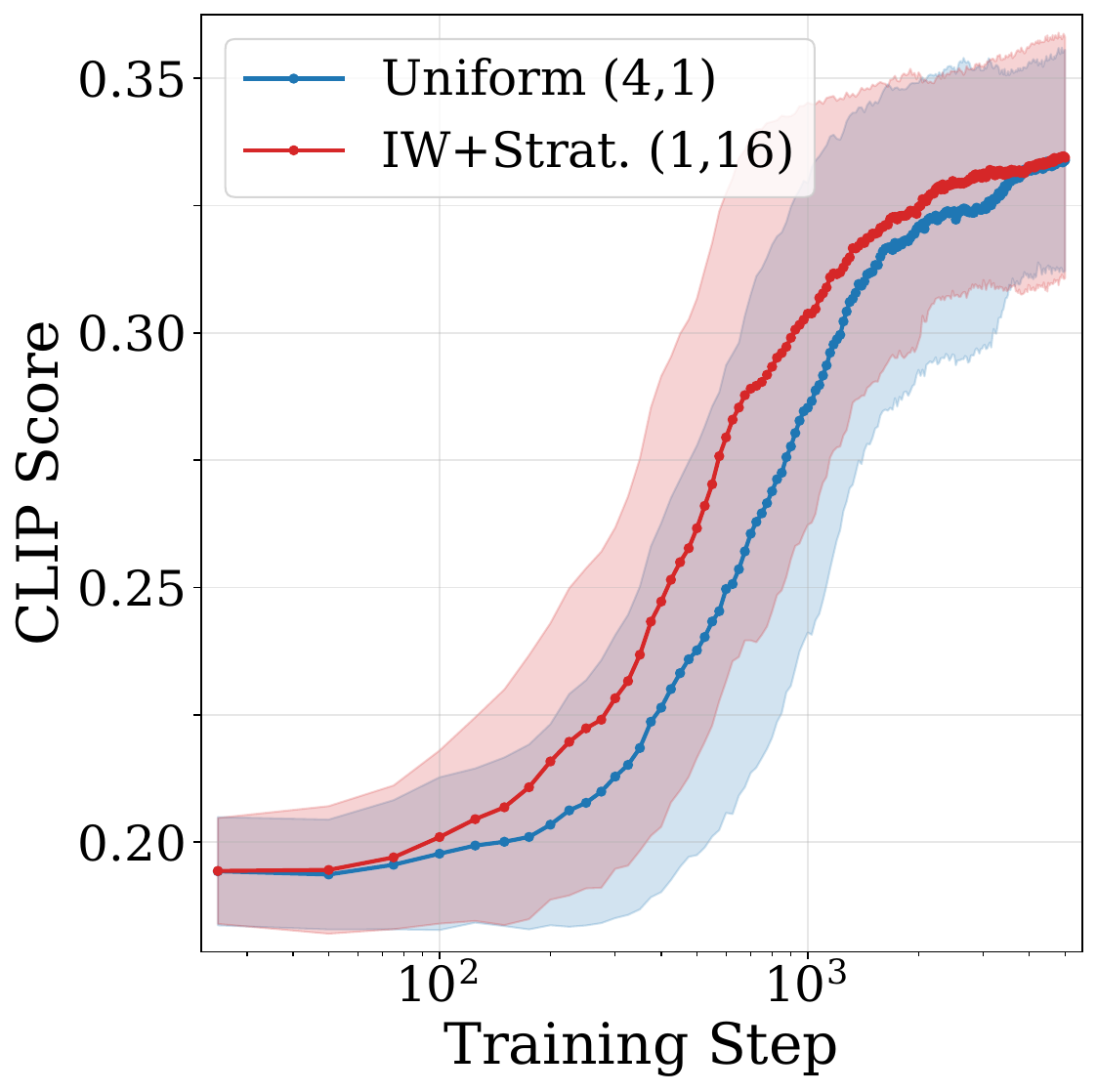}};
                        \node[left=of img11, node distance=0cm, rotate=90, xshift=1.0cm, yshift=-.8cm,  font=\scriptsize\color{black}]{CLIP Score};
                        \node[below=of img11, node distance=0cm, xshift=0.0cm, yshift=1.25cm,  font=\scriptsize\color{black}]{Optimization Iteration};
                    \end{tikzpicture}
                    }
                    }
                    \captionof{figure}{
                        \textbf{Performance Gains from Variance Reduction:}
                        CLIP score versus optimization iteration, averaged across $30$ prompts, $3$ seeds, and multiple views ($\pm$ std. dev.). Baseline $(4,1)$ vs.\ ours $(1,16)$ (stratified+IS+re-noising), whose iterations cost less (\Fig~\ref{fig:quantifying_variance_hierarchical_cost_aware_iw_strat_main}). Higher CLIP at fixed iteration count from lower per-iteration variance (\Fig~\ref{fig:quantifying_variance_hierarchical_cost_aware_iw_strat}); final samples in \App~\Fig~\ref{fig:qualitative_sds}.
                    }
                    \label{fig:clip_sds}
                \end{minipage}\hfill
                \begin{minipage}[b]{0.58\textwidth}
                    \centering
                    \vspace*{-0.0em}
                    \resizebox{\linewidth}{!}{%
                    \begin{tikzpicture}
                        \def\imgw{0.2\linewidth}
                        \def\xdist{-1.3cm}
                        \def\ydist{1.3cm}
                        \def\ydistgroup{\ydist}

                        \node (p1b0) {\includegraphics[trim={0 0 0 0}, clip, width=\imgw]{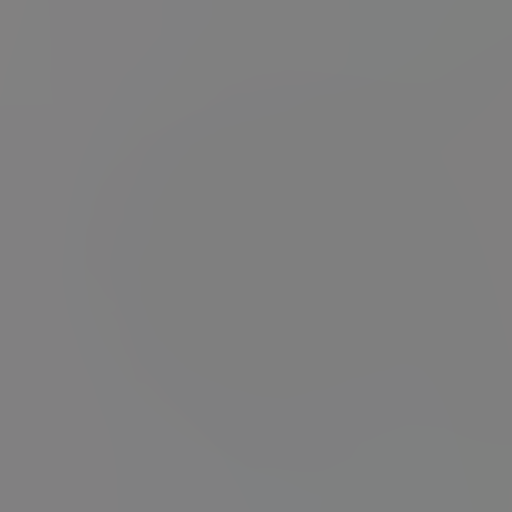}};
                        \node (p1b1) [right=of p1b0, xshift=\xdist]{\includegraphics[trim={0 0 0 0}, clip, width=\imgw]{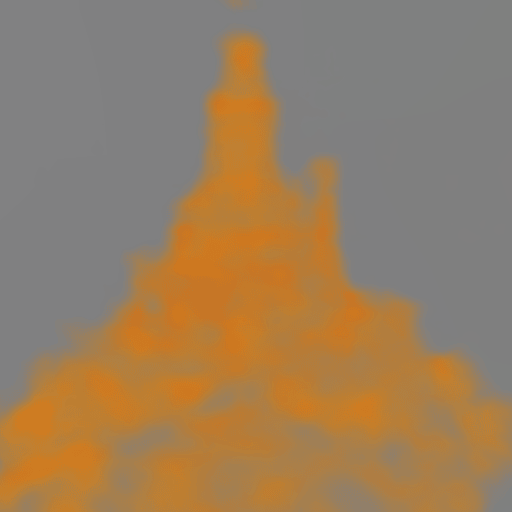}};
                        \node (p1b2) [right=of p1b1, xshift=\xdist]{\includegraphics[trim={0 0 0 0}, clip, width=\imgw]{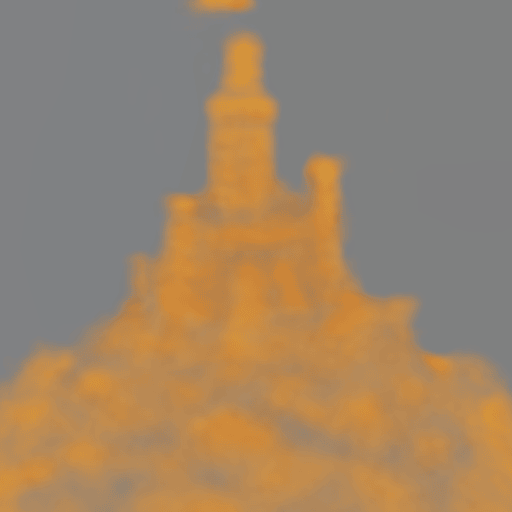}};
                        \node (p1b3) [right=of p1b2, xshift=\xdist]{\includegraphics[trim={0 0 0 0}, clip, width=\imgw]{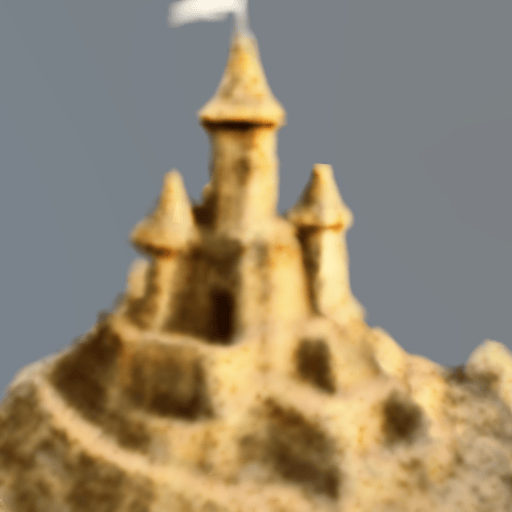}};
                        \node (p1bf) [right=of p1b3, xshift=\xdist]{\includegraphics[trim={0 0 0 0}, clip, width=\imgw]{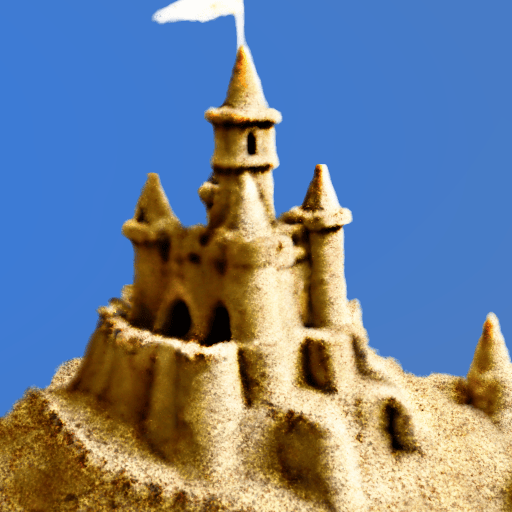}};

                        \node (p1u0) [below=of p1b0, yshift=\ydist]{\includegraphics[trim={0 0 0 0}, clip, width=\imgw]{images/sds/qualitative/training/0.png}};
                        \node (p1u1) [right=of p1u0, xshift=\xdist]{\includegraphics[trim={0 0 0 0}, clip, width=\imgw]{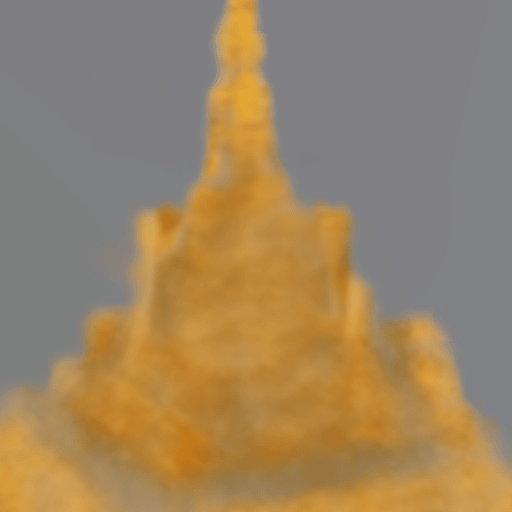}};
                        \node (p1u2) [right=of p1u1, xshift=\xdist]{\includegraphics[trim={0 0 0 0}, clip, width=\imgw]{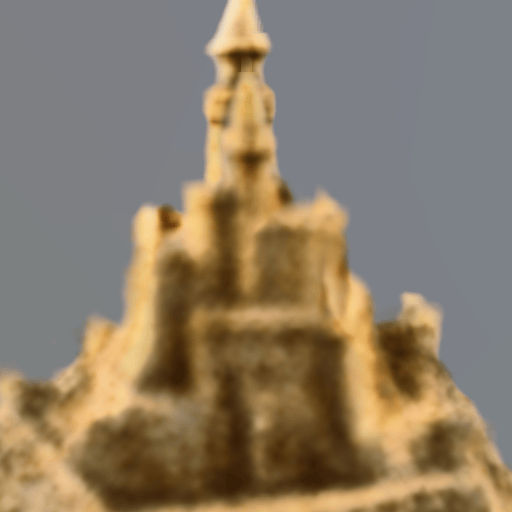}};
                        \node (p1u3) [right=of p1u2, xshift=\xdist]{\includegraphics[trim={0 0 0 0}, clip, width=\imgw]{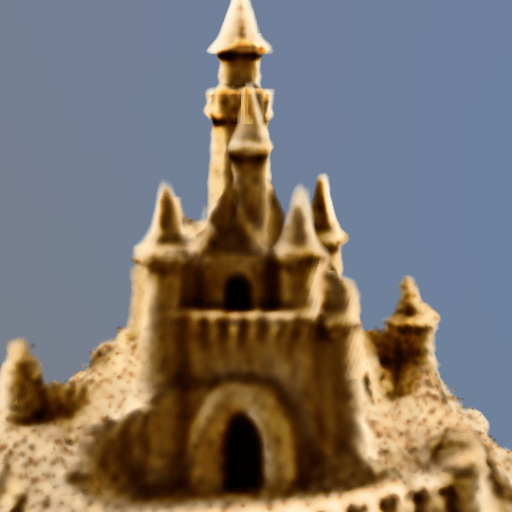}};
                        \node (p1uf) [right=of p1u3, xshift=\xdist]{\includegraphics[trim={0 0 0 0}, clip, width=\imgw]{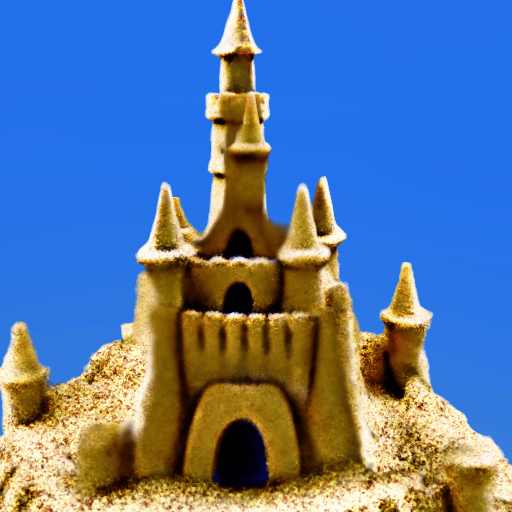}};

                        \node (p2b0) [below=of p1u0, yshift=\ydistgroup]{\includegraphics[trim={0 0 0 0}, clip, width=\imgw]{images/sds/qualitative/training/0.png}};
                        \node (p2b1) [right=of p2b0, xshift=\xdist]{\includegraphics[trim={0 0 0 0}, clip, width=\imgw]{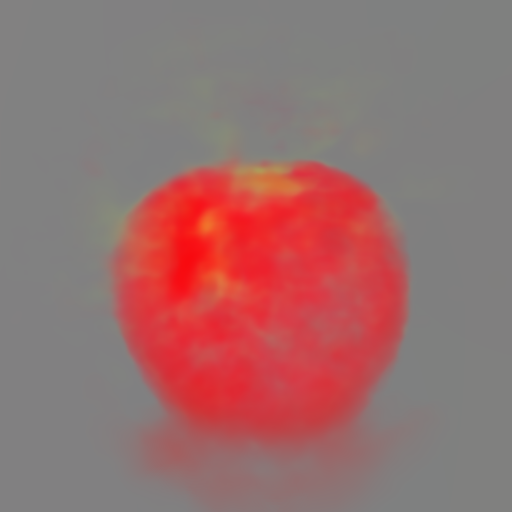}};
                        \node (p2b2) [right=of p2b1, xshift=\xdist]{\includegraphics[trim={0 0 0 0}, clip, width=\imgw]{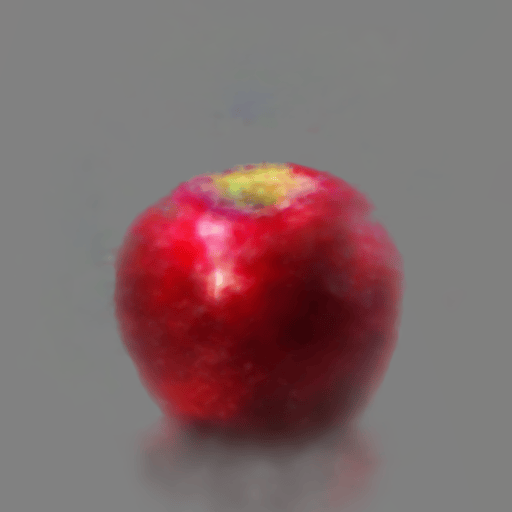}};
                        \node (p2b3) [right=of p2b2, xshift=\xdist]{\includegraphics[trim={0 0 0 0}, clip, width=\imgw]{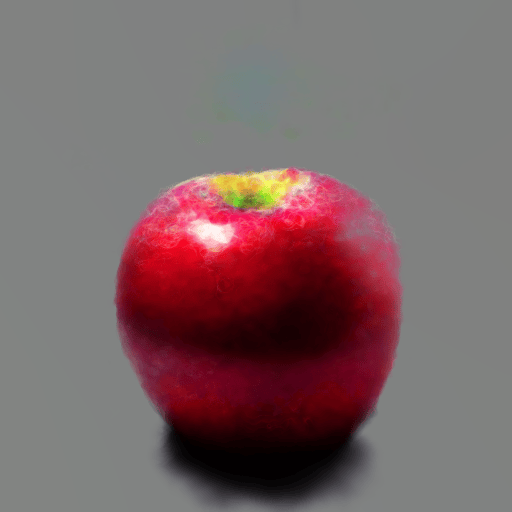}};
                        \node (p2bf) [right=of p2b3, xshift=\xdist]{\includegraphics[trim={0 0 0 0}, clip, width=\imgw]{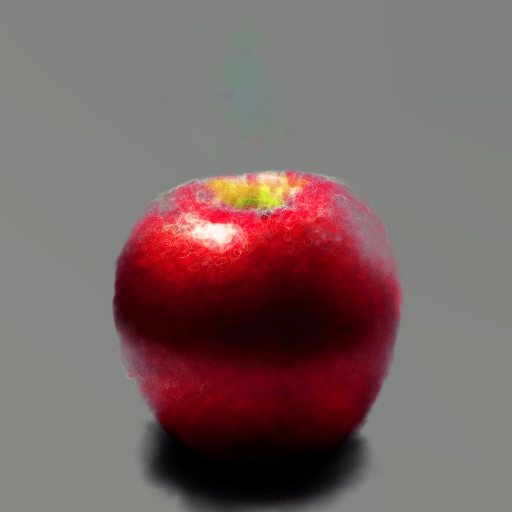}};

                        \node (p2u0) [below=of p2b0, yshift=\ydist]{\includegraphics[trim={0 0 0 0}, clip, width=\imgw]{images/sds/qualitative/training/0.png}};
                        \node (p2u1) [right=of p2u0, xshift=\xdist]{\includegraphics[trim={0 0 0 0}, clip, width=\imgw]{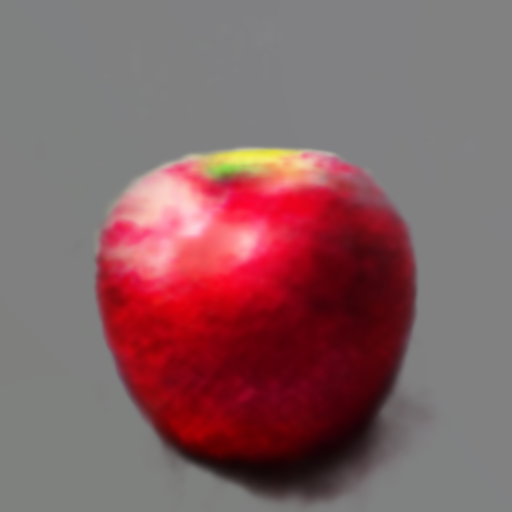}};
                        \node (p2u2) [right=of p2u1, xshift=\xdist]{\includegraphics[trim={0 0 0 0}, clip, width=\imgw]{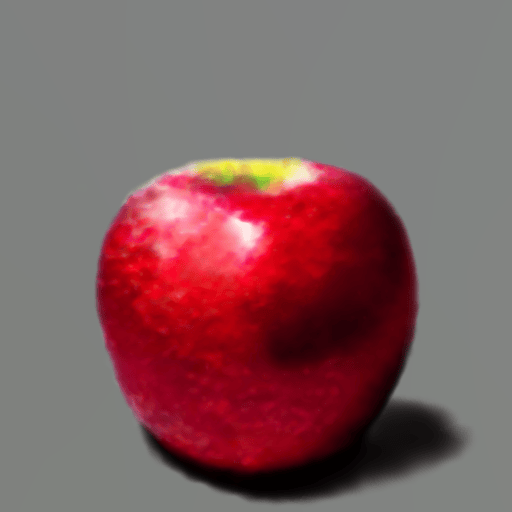}};
                        \node (p2u3) [right=of p2u2, xshift=\xdist]{\includegraphics[trim={0 0 0 0}, clip, width=\imgw]{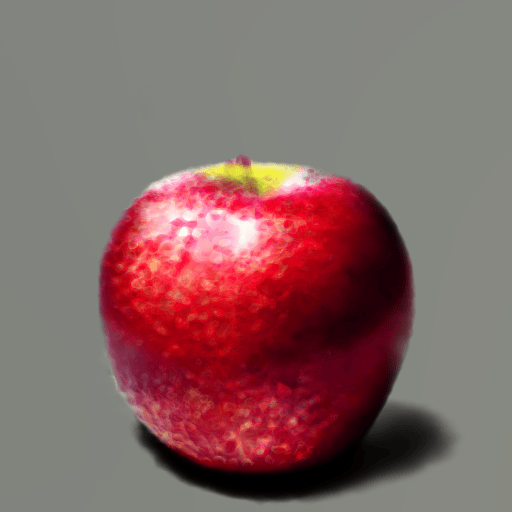}};
                        \node (p2uf) [right=of p2u3, xshift=\xdist]{\includegraphics[trim={0 0 0 0}, clip, width=\imgw]{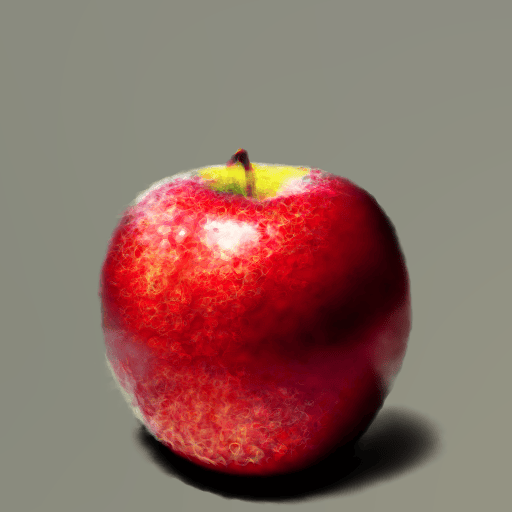}};

                        \node[above=of p1b0, yshift=-1.2cm, font=\scriptsize\color{black}] {Iter 0};
                        \node[above=of p1b1, yshift=-1.2cm, font=\scriptsize\color{black}] {Iter 500};
                        \node[above=of p1b2, yshift=-1.2cm, font=\scriptsize\color{black}] {Iter 1000};
                        \node[above=of p1b3, yshift=-1.2cm, font=\scriptsize\color{black}] {Iter 2000};
                        \node[above=of p1bf, yshift=-1.2cm, font=\scriptsize\color{black}] {Iter 4000};

                        \node[left=of p1b0, xshift=0.9cm, yshift=0.75cm, rotate=90, font=\tiny\color{black}] {Baseline (4, 1)};
                        \node[left=of p1u0, xshift=0.9cm, yshift=0.95cm, rotate=90, font=\tiny\color{black}] {IW+Strat. (1, 16)};
                        \node[left=of p2b0, xshift=0.9cm, yshift=0.75cm, rotate=90, font=\tiny\color{black}] {Baseline (4, 1)};
                        \node[left=of p2u0, xshift=0.9cm, yshift=0.95cm, rotate=90, font=\tiny\color{black}] {IW+Strat. (1, 16)};

                        \node[right=of p1bf, xshift=-0.9cm, yshift=0.75cm, rotate=270, font=\scriptsize\color{black}] {``\emph{A castle-shaped sandcastle}''};
                        \node[right=of p2bf, xshift=-0.9cm, yshift=0.25cm, rotate=270, font=\scriptsize\color{black}] {``\emph{A shiny red apple}''};
                    \end{tikzpicture}
                    }
                    \captionof{figure}{
                        \textbf{Qualitative Optimization Trajectories and Prompt Alignment:}
                        SDS renders over optimization. Baseline uses $(\numRenders,\numReNoises){=}(4,1)$; ours uses $(1,16)$, whose iterations cost less, reaching comparable quality in fewer iterations (ECM averaged over $\numRenders$ peaks at ${\sim}3.3\times$, \Tab~\ref{tab:absolute_ecm_by_m_param}). Columns: iterations 0/500/1000/2000/4000. Qualitative improvements track CLIP-score trends in \Fig~\ref{fig:clip_sds}.
                    }
                    \label{fig:qualitative_sds_trajectory}
                \end{minipage}
                \vspace{-0.00\textheight}
            \end{figure}

        \textbf{Setup:}
            We use threestudio~\citep{threestudio2023} with default hyperparameters per recent work~\citep{xie2024latte3d}. Our \emph{uniform} baseline is the standard SDS configuration used in DreamFusion~\citep{poole2022dreamfusion} and Magic3D~\citep{lin2023magic3d}: uniform timestep sampling on $[\timevar_{\min},\timevar_{\max}]$ with one diffusion draw per render ($\numReNoises\!=\!1$); our methods are drop-in modifications that preserve the SDS objective. We measure: variance of SDS-latent-space updates and parameter gradients (and related dispersion); effective compute multipliers (how much baseline compute matches our variance); and CLIP scores~\citep{hessel2021clipscore} for coarse prompt alignment. We also compare fidelity using renders at matched iterations throughout training (\Fig~\ref{fig:qualitative_sds_trajectory}; final renders in \App~\Fig~\ref{fig:qualitative_sds}). Details in \App~\Sec~\ref{sec:experiments-sds-app}.

        \textbf{Quantitative Results:}
            At $\numReNoises\!=\!8$, ECM averages over $\numRenders$ are $\sim\!2.6\times$ for reuse alone and $\sim\!3.3\times$ for IW+Strat (\Tab~\ref{tab:absolute_ecm_by_m_param}); ECM peaks at $\numReNoises\!=\!8$ and declines at $\numReNoises\!\in\!\{16,32\}$ as the across-unit term $\sigma_A^2/\numRenders$ takes over (\Sec~\ref{sec:method-compute-reuse}). \Fig~\ref{fig:quantifying_variance_hierarchical_cost_aware_iw_strat} shows that, for $\numReNoises\!\in\!\{2,4,8\}$ averaged over $\numRenders$, IW (${\sim}14$--$23\%$) and stratification (${\sim}6$--$12\%$) are complementary (${\sim}25$--$31\%$ combined); \Tabs~\ref{tab:absolute_ecm_by_m_param},~\ref{tab:relative_improvement_by_m_param} break this down. \Fig~\ref{fig:clip_sds}: across $30$ prompts and $3$ seeds, IW+Strat reaches each CLIP score of the standard-SDS baseline in fewer iterations, each costing less (\Fig~\ref{fig:quantifying_variance_hierarchical_cost_aware_iw_strat_main}).
            Per-render stratification beats global (\App~\Fig~\ref{fig:variance_ablation_stratified}); over our camera distribution (which we did not vary), the weight-based IW attains $94$--$97\%$ of the relative efficiency of an empirical oracle that keeps the Jacobian ($\cfgScale\!=\!25$; \App~\Fig~\ref{fig:sds_optimal_importance}, \Tab~\ref{tab:iw_ablation}); IW+Strat reaches ${\sim}91\%$ of the efficiency of a Sinkhorn-optimal pair-probability allocation (per-snapshot, $\numSamples\!=\!2$; \App~\Sec~\ref{app:optimal_pair_distributions}); the IW+Strat $>$ IW $\approx$ Strat $>$ Uniform ranking is stable across $5$ prompts (\App~\Tab~\ref{tab:prompt_ablation}).

            \begin{table}[t!]
                \centering
                \vspace{-0.0\textheight}
                \begin{minipage}[t]{0.55\textwidth}
                    \centering
                    \captionof{table}{Effective compute multiplier (ECM) by re-noisings $\numReNoises$. ECM is baseline cost divided by method cost at equal variance, using the uniform-IID $\numReNoises{=}1$ curve (\Sec~\ref{sec:method-variance-framework}); higher is better. Averaged over the 5 variance prompts and over $\numRenders$.}
                    \label{tab:absolute_ecm_by_m_param}
                    \vspace{0.3em}
                    \scalebox{0.7}{
                    \begin{tabular}{lrrrr}
                    \toprule
                    \numReNoises & \multicolumn{1}{c}{Uniform} & \multicolumn{1}{c}{IW} & \multicolumn{1}{c}{Strat.} & \multicolumn{1}{c}{IW+Strat.} \\
                    \midrule
                    1 & 1.00$\times$ & \textbf{1.24}$\times$ & 1.00$\times$ & \textbf{1.24}$\times$ \\
                    2 & 1.57$\times$ & 1.93$\times$ & 1.66$\times$ & \textbf{2.05}$\times$ \\
                    4 & 2.24$\times$ & 2.68$\times$ & 2.47$\times$ & \textbf{2.94}$\times$ \\
                    8 & 2.63$\times$ & 3.01$\times$ & 2.96$\times$ & \textbf{3.29}$\times$ \\
                    16 & 2.52$\times$ & 2.75$\times$ & 2.75$\times$ & \textbf{2.94}$\times$ \\
                    32 & 1.98$\times$ & 2.09$\times$ & 2.08$\times$ & \textbf{2.18}$\times$ \\
                    \bottomrule
                    \end{tabular}
                    }
                \end{minipage}\hfill
                \begin{minipage}[t]{0.41\textwidth}
                    \centering
                    \captionof{table}{Relative Efficiency (RE) versus $(\numRenders, \numReNoises)$-uniform at each $\numReNoises$, averaged over $\numRenders$. IW and stratification are complementary.}
                    \label{tab:relative_improvement_by_m_param}
                    \vspace{0.3em}
                    \scalebox{0.7}{
                    \begin{tabular}{lrrrr}
                    \toprule
                    \numReNoises & \multicolumn{1}{c}{Uniform} & \multicolumn{1}{c}{IW} & \multicolumn{1}{c}{Strat.} & \multicolumn{1}{c}{IW+Strat.} \\
                    \midrule
                    1 & 1.00$\times$ & \textbf{1.24}$\times$ & 1.00$\times$ & \textbf{1.24}$\times$ \\
                    2 & 1.00$\times$ & 1.23$\times$ & 1.06$\times$ & \textbf{1.30}$\times$ \\
                    4 & 1.00$\times$ & 1.20$\times$ & 1.11$\times$ & \textbf{1.31}$\times$ \\
                    8 & 1.00$\times$ & 1.14$\times$ & 1.12$\times$ & \textbf{1.25}$\times$ \\
                    16 & 1.00$\times$ & 1.09$\times$ & 1.09$\times$ & \textbf{1.17}$\times$ \\
                    32 & 1.00$\times$ & 1.05$\times$ & 1.05$\times$ & \textbf{1.10}$\times$ \\
                    \bottomrule
                    \end{tabular}
                    }
                \end{minipage}
            \end{table}

    \subsection{Single-Step Diffusion Distillation}\label{sec:experiments-dmd}
        \textbf{Setup:}
            We apply our methods to Distribution Matching Distillation (DMD) \cite{yin2024one}, in its DMD2 form \citep{yin2024improved}, via the Monte Carlo estimator of \Eq~\ref{eq:dmd_mc_estimator}, on top of the FastGen~\citep{fastgen2026} reference implementation. We train generators on ImageNet-256~\citep{deng2009imagenet} using the pretrained DiT-XL/2 teacher~\citep{peebles2023scalable}; details in \App~\Sec~\ref{sec:experiments-dmd-app}. The measured cost ratio is $\costRender/\costDenoise\!\approx\!26$, so upstream cost dominates (\Sec~\ref{sec:method-compute-reuse}).

        \textbf{Results:}
            \Tab~\ref{tab:dmd_variance} shows reuse cuts gradient variance by $3.4$--$16\times$ at fixed $\numRenders\!=\!8$, with ${\sim}1.55\times$ per-iteration wall-clock cost (\App~\Sec~\ref{sec:experiments-dmd-app}). Stratification adds $1.0$--$2.0\times$ \emph{at matched compute} within $(8,16)$. The largest variance reduction is in parameter gradients, where combining reuse $(8,16)$ with stratification yields ${\sim}32\times$ over baseline $(8,1)$ (${\sim}20\times$ variance reduction per unit wall-clock). While variance reduction yields similar-or-better per-step FID convergence, no practical improvement remains at matched wall-clock time (\App~\Fig~\ref{fig:student_fid_combined}, \App~\Fig~\ref{fig:dmd2_frontier}). We report DMD as a negative result and treat it as a scope condition of the framework: lower gradient variance does not improve FID at matched wall-clock time (condition (3) in \Sec~\ref{sec:discussion}). \App~\Sec~\ref{sec:experiments-dmd-disconnect} rules out insufficient compute headroom; generator-input diversity is most consistent with the data, while auxiliary-objective effects are plausible but not isolated. Applying reuse and stratification to the fake-score update likewise improves per-step convergence at higher compute, and the benefit fades once adjusted for compute (\App~\Fig~\ref{fig:fake_score_best_fid}). Detailed hypotheses and ablations are in \App~\Sec~\ref{sec:experiments-dmd-app}.

            \begin{table}[t!]
                \centering
                \begin{minipage}[t]{0.58\textwidth}
                    \centering
                    \captionof{table}{DMD variance at iter.\ $20$k: variance (\Eq~\ref{eq:var}) of the Monte Carlo estimates of the teacher score, the score difference, and the parameter gradient. \emph{Reuse} ($(8,16)$ vs $(8,1)$) reduces variance $3.4$--$16\times$ \emph{at higher compute} ($16\times$ more denoiser calls). \emph{Stratification} ($(8,16)$ Strat.\ vs IID) reduces variance $1$--$2\times$ at \emph{matched} compute. FID does not improve at matched wall-clock.}
                    \label{tab:dmd_variance}
                \end{minipage}\hfill
                \begin{minipage}[t]{0.40\textwidth}
                    \centering
                    \captionof{table}{IID vs.\ stratified sampling for data-attribution gradients. Correlations are measured against the full-budget attribution, which stratified sampling reproduces at $768$ timesteps (\App~\Sec~\ref{sec:experiments-motive-app-results}); stratified sampling gives $>\!2\times$ compute multipliers at $64$ to $256$ timesteps.}
                    \label{tab:motive_gradient_estimation_summary}
                \end{minipage}\par
                \nointerlineskip\vspace{0.3em}
                \begin{minipage}[t]{0.58\textwidth}
                    \centering
                    \setlength{\tabcolsep}{4pt}
                    \scalebox{0.78}{
                    \begin{tabular}[t]{@{}lccc@{}}
                    \toprule
                    \textbf{Method} & \textbf{Teacher} & \textbf{Score} & \textbf{Param.} \\
                                    & \textbf{Score Var} & \textbf{Diff.\ Var} & \textbf{Grad.\ Var} \\
                    \midrule
                    $(8, 1)$ IID     & 3954 & 71.3  & 982 \\
                    $(8, 16)$ IID    & 1175 & 13.3  & 59.9 \\
                    $(8, 16)$ Strat. & 1169 & 12.0  & 30.7 \\
                    \bottomrule
                    \end{tabular}
                    }
                \end{minipage}\hfill
                \begin{minipage}[t]{0.40\textwidth}
                    \centering
                    \setlength{\tabcolsep}{4pt}
                    \scalebox{0.78}{
                    \begin{tabular}[t]{@{}rccr@{}}
                    \toprule
                    \textbf{Budget} & \textbf{IID} & \textbf{Stratified} & \textbf{Effective} \\
                    \textbf{(timesteps)} & \textbf{Corr.} & \textbf{Corr.} & \textbf{Multiplier} \\
                    \midrule
                      4 & 0.065 & 0.166 & 2.44$\times$ \\
                     16 & 0.245 & 0.407 & 1.80$\times$ \\
                     64 & 0.616 & 0.805 & 3.82$\times$ \\
                    256 & 0.813 & 0.916 & 2.85$\times$ \\
                    768 & 0.934 & 1.000 & 1.28$\times$ \\
                    \bottomrule
                    \end{tabular}
                    }
                \end{minipage}
            \end{table}

    \subsection{Data Attribution}\label{sec:experiments-motive}
        \textbf{Setup:}
            We follow MOTIVE~\citep{wu2026motion} for video data attribution, using \texttt{Wan2.1-T2V-1.3B}~\citep{wan2025wan}, a flow-matching video model that illustrates our framework's reach beyond noise-prediction diffusion teachers (DiffSynth-Studio implementation), on \texttt{VIDGEN-1M}~\citep{tan2024vidgen}; details in \App~\Sec~\ref{sec:experiments-motive-app}. Data attribution computes the influence of each training datum on a query, which MOTIVE uses to select fine-tuning data. We assess unbiased gradient estimators via gradient variance and the correlation between the rankings they produce and ground-truth rankings.

        \textbf{Results:}
            \Fig~\ref{fig:quantifying_attribution} quantifies gradient variance (top) and ranking correlation (bottom) versus gradient evaluations per data point. Variance decreases with more samples, and stratified sampling consistently beats uniform at equal budget. This yields influence rankings that better match those computed from ground-truth gradients, achieving $>\!2\times$ effective compute multipliers at moderate budgets of $64$ to $256$ timesteps (\Tab~\ref{tab:motive_gradient_estimation_summary}). Re-noising provides less benefit here than in other tasks because encoding cost is moderate relative to denoising, and we require accurate gradients for each fixed training example rather than averaging over sampled inputs. Global stratification is most effective and substantially reduces variance in this setting.

\section{Discussion}\label{sec:discussion}
    The \App~covers extended related work (\App~\Sec~\ref{app:related}), limitations (\App~\Sec~\ref{sec:limitations}), and future directions (\App~\Sec~\ref{sec:future_directions}). To our knowledge, no published work in the cited frozen-teacher SDS, DMD, or attribution lines applies timestep stratification, uses explicit per-loss weights as IS proxies, or measures parameter-gradient variance per unit compute for these tasks.

    \textbf{When Variance Reduction Helps:}
        \ourMethod{} helps when (1) the MC gradient dominates, (2) $\costRender > \costDenoise$, and (3) variance limits convergence. Conditions (1) and (2) can be checked from cost measurements and a short pilot run (\Sec~\ref{sec:method-variance-framework}); DMD, the case that tests (3), needed training runs. A cost-ratio sweep confirms the $\numReNoises^\star$ predictions of \Sec~\ref{sec:method-compute-reuse} and shows that condition (2) matters for re-noising only: at zero render cost, $\numReNoises\!=\!1$ is best for every method, and IW+Strat stays above the baseline at small $\numReNoises$ because of IW (\App~\Sec~\ref{sec:compute_cost_sensitivity}). In SDS, IW+Strat reaches ${\sim}91\%$ of the efficiency of a Sinkhorn-optimized pair allocation (\App~\Sec~\ref{app:optimal_pair_distributions}, \App~\Fig~\ref{fig:pair_prob_comparison}). At low classifier-free guidance, $\cfgScale\!=\!25$, IW+Strat reaches an ECM of $\sim\!3.8\times$ at $(1,8)$ (\App~\Fig~\ref{fig:quantifying_variance_low_guidance}), with CLIP-score and qualitative gains across prompts and seeds (\App~\Sec~\ref{sec:low_guidance_analysis}). Cost selection (thresholds in \App~\Sec~\ref{app:practitioner-guide}): reuse when $\costRender\!\gg\!\costDenoise$ (${\sim}1.6$--$2.6\times$ in SDS), stratification when $\costRender\!\approx\!\costDenoise$ (${\sim}1.3$--$3.8\times$ in attribution), and IS when a known weight dominates how the gradient norm varies with $\timevar$ (${\sim}1.2\times$).

    \textbf{Broader Implications:}
        The framework (\Sec~\ref{sec:method-variance-framework}) is application-agnostic wherever upstream cost dominates the denoiser; the DMD case suggests gradient-variance reduction does not improve FID at matched wall-clock time when auxiliary stabilizers or input diversity bind.

    \subsection{Conclusion}\label{sec:conclusion}
        We presented \ourMethod{}, a compute-aware variance-accounting framework for frozen-teacher Monte Carlo gradients, motivating a hierarchical Monte Carlo estimator with three unbiased drop-ins: timestep importance sampling, stratification, and amortized compute reuse. In our SDS and attribution settings, \ourMethod{} delivers $2$ to $3\times$ effective compute multipliers without changing the objective; in DMD, the same techniques cut gradient variance by an order of magnitude without improving downstream FID, marking the boundary where MC variance no longer governs convergence.

\newpage
\begin{ack}
    We thank the \href{https://research.nvidia.com/labs/genair/}{Fundamental Generative AI Research (GenAIR)} group at NVIDIA for many helpful discussions during this work. We thank Sanja Fidler and the \href{https://research.nvidia.com/labs/sil/}{Spatial Intelligence Lab (SIL)} for hosting the internship that made this collaboration possible.
    This work was funded by NVIDIA; all authors were affiliated with NVIDIA during this work. Jesse Bettencourt also acknowledges support from the University of Toronto and the Vector Institute, and Xindi Wu from Princeton University. Resources used in preparing this research were provided, in part, by the Province of Ontario, the Government of Canada through CIFAR, and companies sponsoring the Vector Institute. The authors declare no other competing interests.
\end{ack}

\bibliographystyle{plainnat}
\bibliography{bib}

\appendix

\section{Broader Impacts}\label{app:broader_impacts}
    Our work improves compute efficiency of pipelines that use pretrained diffusion models as frozen teachers by reducing Monte Carlo estimator variance without changing the target objective. By reallocating samples across noise levels, stratifying timesteps, and reusing expensive upstream computations, practitioners can achieve comparable gradient quality with fewer denoiser, rendering, or encoding evaluations, which reduces energy use and the cost of experiments. The same techniques could also lower the cost of developing or deploying systems that generate synthetic media, which may be misused for deception or harmful content. This paper does not introduce new generative capabilities or datasets, and responsible use should follow the safety, provenance, and content policies of the underlying models. Measuring variance per unit of compute also makes comparisons between estimators fairer under a fixed budget.

\subsection*{LLM Usage}
    We used a large language model as a writing and engineering assistant during the preparation of this manuscript. Specifically, it was used to (i) suggest edits for clarity and concision, (ii) help reorganize prose and LaTeX for readability, and (iii) assist with routine coding tasks (e.g., debugging scripts and preparing plotting utilities). All technical contributions, methodological decisions, experimental design, and results are our own. We verified all generated suggestions and did not rely on the model for new scientific claims or conclusions.

\subsection*{Reproducibility}
    We take several steps to ensure reproducibility of our results. First, we provide complete mathematical specifications of all proposed estimators, including importance sampling (\Sec~\ref{sec:method-noise-schedules}), stratified sampling (\Sec~\ref{sec:method-stratified-sampling}), and compute reuse (\Sec~\ref{sec:method-compute-reuse}), with explicit equations and a combined-pipeline pseudocode (\Algo~\ref{alg:combined}) that can be directly implemented. Second, we build on established open-source codebases: threestudio~\citep{threestudio2023} for SDS (\Sec~\ref{sec:experiments-sds}), the FastGen~\citep{fastgen2026} reference implementation for DMD (\Sec~\ref{sec:experiments-dmd}), and MOTIVE~\citep{wu2026motion} (DiffSynth-Studio backbone) for video data attribution (\Sec~\ref{sec:experiments-motive}); each of our estimators in \Sec~\ref{sec:method-variance-reduction-strategies} is a small change to the per-step sampling and re-noising logic in these pipelines, not a structural change to the surrounding training loop. Third, we report all experimental settings, including batch sizes, compute budgets, number of renders and re-noises, and evaluation metrics, with additional hyperparameters and implementation details provided in \App~\Sec~\ref{app:sec_experiments}. Fourth, our experiments average results over multiple seeds and prompts and report standard deviations to quantify uncertainty (\Fig~\ref{fig:clip_sds}). Fifth, our variance measurement framework (\Sec~\ref{sec:method-variance-framework}) uses standard techniques (Welford's algorithm) and verifies unbiasedness by comparing the MSE and variance, enabling independent validation of our claims. A glossary of all notation is included in \Sec~\ref{app_sec_notation}.

\section{Additional Background}\label{app:sec_background}
    \subsection{Diffusion Models}\label{app:sec_diffusion_models}
        \textbf{Weighted diffusion training objective.}
            Let $\noisePred(\noisedData,\timevar,\textCond)$ be the predicted noise at $(\noisedData,\timevar,\textCond)$ with parameters $\denParams$. A common noise-prediction objective is:\vspace{-0.005\textheight}
            \begin{equation}\label{eq:weighted_diff_loss}
                \smash{\lossDiffusion(\denParams) \!=\! \Eop_{(\dataSample,\textCond),\timevar,\noiseVec}\!\big[\|\residual\|_2^2\big] \!=\! \Eop_{(\dataSample,\textCond),\timevar,\noiseVec}\!\big[\costDiffusion(\encoder(\dataSample),\textCond,\timevar,\noiseVec,\denParams)\big]}
            \end{equation}
            where the residual $\residual \defeq \noisePred(\noisedData,\timevar,\textCond)-\noiseVec$ and the per-sample cost $\costDiffusion \defeq \|\residual\|^2_2$. Following \citet{kingma2023variational}, we use the weighted objective:\vspace{-0.003\textheight}
            \begin{equation}\label{eq:weighted_diffusion_objective}
                \!\!\lossWeighted(\denParams)
                \!=\!\!\!\Eop_{(\dataSample,\textCond),\timevar,\noiseVec}\!\!
                \big[\!\weight(\timevar)\costDiffusion(\encoder(\dataSample),\textCond,\timevar,\noiseVec,\denParams)\!\big].\vspace{-0.003\textheight}
            \end{equation}
            Different prediction parameterizations (e.g., $x$- or $v$-prediction) can be written in this form with an appropriate choice of $\weight(\timevar)$ and the corresponding model output \cite{kingma2023variational, salimans2022progressive}.
            
        \subsubsection{Sampling from Diffusion Models} \label{sec:background-diffusion-models-sampling-app}
            To sample, a pretrained latent diffusion model uses a multi-step sampler that starts from Gaussian noise and iteratively denoises. Let $\{\timevar_\sampleStep\}_{\sampleStep=0}^{\numSampleStep}$ denote a discretization of the continuous noise schedule with $\timevar_0 \approx 0$ and $\timevar_\numSampleStep \approx 1$. The sampler initializes the latent at the highest noise level, $\encodedData_{\timevar_\numSampleStep} \sim \standardNormal$, then applies a sequence of learned transitions:
            \begin{equation}
                \encodedData_{\timevar_{\sampleStep-1}} \sim p_{\denParams}(\encodedData_{\timevar_{\sampleStep-1}} \mid \encodedData_{\timevar_\sampleStep}, \timevar_\sampleStep, \textCond), \qquad \sampleStep = \numSampleStep,\dots,1
            \end{equation}
            where the transition kernels are determined by the denoiser $\noisePred(\encodedData_{\timevar_\sampleStep}, \timevar_\sampleStep, \textCond)$ and the chosen update rule (e.g., DDPM or a DDIM-like update). The composition of these $\numSampleStep$ steps defines a stochastic generator that maps a single Gaussian seed $\encodedData_{\timevar_\numSampleStep}\sim\standardNormal$ to a clean latent $\encodedData_{\timevar_0}$. In \App~\Sec~\ref{sec:background-single-step-distillation-app}, we treat this $\numSampleStep$-step procedure as the teacher and train a one-step generator $\generator$ to match its sample distribution in a single forward pass from noise.

            \textbf{Classifier-free guidance.}
                Many text-conditioned diffusion models are trained with a classifier-free guidance setup, where the conditioning $\textCond$ is randomly dropped during training so that a single network learns both conditional and unconditional predictions. At sampling time, the model is evaluated in both modes and its outputs are combined using a scalar guidance weight $\cfgScale \ge 0$. Using our denoiser notation, the guided noise prediction is:
                \begin{equation}
                    \noisePred(\noisedData, \timevar, \textCond; \cfgScale)
                    \!=\! (1 \!+\! \cfgScale)\noisePred(\noisedData, \timevar, \textCond)
                      \!-\! \cfgScale\noisePred(\noisedData, \timevar, \textCond \!=\! \varnothing)
                \end{equation}
                where $\noisePred(\noisedData, \timevar, \textCond)$ and $\noisePred(\noisedData, \timevar, \varnothing)$ denote the conditional and unconditional outputs of the same network. The guided prediction $\noisePred$ is used in the sampler to update $\encodedData_{\timevar_\sampleStep}$ and appears in the SDS gradients in \Sec~\ref{sec:background-sds} and \Sec~\ref{sec:experiments-sds}.

    \subsection{Reducing Estimator Variance}\label{sec:background-reducing-estimator-variance-app}
        
        \subsubsection{Importance Sampling Theory and Application to Diffusion}\label{sec:background-iw-app}
            We expand the importance-sampling treatment of \Sec~\ref{sec:background-reducing-variance-importance_sampling}.
            
            \textbf{Setup.}
                Let $\timevar\in[0,1]$ be a noise level with base density $p(\timevar)$. Let $\randomness$ denote all other randomness (e.g., sampled input data, Gaussian noise) drawn from a conditional distribution $p(\randomness\mid\timevar)$. For a vector-valued contribution $\gterm(\timevar,\randomness)$ such as a training gradient, define the conditional mean integrand:
                \begin{equation}
                    \testFunc(\timevar)=\E[\gterm(\timevar,\randomness)\mid\timevar]
                \end{equation}
                and its mean:
                \begin{equation}
                  \mean=\E_{\timevar\sim p,\randomness\sim p(\cdot\mid\timevar)}[\gterm(\timevar,\randomness)] = \E_{\timevar\sim p}[\testFunc(\timevar)]
                \end{equation}
            
            \textbf{Importance sampling estimator.}
                For any proposal density $\proposalDensity(\timevar)$ with $\proposalDensity(\timevar)>0$ whenever $p(\timevar)>0$, define the importance weight $\importanceWeight(\timevar)=\frac{p(\timevar)}{\proposalDensity(\timevar)}$ and sample $\timevar^{(\sampleIndex)}\sim\proposalDensity$, $\randomness^{(\sampleIndex)}\sim p(\cdot\mid\timevar^{(\sampleIndex)})$. Then the following is an unbiased estimator for $\mean$:
                \begin{equation}
                    \hat\mean_{\proposalDensity}=\frac{1}{\numSamples}\sum_{\sampleIndex=1}^{\numSamples}\importanceWeight(\timevar^{(\sampleIndex)})\gterm(\timevar^{(\sampleIndex)},\randomness^{(\sampleIndex)})
                \end{equation}
            
            \textbf{Variance and optimal proposals.}
                A direct calculation gives the trace-covariance dispersion
                \begin{equation}
                    \mathrm{tr}(\mathrm{Cov}(\hat\mean_{\proposalDensity}))=\frac{1}{\numSamples}\left(\int \frac{p(\timevar)^2}{\proposalDensity(\timevar)}\E[\|\gterm(\timevar,\randomness)\|_2^2\mid\timevar]\diffd\timevar-\|\mean\|_2^2\right)
                \end{equation}
                which implies the variance-minimizing proposal under this criterion
                \begin{align}\label{eq:opt_proposal}
                    \proposalDensity^\star(\timevar) &\propto p(\timevar)\sqrt{\E[\|\gterm(\timevar,\randomness)\|_2^2\mid\timevar]} \\
                    &= p(\timevar)\sqrt{\|\testFunc(\timevar)\|_2^2+\mathrm{tr}(\mathrm{Cov}(\gterm(\timevar,\randomness)\mid\timevar))};
                \end{align}
                see standard treatments of optimal importance sampling \cite{rubinstein2016simulation}. If $\gterm(\timevar,\randomness)$ is deterministic given $\timevar$, then $\proposalDensity^\star(\timevar)\propto p(\timevar)\|\testFunc(\timevar)\|_2$. For a scalar integrand, this reduces to the familiar form $\proposalDensity^\star(\timevar)\propto p(\timevar)|\testFunc(\timevar)|$. For vector-valued $\testFunc$, even the oracle proposal typically does not yield zero variance because the contribution direction can vary with $\timevar$. Intuitively, importance sampling reallocates samples toward noise levels with large root-mean-square contributions and away from those with small ones.
            
            \textbf{Loss-based proxies and their limitations.}
                Evaluating $\|\testFunc(\timevar)\|_2$ in diffusion is prohibitively expensive. \citet{nichol2021improved} instead set timestep probabilities from per-timestep loss statistics, and \citet{zheng2024non} use a fixed non-uniform schedule derived from an optimal-transport analysis.
            
                Concretely, consider the per-timestep denoising loss from the weighted diffusion objective in \Eq~\ref{eq:weighted_diffusion_objective}. For diffusion model training, the gradient with respect to denoiser parameters $\denParams$ is
                \begin{align}
                    \gterm(\timevar,\randomness) \defeq \nabla_{\denParams}\cost
                    =2\jacobian_{\denParams}^\top\residual
                    \textnormal{ where }
                    \jacobian_{\denParams}=\nabla_{\denParams}\noisePred(\noisedData,\timevar,\textCond)
                \end{align}
                Recall that the optimal importance sampling proposal from \Eq~\ref{eq:opt_proposal} requires $\sqrt{\E[\|\gterm(\timevar,\randomness)\|_2^2\mid\timevar]}$, which equals:
                \begin{align}
                    \sqrt{\E[\|\nabla_{\denParams}\cost\|_2^2\mid\timevar]}
                    =\sqrt{\|\E[\nabla_{\denParams}\cost\mid\timevar]\|_2^2 + \mathrm{tr}(\mathrm{Cov}(\nabla_{\denParams}\cost\mid\timevar))}
                \end{align}
                Since estimating this is expensive, loss statistics serve as a cheap proxy, such as $\sqrt{\E[\cost\mid\timevar]}=\sqrt{\E[\|\residual\|_2^2\mid\timevar]}$; \citet{nichol2021improved} use the root mean square of the per-timestep loss. However, for any single sample
                \begin{align}
                    \|\nabla_{\denParams}\cost\|_2^2
                    =4\|\jacobian_{\denParams}^\top\residual\|_2^2 
                    \leq 4\|\jacobian_{\denParams}\|_2^2\|\residual\|_2^2
                    =4\|\jacobian_{\denParams}\|_2^2\cost
                \end{align}
                
                Equality holds when $\residual$ aligns with the leading left singular vector of $\jacobian_{\denParams}$. The loss proxy $\cost$ captures only $\|\residual\|_2^2$ and ignores $\|\jacobian_{\denParams}\|_2^2$, so a loss-derived schedule can misrank timesteps when $\|\jacobian_{\denParams}\|_2$ varies with $\timevar$ or correlates weakly with $\|\residual\|_2$.
            
            \textbf{The gap in SDS-style optimization.}
                This gap is amplified in the settings we focus on, where the gradient of interest is not with respect to denoiser parameters. For example, in SDS-style optimization (detailed in \Sec~\ref{sec:background-sds}), we optimize parameters $\genParams$ of a differentiable generator using a frozen diffusion teacher. The per-timestep update takes the form:
                \begin{align}
                    \gterm(\timevar,\randomness) &= \sdsWeight(\timevar)\jacobian_{\genParams}^\top\residual \textnormal{ where }\\
                    \jacobian_{\genParams} &=\frac{\partial\encodedData}{\partial\genParams}=\frac{\partial\encoder}{\partial\dataSample}\frac{\partial\prerender}{\partial\genParams}
                \end{align}
                for the known scalar weight $\sdsWeight(\timevar)$ of \Eq~\ref{eq:sds_update}. Here $\prerender(\genParams)$ is the generator output (e.g., a rendered image) and $\encoder$ maps it to latent space; SDS omits the teacher input Jacobian $\nabla_{\encodedData}\noisePred$ from this product. The effective Jacobian $\jacobian_{\genParams}$ thus includes a potentially ill-conditioned encoder-generator chain whose timestep dependence can differ from $\|\residual\|_2$. Combined with $\weight(\timevar)$, this leaves room for proposals that target gradient contributions rather than the loss proxy.

            \textbf{Stratified-IS unbiasedness (\Eq~\ref{eq:stratified-IS}).}
                We verify that the construction in \Sec~\ref{sec:method-stratified-sampling} yields an unbiased estimator of $\E_{\timevar\sim p}[\testFunc(\timevar)]$. Let $F_{\proposalDensity}$ be the CDF of the importance proposal $\proposalDensity$ and partition $[0,1]$ into $\numStrata$ equal-mass strata in $\proposalDensity$-quantile space, $\stratum_{\stratumIndex}=\{\timevar : F_{\proposalDensity}(\timevar)\in[(\stratumIndex{-}1)/\numStrata,\stratumIndex/\numStrata]\}$, so that $\Pr_{\timevar\sim\proposalDensity}[\timevar\in\stratum_{\stratumIndex}]=1/\numStrata$ and the conditional density is $\proposalDensity(\timevar\mid\stratum_{\stratumIndex})=\numStrata\,\proposalDensity(\timevar)\mathbf{1}[\timevar\in\stratum_{\stratumIndex}]$. Drawing $\randomness_{\stratumIndex}\sim\Uniform(0,1)$ and setting $\timevar_{\stratumIndex}=F_{\proposalDensity}^{-1}((\stratumIndex{-}1+\randomness_{\stratumIndex})/\numStrata)$ gives $\timevar_{\stratumIndex}$ distributed according to $\proposalDensity(\cdot\mid\stratum_{\stratumIndex})$. With importance weight $\importanceWeight(\timevar)=p(\timevar)/\proposalDensity(\timevar)$ and contribution $\gterm(\timevar_{\stratumIndex},\randomness'_{\stratumIndex})$ for $\randomness'_{\stratumIndex}\sim p(\cdot\mid\timevar_{\stratumIndex})$, the per-stratum expectation is
                \begin{align}
                    \E\big[\importanceWeight(\timevar_{\stratumIndex})\,\gterm(\timevar_{\stratumIndex},\randomness'_{\stratumIndex})\big]
                    = \int_{\stratum_{\stratumIndex}}\!\frac{p(\timevar)}{\proposalDensity(\timevar)}\,\testFunc(\timevar)\,\numStrata\,\proposalDensity(\timevar)\diffd\timevar
                    = \numStrata\!\int_{\stratum_{\stratumIndex}}\!p(\timevar)\,\testFunc(\timevar)\diffd\timevar.
                \end{align}
                Averaging over the $\numStrata$ strata,
                \begin{align}
                    \E\Big[\tfrac{1}{\numStrata}\!\sum_{\stratumIndex=1}^{\numStrata}\importanceWeight(\timevar_{\stratumIndex})\,\gterm(\timevar_{\stratumIndex},\randomness'_{\stratumIndex})\Big]
                    = \sum_{\stratumIndex=1}^{\numStrata}\int_{\stratum_{\stratumIndex}}\!p(\timevar)\,\testFunc(\timevar)\diffd\timevar
                    = \int_0^1 p(\timevar)\,\testFunc(\timevar)\diffd\timevar
                    = \E_{\timevar\sim p}[\testFunc(\timevar)],
                \end{align}
                so \Eq~\ref{eq:stratified-IS} is unbiased for $\E_{\timevar\sim p}[\testFunc(\timevar)]$ for any proposal $\proposalDensity$ that satisfies $\proposalDensity(\timevar){>}0$ wherever $p(\timevar){>}0$. The variance-reduction argument follows from the standard stratified-sampling decomposition: the variance of the per-stratum-averaged estimator equals $\tfrac{1}{\numStrata}$ times the average within-stratum conditional variance (dropping the between-stratum component carried by simple Monte Carlo; \citet{thompson2012sampling}), composed with the usual importance-reweighting variance formula.
        
\begin{figure}[t]
                \centering
                \begin{minipage}[c]{0.5\textwidth}
                \centering
                \resizebox{\linewidth}{!}{%
                \begin{tikzpicture}
                \centering
                    \node (img11){\includegraphics[trim={0.0cm 0.3cm 0.0cm 1.2cm}, clip, width=.89\linewidth]{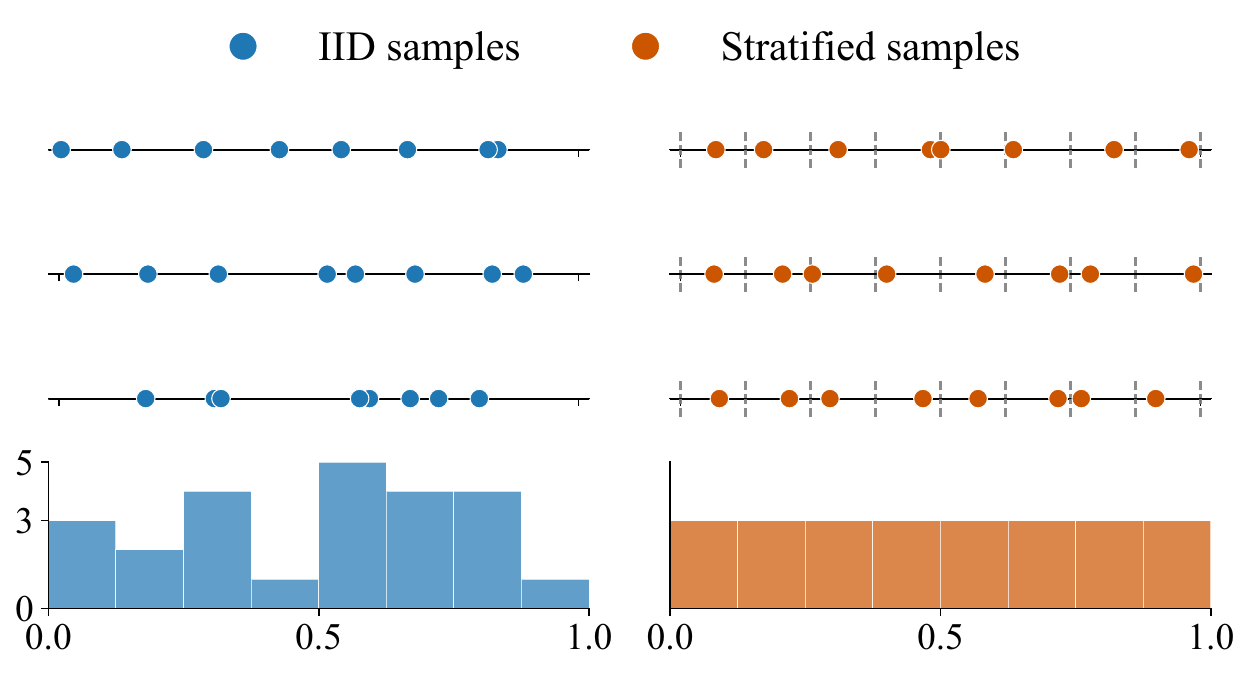}};
                    \node[left=of img11, node distance=0cm, rotate=90, xshift=1.55cm, yshift=-.8cm,  font=\color{black}]{\small{Realizations}};
                    \node[left=of img11, node distance=0cm, rotate=90, xshift=-0.2cm, yshift=-.8cm,  font=\color{black}]{\small{Total Counts}};
                    \node[below=of img11, node distance=0cm, xshift=-1.8cm, yshift=1.2cm,  font=\color{black}]{\normalsize{Timestep $\timevar$}};
                    \node[below=of img11, node distance=0cm, xshift=1.9cm, yshift=1.2cm,  font=\color{black}]{\normalsize{Timestep $\timevar$}};
                    \node[above=of img11, node distance=0cm, xshift=-1.7cm, yshift=-1.55cm,  font=\color{black}]{\normalsize{IID Sampling}};
                    \node[above=of img11, node distance=0cm, xshift=1.9cm, yshift=-1.55cm,  font=\color{black}]{\normalsize{Stratified Sampling}};
                \end{tikzpicture}
                }
                \end{minipage}\hfill
                \begin{minipage}[c]{0.48\textwidth}
                \captionof{figure}{
                    \textbf{Stratified Sampling Visualization:}
                    We show $3$ realizations/batches of $8$ timestep samples for both IID and stratified sampling. For stratified sampling, we create a bin for each sample and require each batch to contain one sample from each bin. This often gives lower-variance estimators.
                }\label{fig:stratifiedVisualization}
                \end{minipage}
                \vspace{-0.008\textheight}
            \end{figure}

            \Fig~\ref{fig:stratifiedVisualization} visualizes the stratified sampling of \Sec~\ref{sec:background-reducing-variance-stratified}.

        \subsubsection{Two-Stage Variance Decomposition and Cost-Optimal Allocation}\label{app:two-stage}
            \Sec~\ref{sec:method-variance-reduction-strategies} derives \Eq~\ref{eq:two-level} from the law of total variance, and \Sec~\ref{sec:method-compute-reuse} substitutes the budget to obtain \Eq~\ref{eq:budget-variance} and its minimizer $\numReNoises^\star$. Given $\dataSample$, $\hat{\mathbf{h}}^{(\renderIndex)}$ has mean $\E[\mathbf{h}\mid\dataSample]$ and variance $\mathrm{Var}(\mathbf{h}\mid\dataSample)/\numReNoises$, so $\mathrm{Var}(\hat{\mathbf{h}}^{(\renderIndex)})=\mathrm{Var}_{\dataSample}(\E[\mathbf{h}\mid\dataSample])+\E_{\dataSample}[\mathrm{Var}(\mathbf{h}\mid\dataSample)]/\numReNoises=\sigma_A^2+\sigma_W^2/\numReNoises$. The estimator is unbiased, $\E[\hat{\nabla}]=\E_{\dataSample}\E[\mathbf{h}\mid\dataSample]$, because each subunit draw has the target law given $\dataSample$; independence across subunits is what gives the $\sigma_W^2/\numReNoises$ term. \Eq~\ref{eq:budget-variance} is convex in $\numReNoises>0$, since its second derivative $2\sigma_W^2\costRender/(\budget\numReNoises^3)$ is nonnegative (positive when $\sigma_W^2\costRender>0$), and minimizing cost at fixed variance gives the same $\numReNoises^\star$ \citep{cochran1977sampling}. As $\costRender\to0$, $\numReNoises^\star\to0$ and $\numReNoises=1$ is best, and $\numReNoises^\star$ grows as $\sqrt{\costRender/\costDenoise}$. These statements assume IID subunits: under per-render stratification the within-unit term is at most $\sigma_W^2/\numReNoises$ and usually falls faster, which shifts the optimum toward larger $\numReNoises$, although $\numReNoises\!=\!1$ remains best for every method in our zero-render-cost sweep (\App~\Sec~\ref{sec:compute_cost_sensitivity}).

            Importance sampling and stratification of the subunits change only $\sigma_W^2$. With proposal $\proposalDensity$ and weight $p/\proposalDensity$, it becomes the reweighted within-unit variance (\App~\Sec~\ref{sec:background-iw-app}, applied conditionally on $\dataSample$); with one draw from each of $\numReNoises$ equal-mass strata of $\proposalDensity$, it becomes the average within-stratum variance. The IID structure across $\renderIndex$, and so the form of \Eq~\ref{eq:two-level}, is unchanged, which is why the gains compose.

        \subsubsection{Diffusion Model Noise Schedules}\label{sec:background-diffusion-models-noise-schedule-app}
            We provide additional details on the connection between noise schedules and importance sampling in diffusion training, expanding on \Sec~\ref{sec:background-reducing-variance-importance_sampling}.
            
            Following \citet{kingma2023variational}, we view the noise schedule as a differentiable, strictly decreasing function
            \begin{equation}
                \logsnr = \logsnrSchedule(\timevar), \qquad \timevar \in [0,1]
            \end{equation}
            that maps continuous time to log signal-to-noise ratio $\logsnr = \log(\signalcoef^2/\noisecoeff^2)$. Monotonicity ensures invertibility, so there is a bijection between time and logSNR.
            
            \textbf{Induced distribution over noise levels.}
                When sampling time uniformly $\timevar \sim \uniformOnUnit$ and evaluating $\logsnr = \logsnrSchedule(\timevar)$, the change-of-variables formula gives a distribution over noise levels
                \begin{equation}
                    p(\logsnr) = \left|\frac{\diffd\logsnr}{\diffd\timevar}\right|^{-1}
                \end{equation}
                where the absolute value accounts for the fact that $\logsnr$ decreases with $\timevar$. Different schedules (linear, cosine, learned) induce different distributions $p(\logsnr)$ even though all sample $\timevar$ uniformly.
            
            \textbf{Noise schedules as importance sampling.}
                If the timestep weight carries the schedule's Jacobian, $\weight(\timevar)=\tfrac{1}{2}\weight(\logsnr)\,|\diffd\logsnr/\diffd\timevar|$ with $\logsnr=\logsnrSchedule(\timevar)$ (the form of \citet{kingma2023variational}), changing variables from $\timevar$ to $\logsnr$ in the weighted objective from \Eq~\ref{eq:weighted_diffusion_objective} gives
                \begin{equation}
                    \lossWeighted(\denParams)
                    = \tfrac{1}{2} \int_{\logsnrMin}^{\logsnrMax} \weight(\logsnr) \E_{\noiseVec} \left[ \|\noisePred(\encodedData_\logsnr,\logsnr,\textCond) - \noiseVec\|_2^2 \right] \diffd\logsnr
                \end{equation}
                where $\logsnrMin = \logsnrSchedule(1)$ and $\logsnrMax = \logsnrSchedule(0)$ are the schedule endpoints and $\encodedData_\logsnr$ denotes the encoded data noised to level $\logsnr$. This integral does not depend on the schedule $\logsnrSchedule$ except through the endpoints.
                
                Equivalently, we can write the objective as an expectation over the induced distribution $p(\logsnr)$:
                \begin{equation}
                    \lossWeighted(\denParams)
                    = \tfrac{1}{2} \E_{\logsnr \sim p(\logsnr), \noiseVec\sim\standardNormal} \left[ \frac{\weight(\logsnr)}{p(\logsnr)} \|\noisePred(\encodedData_\logsnr,\logsnr,\textCond) - \noiseVec\|_2^2 \right]
                \end{equation}
                The noise schedule thus induces an importance distribution $p(\logsnr)$ over noise levels. $\lossWeighted$ is invariant to the schedule (up to endpoints), but estimator variance depends on how $p(\logsnr)$ aligns with the integrand: schedule design for diffusion-model training is an importance-sampling problem.
            
            \textbf{Schedules versus weights in downstream applications.}
                In SDS and DMD (Secs.~\ref{sec:background-sds}, \ref{sec:background-single-step-distillation}), practitioners inherit a fixed teacher schedule and apply an additional weight $\weight(\timevar)$ or $\sdsWeight(\timevar)$ that conflates the intrinsic loss weight $\weight(\logsnr)$ with the schedule-induced density $p(\logsnr)$. The SDS weight $\sdsWeight(\timevar)$ typically bundles modeling choices (e.g., $\signalcoef$) with implicit timestep reweighting; if monotonic, it induces a new effective timestep distribution. DMD's weighting involves data-dependent normalization (e.g., $\|\meanBase(\noisedData,\timevar) - \encodedData\|$) and is non-monotonic, so it does not map to a simple schedule. \Sec~\ref{sec:method-noise-schedules} treats these weights as known functions, building proposals $\proposalDensity(\timevar)$ via likelihood ratios to avoid conflating schedule and reweighting.

    \subsection{Diffusion Model Applications}\label{sec:background-diffusion-applications-app}

        \subsubsection{Diffusion Priors for Optimization}\label{sec:background-sds-app}
            Score Distillation Sampling (SDS) uses a pretrained diffusion model over an observation space (images, videos, audio, or latents) as a frozen conditional prior that supplies gradients to a parametrized generator, renderer, or simulator. Given parameters \(\genParams\) and a sampled rendering condition $\cameraSample$ (for example, camera pose in text-to-3D), we render an observation
            \begin{align}
                \encodedData = \render(\genParams, \cameraSample) = \encoder\!\big(\prerender(\genParams, \cameraSample)\big),
            \end{align}
            where \(\prerender\) is the differentiable renderer (or generator or simulator) and \(\encoder\) maps its output into the teacher's observation space (often latent). We update \(\genParams\) so \(\encodedData\) lies in high-density regions of \(p_{\denParams}(\cdot \mid \textCond)\) (with optional guidance), giving the chain rule
            \begin{align}
                \nabla_{\genParams}\, \Eop_{\cameraSample}\!\left[\log p_{\denParams}(\render(\genParams, \cameraSample) \mid \textCond)\right]
                = \Eop_{\cameraSample}\!\left[
                    \frac{\diffd \log p_{\denParams}(\encodedData \mid \textCond)}{\diffd \encodedData}
                    \frac{\diffd \encodedData}{\diffd \genParams}
                \right].
                \label{eq:sds_chain_rule}
            \end{align}
            
            \textbf{Diffusion teacher and the SDS residual.}
            Write the forward noising process (in the teacher's observation space) as
            \begin{align}
                \noisedData = \noisedData(\encodedData, \timevar, \noiseVec) = \alpha_{\timevar}\,\encodedData + \sigma_{\timevar}\,\noiseVec,
                \qquad \noiseVec \sim \mathcal{N}(0, \identity), \qquad \timevar \sim p(\timevar),
                \label{eq:forward_noising}
            \end{align}
            where \(\alpha_{\timevar}, \sigma_{\timevar}\) are the usual diffusion coefficients (the exact parameterization is model-dependent). Let the teacher predict noise (optionally with classifier-free guidance scale \(\cfgScale\)):
            \begin{align}
                \noisePred = \noisePred(\noisedData, \timevar, \textCond; \cfgScale).
            \end{align}
            We define the per-sample denoising residual
            \begin{align}
                \residual_{\!\denParams}\!(\genParams, \cameraSample,\timevar,\noiseVec, \textCond, \cfgScale) \defeq \noisePred(\noisedData, \timevar, \textCond; \cfgScale) - \noiseVec,
                \label{eq:sds_residual_def}
            \end{align}
            where \(\noisedData\) is understood to be \(\noisedData(\encodedData(\genParams,\cameraSample),\timevar,\noiseVec)\).
            
            \textbf{SDS gradient estimator (with stop-gradient through the teacher).}
                SDS uses a simple surrogate for the score term \(\diffd \log p_{\denParams}(\encodedData \mid \textCond) / \diffd \encodedData\) by differentiating through the noising map but \emph{not} through the teacher prediction. Concretely, in the backward pass we treat \(\noisePred(\cdot)\) as a constant with respect to \(\noisedData\) (equivalently, we drop the Jacobian \(\diffd \noisePred / \diffd \noisedData\)). This yields the estimator
                \begin{align}
                    \sdsupdate(\genParams) = \Eop_{\cameraSample}\!\left[
                    \Eop_{\timevar \sim p(\timevar),\, \noiseVec \sim \mathcal{N}(0,I)} \!\left[
                        \weight(\timevar)\,
                        \operatorname{sg}\!\big(\residual_{\!\denParams}\big)\,
                        \frac{\diffd \noisedData}{\diffd \encodedData}
                    \right]
                    \frac{\diffd \encodedData}{\diffd \genParams}
                    \right].
                    \label{eq:sds_update_app}
                \end{align}
                Here \(\weight(\timevar)\) is the timestep weight of the SDS loss (\Sec~\ref{sec:background-sds}). Under \Eq~\ref{eq:forward_noising}, \(\diffd \noisedData / \diffd \encodedData = \alpha_{\timevar} \identity\), so with \(\sdsWeight(\timevar) = \weight(\timevar)\alpha_{\timevar}\) this is \Eq~\ref{eq:sds_update}, the common implementation pattern where the timestep weight includes the \(\alpha_{\timevar}\) factor.
            
            \textbf{Equivalent surrogate MSE form used in code.}
                The same update can be obtained as the gradient of a mean-squared error objective with a stop-gradient target. Define the (detached) per-sample gradient direction in observation space
                \begin{align}
                    \widehat{g}_{\encodedData} \defeq \sdsWeight(\timevar)\,
                    \operatorname{sg}\!\big(\residual_{\!\denParams}\big)\,
                    \label{eq:sds_g_hat}
                \end{align}
                and set the target as
                \begin{align}
                    \encodedData_{\text{tgt}} \defeq \operatorname{sg}\!\left(\encodedData - \widehat{g}_{\encodedData}\right).
                    \label{eq:sds_target}
                \end{align}
                Then the surrogate loss
                \begin{align}
                    \loss_{\text{SDS}}(\genParams) \defeq \tfrac{1}{2}\,
                    \Eop_{\cameraSample, \timevar, \noiseVec}
                    \!\left[
                        \left\|\encodedData(\genParams,\cameraSample) - \encodedData_{\text{tgt}}\right\|_2^2
                    \right]
                    \label{eq:sds_surrogate}
                \end{align}
                has gradient
                \begin{align}
                    \nabla_{\genParams}\loss_{\text{SDS}}(\genParams) = \Eop_{\cameraSample, \timevar, \noiseVec}\!\left[
                        \widehat{g}_{\encodedData}\,
                        \frac{\diffd \encodedData}{\diffd \genParams}
                    \right],
                \end{align}
                which matches \Eq~\ref{eq:sds_update}. This is the form we implement: we compute \(\widehat{g}_{\encodedData}\) using the frozen teacher (with stop-gradient through \(\noisePred\) and forward noising), form the detached target \Eq~\ref{eq:sds_target}, and optimize the MSE \Eq~\ref{eq:sds_surrogate}.

        \subsubsection{Single-Step Diffusion Distillation}\label{sec:background-single-step-distillation-app}
            \textbf{Objective and score-based formulation.}
                Distribution Matching Distillation (DMD) \cite{yin2024one} distills a pretrained multi-step diffusion teacher into a one-step generator $\generator: \mathbb{R}^d \to \mathbb{R}^d$ parameterized by $\genParams$. Given $\noiseVec \sim \standardNormal$, the generator produces $\encodedData = \generator(\noiseVec)$ and induces a distribution $\fakeDensity(\encodedData)$. The goal is to match $\fakeDensity$ to the real data distribution $\realDensity$ by minimizing the reverse KL divergence:
                \begin{equation}
                    \KL(\fakeDensity \| \realDensity) = \E_{\encodedData \sim \fakeDensity}[\log \fakeDensity(\encodedData) - \log \realDensity(\encodedData)]
                \end{equation}
                Taking the gradient with respect to $\genParams$ and applying the chain rule gives:
                \begin{equation}
                    \nabla_\genParams \KL(\fakeDensity \| \realDensity) = \E_{\noiseVec \sim \standardNormal}[(\scoreFake(\encodedData) - \scoreReal(\encodedData)) \tfrac{\partial \generator(\noiseVec)}{\partial \genParams}]
                \end{equation}
                where $\encodedData = \generator(\noiseVec)$ and the score functions are $\scoreReal(\encodedData) = \nabla_{\encodedData} \log \realDensity(\encodedData)$ and $\scoreFake(\encodedData) = \nabla_{\encodedData} \log \fakeDensity(\encodedData)$.
            
            \textbf{Score estimation via diffusion noising.}
                Direct score evaluation is intractable and unstable when $\fakeDensity$ and $\realDensity$ have disjoint support. DMD estimates scores by perturbing samples with forward diffusion noise and denoising them with diffusion models. For a clean sample $\encodedData$ and timestep $\timevar$, form the noised sample:
                \begin{equation}
                    \noisedData = \signalcoef \encodedData + \noisecoeff \noiseVec', \qquad \noiseVec' \sim \standardNormal
                \end{equation}
                where $\signalcoef, \noisecoeff$ are the same diffusion schedule coefficients used in \Sec~\ref{sec:background-diffusion-models}.
                The score of the noised distribution can be related to a denoising mean predictor. If $\mean(\noisedData, \timevar)$ predicts $\E[\encodedData \mid \noisedData, \timevar]$, then:
                \begin{equation}
                    \nabla_{\noisedData} \log p(\noisedData \mid \timevar) = - \frac{\noisedData - \signalcoef \mean(\noisedData, \timevar)}{\noisecoeff^2}
                \end{equation}
                
                DMD uses two mean predictors:
                \begin{itemize}[nosep]
                    \item $\meanBase(\noisedData, \timevar)$: the frozen pretrained teacher, which estimates $\E[\encodedData \mid \noisedData, \timevar]$ under $\realDensity$
                    \item $\meanFake(\noisedData, \timevar)$: a learned model parameterized by $\denParams$, which estimates $\E[\encodedData \mid \noisedData, \timevar]$ under $\fakeDensity$
                \end{itemize}
                The approximate scores are:
                \begin{equation}
                \begin{aligned}
                    \scoreReal(\noisedData, \timevar) &= - \frac{\noisedData - \signalcoef \meanBase(\noisedData, \timevar)}{\noisecoeff^2} \\
                    \scoreFake(\noisedData, \timevar) &= - \frac{\noisedData - \signalcoef \meanFake(\noisedData, \timevar)}{\noisecoeff^2}
                \end{aligned}
                \end{equation}
                Noising ensures that both distributions have overlapping support in $\noisedData$-space, stabilizing training.
            
            \textbf{Practical gradient estimator.}
                Substituting the noised-score approximations and integrating over timesteps gives the DMD generator gradient:
                \begin{equation}\label{eq:dmd_mc_estimator}
                \begin{aligned}
                    \nabla_\genParams \KL \simeq \Eop_{\substack{\noiseVec \sim \standardNormal \\ \timevar \sim p(\timevar) \\ \noiseVec' \sim \standardNormal}} \bigg[ \weight(\timevar) \signalcoef \left( \scoreFake(\noisedData, \timevar) - \scoreReal(\noisedData, \timevar) \right)
                    \times \tfrac{\partial \generator(\noiseVec)}{\partial \genParams} \bigg]
                \end{aligned}
                \end{equation}
                where $\encodedData = \generator(\noiseVec)$, $\noisedData = \signalcoef \encodedData + \noisecoeff \noiseVec'$, and $\weight(\timevar)$ is a weighting function. A common choice is:
                \begin{equation}
                    \weight(\timevar) = \frac{\noisecoeff^2}{\signalcoef} \frac{\numChannels \numSpatial}{\|\meanBase(\noisedData, \timevar) - \encodedData\|_1}
                \end{equation}
                where $\numChannels$ and $\numSpatial$ are the numbers of channels and spatial locations, respectively. This weight normalizes scale variations across timesteps to stabilize training. Intuitively, $\scoreReal$ pulls generator samples toward the data manifold, while $-\scoreFake$ discourages mode collapse by repelling samples from regions of excessive fake density.
            
            \textbf{Auxiliary losses.}
                To track the evolving fake distribution during training, DMD updates $\meanFake$ online using a standard diffusion denoising loss on stop-gradient generator outputs:
                \begin{equation}
                    \lossDenoise(\denParams) = \Eop_{\noiseVec, \timevar, \noiseVec'} \left[ \|\meanFake(\noisedData, \timevar) - \operatorname{sg}(\encodedData)\|_2^2 \right]
                \end{equation}
                where $\encodedData = \generator(\noiseVec)$ and $\noisedData = \signalcoef \encodedData + \noisecoeff \noiseVec'$.
                
                Additionally, an optional regression loss aligns the one-step generator with deterministic samples from the teacher on a small paired dataset $\dataset = \{(\dataSample, \dataLabel)\}$:
                \begin{equation}
                    \lossReg(\genParams) = \E_{(\dataSample, \dataLabel) \sim \dataset} \left[ \ell(\generator(\dataSample), \dataLabel) \right]
                \end{equation}
                where $\ell$ is a perceptual distance such as LPIPS. As in \citet{yin2024improved}, we do not use this loss in our experiments. \citet{yin2024improved} also introduced a discriminator trained on the fake model's features to distinguish real data from generator outputs. We also utilize this objective in our experiments. The generator is trained with the combined objective while the fake model minimizes $\lossDenoise(\denParams)$ and remains detached in the generator gradient.
            
            \textbf{Classifier-free guidance.}
                For conditional generation with text conditioning $\textCond$ and classifier-free guidance scale $\cfgScale$, the same construction applies. The real score uses the guided teacher prediction:
                \begin{equation}
                    \meanBase(\noisedData, \timevar, \textCond; \cfgScale) = (1 + \cfgScale) \meanBase(\noisedData, \timevar, \textCond) - \cfgScale \meanBase(\noisedData, \timevar, \varnothing)
                \end{equation}
                while the fake score is unchanged. The generator trains at a fixed guidance scale to match the guided teacher distribution.
            
            \textbf{Connection to variance reduction.}
                Like SDS, the DMD gradient is an expectation over generator input $\noiseVec$, timestep $\timevar$, and forward noise $\noiseVec'$, which we estimate by Monte Carlo. Each gradient sample~requires:
                \begin{enumerate}[nosep]
                    \item Generating $\encodedData = \generator(\noiseVec)$ (potentially expensive)
                    \item Forward noising to form $\noisedData$ (cheap)
                    \item Evaluating both $\meanBase$ and $\meanFake$ (moderate cost)
                    \item Backpropagating through $\generator$ (expensive)
                \end{enumerate}
                Since step (1) is independent of $(\timevar, \noiseVec')$, the amortized resampling strategy from \Sec~\ref{sec:method-compute-reuse} can cache $\encodedData = \generator(\noiseVec)$ and resample $(\timevar, \noiseVec')$ multiple times per generator forward pass. Similarly, timestep stratification (\Sec~\ref{sec:method-stratified-sampling}) and importance sampling (\Sec~\ref{sec:method-noise-schedules}) reduce variance over $\timevar$. Unlike SDS, generator input variability often dominates in DMD, so allocating budget to more independent samples $\noiseVec$ (rather than many re-noisings per sample) can be more effective, as discussed in \Sec~\ref{sec:experiments-dmd}.

        \subsubsection{Data Attribution for Video Generation}\label{app:background-motive}
            Data attribution quantifies how training examples contribute to a model's outputs. In generative modeling, it identifies influential or harmful fine-tuning clips for data selection, debugging, and efficient specialization.

            \textbf{Influence functions and scalable approximations.}
                Let $\loss(\denParams;\dataSample,\textCond)$ be a per-example training loss with query $(\dataSample_{\query},\textCond_{\query})$. For training example $(\dataSample_\ntrain,\textCond_\ntrain)$, infinitesimally upweighting it changes the query loss (under regularity) as in \cite{koh2017understanding}:
                \begin{equation}
                    \influence((\dataSample_\ntrain,\textCond_\ntrain),(\dataSample_{\query},\textCond_{\query}))
                    =-\nabla_{\denParams}\loss(\denParams;\dataSample_{\query},\textCond_{\query})^\top
                    \hessian(\denParams)^{-1}
                    \nabla_{\denParams}\loss(\denParams;\dataSample_\ntrain,\textCond_\ntrain)
                \end{equation}
                where $\hessian(\denParams)=\nabla_{\denParams}^2\frac{1}{|\dataset|}\sum_{(\dataSample,\textCond)\in\dataset}\loss(\denParams;\dataSample,\textCond)$. Since applying $\hessian(\denParams)^{-1}$ is infeasible at modern scales, practical methods approximate influence using gradient similarity computed across checkpoints (TracIn) or via projected gradient features (TRAK) \cite{pruthi2020estimating,park2023trak}.
            
            \textbf{Diffusion attribution as gradient similarity over $(\timevar,\noiseVec)$.}
                For diffusion/flow-matching teachers, training losses and gradients are expectations over noise levels and Gaussian noise. Using notation from \Sec~\ref{sec:background-diffusion-models}, define the per-example, per-draw diffusion gradient:
                \begin{equation}
                    \motiveGrad(\denParams;\dataSample,\textCond,\timevar,\noiseVec)
                    =\nabla_{\denParams}\costDiffusion(\encoder(\dataSample),\textCond,\timevar,\noiseVec,\denParams)
                \end{equation}
                and let $\sampleSet$ denote a multiset of $(\timevar,\noiseVec)$ shared across query and training. Following the gradient normalization of \cite{xie2024data}, we score attribution by the cosine similarity of normalized gradients, averaged over $\sampleSet$
                \begin{equation}\label{eq:diffusion_attrib_cosine}
                    \influence_{\mathrm{diff}}(\ntrain,\query)
                    =\frac{1}{|\sampleSet|}\sum_{(\timevar,\noiseVec)\in\sampleSet}
                    \frac{\motiveGrad(\denParams;\dataSample_{\query},\textCond_{\query},\timevar,\noiseVec)}{\|\motiveGrad(\denParams;\dataSample_{\query},\textCond_{\query},\timevar,\noiseVec)\|_2}^\top
                    \frac{\motiveGrad(\denParams;\dataSample_\ntrain,\textCond_\ntrain,\timevar,\noiseVec)}{\|\motiveGrad(\denParams;\dataSample_\ntrain,\textCond_\ntrain,\timevar,\noiseVec)\|_2}
                \end{equation}
                where $\influence_{\mathrm{diff}}(\ntrain,\query)$ abbreviates influence between $(\dataSample_\ntrain,\textCond_\ntrain)$ and $(\dataSample_{\query},\textCond_{\query})$. Sharing $(\timevar,\noiseVec)$ couples the query and candidate gradients through common random numbers, and per-draw normalization mitigates scale effects. Substantial Monte Carlo variance in \Eq~\ref{eq:diffusion_attrib_cosine} over $(\timevar,\noiseVec)$ limits influence ranking stability at fixed compute, making stable rankings expensive.

            \textbf{Why video is different: appearance-motion entanglement and length effects.}
                For video, $\dataSample\in\mathbb{R}^{F\times H\times W\times 3}$, and the diffusion loss aggregates frame and spatial contributions. Two issues arise. First, whole-video gradients overemphasize static appearance (objects, backgrounds) over temporal dynamics. Second, gradient magnitudes scale with clip length $F$, biasing similarity and selection toward longer clips.
            
            \textbf{Motion-centric attribution via loss-space masking (MOTIVE).}
                MOTIVE \cite{wu2026motion} introduces a motion-weighted attribution loss that emphasizes dynamic regions (e.g., using optical-flow-derived motion magnitude) while suppressing static backgrounds, and corrects dominant length scaling. It also projects gradients efficiently for scalability. Let $\mathbf{M}(\dataSample)\in[0,1]^{F\times H'\times W'}$ be a motion mask aligned with the latent grid (after any required downsampling), and let $\tilde{\cost}_{\mathrm{Diff}}(\encoder(\dataSample),\textCond,\timevar,\noiseVec,\denParams)\in\mathbb{R}^{F\times H'\times W'}$ denote a per-location squared-error form of the diffusion cost.
                MOTIVE defines the motion-weighted per-example cost:
                \begin{equation}
                    \cost_{\mathrm{mot}}(\denParams;\dataSample,\textCond,\timevar,\noiseVec)
                    =\frac{1}{F}\mathrm{mean}_{f,h,w}
                    \big[
                        \mathbf{M}(\dataSample)_{f,h,w}\tilde{\cost}_{\mathrm{Diff}}(\encoder(\dataSample),\textCond,\timevar,\noiseVec,\denParams)_{f,h,w}
                    \big]
                \end{equation}
                and the corresponding motion-weighted gradient:
                \begin{equation}
                    \motiveGrad_{\mathrm{mot}}(\denParams;\dataSample,\textCond,\timevar,\noiseVec)
                    =\nabla_{\denParams}\cost_{\mathrm{mot}}(\denParams;\dataSample,\textCond,\timevar,\noiseVec)
                \end{equation}
                which is substituted for $\motiveGrad$ in \Eq~\ref{eq:diffusion_attrib_cosine}. This isolates temporal dynamics while leaving the forward noising process unchanged, since the reweighting occurs only in the attribution loss.
            
            \textbf{Connection to variance reduction.}
                Both $\influence_{\mathrm{diff}}$ and its motion-weighted variant are Monte Carlo estimators over $(\timevar,\noiseVec)$ with expensive upstream encoding $\encoder(\dataSample)$, a natural target for the strategies in \Sec~\ref{sec:method}: timestep IS and stratification reduce variance over $\timevar$, and amortized re-noising reuses cached $\encoder(\dataSample)$ across $(\timevar,\noiseVec)$ draws. In our experiments (\Sec~\ref{sec:experiments-motive}), we adopt this motion-centric setup and focus on variance reduction for these estimators.

\section{Additional Method Details}\label{app:sec_method}

    \begin{figure}[h!]
        \centering
        \scalebox{1.0}{
        \begin{tikzpicture}
        \centering
            \node (img11){\includegraphics[trim={0.8cm 0.8cm 0cm 0.8cm}, clip, width=.7\linewidth]{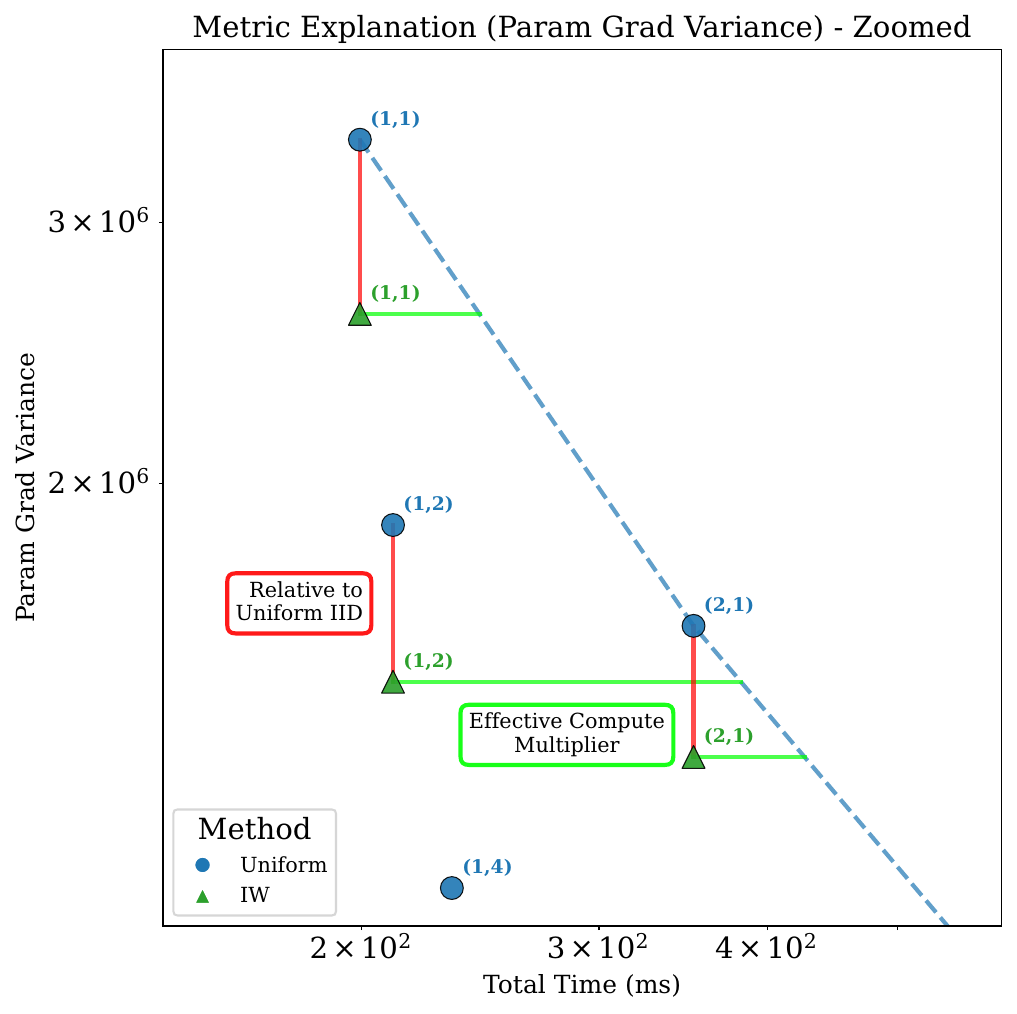}};
            \node[left=of img11, node distance=0cm, rotate=90, xshift=2.5cm, yshift=-.8cm,  font=\color{black}]{\normalsize{Parameter Gradient Variance}};
            \node[below=of img11, node distance=0cm, xshift=0.0cm, yshift=1.15cm,  font=\color{black}]{\normalsize{Total time (ms)}};
        \end{tikzpicture}
        }
        \caption{
            \textbf{Geometric intuition for efficiency metrics:}
            We visualize parameter-gradient variance versus compute cost (wall-clock time) to provide intuition into our performance metrics. The \emph{effective compute multiplier} (ECM) compares a method to the baseline (uniform-IID with $\numReNoises=1$) at iso-variance: at a fixed variance level (horizontal line), ECM equals the ratio of baseline cost to method cost along their respective Pareto curves. The \emph{relative efficiency} (RE) compares methods at identical $(\numRenders, \numReNoises)$, isolating the variance benefit from sampling strategies (importance weighting, stratification) independent of batch size. Higher ECM and RE indicate better efficiency. This geometric view clarifies how our variance-reduction methods achieve $2$--$3\times$ compute multipliers across tasks by shifting the variance-cost frontier to the left.
        }\label{fig:compute_mult_explanation}
    \end{figure}

    \subsection{Variance Measurement Framework}\label{sec:method-variance-framework-app}
        Implementation details for the framework are summarized in \Sec~\ref{sec:method-variance-framework}.
    
        \subsubsection{Ground-Truth Estimation and Dispersion Metrics}
            For an unbiased estimator $\hat\mean$ of a target mean $\mean$, the variance $\mathrm{Var}(\hat\mean)=\E[\|\hat\mean-\mean\|_2^2]$ equals the mean squared error. We approximate $\mean$ using a high-sample Monte Carlo reference $\hat\mean_{\mathrm{GT}}$ formed by averaging $\numSamples_{\mathrm{GT}}$ independent draws ($\numSamples_{\mathrm{GT}}{=}\num{20000}$ in the SDS experiments; \App~\Sec~\ref{sec:experiments-sds-app-details}). For a test estimator $\hat\mean$ with $\numSamples$ samples, we compute
            \begin{equation}
                \widehat{\mathrm{Var}}(\hat\mean) = \|\hat\mean - \hat\mean_{\mathrm{GT}}\|_2^2
            \end{equation}
            and average it over independent realizations. For an independent, unbiased reference this estimates $\mathrm{Var}(\hat\mean)+\mathrm{Var}(\hat\mean_{\mathrm{GT}})$, so the reference's own variance enters as an additive bias; we therefore use it to validate the online variance estimates (\App~\Sec~\ref{app:caching}, \App~\Sec~\ref{sec:experiments-sds-app-details}) rather than as the primary measurement. We validate ground-truth quality by repeating the procedure with different random seeds and confirming agreement to within a small fraction of the standard error.
    
            We measure variance at two stages of the SDS gradient pipeline:
            \begin{enumerate}[nosep]
                \item \textbf{Latent-space update (SDS residual):} For render $\renderIndex$, the vector $\bar\gterm^{(\renderIndex)} = \frac{1}{\numReNoises}\sum_{k=1}^{\numReNoises} \sdsWeight(\timevar^{(\renderIndex,k)})\residual^{(\renderIndex,k)}$ that multiplies its renderer Jacobian. This is the cheapest to compute because it does not require backpropagation through the renderer.
                \item \textbf{Parameter gradient:} The full gradient $\nabla_{\genParams}\lossSDS = \frac{1}{\numRenders}\sum_{\renderIndex=1}^{\numRenders} \bar\gterm^{(\renderIndex)} \frac{\partial \encodedData^{(\renderIndex)}}{\partial \genParams}$, which includes the renderer Jacobian. This is more expensive but directly measures the quantity used for optimization.
            \end{enumerate}
            Most prior work reports variance of the latent-space residual because it is easy to compute in a batched loop. However, the parameter gradient variance depends on the interaction between the residual and the renderer Jacobian, which can differ substantially across timesteps. We find that optimal importance-sampling proposals differ between the two metrics (see \App~\Fig~\ref{fig:iwvis_latent_vs_param}), so we report both and, when feasible, use parameter-gradient variance as our primary design criterion.
    
            For parameter gradients, standard batched backpropagation aggregates contributions from all samples in a batch, so we cannot isolate individual sample gradients $\nabla_{\genParams}\lossSDS^{(\sampleIndex)}$ without re-running backpropagation. To populate per-sample statistics, we use a batch size of $1$ and loop over samples, which is expensive but necessary for accurate measurement. For DMD and data attribution, we follow a similar procedure, isolating per-sample contributions to the generator or denoiser gradients.
    
            We also report cosine similarity between estimators and the ground truth,
            \begin{equation}
                \mathrm{CosineSim}(\hat\mean, \hat\mean_{\mathrm{GT}}) = \frac{\hat\mean^\top \hat\mean_{\mathrm{GT}}}{\|\hat\mean\|_2 \|\hat\mean_{\mathrm{GT}}\|_2}
            \end{equation}
            which captures directional alignment independent of magnitude. Like the reference-based MSE, cosine similarity requires the reference to be precomputed; the online variance estimate does not.
    
        \subsubsection{Cost Metrics and Extrapolation}
            We measure compute cost using three complementary metrics:
            \begin{itemize}[nosep]
                \item \textbf{Wall-clock time (ms):} End-to-end time for a gradient step, including parallelization and GPU scheduling effects. This is the metric we use for effective compute multipliers in the main text.
                \item \textbf{GPU memory (MB):} Peak memory usage, which bounds feasible batch sizes and reveals parallelization headroom.
                \item \textbf{Number of function evaluations (NFE):} Counts of expensive operations (renders, denoiser calls, encoder calls) independent of hardware. For SDS, NFE is typically $(\numRenders, \numReNoises)$ denoting the number of renders and re-noisings per render.
            \end{itemize}
            When variance scales as $\widehat{\mathrm{Var}}(\hat\mean) \approx C/\numSamples$, we fit constant $C$ and extrapolate to larger $\numSamples$ assuming no parallelization benefit (cost scales linearly with $\numSamples$). This gives the best-case variance estimates at higher compute budgets. We also measure wall-clock time at higher parallelism to capture GPU scheduling, memory bandwidth, and batching effects, which can cause the cost to plateau or increase as parallelism is reduced.
    
        \subsubsection{Caching and Computational Tricks}\label{app:caching}
            To reduce the cost of variance measurement, we exploit conditional independence in the estimators. For SDS, the renderer output $\dataSample=\prerender(\genParams,\cameraSample)$ is independent of the diffusion noise $(\timevar,\noiseVec)$, so we can cache $\dataSample$ and its forward-pass encoding $\encodedData=\encoder(\dataSample)$ and reuse them across many $(\timevar,\noiseVec)$ draws. This amortizes the expensive render-and-encode step over many cheap denoiser calls, making it feasible to estimate $\hat\mean_{\mathrm{GT}}$ with thousands of samples. For parameter gradient variance, we cannot fully cache the backward pass because the render-and-encode Jacobian $\frac{\partial \encodedData}{\partial \genParams}$ depends on $\dataSample$, but we can still cache the forward computation and re-noise multiple times before backpropagating.
    
            For online variance estimation, we use Welford's algorithm to update the mean and variance without storing samples; it is useful for monitoring variance trends during training, but it is not applicable to cosine similarity or any metric that requires the full ground truth.
    
        \subsubsection{Validation and Reproducibility}
            We validate variance estimates by repeating measurements with different random seeds and confirming consistency. For each configuration (e.g., batch size, importance proposal, stratification), we repeat measurements across prompts and seeds and report standard errors where appropriate. We also check that ground-truth estimates $\hat\mean_{\mathrm{GT}}$ from different seeds agree to within their Monte Carlo error, ensuring that $\numSamples_{\mathrm{GT}}$ is large enough.

        \subsubsection{Worked Example: ECM Computation for SDS}\label{app:ecm-worked-example}
            We illustrate how the two efficiency metrics of \Sec~\ref{sec:method-variance-framework} combine, using the SDS values in \Tabs~\ref{tab:absolute_ecm_by_m_param} and~\ref{tab:relative_improvement_by_m_param}.
        
            \textbf{ECM factorizes into a reuse term and a sampling term.}
                Along the uniform $(\numRenders,1)$ baseline, variance scales as $1/\numRenders$ and cost as $\numRenders$, so the baseline reaches variance $V$ at cost $c_{\mathrm{b}}(V)\!=\!C_{\mathrm{b}}/V$ for a constant $C_{\mathrm{b}}$. An estimator with variance $V_{\mathrm{m}}$ and per-iteration cost $c_{\mathrm{m}}$ therefore has $\mathrm{ECM}\!=\!c_{\mathrm{b}}(V_{\mathrm{m}})/c_{\mathrm{m}}\!=\!C_{\mathrm{b}}/(V_{\mathrm{m}}c_{\mathrm{m}})$, which is also its variance reduction relative to the baseline at matched cost. Importance weighting and stratification reallocate the same draws, so they leave the per-iteration cost at a given $(\numRenders,\numReNoises)$ unchanged. The ratio of the ECMs of IW$+$Strat and uniform sampling at the same $(\numRenders,\numReNoises)$ is then their variance ratio, which is RE:
                \[
                    \mathrm{ECM}_{\mathrm{IW+Strat}}=\mathrm{ECM}_{\mathrm{uniform}}\cdot\mathrm{RE}_{\mathrm{IW+Strat}}.
                \]
        
            \textbf{Numbers.}
                At $\numReNoises\!=\!8$, compute reuse alone gives $\mathrm{ECM}_{\mathrm{uniform}}\!\approx\!2.63$ (\Tab~\ref{tab:absolute_ecm_by_m_param}) and IW$+$Strat gives $\mathrm{RE}\!\approx\!1.25$ (\Tab~\ref{tab:relative_improvement_by_m_param}), so $\mathrm{ECM}_{\mathrm{IW+Strat}}\!\approx\!2.63\times1.25\!\approx\!3.29$, the ${\sim}3.3\times$ headline in \Sec~\ref{sec:experiments-sds}. The same product reproduces the other rows of \Tab~\ref{tab:absolute_ecm_by_m_param} to within rounding, e.g., $2.24\times1.31\!\approx\!2.94$ at $\numReNoises\!=\!4$ and $2.52\times1.17\!\approx\!2.94$ at $\numReNoises\!=\!16$. Compute reuse thus supplies most of the multiplier, and IW$+$Strat a further ${\sim}25\%$.

    \subsection{Algorithm: Combined IW + Stratified + Re-noising Estimator}\label{app:algorithm}
        \Algo~\ref{alg:combined} compiles \Eq~\ref{eq:reuse-estimator}, \Eq~\ref{eq:per-render-strat}, and \Eq~\ref{eq:stratified-IS} into a single drop-in pseudocode, exactly matching the SDS configuration we recommend. The DMD variant replaces the render with the generator forward pass and the residual with the weighted score difference (\Eq~\ref{eq:dmd_mc_estimator}); the attribution variant differentiates the loss with respect to $\denParams$ for each fixed example, with no generator Jacobian. Both keep the same outer loop.

        \textbf{Per-task instantiation.}
            \emph{SDS.} Per step, sample $\numRenders$ poses $\cameraSample^{(\renderIndex)}$, render $\dataSample^{(\renderIndex)}=\prerender(\genParams,\cameraSample^{(\renderIndex)})$ once each, and (for latent diffusion) encode $\encodedData^{(\renderIndex)}$ once; for each $\renderIndex$ draw $\numReNoises$ pairs $(\timevar,\noiseVec)$ and backpropagate once per render via \Eq~\ref{eq:reuse-estimator}. \emph{One-step distillation.} Treat the generator output $\encodedData^{(\renderIndex)}=\generator(\noiseVec^{(\renderIndex)})$ as the expensive state and apply the same re-noising pattern with fresh $(\timevar,\noiseVec')$.

        \textbf{Timestep stratification details.}
            For a uniform $p(\timevar)$ (as in SDS, on $[\timevar_{\min},\timevar_{\max}]$), equal-width bins each have probability $1/\numStrata$, and one sample per bin gives an unbiased estimator compatible with \Eq~\ref{eq:reuse-estimator}. For non-uniform densities, we use the quantile construction of \Sec~\ref{sec:method-stratified-sampling} (\Fig~\ref{fig:inverseTransformVisualization}). If $\timevar\in\{0,\dots,\maxTimevar{-}1\}$ (e.g., $\maxTimevar{=}1000$), we stratify in $[0,1]$ using $\numStrata$ equal bins with one draw per bin, then snap to the nearest index; this matches the discrete grid the teacher was trained on and moves each draw by at most half a grid step. We switch between continuous $[0,1]$ and discrete $\{0,\dots,\maxTimevar{-}1\}$ views for notation. With importance proposal $\proposalDensity$, we stratify in $[0,1]$ then apply inverse-CDF sampling for stratified $\timevar$.

        \textbf{Global stratification across renders.}
                At $\numReNoises\!=\!1$ there is nothing to stratify within a unit, so timesteps are stratified across the whole batch instead, with bins $\stratum_{\stratumIndex}$ constructed as in \Sec~\ref{sec:method-stratified-sampling} but with $\numStrata\defeq\numRenders\times\numReNoises$. For a uniform base density and equal-width bins, the global stratified estimator is:\vspace{-0.002\textheight}
                \begin{equation}\label{eq:global-strat}
                    \smash{
                    \bar{\gterm}_{\mathrm{global}} =\nicefrac{1}{\numStrata}\sum\nolimits_{\stratumIndex=1}^{\numStrata} \gterm(\dataSample_{\stratumIndex},\timevar_{\stratumIndex},\noiseVec_{\stratumIndex})
                    }
                \end{equation}
                where $\timevar_{\stratumIndex}\!\sim\!\Uniform(\stratum_{\stratumIndex})$ and $\dataSample_{\stratumIndex}$ is a (potentially reused) render; for a non-uniform proposal $\proposalDensity$, the bins come from its inverse CDF and each term carries the weight $p/\proposalDensity$, as in \Eq~\ref{eq:stratified-IS}.

        \begin{algorithm}[!b]
            \caption{Combined IW $+$ stratified $+$ re-noising estimator (per gradient step, SDS).}\label{alg:combined}
            \begin{algorithmic}[1]
                \Require parameters $\genParams$; renders/step $\numRenders$; re-noisings/render $\numReNoises\!=\!\numStrata$; base timestep density $p$; importance proposal $\proposalDensity\!\propto\!p\,\sdsWeight$; frozen teacher $\noisePred(\cdot;\denParams)$; encoder $\encoder$; renderer $\prerender$; conditioning $\textCond$; guidance scale $\cfgScale$.
                \For{$\renderIndex = 1, \dots, \numRenders$}
                    \State Sample render condition $\cameraSample^{(\renderIndex)}$; compute $\dataSample^{(\renderIndex)} \!=\! \prerender(\genParams, \cameraSample^{(\renderIndex)})$, $\encodedData^{(\renderIndex)}\!=\!\encoder(\dataSample^{(\renderIndex)})$
                    \For{$\stratumIndex = 1, \dots, \numStrata$}
                        \State Draw $\randomness^{(\renderIndex)}_{\stratumIndex}\sim\Uniform(0,1)$; set quantile $\quantile^{(\renderIndex)}_{\stratumIndex}\!=\!(\stratumIndex - 1 + \randomness^{(\renderIndex)}_{\stratumIndex})/\numStrata$
                        \State $\timevar^{(\renderIndex)}_{\stratumIndex} \gets \mathrm{CDF}_{\proposalDensity}^{-1}(\quantile^{(\renderIndex)}_{\stratumIndex})$ \Comment{stratified inverse-CDF; \Fig~\ref{fig:inverseTransformVisualization}}
                        \State Draw $\noiseVec^{(\renderIndex,\stratumIndex)}\sim\standardNormal$; form $\noisedData^{(\renderIndex,\stratumIndex)}\!=\!\signalcoef \encodedData^{(\renderIndex)} + \noisecoeff\noiseVec^{(\renderIndex,\stratumIndex)}$
                        \State $\residual^{(\renderIndex,\stratumIndex)} \gets \noisePred(\noisedData^{(\renderIndex,\stratumIndex)},\timevar^{(\renderIndex)}_{\stratumIndex},\textCond;\cfgScale) - \noiseVec^{(\renderIndex,\stratumIndex)}$
                        \State $\importanceWeight^{(\renderIndex)}_{\stratumIndex} \gets p(\timevar^{(\renderIndex)}_{\stratumIndex})\,/\,\proposalDensity(\timevar^{(\renderIndex)}_{\stratumIndex})$
                    \EndFor
                    \State $\bar{\gterm}^{(\renderIndex)} \gets \frac{1}{\numStrata}\!\sum_{\stratumIndex} \importanceWeight^{(\renderIndex)}_{\stratumIndex}\,\sdsWeight(\timevar^{(\renderIndex)}_{\stratumIndex})\,\residual^{(\renderIndex,\stratumIndex)}$ \Comment{per-render IS+strat avg.}
                \EndFor
                \State \Return $\hat\nabla_\genParams \gets \frac{1}{\numRenders}\!\sum_{\renderIndex}\bar{\gterm}^{(\renderIndex)}\, \partial \encodedData^{(\renderIndex)}/\partial \genParams$ \Comment{single backward pass per render, through the encoder}
            \end{algorithmic}
        \end{algorithm}

\section{Additional Experimental Details}\label{app:sec_experiments}

    \subsection{Diffusion Priors for Optimization}\label{sec:experiments-sds-app}

        \subsubsection{Details}\label{sec:experiments-sds-app-details}

            \textbf{Model and Architecture.}
                We use \texttt{stable-diffusion-2-1-base}~\citep{rombach2022high} as the 2D diffusion prior within the threestudio framework~\citep{threestudio2023}.
                The 3D representation is an implicit volume with a ProgressiveBandHashGrid encoder (instant-NGP style~\citep{muller2022instant}), using $16$ levels, $2$ features per level, and $\log_2(\text{hashmap size}) = 19$.
                The density and color networks are VanillaMLP with 64 neurons and 1 hidden layer.
                Images are rendered at $256 \times 256$ resolution and bilinearly interpolated to $512 \times 512$ before being encoded by the frozen VAE encoder, yielding $4 \times 64 \times 64$ latents.
            
            \textbf{Diffusion Guidance.}
                We use classifier-free guidance with default scale $\omega = 100$.
                For low guidance ablation experiments (\Sec~\ref{sec:low_guidance_analysis}, \Fig~\ref{fig:quantifying_variance_low_guidance}) we use $\omega = 25$.
                Timesteps are sampled from the range $[\timevar_{\min}, \timevar_{\max}] = [20, 980]$ out of $1000$ total steps, corresponding to $[\texttt{min\_step\_percent}, \texttt{max\_step\_percent}] = [0.02, 0.98]$.
                The noise schedule uses scaled-linear spacing in $\sqrt{\beta}$ space with $\beta_{\text{start}} = 0.00085$ and $\beta_{\text{end}} = 0.012$. With threestudio's default SDS weighting, $\sdsWeight(\timevar)=\noisecoeff^2$; the curves labelled $\weight(\timevar)=\noisecoeff^2$ in \Fig~\ref{fig:toyIWVisualization} (right) and \Fig~\ref{fig:inverseTransformVisualization} show this effective weight.
            
            \textbf{Camera Sampling.}
                Camera poses are sampled uniformly with elevation in $[-10^\circ, 45^\circ]$, azimuth in $[0^\circ, 360^\circ]$, field of view in $[15^\circ, 80^\circ]$, and fixed camera distance of 2.0.
            
            \textbf{Optimization.}
                We use the Adam optimizer~\citep{kingma2014adam} with $\beta_1 = 0.9$, $\beta_2 = 0.99$, $\epsilon = 10^{-15}$, and learning rates of $0.005$ for geometry and $0.0001$ for background.
                Training runs for 5000 iterations at $\cfgScale\!=\!100$ (longer at low guidance; \App~\Sec~\ref{sec:low_guidance_analysis}), saving checkpoints every $\num{1000}$ steps.
            
            \textbf{Variance Evaluation Protocol.}
                We evaluate variance reduction on a subset of the trained NeRF checkpoints, using 5 prompts at high guidance ($\omega=100$; see \Tab~\ref{tab:prompts}) across 3 seeds $\{1, 2, 3\}$. The ground-truth gradient $\hat\mean_{\mathrm{GT}}$ is estimated by averaging 20{,}000 independent samples under uniform timestep sampling. It validates the online variance estimates (Variance Estimation, below) and is the reference for cosine similarity.
            
            \textbf{Timestep Sampling Strategies.}
                We evaluate combinations of timestep distributions (uniform and importance-weighted) with batch sampling strategies (IID and stratified), described below.
                We additionally ablate a parameter-gradient-weighted proposal (\texttt{param\_iw}) described in \Sec~\ref{sec:iw_ablation}.
  
            \textbf{Training Runs for 3D NeRF Checkpoints.}
                We train NeRF models on 30 text prompts (listed in \Tab~\ref{tab:prompts}) using two configurations:
                (i)~\emph{Baseline}: uniform-iid timestep sampling with $(\numRenders, \numReNoises) = (4, 1)$ (4 renders, 1 re-noising each);
                (ii)~\emph{Ours}: importance-weighted + stratified sampling with $(\numRenders, \numReNoises) = (1, 16)$ (1 render, 16 re-noisings each).
                Ours costs less per iteration (\Fig~\ref{fig:quantifying_variance_hierarchical_cost_aware_iw_strat}).
                Each prompt is trained for 5000 iterations across 3 seeds $\{1,2,3\}$, yielding 180 final checkpoints (30 prompts $\times$ 2 methods $\times$ 3 seeds). \Fig~\ref{fig:qualitative_sds_trajectory} shows validation renders throughout training.
                We save intermediate checkpoints every 1000 iterations for use in variance~\mbox{experiments}.

            \textbf{CLIP Score Evaluation.}
                We use ViT-B/32~\citep{radford2021learning} on renders produced every 25 NeRF training steps. Each validation uses 10 views at fixed elevation ($12.5^\circ$), camera distance ($2.0$), and FOV ($40^\circ$), with azimuth uniform around the object. Scores are averaged across views, prompts, and seeds for the curves in \Fig~\ref{fig:clip_sds}.
            
            \textbf{Variance Estimation.}
                We measure variance at two levels: (i)~\emph{SDS gradient variance}, the variance of the latent-space gradient output by the diffusion model, and (ii)~\emph{parameter gradient variance}, variance after backpropagation through the renderer and VAE encoder.
                
                We estimate variance using Welford's online algorithm with early stopping on the parameter-gradient estimate: after $\num{1000}$ iters, check every $50$ steps and terminate when relative change stays $<\!0.1\%$ for $3$ consecutive checks. We validated this rule against a $\num{20000}$-sample MSE reference and found agreement within MC noise. Runs are capped at $\num{20000}$ iterations, but most stop before $\num{4000}$.

            \begin{table}[h]
                \centering
                \caption{Text prompts used for NeRF training and CLIP score evaluation. Prompts marked with $\dagger$ were used for variance experiments.}\label{tab:prompts}
                \small
                \begin{tabular}{l@{\hspace{1em}}c}
                \toprule
                Prompt & Variance Experiments \\
                \midrule
               A cactus with pink flowers & \\
                An antique wooden rocking horse & \\
                A golden retriever with a blue bowtie & \\
                An ivory candlestick holder & \\
                A bright red fire hydrant & \\
                A castle-shaped sandcastle & \\
                A plush teddy bear with a satin bow & $\dagger$ \\
                A vintage porcelain doll with a frilly dress & \\
                A tarnished brass pocket watch & \\
                A ceramic teapot with floral patterns & \\
                An antique ruby-studded brooch & \\
                A shiny emerald green beetle & $\dagger$\\
                A crumpled silver aluminum soda can & \\
                A shimmering emerald pendant necklace & \\
                An antique glass perfume bottle & \\
                A polished mahogany grand piano & $\dagger$ \\
                A pristine white wedding gown & \\
                A chipped porcelain teacup &  \\
                A rustic wrought-iron candle holder &  \\
                A rusty, vintage metal key &  \\
                A delicate, handmade lace doily & \\
                A glossy grand black piano & \\
                A gold glittery carnival mask & $\dagger$ \\
                A shiny red apple & \\
                An antique gold pocket watch & \\
                A fluffy, orange cat & \\
                A scuffed up soccer ball & \\
                A sleek stainless steel teapot & \\
                An intricate ceramic vase with peonies painted on it &  \\
                A blooming potted orchid with purple flowers & $\dagger$ \\
                \bottomrule
                \end{tabular}
            \end{table}
        
            \textbf{Variance Experiment Configurations.}
                We evaluate SDS gradient variance across a grid of timestep sampling methods and batch configurations.
            
            \textbf{Batch configurations.}
                We test all $(\numRenders, \numReNoises)$ pairs satisfying $\numRenders \times \numReNoises \leq 32$, where $\numRenders \in \{1, 2, 4, 8\}$ is the number of rendered views and $\numReNoises \in \{1, 2, 4, 8, 16, 32\}$ is the number of timesteps per view, yielding 18 unique configurations.
            
            \textbf{Timestep sampling methods.}\label{sec:timestep-sampling-methods}
                We compare two main sampling distributions over $[\timevar_{\min}, \timevar_{\max}]$:
                \begin{enumerate}
                    \item \emph{Uniform}: $p(\timevar) = \text{Uniform}[\timevar_{\min}, \timevar_{\max}]$
                    \item \emph{Importance-weighted} (IW): $q(\timevar) \propto p(\timevar) \sdsWeight(\timevar)$ using the SDS weighting function from \Eq~\ref{eq:sds_update}.
                \end{enumerate}
                For importance-weighted sampling, gradient contributions are scaled by $\nicefrac{p(\timevar)}{q(\timevar)}$ so the estimator is unbiased for the expectation under uniform $p(\timevar)$.
            
            \textbf{Stratification.}\label{sec:stratification-methods}
                For each sampling method, we compare:
                \begin{enumerate}
                    \item \emph{iid}: independent sampling from $q(\timevar)$ for each timestep
                    \item \emph{stratified}: $\numReNoises$ timesteps per view are stratified into $\numReNoises$ equal-probability strata (for importance-weighted sampling, this uses inverse-CDF sampling; see \Sec~\ref{sec:method-stratified-sampling} and \Fig~\ref{fig:inverseTransformVisualization}).
                \end{enumerate}
            
\textbf{Experimental matrix.}
    The full matrix of 4 method combinations (2 timestep sampling methods $\times$ 2 batch sampling strategies) across 18 batch configurations is evaluated at training step 5{,}000 on 5 text prompts; see \Sec~\ref{sec:experiments-sds} for results.
    Variance experiments are conducted on checkpoints trained with 3 different seeds. For each checkpoint, variance is measured using 4 independent Monte Carlo seeds (3 for the oracle ablation), yielding $5$ prompts $\times$ $3$ training seeds $\times$ $4$ MC seeds $= 60$ independent variance measurements per method configuration. The same configurations are additionally tested at intermediate training checkpoints (steps $\num{1000}$--$\num{4000}$) to verify that the conclusions hold throughout optimization; \Fig~\ref{fig:quantifying_variance_low_guidance_over_training} shows the low-guidance runs at steps $\num{1000}$, $\num{2000}$, and $\num{9000}$.
            
\begin{figure}
                    \centering
                    \vspace{-0.01\textheight}
                    \resizebox{\linewidth}{!}{
                    \begin{tikzpicture}
                    \centering
                        \node (img11){\includegraphics[trim={1.39cm 1.21cm 0cm 0.83cm}, clip, width=0.2978\linewidth]{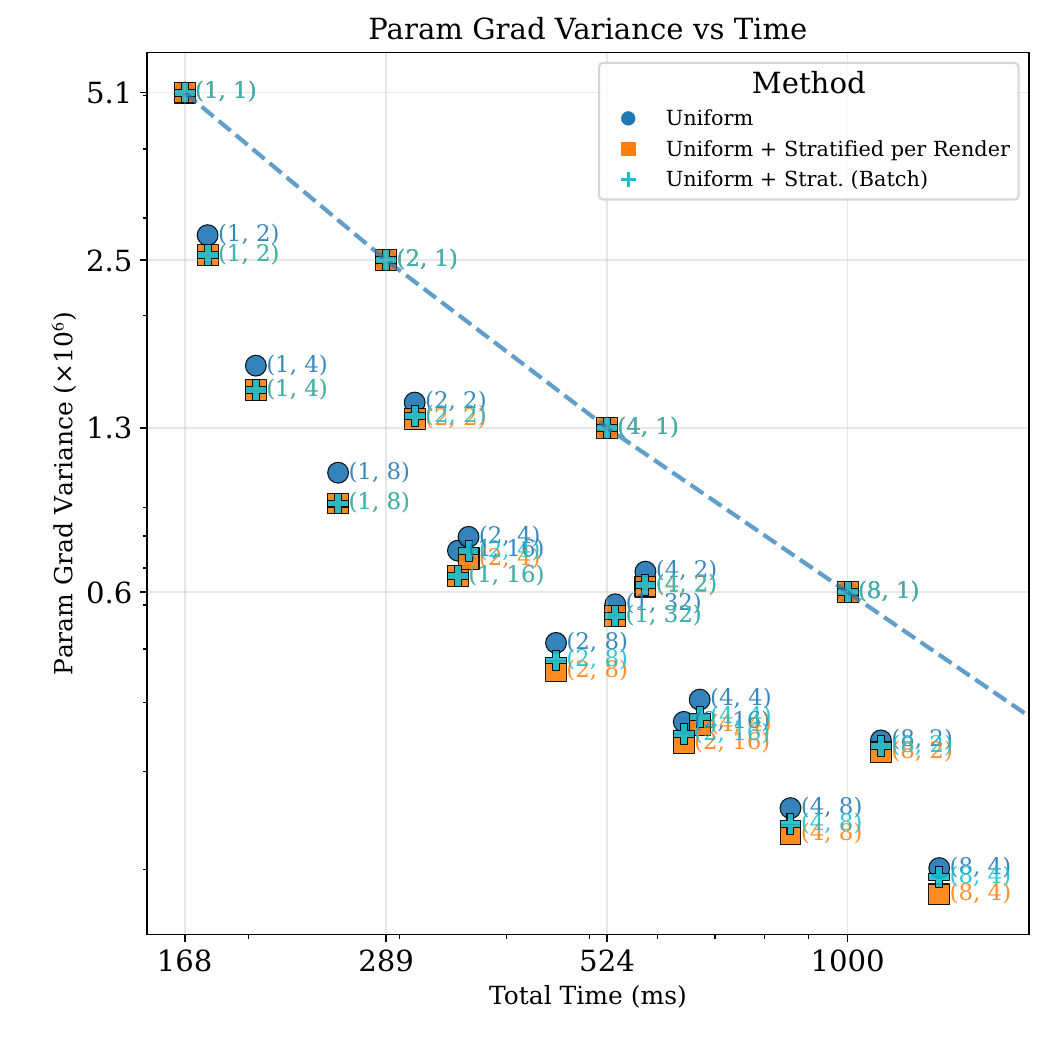}};
                        \node[left=of img11, node distance=0cm, rotate=90, xshift=1.0cm, yshift=-1.0cm,  font=\color{black}]{\scriptsize{Variance $(\times 10^6)$}};
                        \node[below=of img11, node distance=0cm, xshift=-0.0cm, yshift=1.15cm,  font=\color{black}]{\scriptsize{Per-Iteration Compute (ms)}};
        
                        \node [right=of img11, node distance=0cm, xshift=-0.75cm](img21){\includegraphics[trim={1.09cm 1.21cm 0cm 0.83cm}, clip, width=0.3020\linewidth]{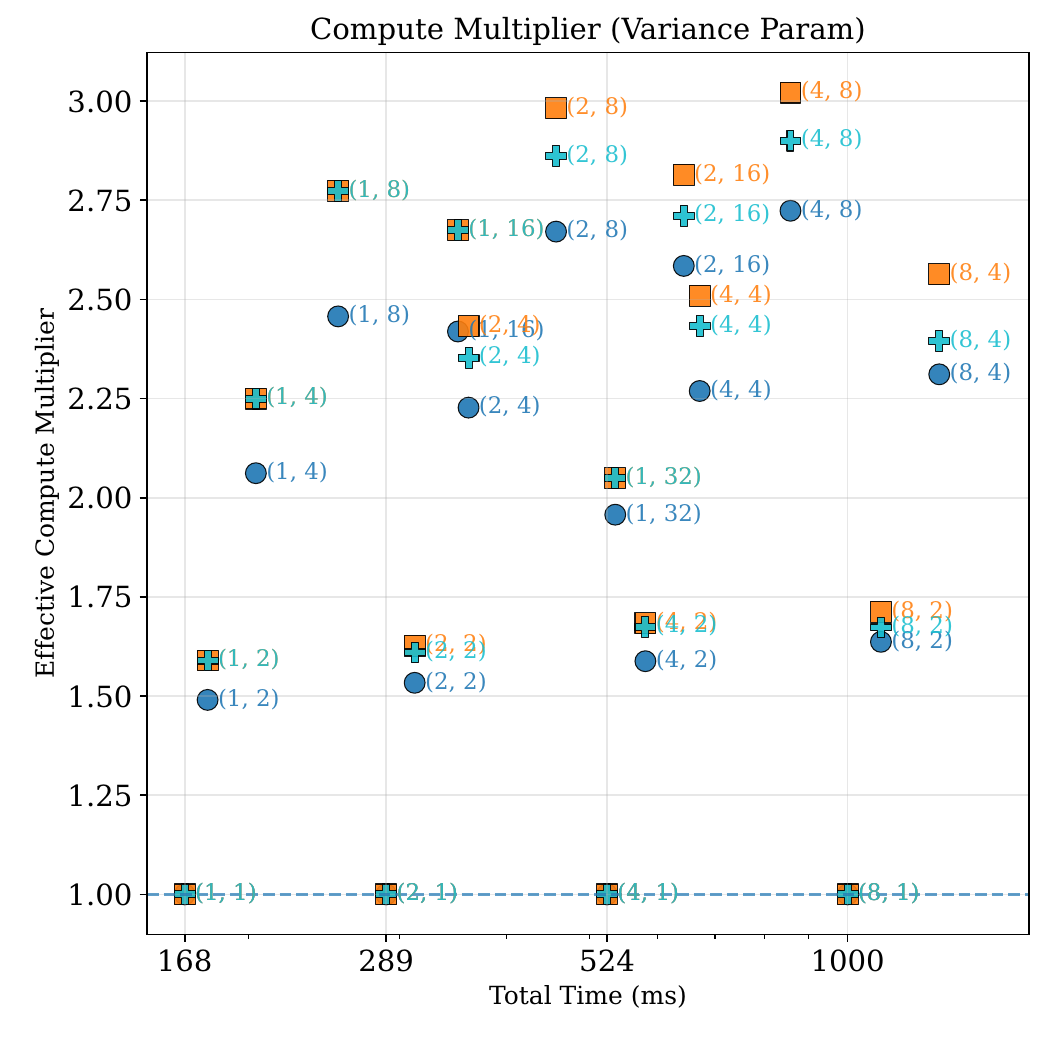}};
                        \node[left=of img21, node distance=0cm, rotate=90, xshift=1.8cm, yshift=-1.0cm,  font=\color{black}]{\scriptsize{Effective Compute Multiplier}};
                        \node[below=of img21, node distance=0cm, xshift=0.0cm, yshift=1.15cm,  font=\color{black}]{\scriptsize{Per-Iteration Compute (ms)}};
        
                        \node [below=of img11, xshift=2.22cm, yshift=0.6cm](img31){\includegraphics[trim={1.2cm 1.21cm 0cm 0.83cm}, clip, width=.3\linewidth]{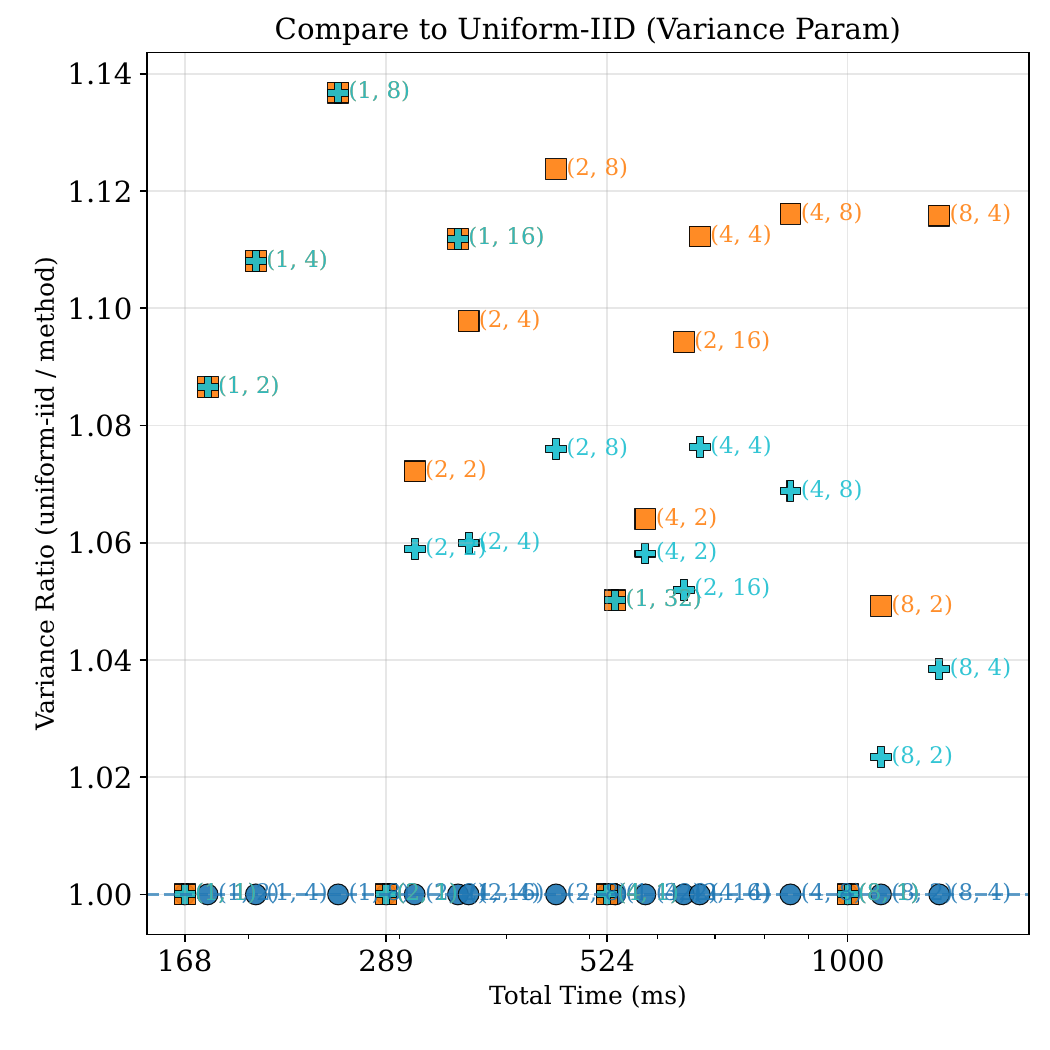}};
                        \node[left=of img31, node distance=0cm, rotate=90, xshift=2.0cm, yshift=-1.0cm,  font=\color{black}]{\scriptsize{Relative Efficiency to Uniform}};
                        \node[below=of img31, node distance=0cm, xshift=0.0cm, yshift=1.15cm,  font=\color{black}]{\scriptsize{Per-Iteration Compute (ms)}};
                    \end{tikzpicture}
                    }
                    \vspace{-0.01\textheight}
                    \caption{
                        \textbf{Comparing Per-Render and Global Stratification Strategies.}
                        This figure ablates two stratified sampling approaches: per-render stratification (\Eq~\ref{eq:per-render-strat}), which stratifies timesteps independently within each render's $\numReNoises$ re-noisings, versus global stratification (\Eq~\ref{eq:global-strat}), which stratifies timesteps across all $\numRenders \times \numReNoises$ samples in the batch.
                        \emph{Top left:} Variance versus compute budget for the uniform baseline (blue circles), global stratification (cyan crosses; ``Batch'' in the legend), and per-render stratification (orange squares). Points are annotated by $(\numRenders,\numReNoises)$ configurations.
                        \emph{Top right:} Effective compute multiplier relative to the uniform $\numReNoises\!=\!1$ baseline curve.
                        \emph{Bottom:} Relative efficiency (RE) isolating the gain from stratification by comparing to uniform sampling at the same $(\numRenders,\numReNoises)$ configuration.
                        Per-render stratification matches or outperforms global stratification when $\numReNoises\!>\!1$, as it exploits the hierarchical structure by reducing within-render variance. When $\numReNoises\!=\!1$, per-render stratification degenerates to uniform sampling (no timesteps to stratify within a single re-noising), so global stratification is the only stratified option there; this sweep shows no gain from it at $\numReNoises\!=\!1$ (bottom). This motivates our choice of per-render stratification for SDS experiments, where renders are expensive, and configurations with $\numReNoises\!>\!1$ are most efficient.
                    }\label{fig:variance_ablation_stratified}
                \end{figure}

            \textbf{Ablations.} We additionally evaluate two ablations on a limited subset of prompts and steps:
                (i)~\emph{parameter-gradient-weighted (oracle) sampling}, where $q(\timevar)$ is proportional to pre-computed parameter gradient norm per timestep (estimated from prior experiments); see \Sec~\ref{sec:iw_ablation}, \Fig~\ref{fig:sds_optimal_importance}, and \Tab~\ref{tab:iw_ablation}.
                (ii)~\emph{global vs. per-render stratification}, where all $\numRenders \times \numReNoises$ timesteps are jointly stratified across the entire batch (\Eq~\ref{eq:global-strat}) rather than per render (\Eq~\ref{eq:per-render-strat}); see \Fig~\ref{fig:variance_ablation_stratified}.

        \textbf{Compute Usage.}
            All experiments are conducted on NVIDIA A100-80GB GPUs (1 GPU per job).
            
            \emph{Checkpoint training:}
            We train $30$ prompts $\times$ $2$ training methods $\times$ $3$ seeds $=$ $180$ runs, each running $5{,}000$ steps in ${\sim}2$ hours, totaling ${\sim}360$ GPU-hours.
            
            \emph{Main variance experiments:}
            We evaluate $2$ timestep methods (uniform, importance-weighted) $\times$ $2$ batch methods (IID, stratified) $\times$ $18$ $(\numRenders, \numReNoises)$ configurations (all pairs where $\numRenders \times \numReNoises \leq 32$) $\times$ $4$ variance seeds $\times$ $5$ prompts $\times$ $3$ training seeds $=$ $4{,}320$ runs, totaling ${\sim}2{,}735$ GPU-hours.
            Importance-weighted runs are slightly faster (${\sim}662$ GPU-hours per method-batch combination) than uniform runs (${\sim}705$ GPU-hours) due to variance reduction.
            
            \emph{Oracle ablation (\Sec~\ref{sec:iw_ablation}):}
            We compare the weight heuristic against the intractable oracle proposal using $1$ timestep method (oracle IW) $\times$ $1$ batch method (stratified) $\times$ $18$ $(\numRenders, \numReNoises)$ configurations $\times$ $3$ variance seeds $\times$ $5$ prompts $\times$ $3$ training seeds $=$ $810$ runs, totaling ${\sim}286$ GPU-hours.
            
            \emph{Total:}
            ${\sim}3{,}400$ A100 GPU-hours (${\sim}142$ GPU-days) for the three items above; the low-guidance runs (\App~\Sec~\ref{sec:low_guidance_analysis}) and the global-stratification ablation are not included.

        \subsubsection{Results}\label{sec:experiments-sds-app-results}
\begin{figure}[h!]
                \centering
                \resizebox{\linewidth}{!}{
                \begin{tikzpicture}
                \centering
                    \node (img11){\includegraphics[trim={1.3cm 1.21cm 0cm 0.8cm}, clip, width=.3\linewidth]{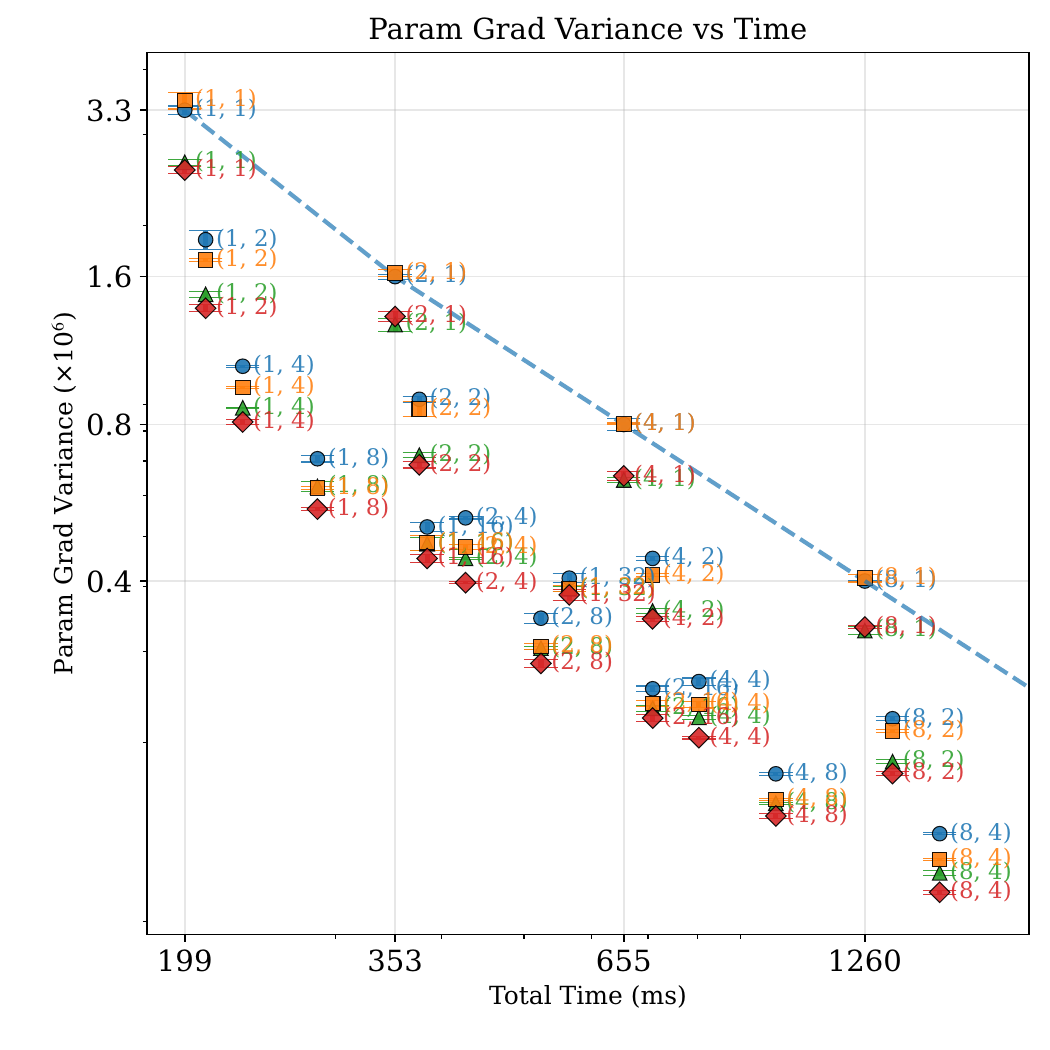}};
                    \node[left=of img11, node distance=0cm, rotate=90, xshift=1.25cm, yshift=-.9cm,  font=\color{black}]{\scriptsize{Variance $(\times10^6)$}};
                    \node[below=of img11, node distance=0cm, xshift=0.0cm, yshift=1.15cm,  font=\color{black}]{\scriptsize{Per-Iteration Compute (ms)}};

                    \node [right=of img11, node distance=0cm, xshift=-0.75cm](img21){\includegraphics[trim={1.4cm 1.21cm 0cm 0.8cm}, clip, width=.3\linewidth]{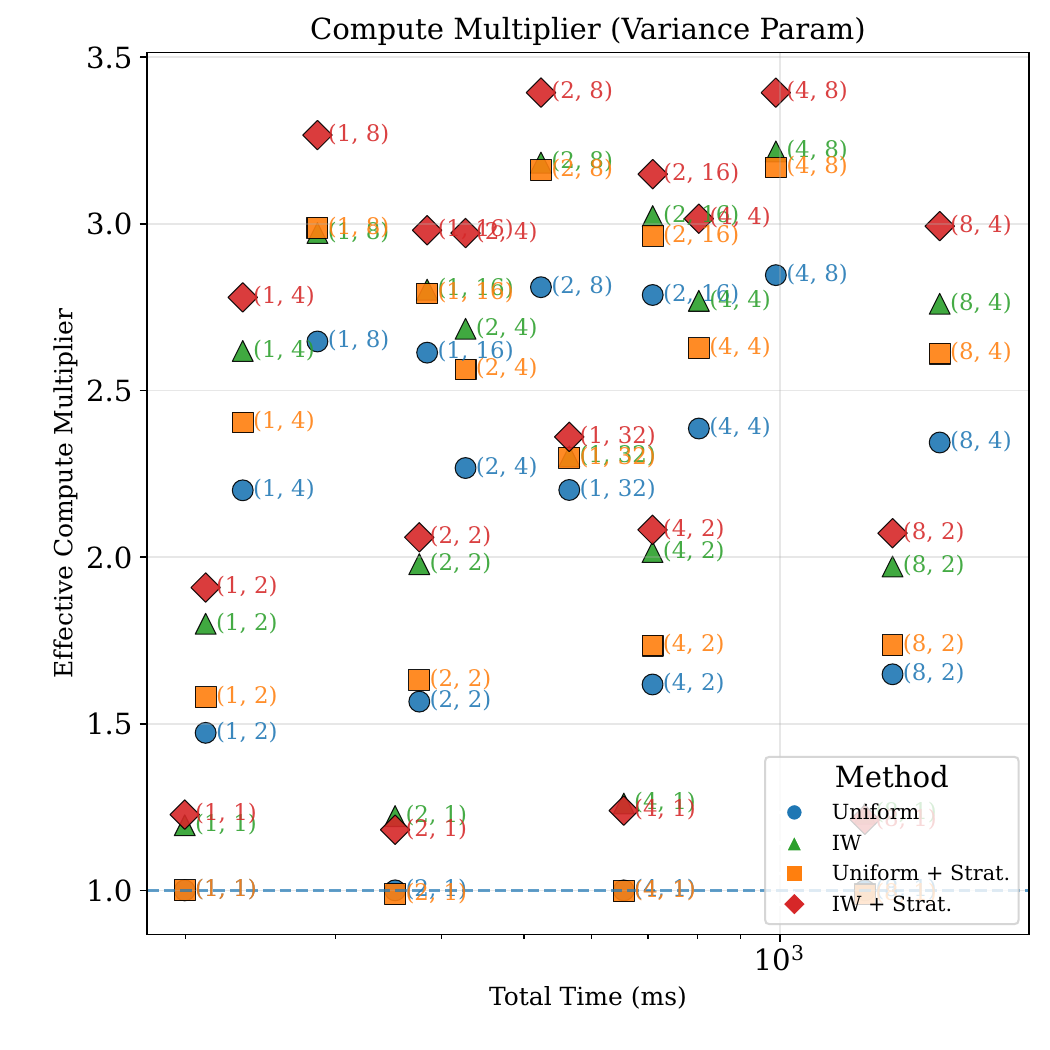}};
                    \node[left=of img21, node distance=0cm, rotate=90, xshift=2.1cm, yshift=-.9cm,  font=\color{black}]{\scriptsize{Effective Compute Mult. to Baseline}};
                    \node[below=of img21, node distance=0cm, xshift=0.0cm, yshift=1.15cm,  font=\color{black}]{\scriptsize{Per-Iteration Compute (ms)}};

                    \node [below=of img11, xshift=2.22cm, yshift=0.6cm](img31){\includegraphics[trim={1.39cm 1.21cm 0cm 0.8cm}, clip, width=.3\linewidth]{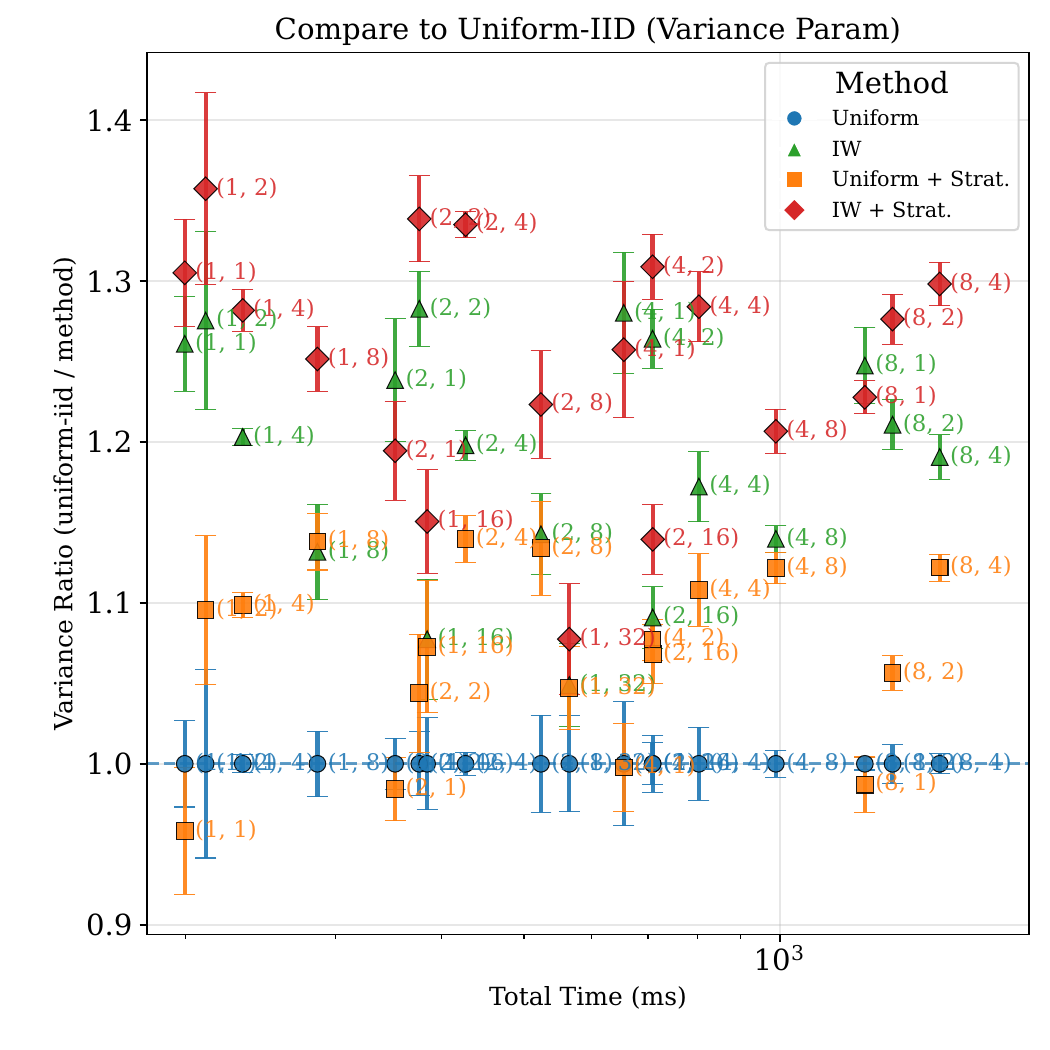}};
                    \node[left=of img31, node distance=0cm, rotate=90, xshift=2.1cm, yshift=-.9cm,  font=\color{black}]{\scriptsize{Relative Efficiency to Uniform}}; 
                    \node[below=of img31, node distance=0cm, xshift=0.0cm, yshift=1.15cm,  font=\color{black}]{\scriptsize{Per-Iteration Compute (ms)}};
                \end{tikzpicture}
                }
                \caption{
                    \textbf{Variance reduction with Monte Carlo seed error bars (single SDS prompt).}
                    Same axes as \Fig~\ref{fig:quantifying_variance_hierarchical_cost_aware_iw_strat}, restricted to one prompt and overlaid with error bars of $\pm 1$ s.d.\ across $4$ independent Monte Carlo seeds for the variance estimator.
                    \emph{Top left:} variance vs.\ compute. \emph{Top right:} effective compute multiplier relative to the uniform $\numReNoises\!=\!1$ baseline curve. \emph{Bottom:} relative efficiency vs.\ uniform at matched $(\numRenders,\numReNoises)$.
                    The relative ranking of methods is stable across seeds: IW+Strat consistently dominates and the per-seed dispersion is small relative to the gap to uniform, so the conclusions are not artifacts of estimator randomness.
                }\label{fig:error_quantifying_variance_hierarchical_cost_aware_iw_strat}
            \end{figure}

\begin{figure}[h!]
                \centering
                \resizebox{\linewidth}{!}{
                \begin{tikzpicture}
                \centering
                    \node (img11){\includegraphics[trim={1.3cm 1.21cm 0cm 0.8cm}, clip, width=.3\linewidth]{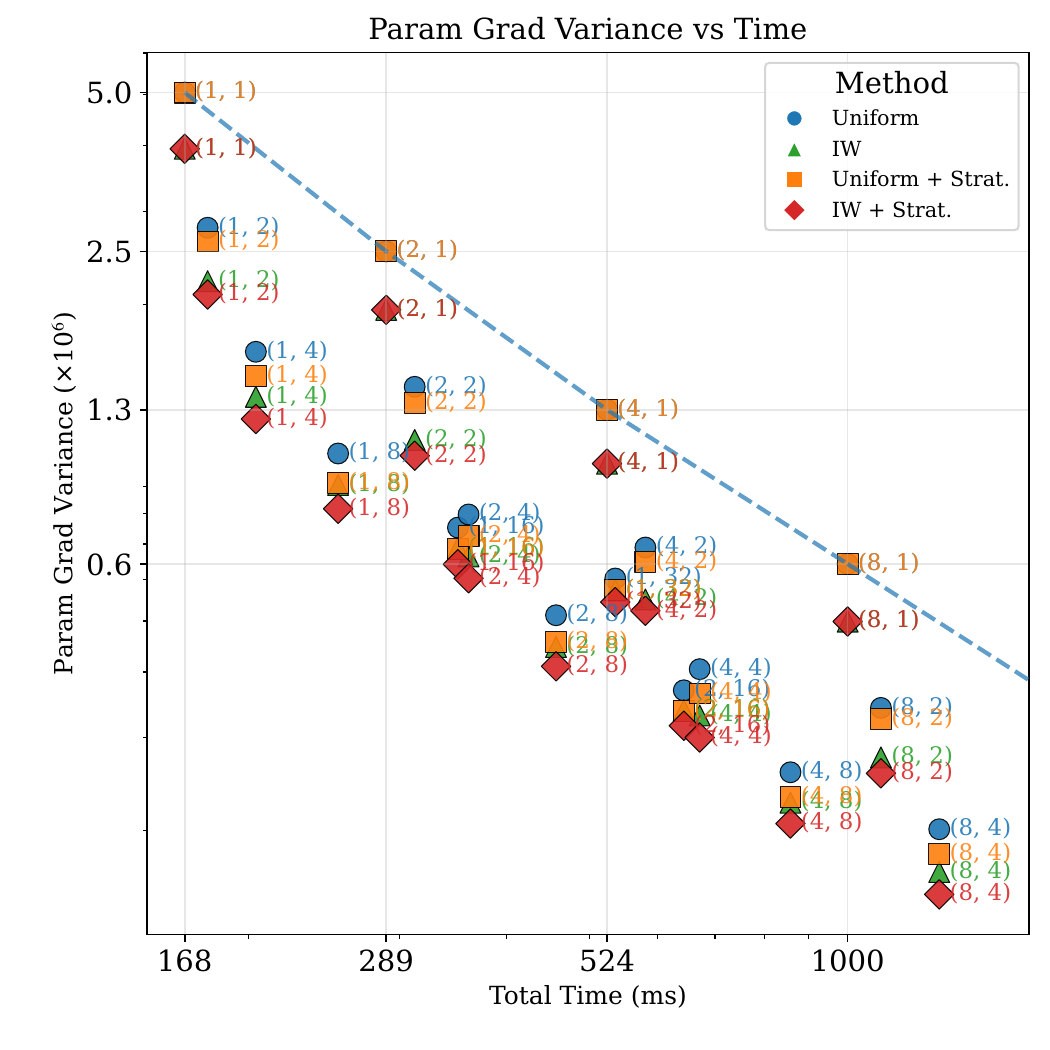}};
                    \node[left=of img11, node distance=0cm, rotate=90, xshift=1.25cm, yshift=-.9cm,  font=\color{black}]{\scriptsize{Variance $(\times10^6)$}};
                    \node[below=of img11, node distance=0cm, xshift=0.0cm, yshift=1.15cm,  font=\color{black}]{\scriptsize{Per-Iteration Compute (ms)}};

                    \node [right=of img11, node distance=0cm, xshift=-0.75cm](img21){\includegraphics[trim={1.4cm 1.21cm 0cm 0.8cm}, clip, width=.3\linewidth]{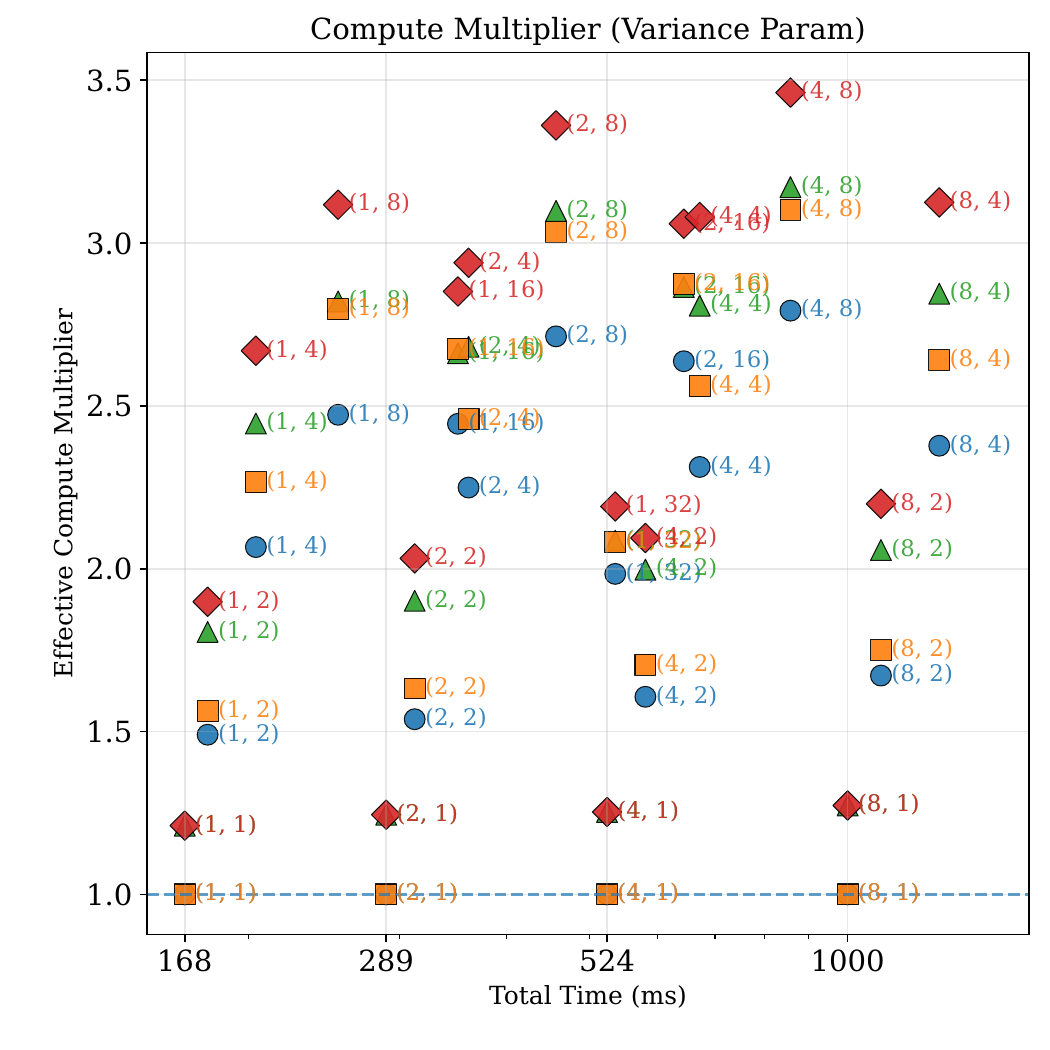}};
                    \node[left=of img21, node distance=0cm, rotate=90, xshift=2.1cm, yshift=-.9cm,  font=\color{black}]{\scriptsize{Effective Compute Mult. to Baseline}};
                    \node[below=of img21, node distance=0cm, xshift=0.0cm, yshift=1.15cm,  font=\color{black}]{\scriptsize{Per-Iteration Compute (ms)}};

                    \node [below=of img11, xshift=2.22cm, yshift=0.6cm](img31){\includegraphics[trim={1.2cm 1.21cm 0cm 0.8cm}, clip, width=.3\linewidth]{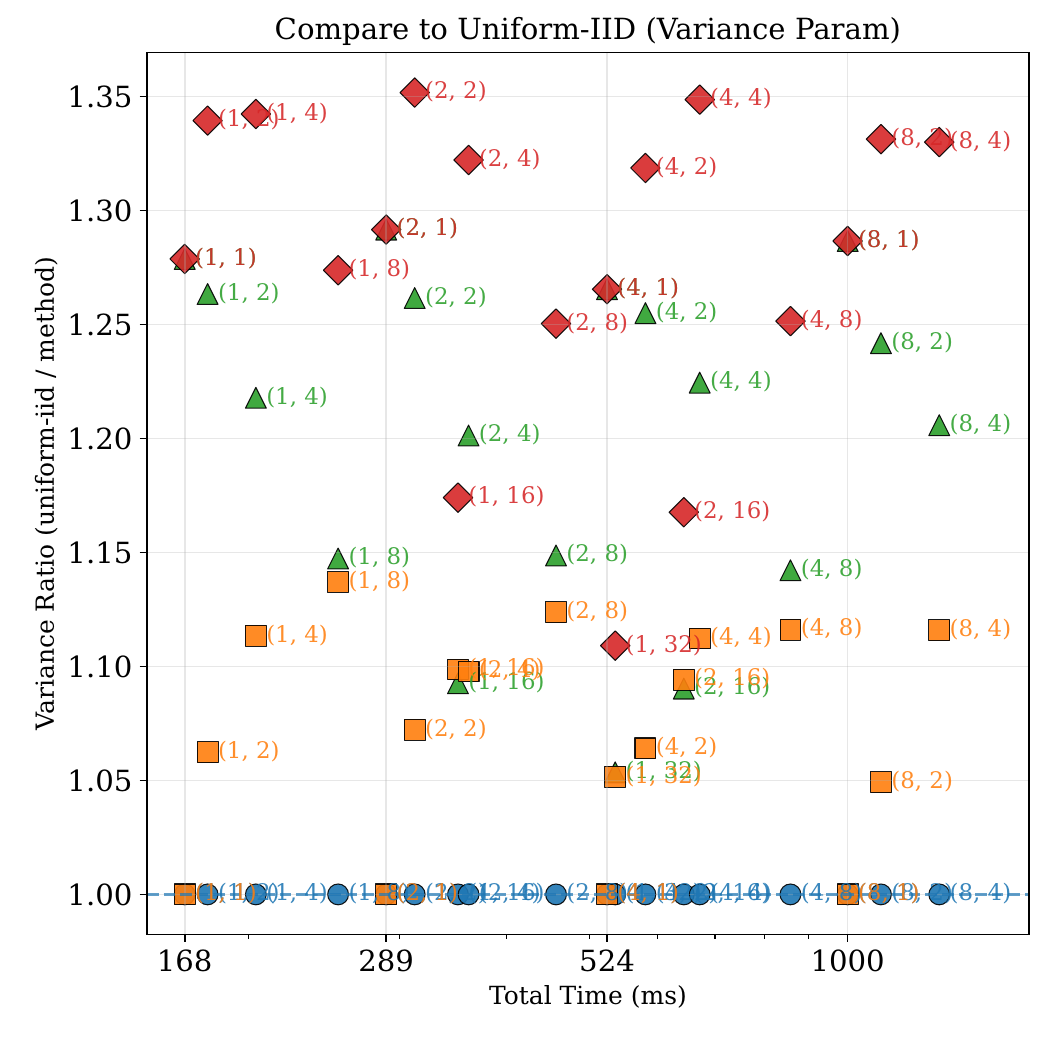}};
                    \node[left=of img31, node distance=0cm, rotate=90, xshift=2.1cm, yshift=-.9cm,  font=\color{black}]{\scriptsize{Relative Efficiency to Uniform}}; 
                    \node[below=of img31, node distance=0cm, xshift=0.0cm, yshift=1.15cm,  font=\color{black}]{\scriptsize{Per-Iteration Compute (ms)}};
                \end{tikzpicture}
                }
                \caption{
                    \textbf{Quantifying variance reduction from hierarchical cost awareness with importance weighting (IW) and stratification (Strat.).}
                    Combined effect of IW, stratification, and compute reuse on variance and ECM.
                    \emph{Top left:} Variance (MSE to the ground-truth gradient late in SDS training, which approximates the variance for unbiased estimators) versus compute. Markers: uniform (blue circles), IW (green triangles), Strat (orange squares), IW+Strat (red diamonds); points annotated by $(\numRenders,\numReNoises)$. Compared with uniform $(\numRenders\!=\!2, \numReNoises\!=\!1)$, IW+Strat $(\numRenders\!=\!1, \numReNoises\!=\!4)$ achieves about two-thirds of the cost and about half the variance.
                    \emph{Top right:} ECM relative to the uniform $\numReNoises\!=\!1$ baseline curve (\Sec~\ref{sec:method-variance-framework}). Averaged over $\numRenders$, $\numReNoises\!=\!8$ gives $\sim\!2.6\times$ (uniform), $\sim\!3.0\times$ (IW), $\sim\!3.0\times$ (Strat.), $\sim\!3.3\times$ (IW+Strat; \Tab~\ref{tab:absolute_ecm_by_m_param}).
                    \emph{Bottom:} RE isolating IW/Strat gains at fixed $(\numRenders,\numReNoises)$: Strat $\sim\!6$--$12\%$, IW $\sim\!14$--$23\%$, combined $\sim\!25$--$31\%$ over the recommended sweet spot $\numReNoises\!\in\!\{2,4,8\}$. The \Tab~\ref{tab:relative_improvement_by_m_param} envelope $\numReNoises\!\in\!\{1,\dots,32\}$ widens to $\sim\!5$--$24\%$ (IW), $\sim\!0$--$12\%$ (Strat), $\sim\!10$--$31\%$ (combined); gains shrink at very large $\numReNoises$ as within-render variance saturates and across-render variability binds.
                    Main-body \Fig~\ref{fig:quantifying_variance_hierarchical_cost_aware_iw_strat_main} keeps only uniform and IW+Strat for clarity.
                }\label{fig:quantifying_variance_hierarchical_cost_aware_iw_strat}
            \end{figure}

\begin{figure}[h!]
                \centering
                \scalebox{1.0}{
                \begin{tikzpicture}
                    \def\imgw{.28\linewidth}
                    \def\xsep{-0.9cm}
                    \def\ysep{1.1cm}
                
                    \node (img11) {\includegraphics[trim={0 0 0 0}, clip, width=\imgw]{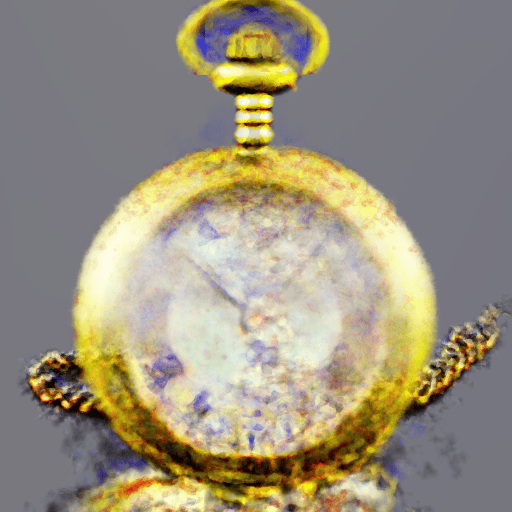}};
                    \node (img12) [right=of img11, xshift=\xsep] {\includegraphics[trim={0 0 0 0}, clip, width=\imgw]{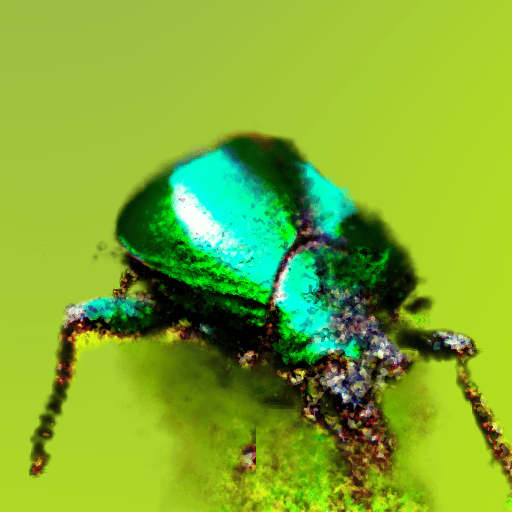}};
                    \node (img13) [right=of img12, xshift=\xsep] {\includegraphics[trim={0 0 0 0}, clip, width=\imgw]{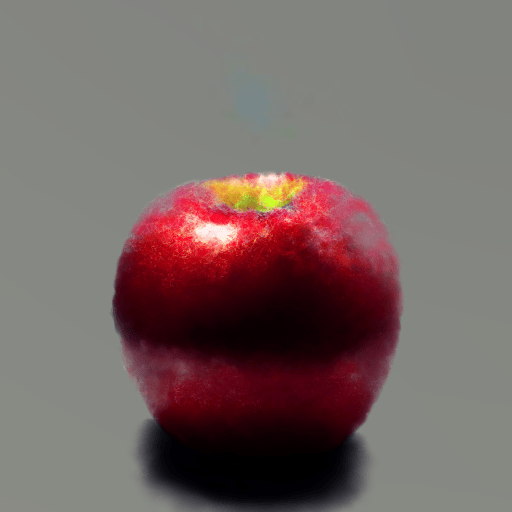}};
    
                    \node (img21) [below=of img11, yshift=\ysep] {\includegraphics[trim={0 0 0 0}, clip, width=\imgw]{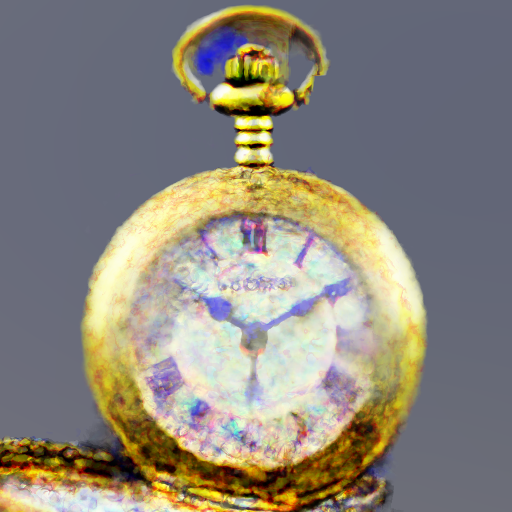}};
                    \node (img22) [right=of img21, xshift=\xsep] {\includegraphics[trim={0 0 0 0}, clip, width=\imgw]{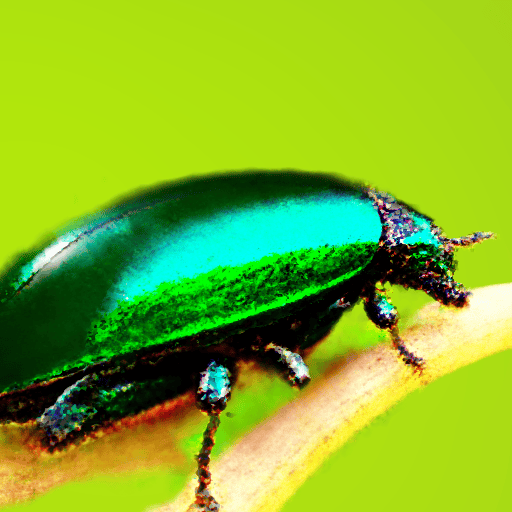}};
                    \node (img23) [right=of img22, xshift=\xsep] {\includegraphics[trim={0 0 0 0}, clip, width=\imgw]{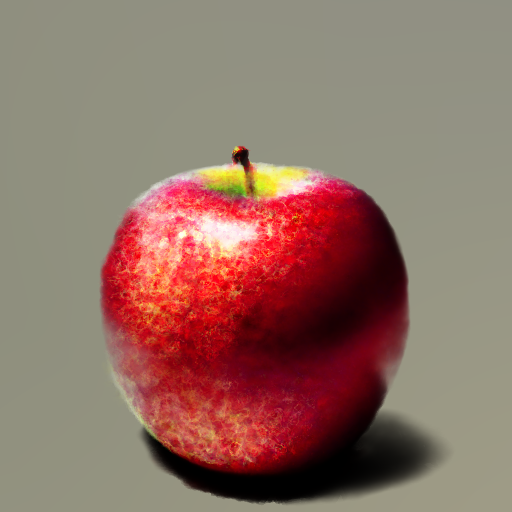}};
                
                    \node[above=of img11, node distance=0cm, yshift=-1.15cm, font=\scriptsize\color{black}] {``\emph{An antique gold pocket watch}''};
                    \node[above=of img12, node distance=0cm, yshift=-1.15cm, font=\scriptsize\color{black}] {``\emph{A shiny emerald green beetle}''};
                    \node[above=of img13, node distance=0cm, yshift=-1.15cm, font=\scriptsize\color{black}] {``\emph{A shiny red apple}''};
    
                    \node[left=of img11, rotate=90, node distance=0cm, xshift=1.0cm, yshift=-.9cm, font=\color{black}] {\scriptsize{Baseline, $(4, 1)$}};
                    \node[left=of img21, rotate=90, node distance=0cm, xshift=1.15cm, yshift=-.9cm, font=\color{black}] {\scriptsize{Ours, IW+Strat, $(1, 16)$}};
                \end{tikzpicture}
                }
                \caption{
                    \textbf{Qualitative Results from Variance Reduction:}
                    We show renders for several prompts at the end of training from \Fig~\ref{fig:clip_sds}. On the top, we show renders from a baseline method, while on the bottom, we display a reduced-variance method that combines stratified sampling, importance sampling, and re-noising. Ours costs less per iteration (\Fig~\ref{fig:quantifying_variance_hierarchical_cost_aware_iw_strat_main}); both methods run the same number of iterations and are unbiased estimators, yet our reduced-variance strategy (\Fig~\ref{fig:quantifying_variance_hierarchical_cost_aware_iw_strat}) yields higher visual quality. \Fig~\ref{fig:qualitative_sds_trajectory} shows renders throughout the optimization trajectory.
                }
              \label{fig:qualitative_sds}
            \end{figure}

            \Fig~\ref{fig:quantifying_variance_hierarchical_cost_aware_iw_strat} compares all methods by variance, ECM and RE; \Fig~\ref{fig:error_quantifying_variance_hierarchical_cost_aware_iw_strat} repeats it for one prompt with seed error bars, \Fig~\ref{fig:quantifying_variance_low_guidance_over_training} tracks variance through training at low guidance, and \Fig~\ref{fig:qualitative_sds} shows final renders.

                    \begin{figure}[h!]
            \centering
            \resizebox{0.92\linewidth}{!}{
            \begin{tikzpicture}
            \centering
                \node (img11){\includegraphics[trim={1.3cm 1.21cm 0cm 0.8cm}, clip, width=.3\linewidth]{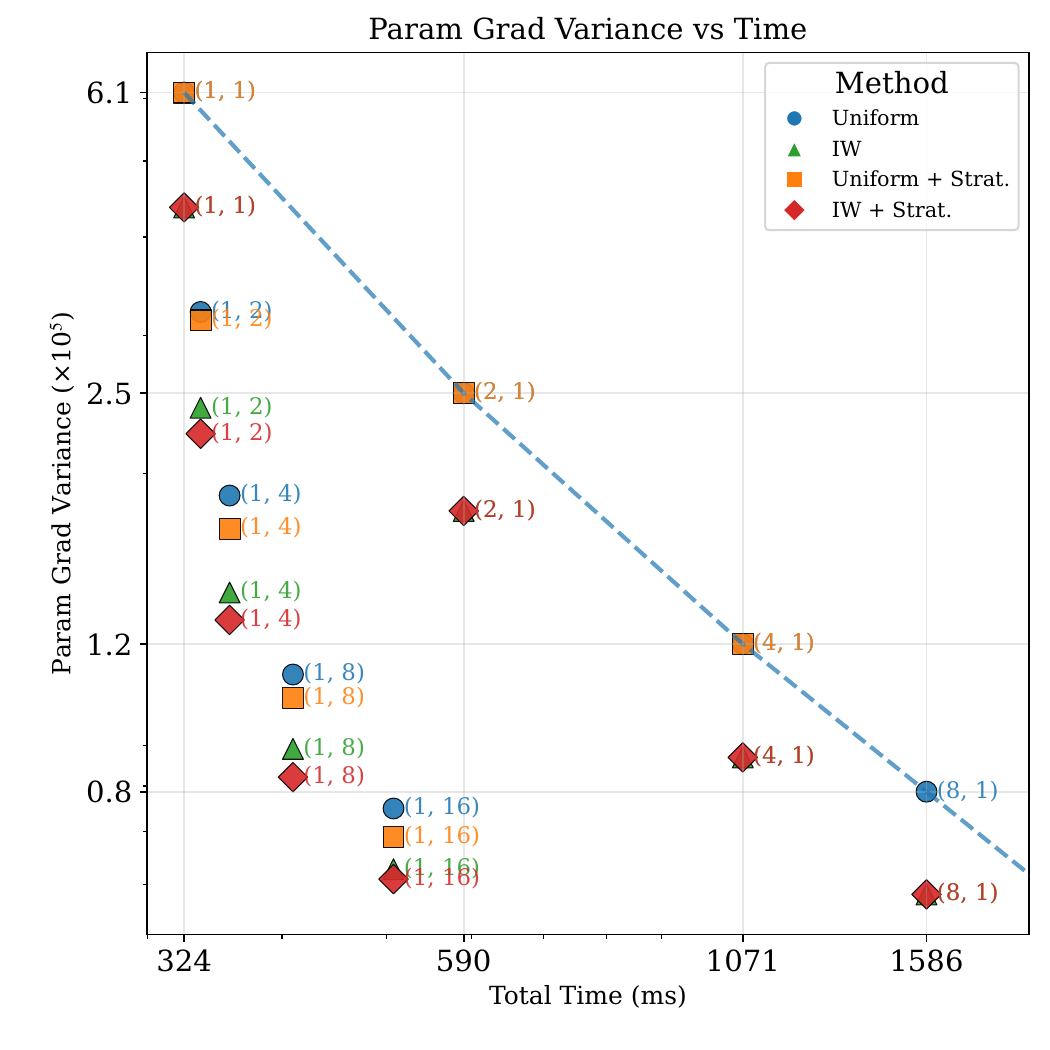}};
                \node [right=of img11, node distance=0cm, xshift=-0.75cm](img12){\includegraphics[trim={1.4cm 1.21cm 0cm 0.8cm}, clip, width=.3\linewidth]{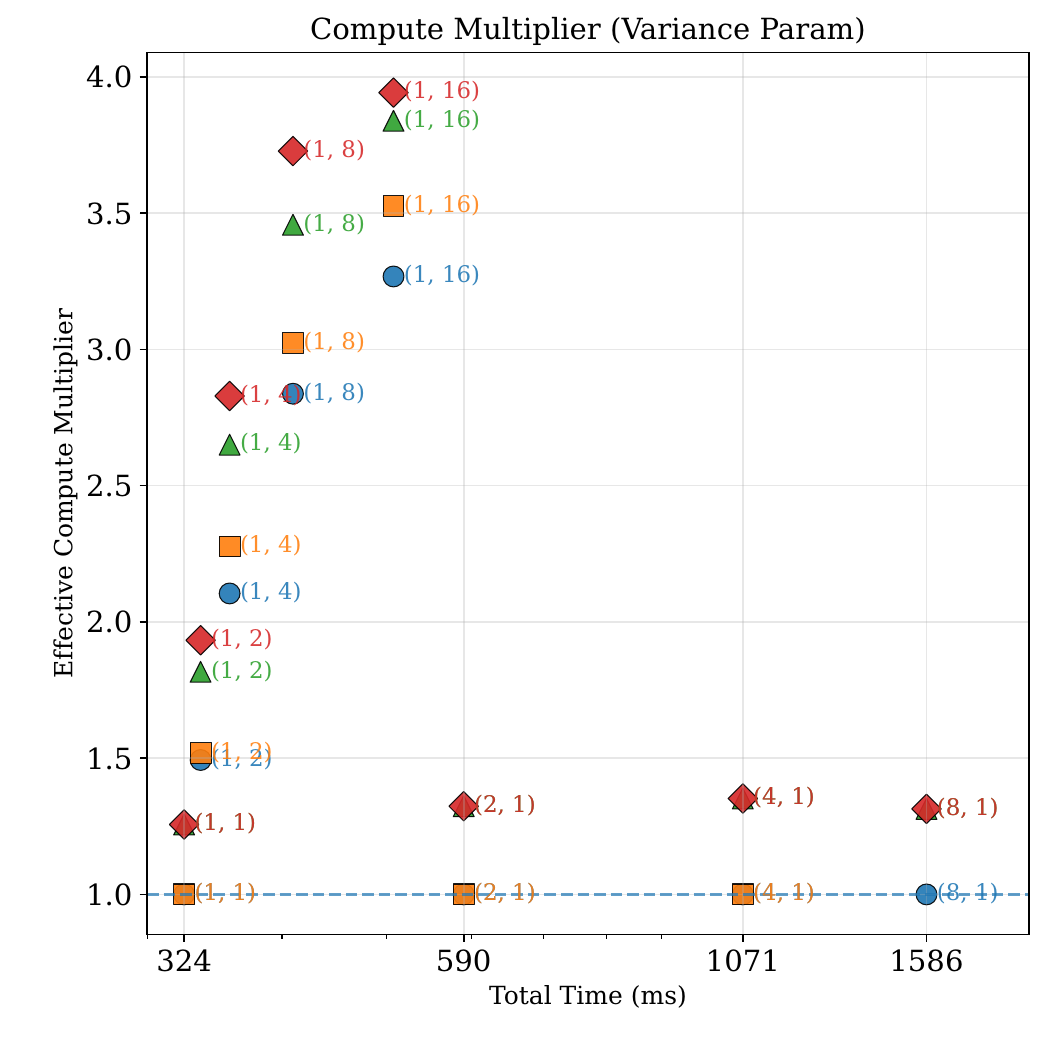}};
                \node [below=of img11, node distance=0cm, yshift=1.2cm](img21){\includegraphics[trim={1.3cm 1.21cm 0cm 0.8cm}, clip, width=.3\linewidth]{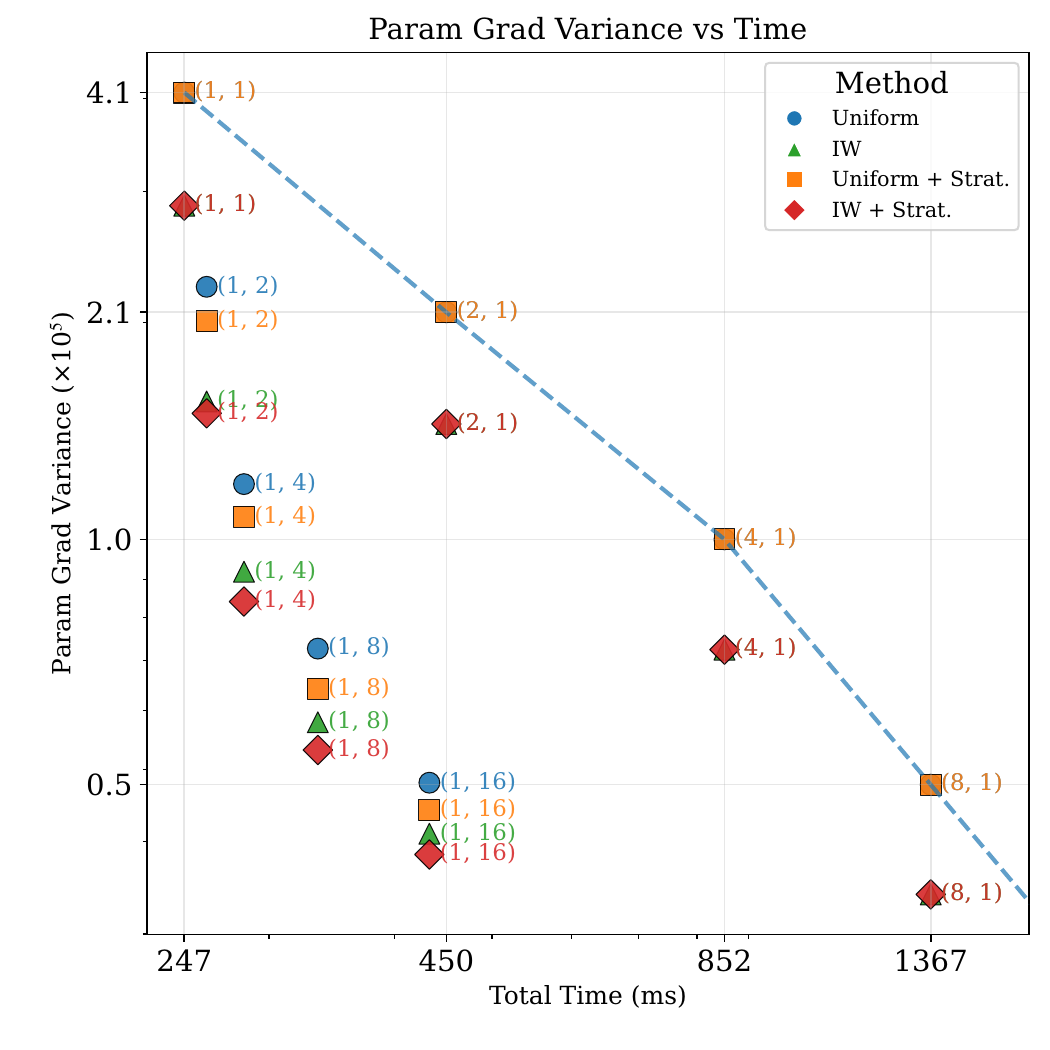}};
                \node [right=of img21, node distance=0cm, xshift=-0.75cm](img22){\includegraphics[trim={1.4cm 1.21cm 0cm 0.8cm}, clip, width=.3\linewidth]{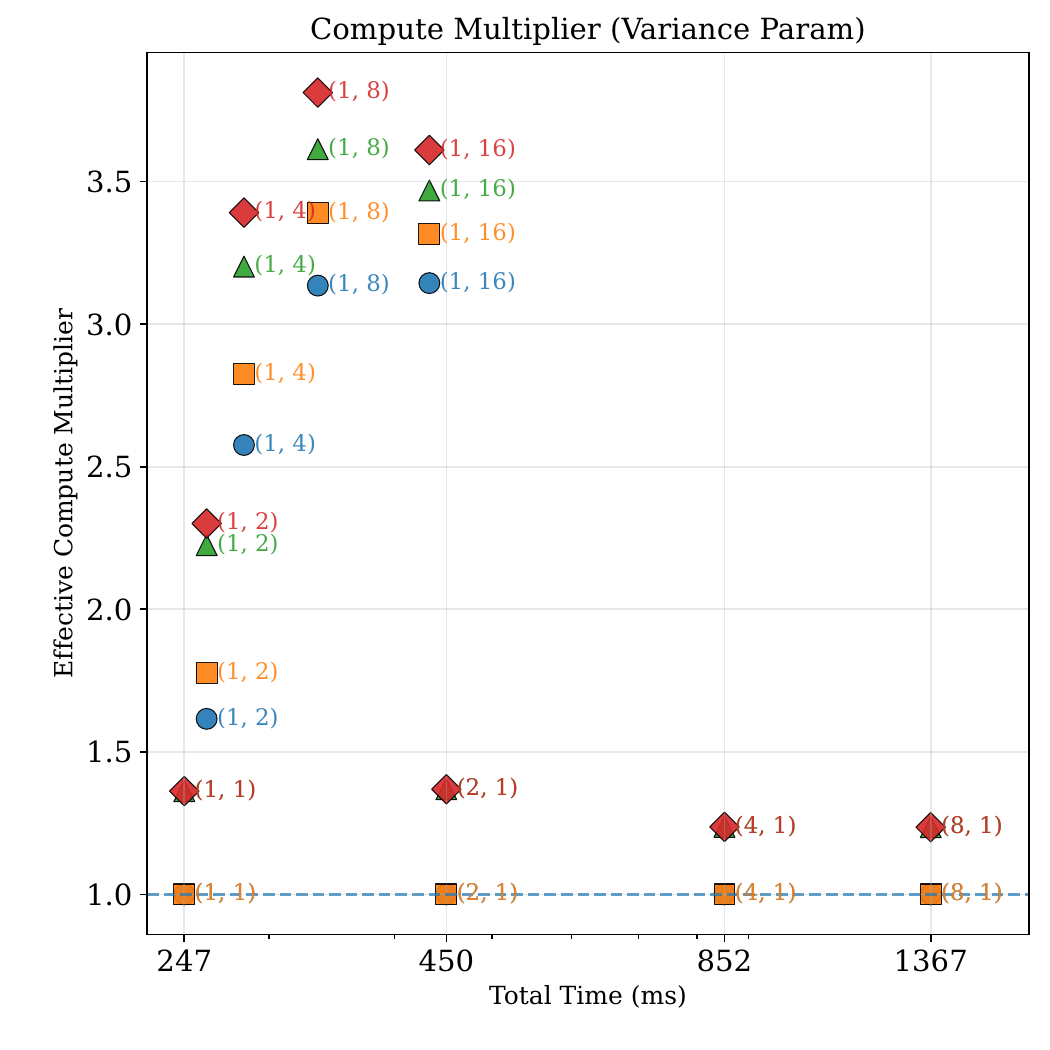}};
                \node [below=of img21, node distance=0cm, yshift=1.2cm](img31){\includegraphics[trim={1.3cm 1.21cm 0cm 0.8cm}, clip, width=.3\linewidth]{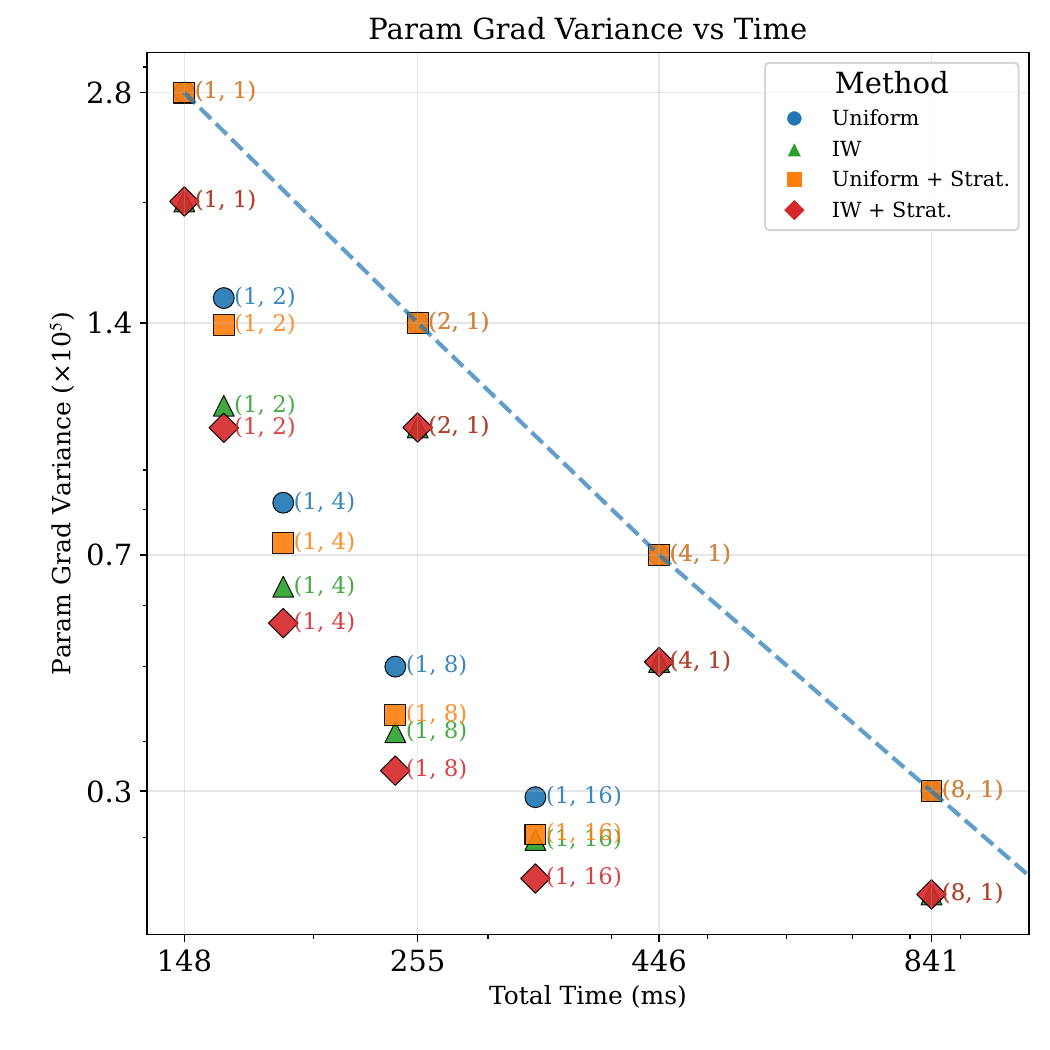}};
                \node [right=of img31, node distance=0cm, xshift=-0.75cm](img32){\includegraphics[trim={1.4cm 1.21cm 0cm 0.8cm}, clip, width=.3\linewidth]{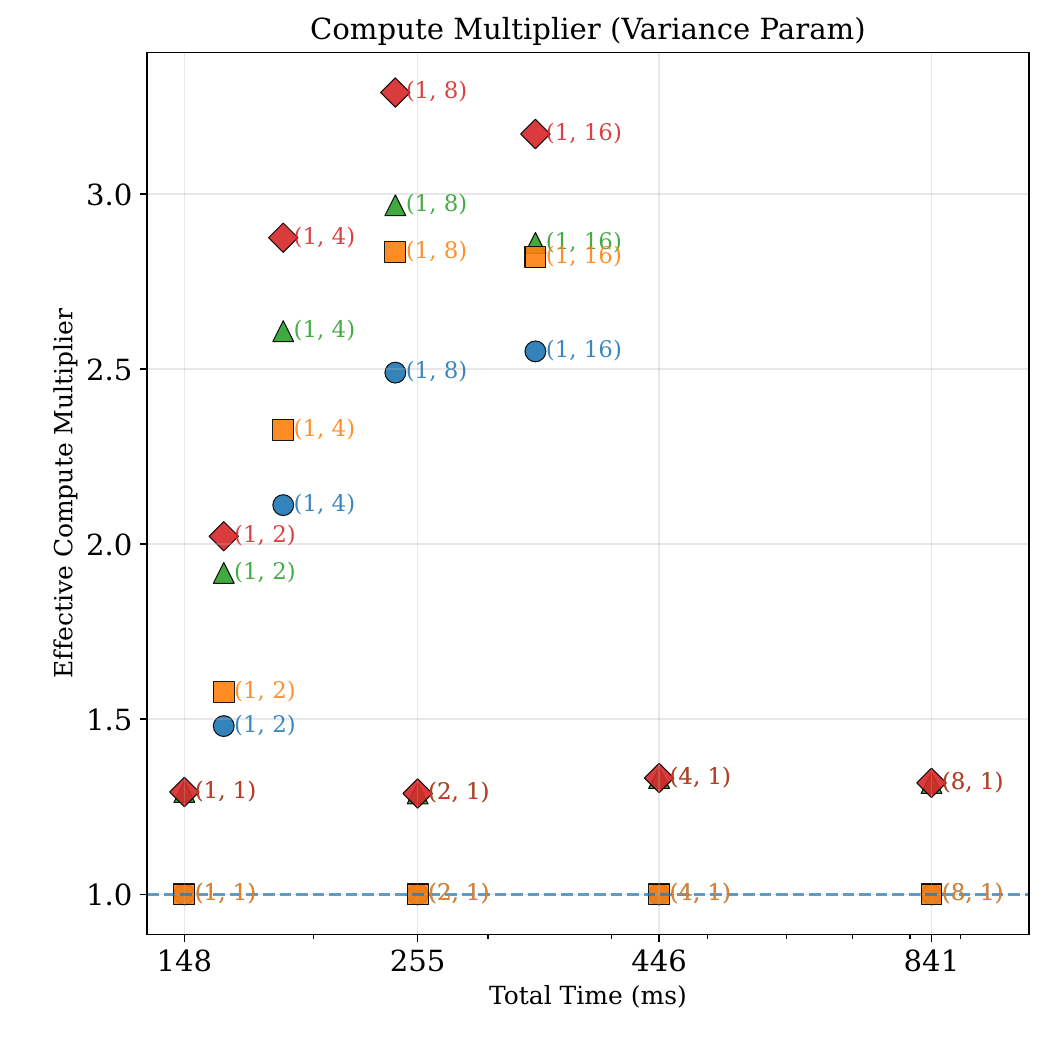}};
                \node[above=of img11, node distance=0cm, xshift=0.0cm, yshift=-1.2cm, font=\fontsize{7.4}{8.9}\selectfont\color{black}]{Variance $(\times10^5)$};
                \node[above=of img12, node distance=0cm, xshift=0.0cm, yshift=-1.2cm, font=\fontsize{7.4}{8.9}\selectfont\color{black}]{Effective Compute Mult.};
                \node[left=of img11, node distance=0cm, xshift=1.0cm, yshift=0.75cm, rotate=90, font=\fontsize{7.4}{8.9}\selectfont\color{black}]{Step $\num{1000}$};
                \node[left=of img21, node distance=0cm, xshift=1.0cm, yshift=0.75cm, rotate=90, font=\fontsize{7.4}{8.9}\selectfont\color{black}]{Step $\num{2000}$};
                \node[left=of img31, node distance=0cm, xshift=1.0cm, yshift=0.75cm, rotate=90, font=\fontsize{7.4}{8.9}\selectfont\color{black}]{Step $\num{9000}$};
                \node[below=of img31, node distance=0cm, xshift=0.0cm, yshift=1.2cm, font=\fontsize{7.4}{8.9}\selectfont\color{black}]{Per-Iteration Compute (ms)};
                \node[below=of img32, node distance=0cm, xshift=0.0cm, yshift=1.2cm, font=\fontsize{7.4}{8.9}\selectfont\color{black}]{Per-Iteration Compute (ms)};
            \end{tikzpicture}
            }
            \caption{
                \textbf{Variance reduction across training, low classifier-free guidance ($\cfgScale\!=\!25$).}
                Analogous to \Fig~\ref{fig:quantifying_variance_hierarchical_cost_aware_iw_strat}, measured at three optimization checkpoints. Rows: training step $\num{1000}$, $\num{2000}$, and $\num{9000}$.
                \emph{Left:} variance vs.\ compute. \emph{Right:} effective compute multiplier relative to the uniform $\numReNoises\!=\!1$ baseline curve. The continuation shows relative efficiency vs.\ uniform at matched $(\numRenders,\numReNoises)$.
                Higher $\numReNoises$ wins more strongly early in training, when rendering is more expensive relative to denoising and re-noising amortizes that cost most efficiently; the gap narrows in late training but variance reduction continues to dominate the uniform baseline at every checkpoint.
            }\label{fig:quantifying_variance_low_guidance_over_training}
            \end{figure}

            \begin{figure}[h!]
            \ContinuedFloat
            \centering
            \resizebox{\linewidth}{!}{
            \begin{tikzpicture}
            \centering
                \node (img13){\includegraphics[trim={1.3cm 1.21cm 0cm 0.8cm}, clip, width=.3\linewidth]{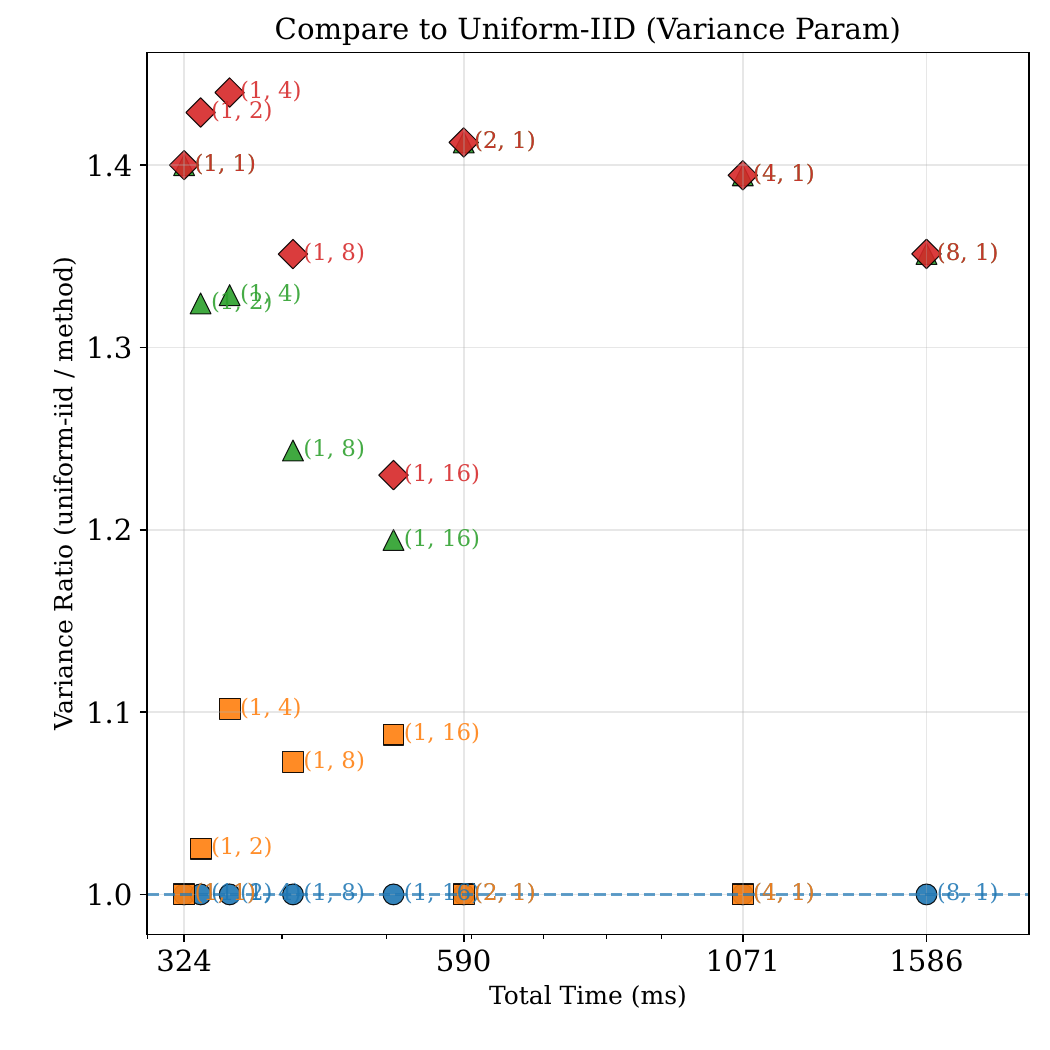}};
                \node [right=of img13, node distance=0cm, xshift=-0.75cm](img23){\includegraphics[trim={1.3cm 1.21cm 0cm 0.8cm}, clip, width=.3\linewidth]{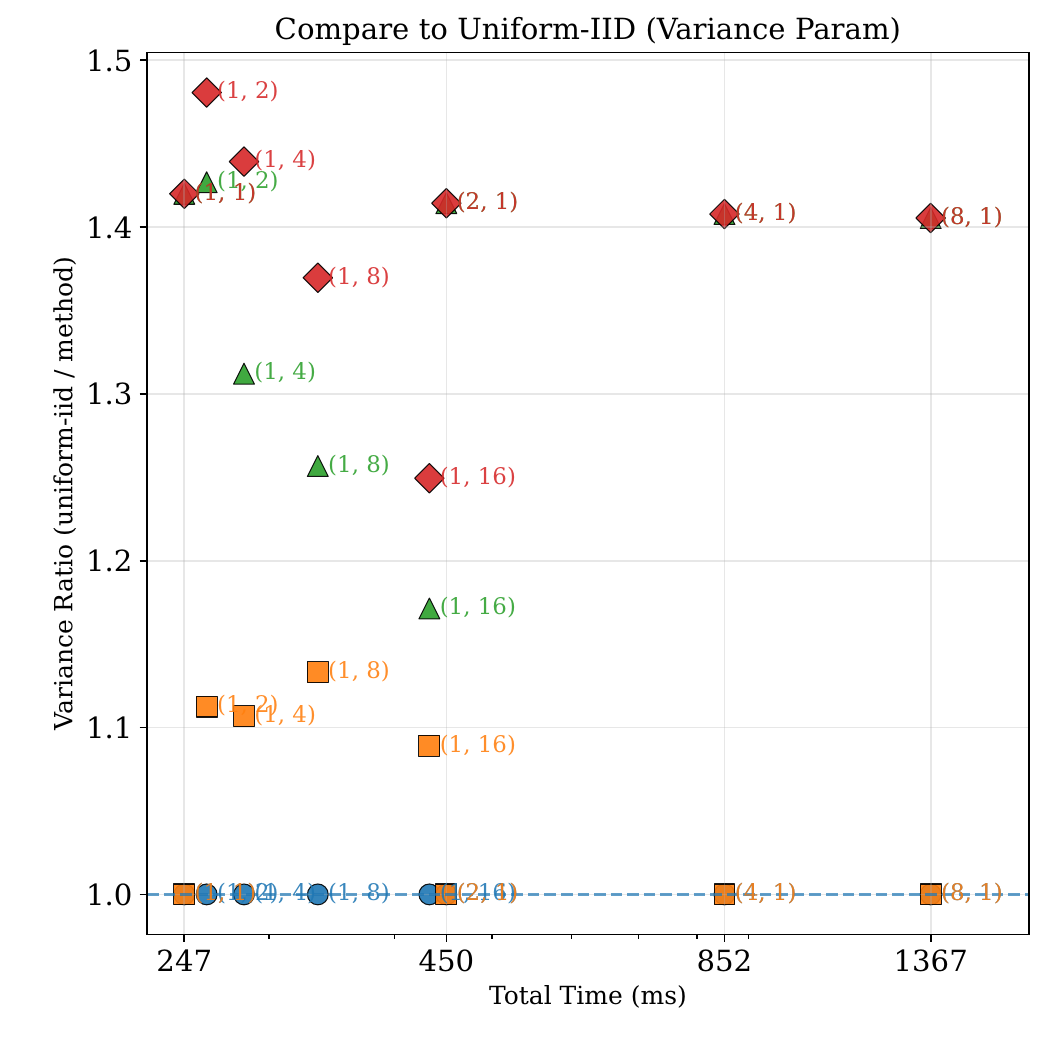}};
                \node [below=of img13, xshift=2.22cm, yshift=0.3cm](img33){\includegraphics[trim={1.3cm 1.21cm 0cm 0.8cm}, clip, width=.3\linewidth]{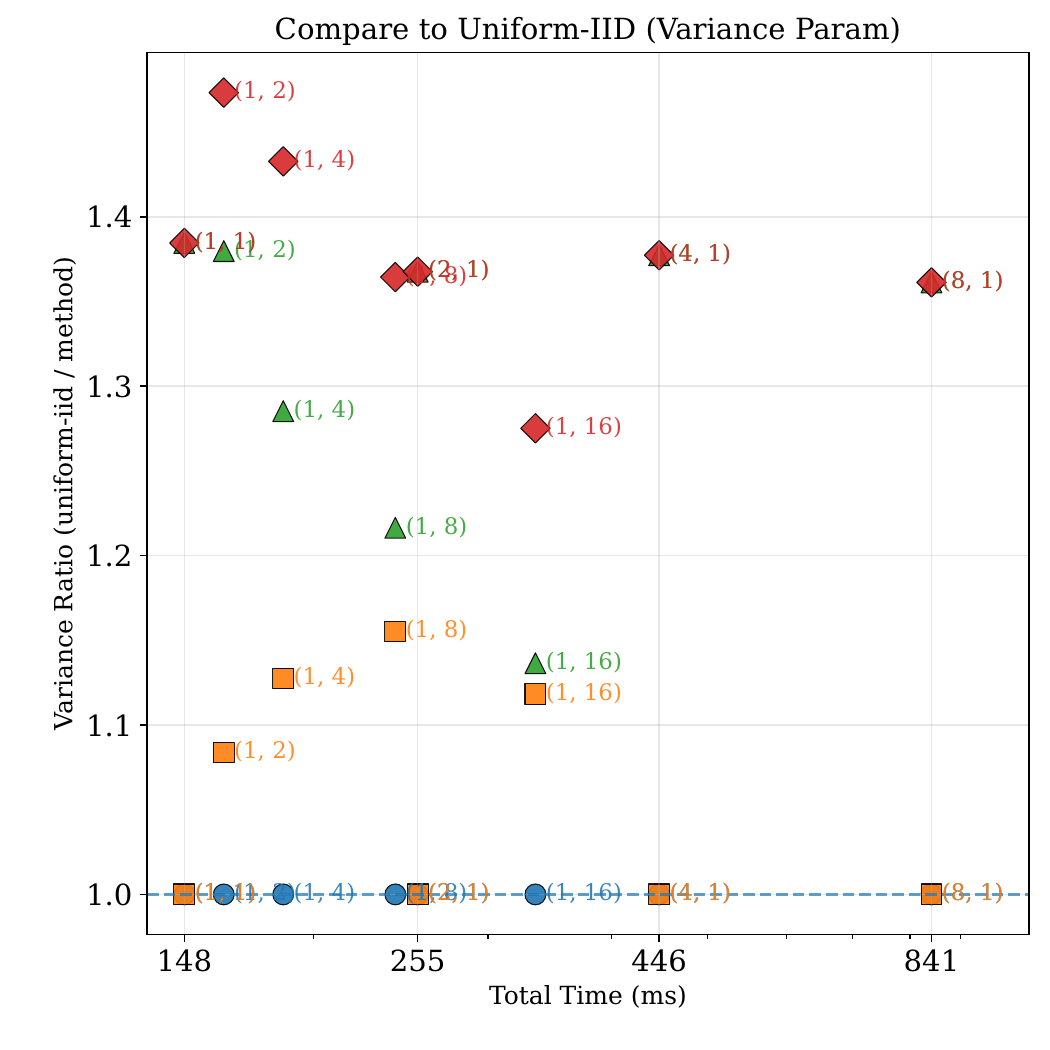}};
                \node[left=of img13, node distance=0cm, rotate=90, xshift=2.1cm, yshift=-.9cm, font=\color{black}]{\scriptsize{Relative Efficiency}};
                \node[above=of img13, node distance=0cm, xshift=0.0cm, yshift=-1.2cm, font=\color{black}]{\scriptsize{Step $\num{1000}$}};
                \node[below=of img13, node distance=0cm, xshift=0.0cm, yshift=1.15cm, font=\color{black}]{\scriptsize{Per-Iteration Compute (ms)}};
                \node[left=of img23, node distance=0cm, rotate=90, xshift=2.1cm, yshift=-.9cm, font=\color{black}]{\scriptsize{Relative Efficiency}};
                \node[above=of img23, node distance=0cm, xshift=0.0cm, yshift=-1.2cm, font=\color{black}]{\scriptsize{Step $\num{2000}$}};
                \node[below=of img23, node distance=0cm, xshift=0.0cm, yshift=1.15cm, font=\color{black}]{\scriptsize{Per-Iteration Compute (ms)}};
                \node[left=of img33, node distance=0cm, rotate=90, xshift=2.1cm, yshift=-.9cm, font=\color{black}]{\scriptsize{Relative Efficiency}};
                \node[above=of img33, node distance=0cm, xshift=0.0cm, yshift=-1.2cm, font=\color{black}]{\scriptsize{Step $\num{9000}$}};
                \node[below=of img33, node distance=0cm, xshift=0.0cm, yshift=1.15cm, font=\color{black}]{\scriptsize{Per-Iteration Compute (ms)}};
            \end{tikzpicture}
            }
            \caption[]{\textbf{(continued)} Relative efficiency vs.\ uniform at matched $(\numRenders,\numReNoises)$ at training steps $\num{1000}$, $\num{2000}$, and $\num{9000}$ ($\cfgScale\!=\!25$).}
            \end{figure}

        \subsubsection{Residual-Norm Variance Analysis}\label{sec:residual-norm-variance}
            The preceding analysis (\Fig~\ref{fig:quantifying_variance_hierarchical_cost_aware_iw_strat}) measures variance of the full parameter gradient $\nabla_{\genParams}\lossSDS$, which includes backpropagation through the renderer Jacobian. Here we present the analogous analysis for the latent-space residual $\sdsWeight(\timevar)\residual$, a commonly used proxy that avoids the cost of backpropagation.
            
\begin{figure}[h!]
                \centering
                \resizebox{\linewidth}{!}{
                \begin{tikzpicture}
                \centering
                    \node (img11){\includegraphics[trim={1.3cm 1.21cm 0cm 0.8cm}, clip, width=.3\linewidth]{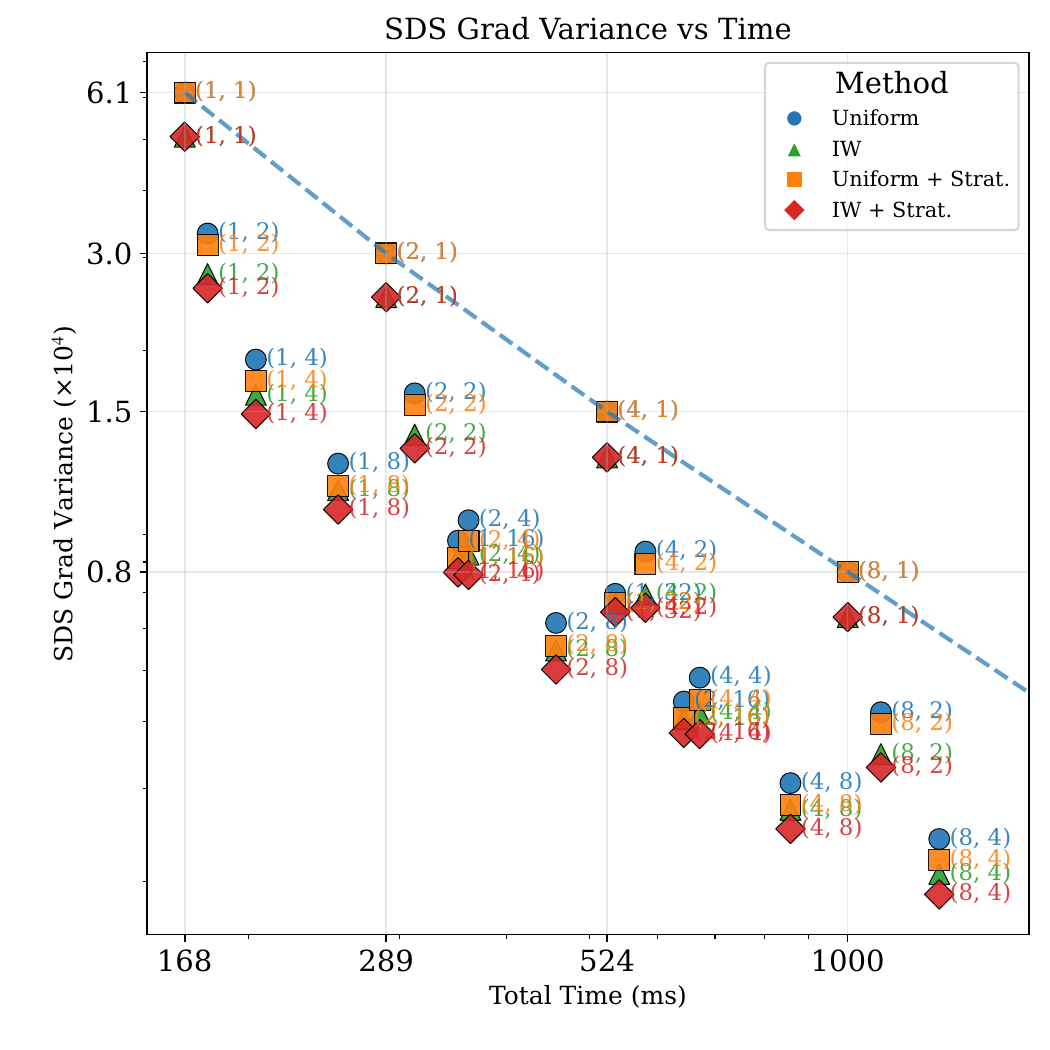}};
                    \node[left=of img11, node distance=0cm, rotate=90, xshift=1.25cm, yshift=-.9cm,  font=\color{black}]{\scriptsize{Variance $(\times10^4)$}};
                    \node[below=of img11, node distance=0cm, xshift=0.0cm, yshift=1.15cm,  font=\color{black}]{\scriptsize{Per-Iteration Compute (ms)}};
        
                    \node [right=of img11, node distance=0cm, xshift=-0.75cm](img21){\includegraphics[trim={1.4cm 1.21cm 0cm 0.8cm}, clip, width=.3\linewidth]{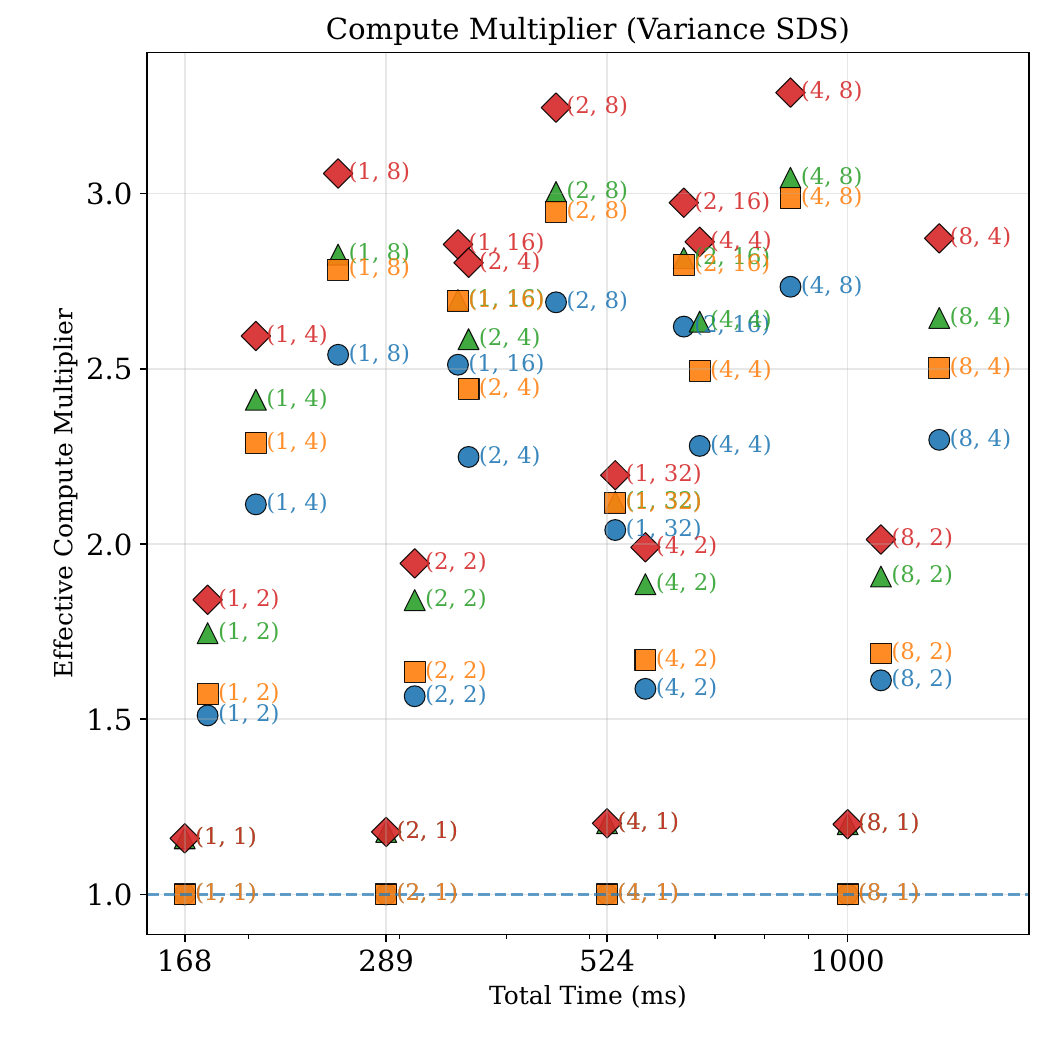}};
                    \node[left=of img21, node distance=0cm, rotate=90, xshift=2.1cm, yshift=-.9cm,  font=\color{black}]{\scriptsize{Effective Compute Mult. to Baseline}};
                    \node[below=of img21, node distance=0cm, xshift=0.0cm, yshift=1.15cm,  font=\color{black}]{\scriptsize{Per-Iteration Compute (ms)}};
        
                    \node [below=of img11, xshift=2.22cm, yshift=0.6cm](img31){\includegraphics[trim={1.2cm 1.21cm 0cm 0.8cm}, clip, width=.3\linewidth]{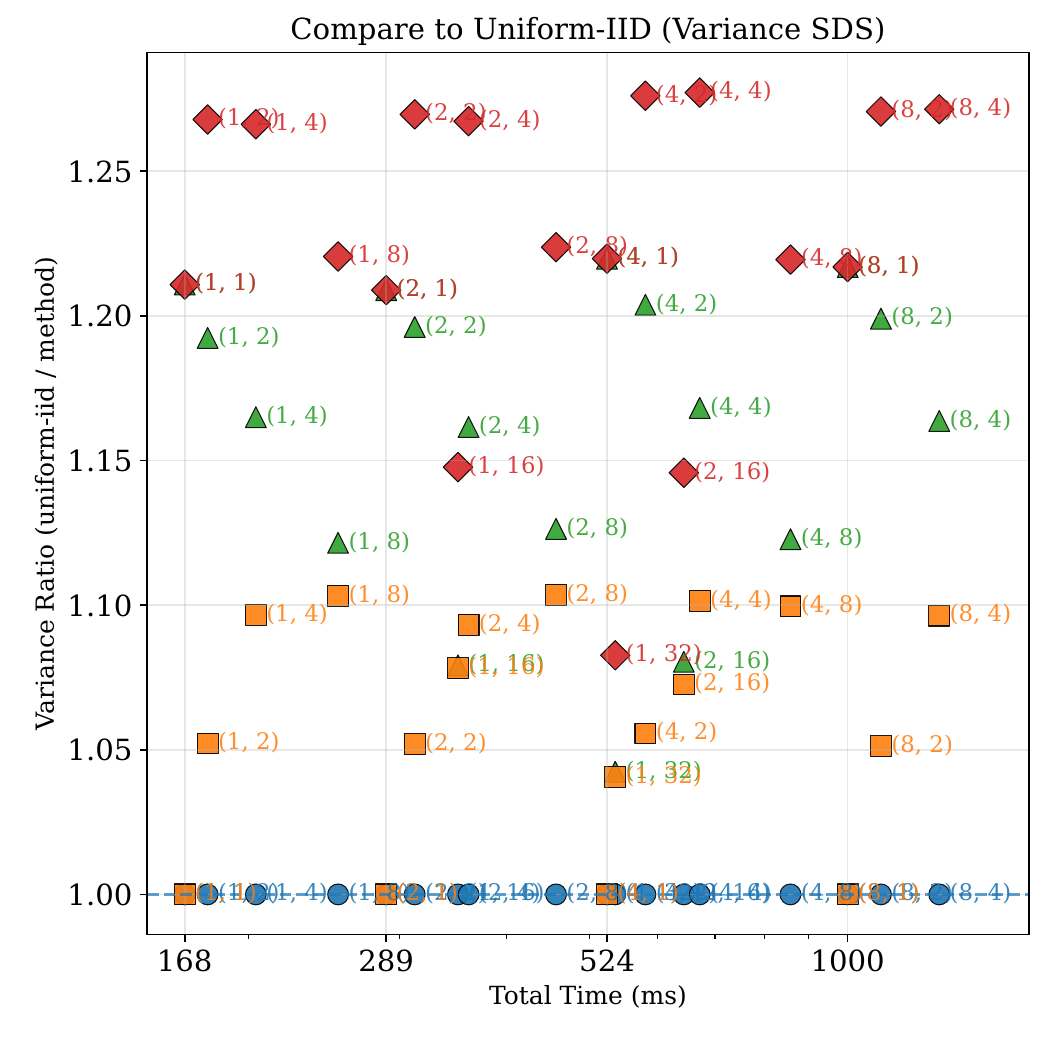}};
                    \node[left=of img31, node distance=0cm, rotate=90, xshift=2.1cm, yshift=-.9cm,  font=\color{black}]{\scriptsize{Relative Efficiency to Uniform}}; 
                    \node[below=of img31, node distance=0cm, xshift=0.0cm, yshift=1.15cm,  font=\color{black}]{\scriptsize{Per-Iteration Compute (ms)}};
                \end{tikzpicture}
                }
                \caption{
                    \textbf{Variance reduction measured via latent-space residual norm.}
                    Analogous to \Fig~\ref{fig:quantifying_variance_hierarchical_cost_aware_iw_strat}, but measuring variance of the weighted residual $\sdsWeight(\timevar)\residual$ rather than the full parameter gradient.
                    \emph{Top left:} Variance versus compute budget. Colors denote uniform baseline, IW only, stratification only, and IW+Strat combined. Points are annotated by $(\numRenders,\numReNoises)$ tuples.
                    \emph{Top right:} Effective compute multiplier relative to the uniform $\numReNoises\!=\!1$ baseline curve.
                    \emph{Bottom:} Relative efficiency to uniform sampling at matched $(\numRenders,\numReNoises)$ configurations.
                    The qualitative trends match \Fig~\ref{fig:quantifying_variance_hierarchical_cost_aware_iw_strat}: importance weighting and stratification both reduce variance, and their benefits combine. However, the residual-norm metric does not account for how the renderer Jacobian modulates per-timestep contributions, so absolute efficiency gains and optimal configurations may differ from the parameter-gradient analysis. This supports using residual-norm variance as an inexpensive diagnostic during development, while validating final design choices with parameter-gradient variance.
                }\label{fig:quantifying_variance_hierarchical_cost_aware_iw_strat_residual}
            \end{figure}

            As discussed in \Sec~\ref{sec:method-variance-framework-app}, the residual-norm metric captures variance at an intermediate stage of the gradient pipeline. While easier to compute, it does not account for how the renderer Jacobian modulates contributions across timesteps. Comparing \Fig~\ref{fig:quantifying_variance_hierarchical_cost_aware_iw_strat_residual} to \Fig~\ref{fig:quantifying_variance_hierarchical_cost_aware_iw_strat} reveals that relative efficiency gains from importance weighting and stratification are qualitatively similar, but the absolute rankings and magnitudes can differ. This confirms that residual-norm variance is a reasonable heuristic for initial guidance on estimator design, but practitioners targeting parameter-space efficiency should validate with full gradient variance when feasible.
        
        \subsubsection{Alternative Dispersion Metrics}\label{app:alternative-dispersion-metrics}
\begin{figure}[h!]
                \centering
                \scalebox{1.0}{
                \begin{tikzpicture}
                \centering
                    \node (img11){\includegraphics[trim={0.8cm 0.8cm 0cm 0.8cm}, clip, width=.7\linewidth]{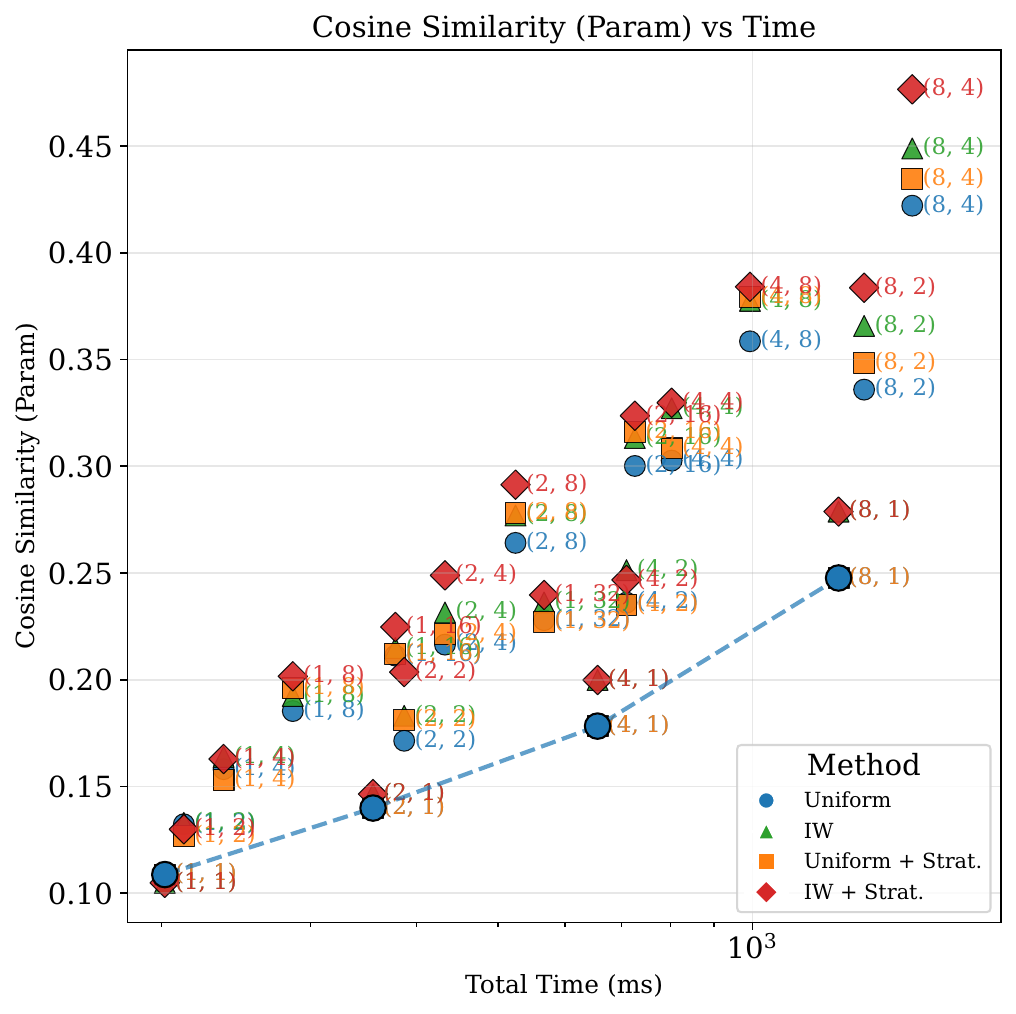}};
                    \node[left=of img11, node distance=0cm, rotate=90, xshift=1.5cm, yshift=-.9cm, font=\color{black}]{\footnotesize{Expected Cosine Similarity}};
                    \node[below=of img11, node distance=0cm, xshift=0.0cm, yshift=1.15cm, font=\color{black}]{\footnotesize{Per-Iteration Compute (ms)}};
                \end{tikzpicture}
                }
                \caption{
                    \textbf{Cosine similarity to ground-truth gradient versus compute budget.}
                    Cosine similarity between estimated and ground-truth parameter gradients (SDS) for uniform, IW, Strat.\ and IW+Strat across $(\numRenders, \numReNoises)$. Higher is better. Ranking matches the variance analysis in \Fig~\ref{fig:quantifying_variance_hierarchical_cost_aware_iw_strat_main}: IW+Strat wins at matched compute, and reuse ($\numReNoises > 1$) helps. ECM magnitudes differ from the variance metric because cosine similarity is~\mbox{scale-invariant}.
                }\label{fig:cossim_vs_time}
            \end{figure}

            Throughout the main text, we measure estimator quality using the trace-covariance variance (\Eq~\ref{eq:var}), which captures the mean-squared error relative to the ground-truth gradient. However, practitioners may also care about directional alignment between estimated and ground-truth gradients, motivating cosine similarity as an alternative dispersion metric. Let $\hat\mean$ denote the estimated gradient and $\mean$ denote the ground truth. The cosine similarity
            \begin{equation}
                \mathrm{CosSim}(\hat\mean, \mean) = \frac{\hat\mean^\top \mean}{\|\hat\mean\|_2 \|\mean\|_2}
            \end{equation}
            measures directional agreement independent of magnitude, which may be relevant when gradients are subsequently normalized or clipped. \Fig~\ref{fig:cossim_vs_time} shows expected cosine similarity to the ground-truth gradient as a function of compute budget for the SDS experiments in \Sec~\ref{sec:experiments-sds}.
        
            \textbf{Takeaways.}
                Cosine and variance metrics agree qualitatively: IW, stratification, and reuse all improve estimator quality per unit compute. ECM magnitudes can differ because cosine ignores scale, while variance captures both directional and magnitude error. When gradients are normalized downstream (e.g., Adam), cosine is the more relevant metric; we report both for completeness.

        \subsubsection{Analysis of the Low Guidance Regime}\label{sec:low_guidance_analysis}
            Low classifier-free guidance ($\cfgScale$) is an increasingly popular regime for diffusion-guided optimization, as it enables more diverse generations by reducing over-reliance on the text conditioning signal. We investigate whether our variance-reduction strategies provide amplified benefits in this regime.
            
            \textbf{Setup.}
                We repeat the variance analysis of \Fig~\ref{fig:quantifying_variance_hierarchical_cost_aware_iw_strat} at $\cfgScale\!=\!25$ (vs.\ $\cfgScale\!=\!100$ in the main run) on a subset of $11$ prompts, with longer training (\Fig~\ref{fig:clip_sds_low_guidance} runs to $\num{20000}$ steps) and other hyperparameters fixed. We measure parameter-gradient variance at step $\num{5000}$ and CLIP scores over the trajectory.
            
\begin{figure}[h!]
                \centering
                \resizebox{\linewidth}{!}{
                \begin{tikzpicture}
                \centering
                    \node (img11){\includegraphics[trim={1.3cm 1.21cm 0cm 0.8cm}, clip, width=.3\linewidth]{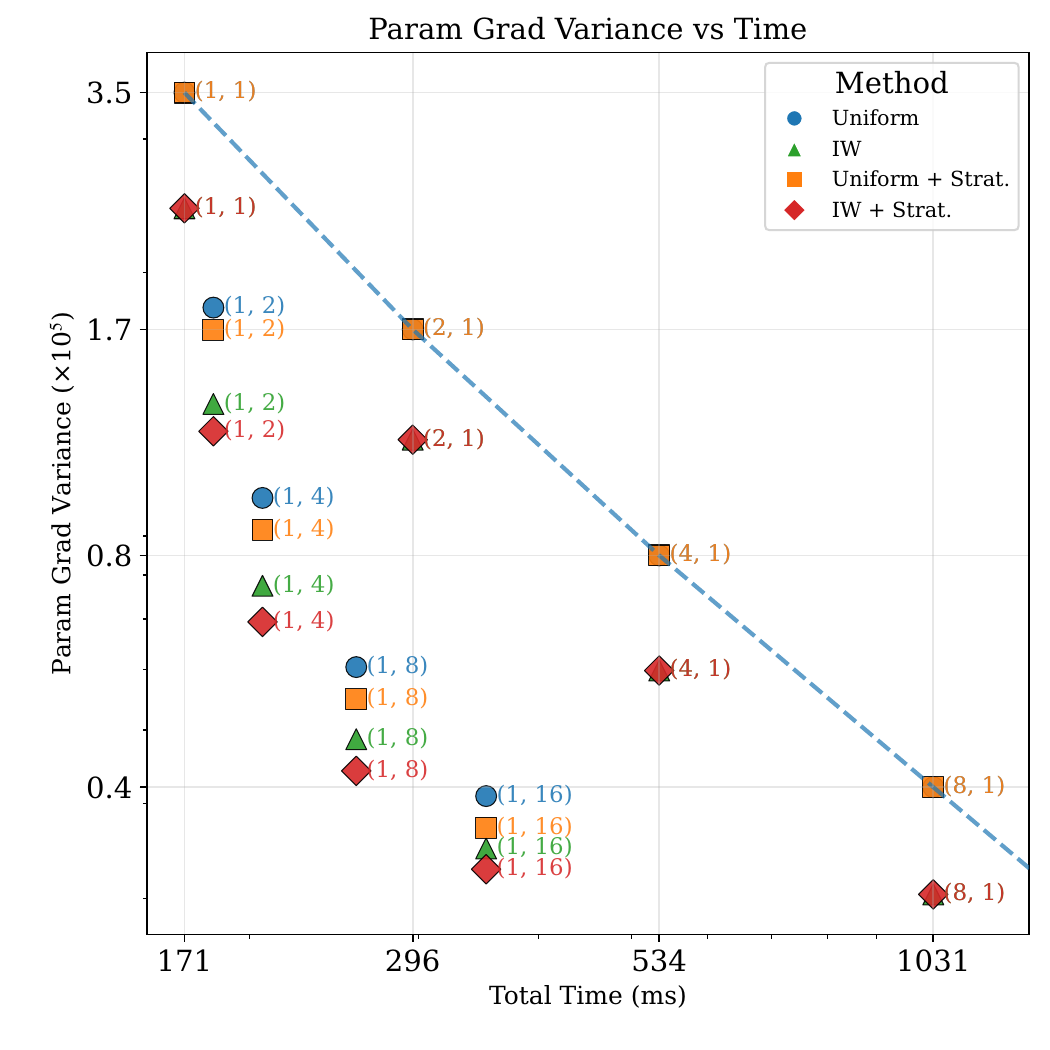}};
                    \node[left=of img11, node distance=0cm, rotate=90, xshift=1.25cm, yshift=-.9cm,  font=\color{black}]{\scriptsize{Variance $(\times10^5)$}};
                    \node[below=of img11, node distance=0cm, xshift=0.0cm, yshift=1.15cm,  font=\color{black}]{\scriptsize{Per-Iteration Compute (ms)}};

                    \node [right=of img11, node distance=0cm, xshift=-0.75cm](img21){\includegraphics[trim={1.4cm 1.21cm 0cm 0.8cm}, clip, width=.3\linewidth]{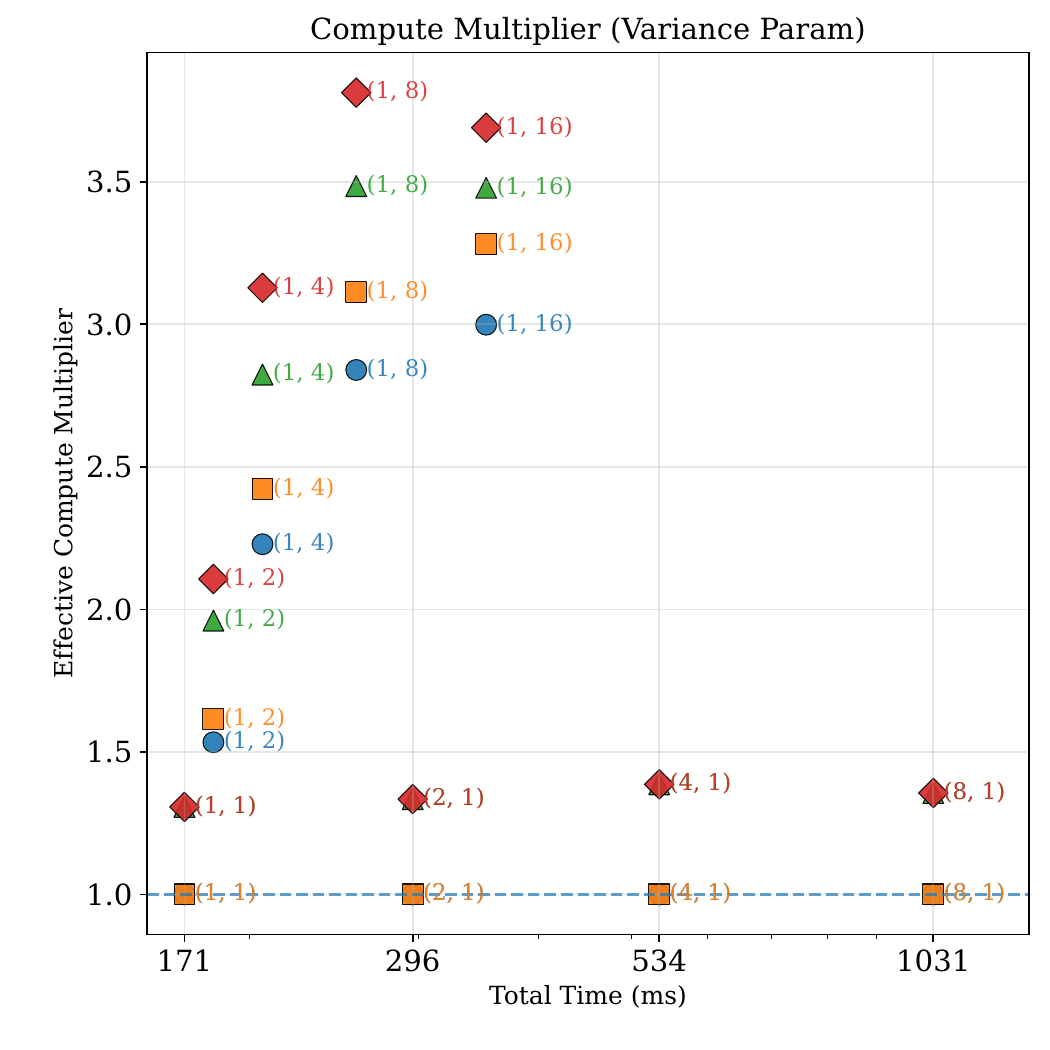}};
                    \node[left=of img21, node distance=0cm, rotate=90, xshift=2.1cm, yshift=-.9cm,  font=\color{black}]{\scriptsize{Effective Compute Mult. to Baseline}};
                    \node[below=of img21, node distance=0cm, xshift=0.0cm, yshift=1.15cm,  font=\color{black}]{\scriptsize{Per-Iteration Compute (ms)}};

                    \node [below=of img11, xshift=2.22cm, yshift=0.6cm](img31){\includegraphics[trim={1.41cm 1.21cm 0cm 0.8cm}, clip, width=.3\linewidth]{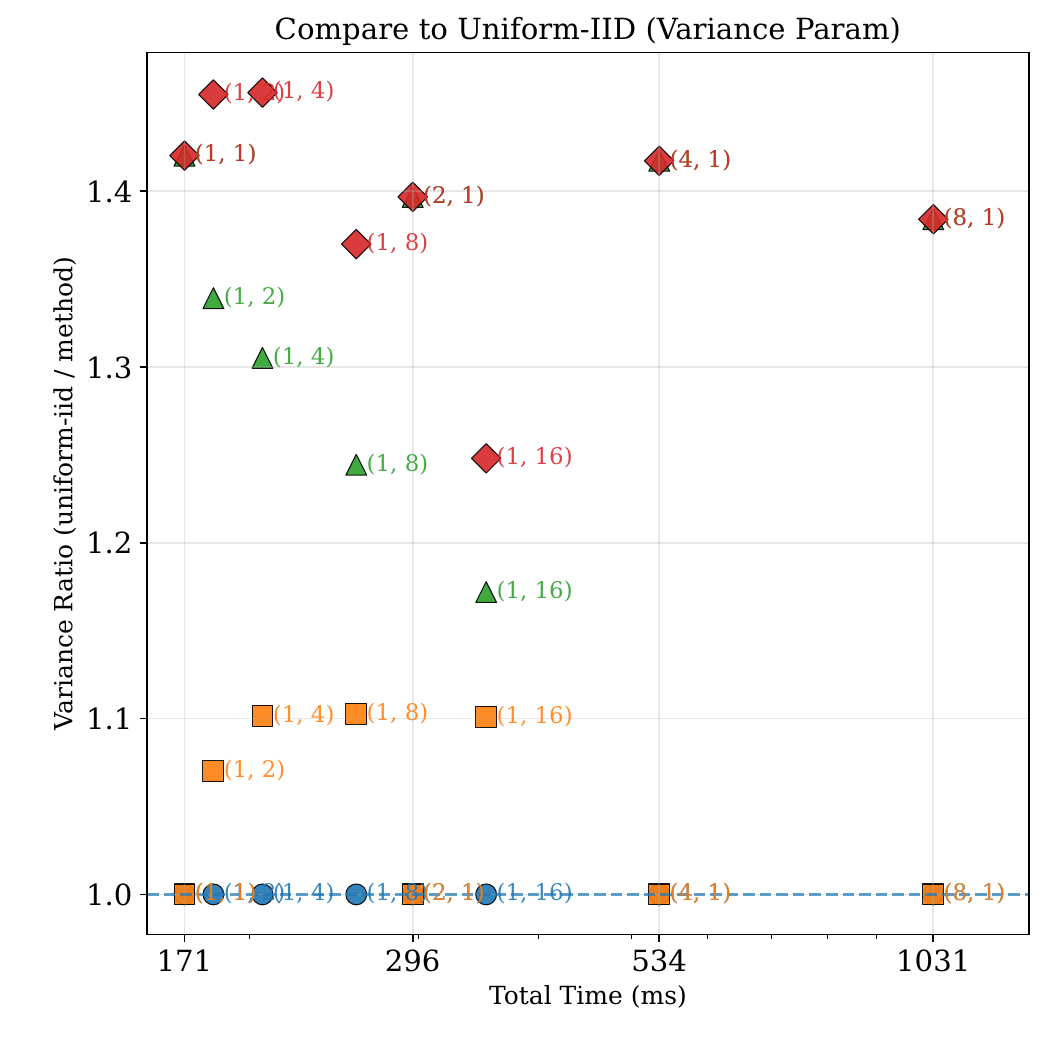}};
                    \node[left=of img31, node distance=0cm, rotate=90, xshift=2.1cm, yshift=-.9cm,  font=\color{black}]{\scriptsize{Relative Efficiency to Uniform}}; 
                    \node[below=of img31, node distance=0cm, xshift=0.0cm, yshift=1.15cm,  font=\color{black}]{\scriptsize{Per-Iteration Compute (ms)}};
                \end{tikzpicture}
                }
                \caption{
                    \textbf{Variance reduction in the low guidance regime ($\cfgScale\!=\!25$).}
                    Analogous to \Fig~\ref{fig:quantifying_variance_hierarchical_cost_aware_iw_strat}, measured at training step $\num{5000}$ with reduced classifier-free guidance.
                    \emph{Top left:} Variance versus compute budget. Colors denote uniform baseline, IW only, stratification only, and IW+Strat combined (red). Points are annotated by $(\numRenders,\numReNoises)$ tuples.
                    \emph{Top right:} Effective compute multiplier relative to the uniform $\numReNoises\!=\!1$ baseline curve. At $(\numRenders,\numReNoises)\!=\!(1,8)$, the only $\numReNoises\!=\!8$ configuration plotted, the multipliers are ${\sim}2.8\times$ (uniform), ${\sim}3.5\times$ (IW), ${\sim}3.1\times$ (Strat.), and ${\sim}3.8\times$ (IW+Strat).
                    \emph{Bottom:} Relative efficiency to uniform at matched $(\numRenders,\numReNoises)$ configurations. For $\numReNoises\!\ge\!2$ ($\numRenders\!=\!1$), importance weighting gives ${\sim}17$--$34\%$ and stratification ${\sim}7$--$10\%$, and their combination ${\sim}25$--$46\%$.
                }\label{fig:quantifying_variance_low_guidance}
            \end{figure}

\begin{figure}[h!]
                \centering
                \vspace{-0.0\textheight}
                \scalebox{1.0}{
                \begin{tikzpicture}
                \centering
                    \node (img11){\includegraphics[trim={0.7cm 0.8cm 0cm 2.3cm}, clip, width=.77\linewidth]{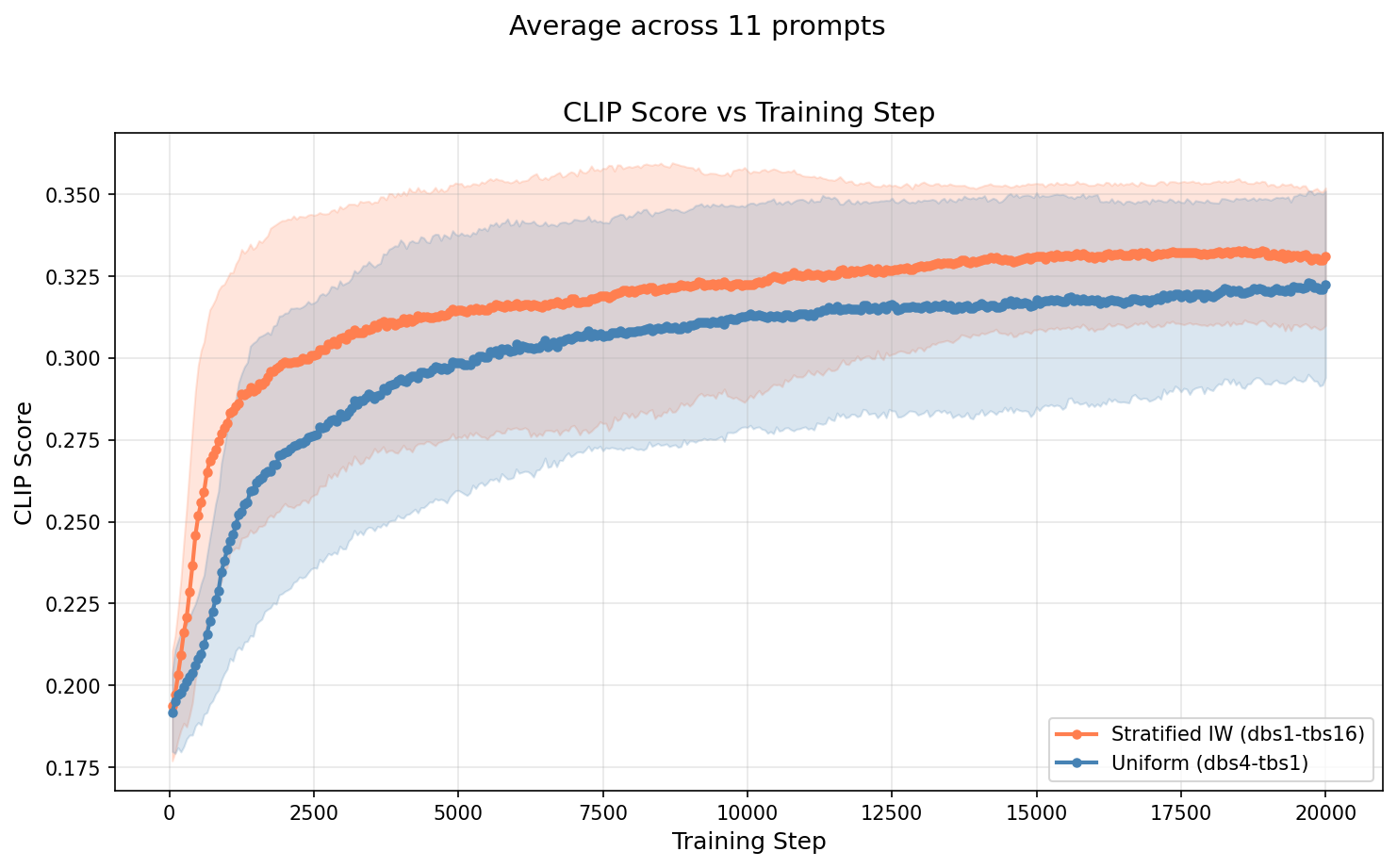}};
                    \node[left=of img11, node distance=0cm, rotate=90, xshift=1.0cm, yshift=-.8cm,  font=\color{black}]{\normalsize{CLIP Score}};
                    \node[below=of img11, node distance=0cm, xshift=0.0cm, yshift=1.15cm,  font=\color{black}]{\normalsize{Optimization Iteration}};
                \end{tikzpicture}
                }
                \vspace{-0.00\textheight}
                \caption{
                    \textbf{Performance gains from variance reduction at low guidance ($\cfgScale\!=\!25$).}
                    CLIP score vs.\ iteration at $\cfgScale\!=\!25$, averaged over $11$ prompts, $3$ seeds, and multiple views ($\pm$ std.\ dev.). Baseline $(4,1)$ vs.\ ours $(1,16)$ (Strat+IW+re-noising), whose iterations cost less (\Fig~\ref{fig:quantifying_variance_low_guidance}); the legend's dbs and tbs are $\numRenders$ and $\numReNoises$. The gap is larger than at standard guidance (\Fig~\ref{fig:clip_sds}), so variance reduction matters more at low guidance.
                }\label{fig:clip_sds_low_guidance}
            \end{figure}

            \textbf{Variance reduction benefits.}
                \Fig~\ref{fig:quantifying_variance_low_guidance} repeats the variance analysis at $\cfgScale\!=\!25$. At $\cfgScale\!=\!100$, IW+Strat has an effective compute multiplier of ${\sim}3.3\times$ averaged over $\numRenders$ at $\numReNoises\!=\!8$ (\Tab~\ref{tab:absolute_ecm_by_m_param}); at $\cfgScale\!=\!25$, it reaches ${\sim}3.8\times$ at $(\numRenders\!=\!1,\numReNoises\!=\!8)$. The first value averages over $\numRenders$; the second is a single configuration.
            
\begin{figure}[!t]
\centering
\footnotesize
\setlength{\tabcolsep}{0pt}
\renewcommand{\arraystretch}{0}
\begin{tabular}{@{}m{0.042\textwidth}@{}m{0.083\textwidth}@{}m{0.083\textwidth}@{}m{0.083\textwidth}@{}m{0.083\textwidth}@{}m{0.083\textwidth}@{}m{0.083\textwidth}@{}m{0.083\textwidth}@{}m{0.083\textwidth}@{}}
 & \centering\arraybackslash 100 & \centering\arraybackslash 200 & \centering\arraybackslash 400 & \centering\arraybackslash 500 & \centering\arraybackslash 1000 & \centering\arraybackslash 2000 & \centering\arraybackslash 5000 & \centering\arraybackslash 10000 \\
\rotatebox{90}{\scriptsize\shortstack{Uniform\\(2,1)}} & \centering\arraybackslash\includegraphics[width=0.083\textwidth]{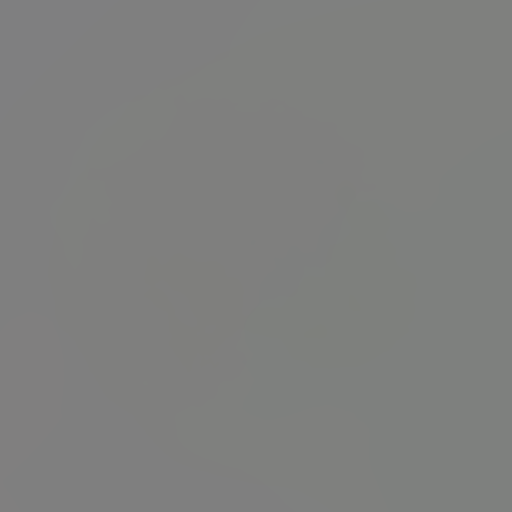} & \centering\arraybackslash\includegraphics[width=0.083\textwidth]{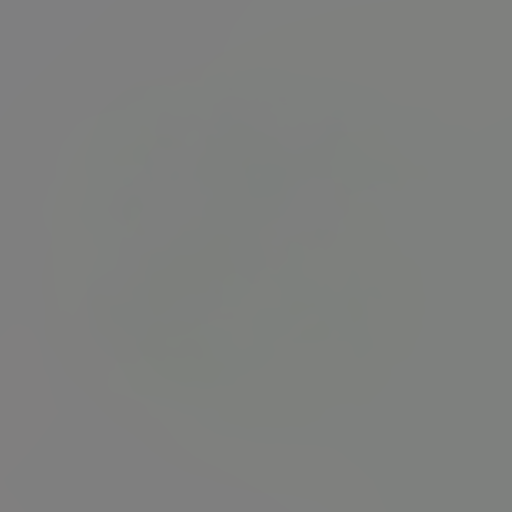} & \centering\arraybackslash\includegraphics[width=0.083\textwidth]{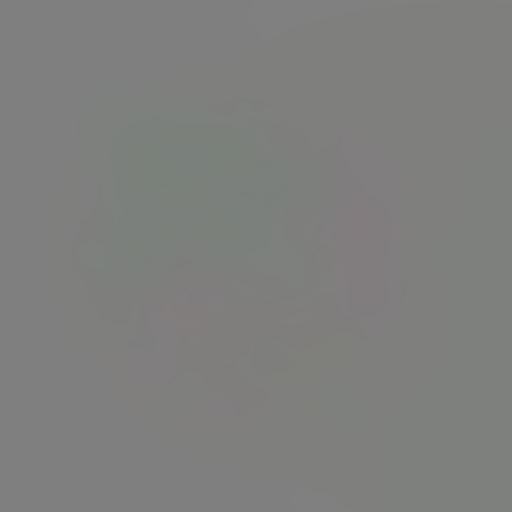} & \centering\arraybackslash\includegraphics[width=0.083\textwidth]{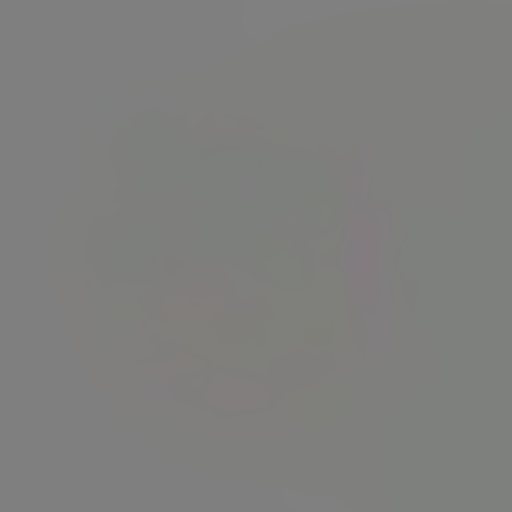} & \centering\arraybackslash\includegraphics[width=0.083\textwidth]{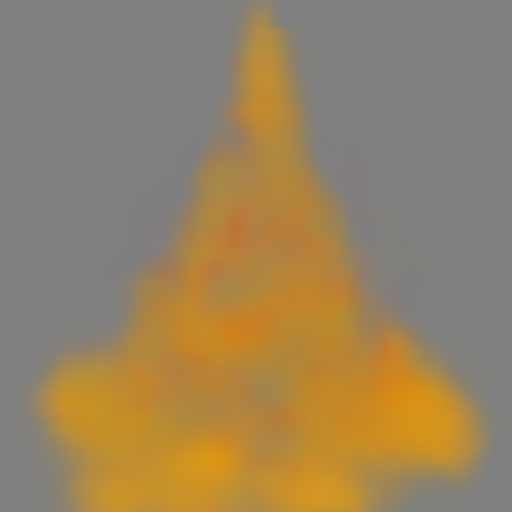} & \centering\arraybackslash\includegraphics[width=0.083\textwidth]{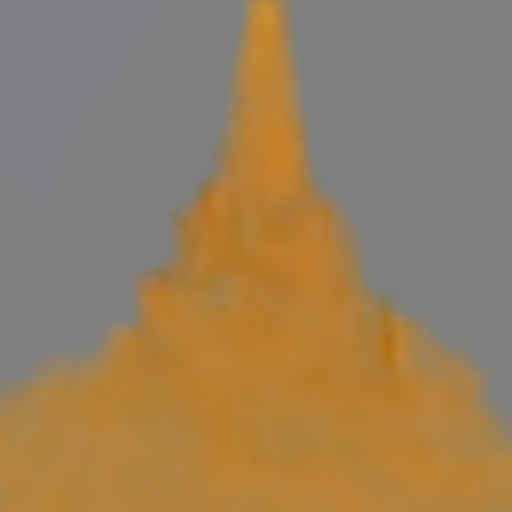} & \centering\arraybackslash\includegraphics[width=0.083\textwidth]{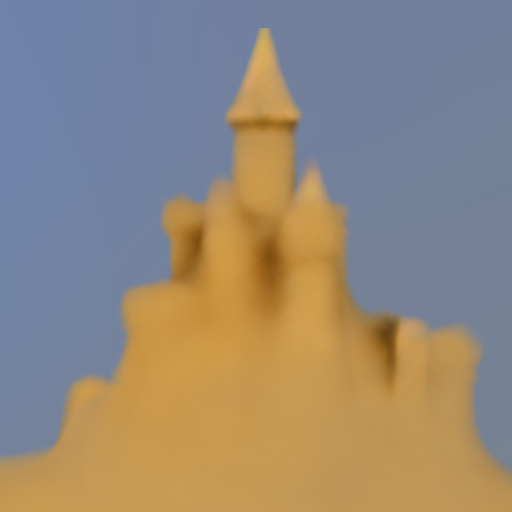} & \centering\arraybackslash\includegraphics[width=0.083\textwidth]{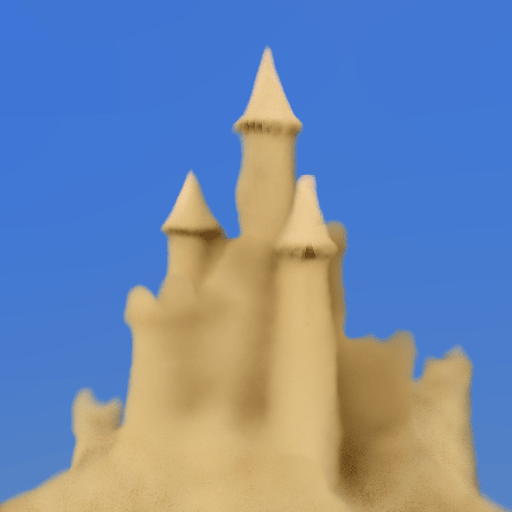} \\
\rotatebox{90}{\scriptsize\shortstack{Uniform\\(2,1)}} & \centering\arraybackslash\includegraphics[width=0.083\textwidth]{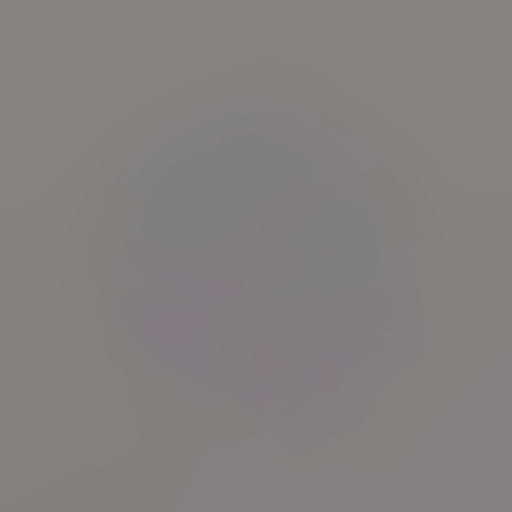} & \centering\arraybackslash\includegraphics[width=0.083\textwidth]{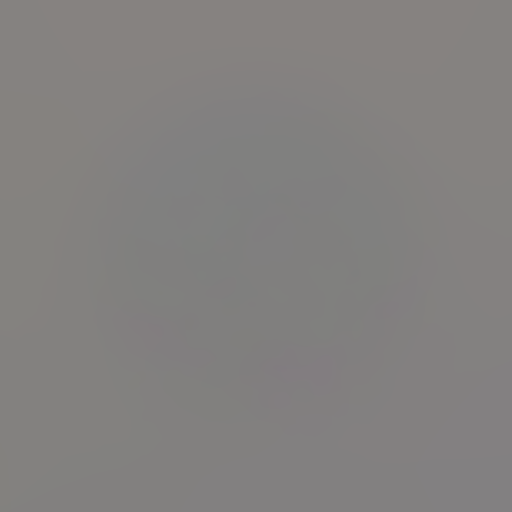} & \centering\arraybackslash\includegraphics[width=0.083\textwidth]{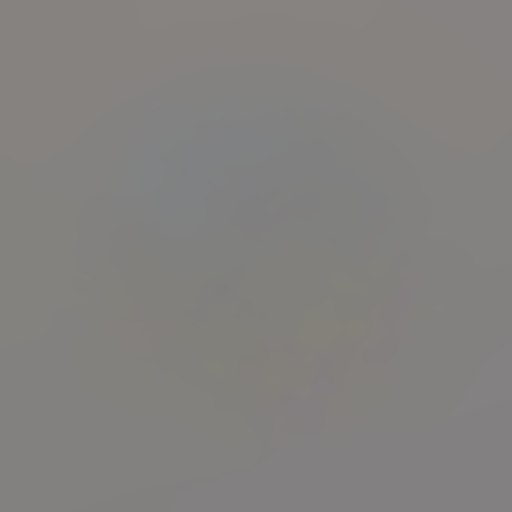} & \centering\arraybackslash\includegraphics[width=0.083\textwidth]{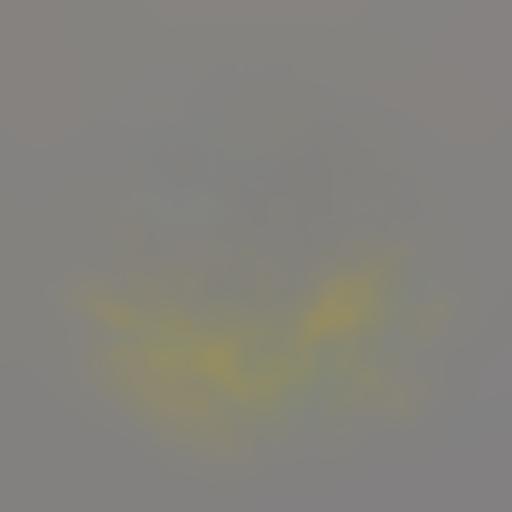} & \centering\arraybackslash\includegraphics[width=0.083\textwidth]{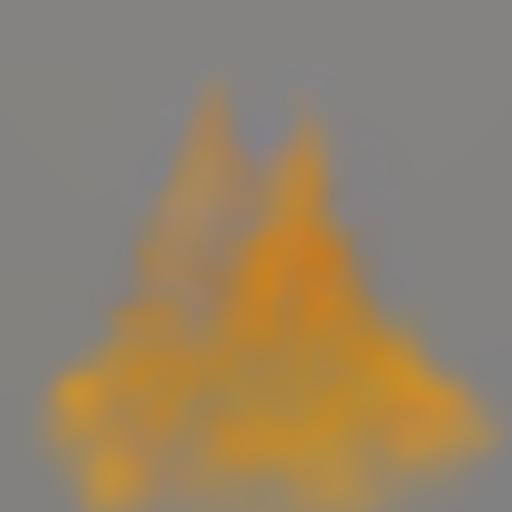} & \centering\arraybackslash\includegraphics[width=0.083\textwidth]{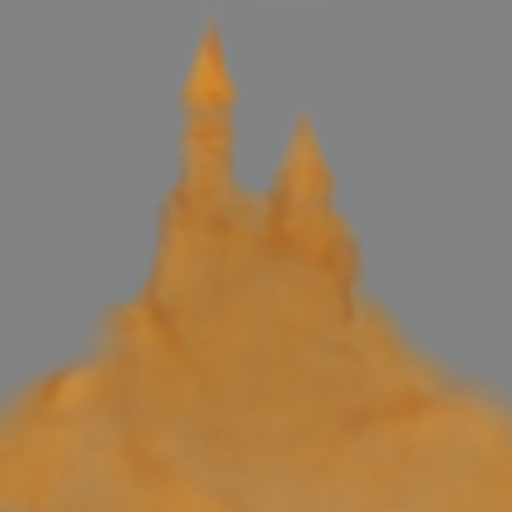} & \centering\arraybackslash\includegraphics[width=0.083\textwidth]{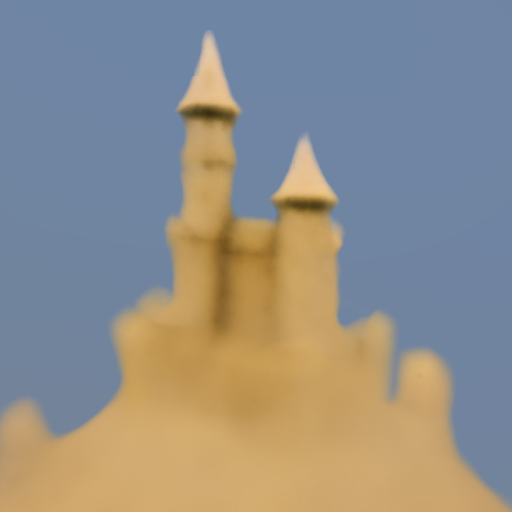} & \centering\arraybackslash\includegraphics[width=0.083\textwidth]{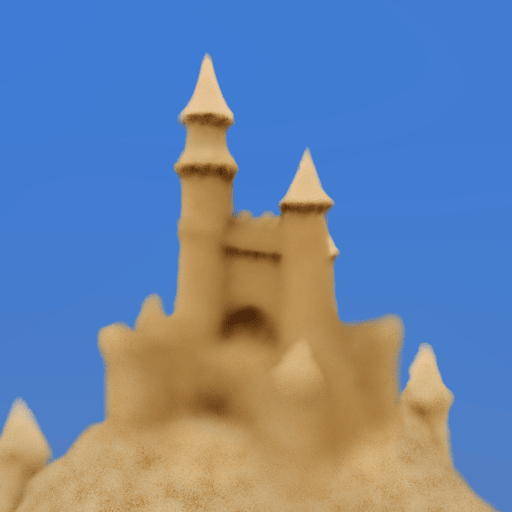} \\
\rotatebox{90}{\scriptsize\shortstack{IW+Strat\\(1,16)}} & \centering\arraybackslash\includegraphics[width=0.083\textwidth]{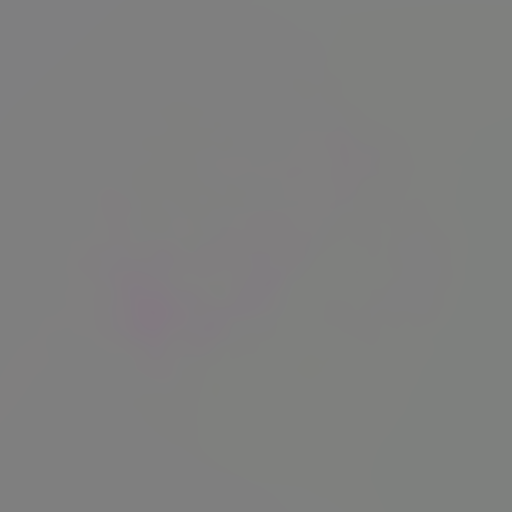} & \centering\arraybackslash\includegraphics[width=0.083\textwidth]{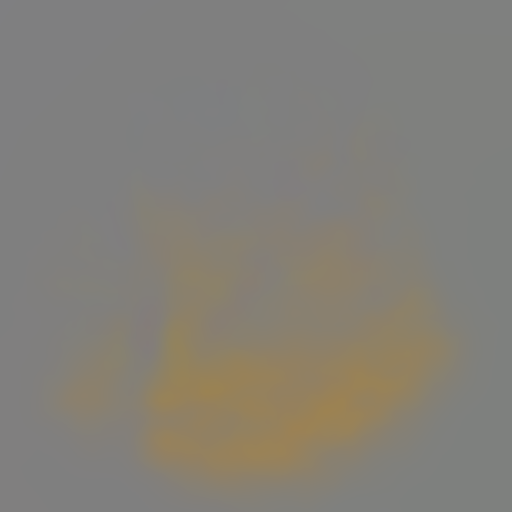} & \centering\arraybackslash\includegraphics[width=0.083\textwidth]{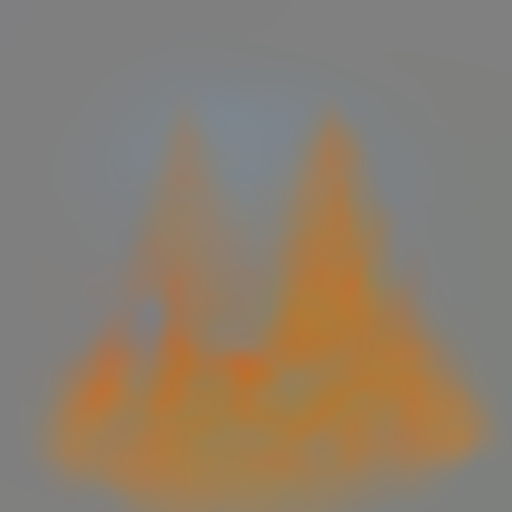} & \centering\arraybackslash\includegraphics[width=0.083\textwidth]{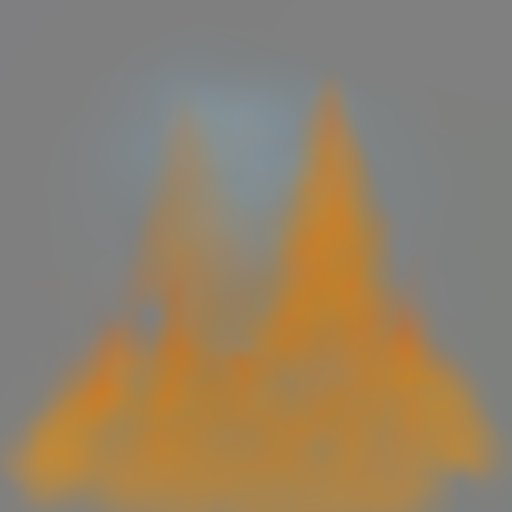} & \centering\arraybackslash\includegraphics[width=0.083\textwidth]{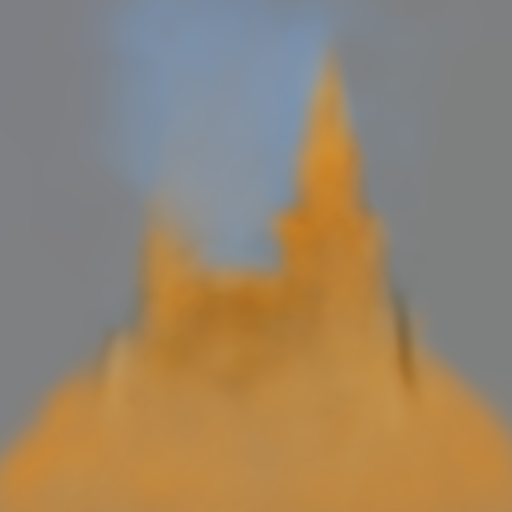} & \centering\arraybackslash\includegraphics[width=0.083\textwidth]{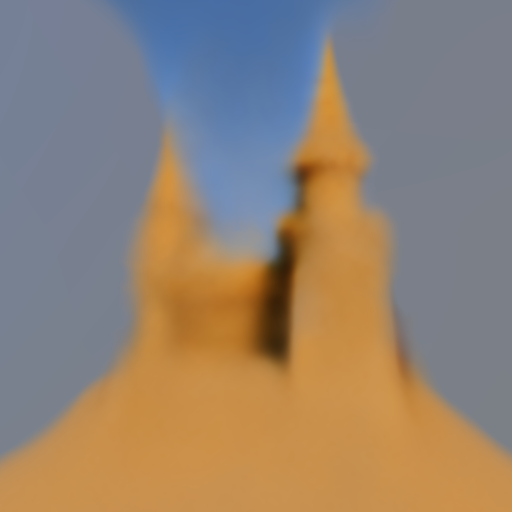} & \centering\arraybackslash\includegraphics[width=0.083\textwidth]{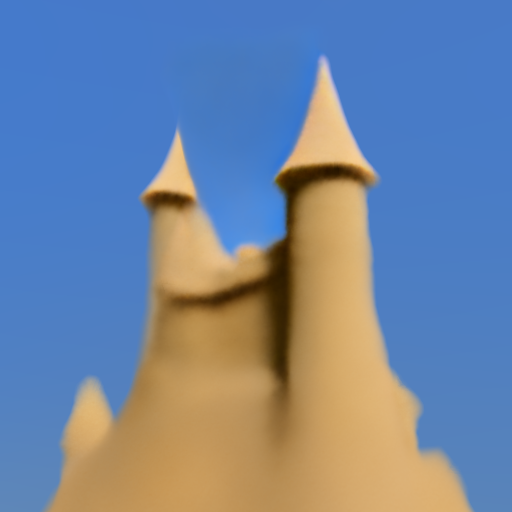} & \centering\arraybackslash\includegraphics[width=0.083\textwidth]{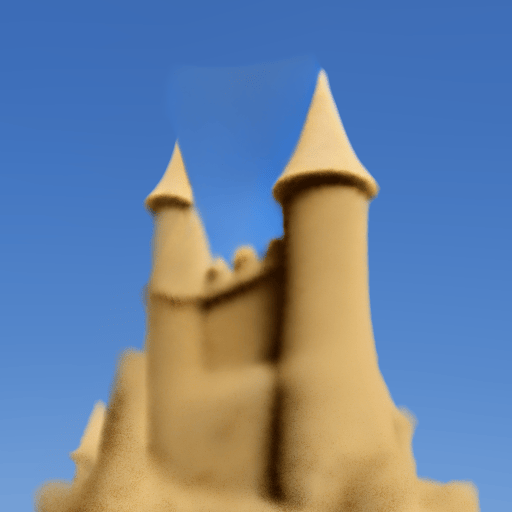} \\
\rotatebox{90}{\scriptsize\shortstack{IW+Strat\\(1,16)}} & \centering\arraybackslash\includegraphics[width=0.083\textwidth]{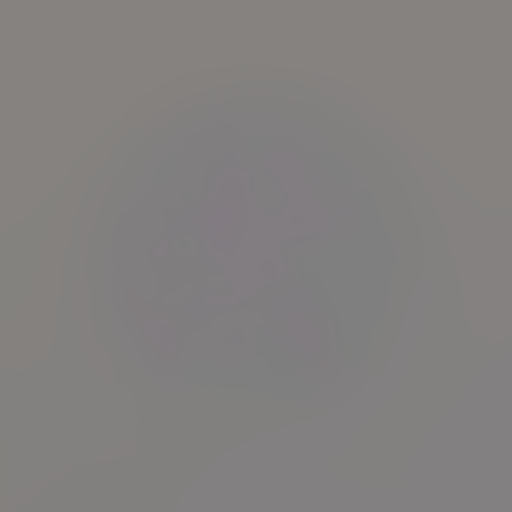} & \centering\arraybackslash\includegraphics[width=0.083\textwidth]{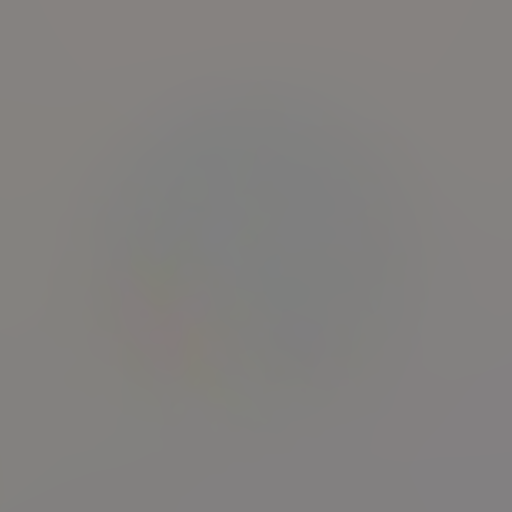} & \centering\arraybackslash\includegraphics[width=0.083\textwidth]{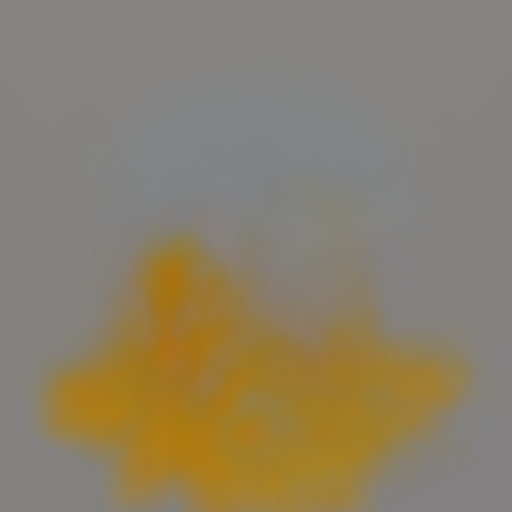} & \centering\arraybackslash\includegraphics[width=0.083\textwidth]{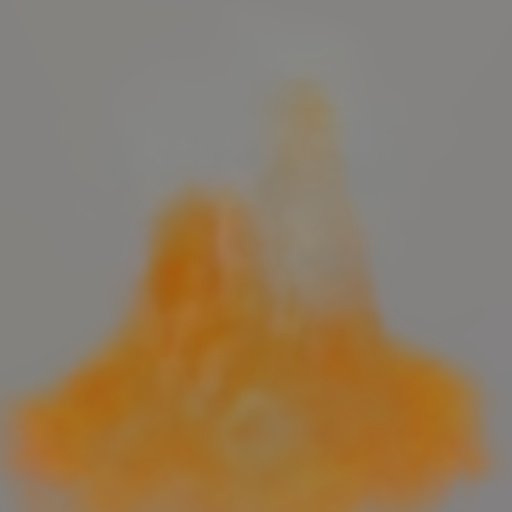} & \centering\arraybackslash\includegraphics[width=0.083\textwidth]{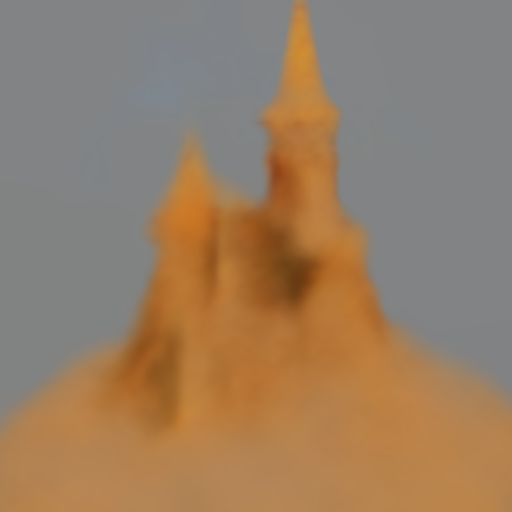} & \centering\arraybackslash\includegraphics[width=0.083\textwidth]{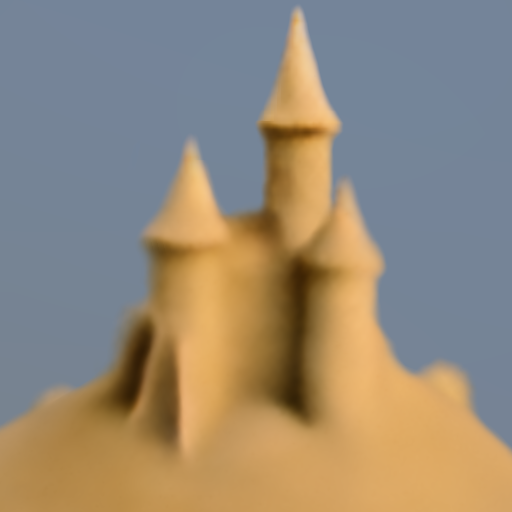} & \centering\arraybackslash\includegraphics[width=0.083\textwidth]{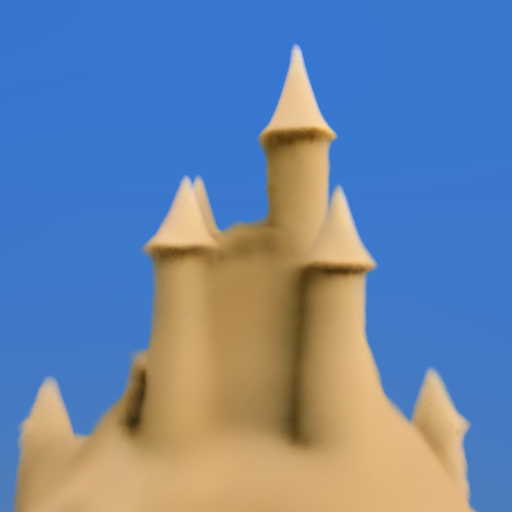} & \centering\arraybackslash\includegraphics[width=0.083\textwidth]{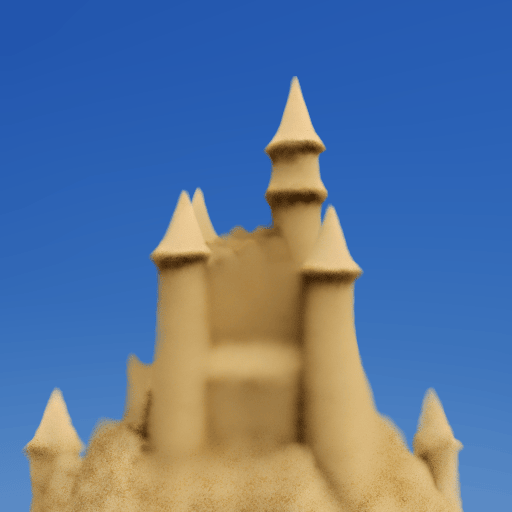} \\
\end{tabular}
\caption{
    \textbf{Qualitative SDS trajectories at low classifier-free guidance ($\cfgScale\!=\!25$):} comparison of the uniform $(\numRenders,\numReNoises)\!=\!(2,1)$ baseline (top two rows) and our IW+Strat $(1,16)$ method (bottom two rows) for the prompt ``\emph{A castle-shaped sandcastle}'' across $2$ training seeds. Columns show renders at training steps $\num{100}$, $\num{200}$, $\num{400}$, $\num{500}$, $\num{1000}$, $\num{2000}$, $\num{5000}$, $\num{10000}$. Consistent with the CLIP curves in \Fig~\ref{fig:clip_sds_low_guidance}, our method reaches the baseline's converged geometry in substantially fewer iterations.
}
\label{fig:castle_sandcastle_uniform_vs_iw_all_seeds}
\end{figure}

            \textbf{Downstream performance.}
                \Fig~\ref{fig:clip_sds_low_guidance} shows CLIP score versus optimization iteration at low guidance. The performance gap between our method and the baseline is larger than at standard guidance (\Fig~\ref{fig:clip_sds}), so variance reduction matters more for convergence at low guidance. Per-step qualitative trajectories on two prompts (\Fig~\ref{fig:low_guidance_qualitative_comparison}) and across two seeds for a third (\Fig~\ref{fig:castle_sandcastle_uniform_vs_iw_all_seeds}) corroborate this: the reduced-variance estimator reaches the baseline's later-stage geometry and appearance in fewer iterations. This suggests that variance reduction may be a complementary strategy for enabling lower guidance settings in practice, potentially reducing high-guidance artifacts without sacrificing convergence speed.

            {  \graphicspath{{images/sds/qualitative/main_low_guidance_per_step_main/}}
            \begin{figure}[h!]
    \centering
    \begin{tikzpicture}
        \node[inner sep=0pt] (img) {\includegraphics[width=0.78\linewidth]{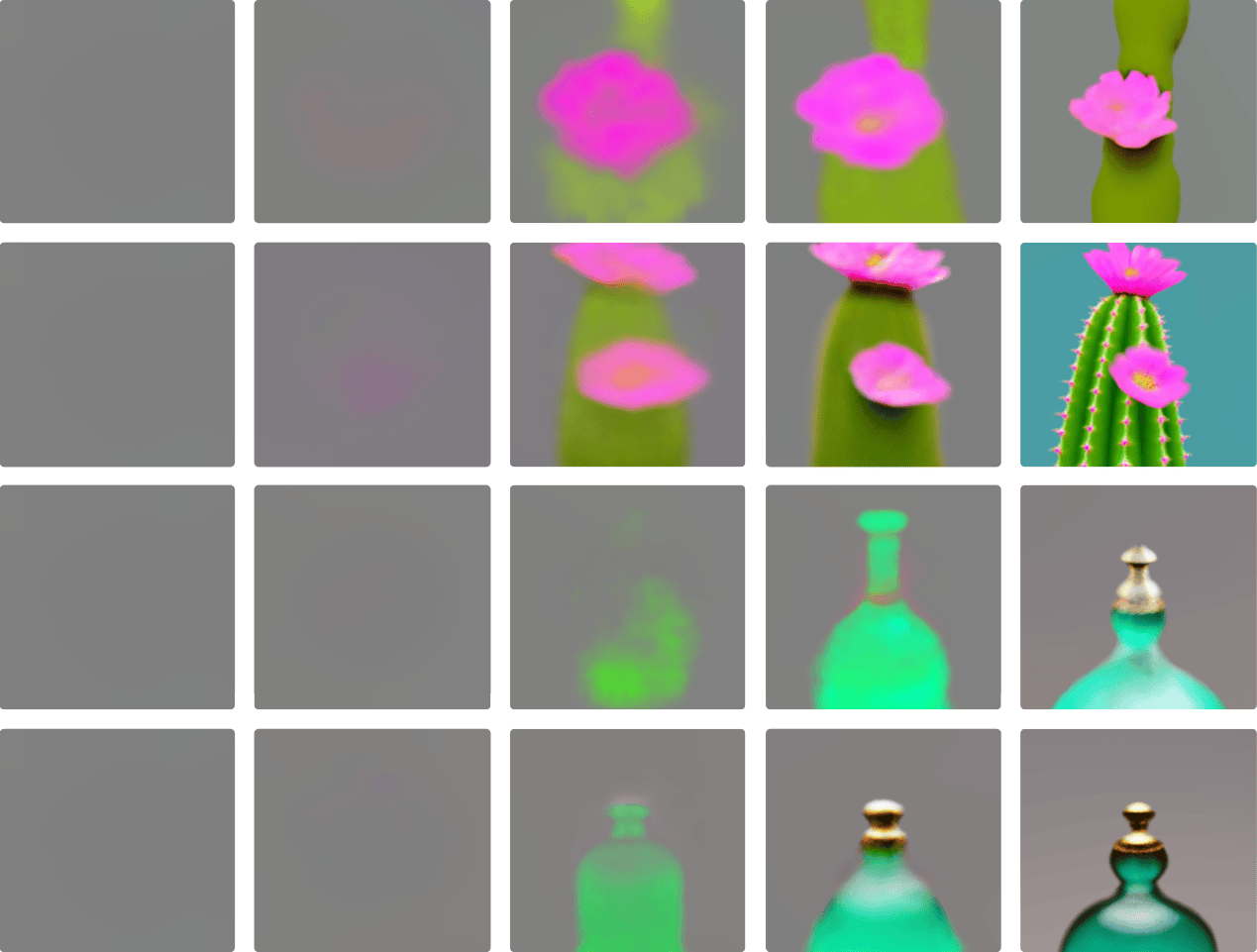}};

        \node[anchor=south, yshift=3pt, font=\footnotesize] at ($(img.north west)!0.093!(img.north east)$) {Step 0};
        \node[anchor=south, yshift=3pt, font=\footnotesize] at ($(img.north west)!0.296!(img.north east)$) {Step 100};
        \node[anchor=south, yshift=3pt, font=\footnotesize] at ($(img.north west)!0.500!(img.north east)$) {Step 500};
        \node[anchor=south, yshift=3pt, font=\footnotesize] at ($(img.north west)!0.704!(img.north east)$) {Step 1000};
        \node[anchor=south, yshift=3pt, font=\footnotesize] at ($(img.north west)!0.907!(img.north east)$) {Step 5000};

        \node[anchor=east, xshift=-4pt, font=\footnotesize] at ($(img.north west)!0.1167!(img.south west)$) {Uniform $(2,1)$};
        \node[anchor=east, xshift=-4pt, font=\footnotesize] at ($(img.north west)!0.3724!(img.south west)$) {IW+Strat $(1,16)$};
        \node[anchor=east, xshift=-4pt, font=\footnotesize] at ($(img.north west)!0.6276!(img.south west)$) {Uniform $(2,1)$};
        \node[anchor=east, xshift=-4pt, font=\footnotesize] at ($(img.north west)!0.8833!(img.south west)$) {IW+Strat $(1,16)$};

        \node[rotate=270, font=\scriptsize] at ([xshift=10pt]$(img.north east)!0.2446!(img.south east)$) {``\emph{A cactus with pink flowers}''};
        \node[rotate=270, font=\scriptsize] at ([xshift=10pt]$(img.north east)!0.7555!(img.south east)$) {``\emph{An antique glass perfume bottle}''};
    \end{tikzpicture}
    \vspace{4pt}
    \caption{
        \textbf{Qualitative SDS trajectories at low classifier-free guidance ($\cfgScale\!=\!25$).}
        Comparison of the uniform $(\numRenders,\numReNoises)\!=\!(2,1)$ baseline and our IW+Strat $(1,16)$ method on two prompts.
        Consistent with the CLIP curves in \Fig~\ref{fig:clip_sds_low_guidance}, the reduced-variance estimator reaches the baseline's later-stage geometry and appearance in fewer iterations, with the advantage becoming clearer in this lower-guidance regime where the SDS gradient is noisier.
        Multi-seed trajectories for an additional prompt are in \Fig~\ref{fig:castle_sandcastle_uniform_vs_iw_all_seeds}.
    }\label{fig:low_guidance_qualitative_comparison}
\end{figure}

            }

        \subsubsection{Optimal Pair Probability Distributions}\label{app:optimal_pair_distributions}
            We study two-sample ($\numReNoises\!=\!2$) without-replacement estimators, complementing the empirical analysis in \Sec~\ref{sec:method-stratified-sampling}. The Horvitz-Thompson (HT) estimator is governed by a pair-probability matrix $\tilde{Q}(i,j)$ for unordered pairs $(i,j)$ \citep{thompson2012sampling, owen2013monte}; we compare classical choices for $\tilde{Q}$ with a numerically computed optimum.
        
            \textbf{Setup.}
                Consider $\numSamples$ timestep indices with target values $\{\mathbf{y}_i\}_{i=1}^{\numSamples}$, where $\mathbf{y}_i = p_i \gterm_i$ for base probabilities $p_i$ and per-timestep gradient contributions $\gterm_i$ (as in \Sec~\ref{sec:method-variance-reduction-strategies}). With $\tilde{Q}$ symmetric and normalized over unordered pairs, $\sum_{i<j}\tilde{Q}(i,j)=1$, the marginal inclusion probability $\kappa_i = \sum_{j \neq i} \tilde{Q}(i,j)$ (so $\sum_i\kappa_i=2$) determines the HT estimator for a sampled pair $(i,j) \sim \tilde{Q}$:
                \begin{equation}
                    \estimatedMean = \mathbf{y}_i / \kappa_i + \mathbf{y}_j / \kappa_j
                \end{equation}
                Following our variance definition (\Eq~\ref{eq:var}), the theoretical variance is
                \begin{equation}
                    \mathrm{Var}(\estimatedMean) = \sum_{i < j} \tilde{Q}(i,j) \|\estimatedMean_{ij} - \boldsymbol{\mu}_{\mathbf{y}}\|_2^2
                \end{equation}
                where $\boldsymbol{\mu}_{\mathbf{y}} = \sum_i \mathbf{y}_i$ is the target sum and $\estimatedMean_{ij}$ denotes the estimator value for pair $(i,j)$.

            \textbf{Asymptotic rates as context.}
                Stratified sampling on $[0,1]$ is never worse than IID at the same sample count, and for sufficiently smooth integrands (e.g., Lipschitz) it improves the variance rate from $\mathcal{O}(\numSamples^{-1})$ to $\mathcal{O}(\numSamples^{-3})$ in $1$D with one sample per equal-width stratum \citep{owen2013monte}. Importance sampling with a fixed proposal changes the leading constant (a good proposal shrinks it) but keeps the IID rate. The two-sample analysis below isolates the constant-factor structure these asymptotic arguments~hide.
        
\begin{figure}[h!]
                \centering
                \scalebox{1.0}{
                \begin{tikzpicture}
                    \def\imgw{0.27\linewidth}
                    \def\xsep{-1.2cm}
                    
                    \node (q1) {\includegraphics[trim={0.7cm 0.7cm 3.0cm 1.1cm}, clip, width=0.287\linewidth]{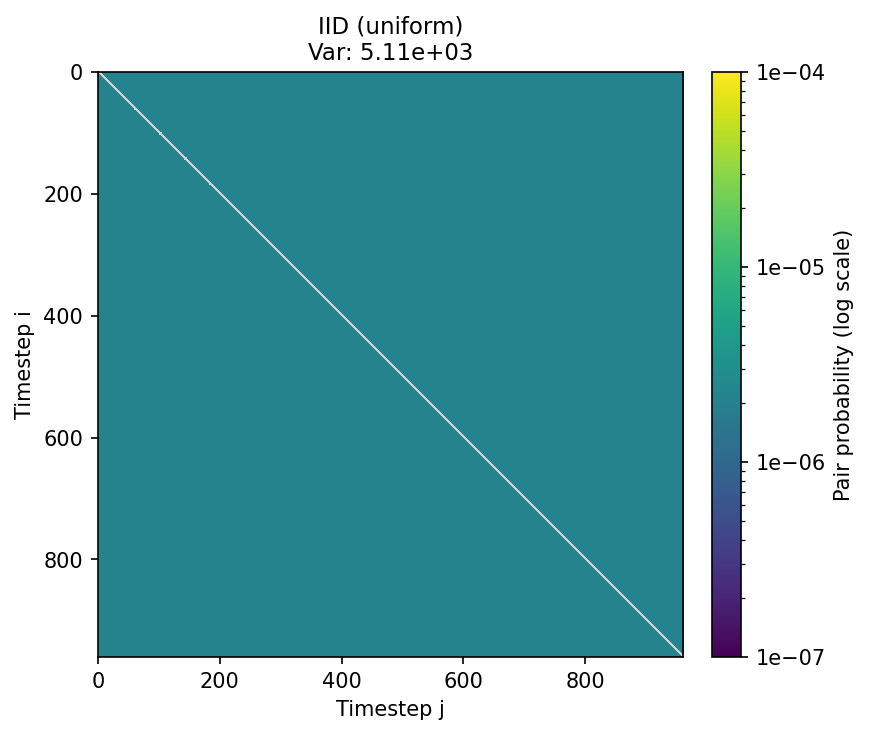}};
                    \node (q2) [right=of q1, xshift=\xsep] {\includegraphics[trim={1.5cm 0.7cm 3.0cm 1.1cm}, clip, width=\imgw]{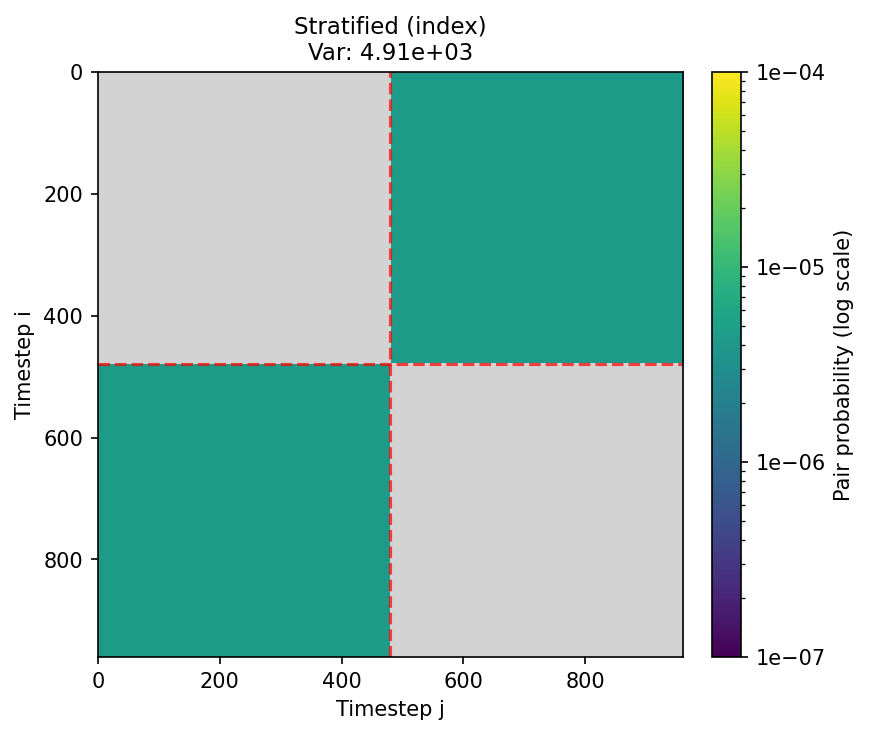}};
                    \node (q3) [right=of q2, xshift=\xsep] {\includegraphics[trim={1.5cm 0.7cm 3.0cm 1.2cm}, clip, width=\imgw]{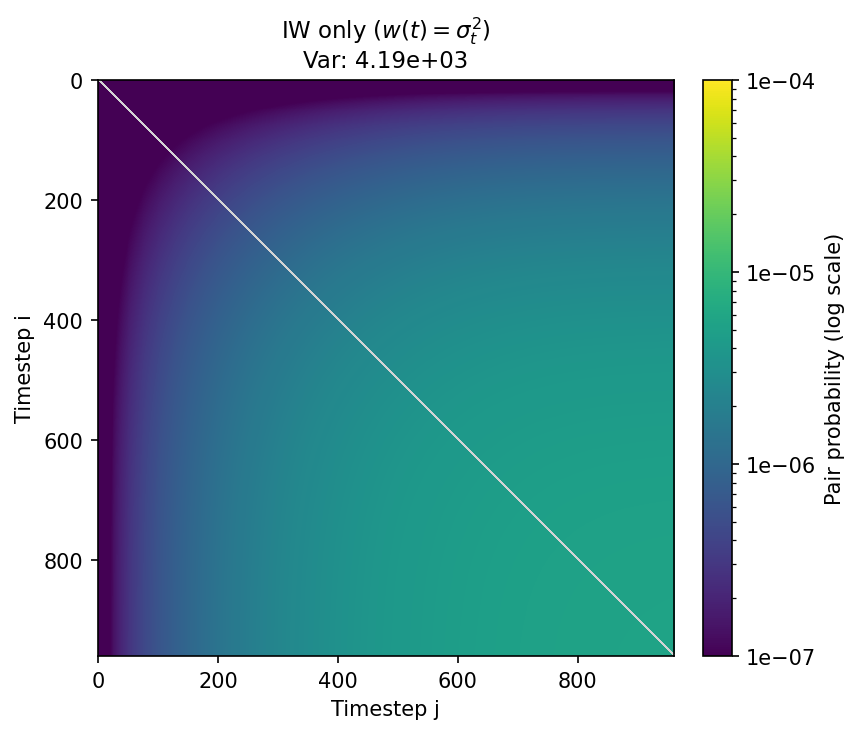}};
                    \node (q4) [below=of q1, yshift=0.55cm] {\includegraphics[trim={0.7cm 0.7cm 3.0cm 1.1cm}, clip, width=0.287\linewidth]{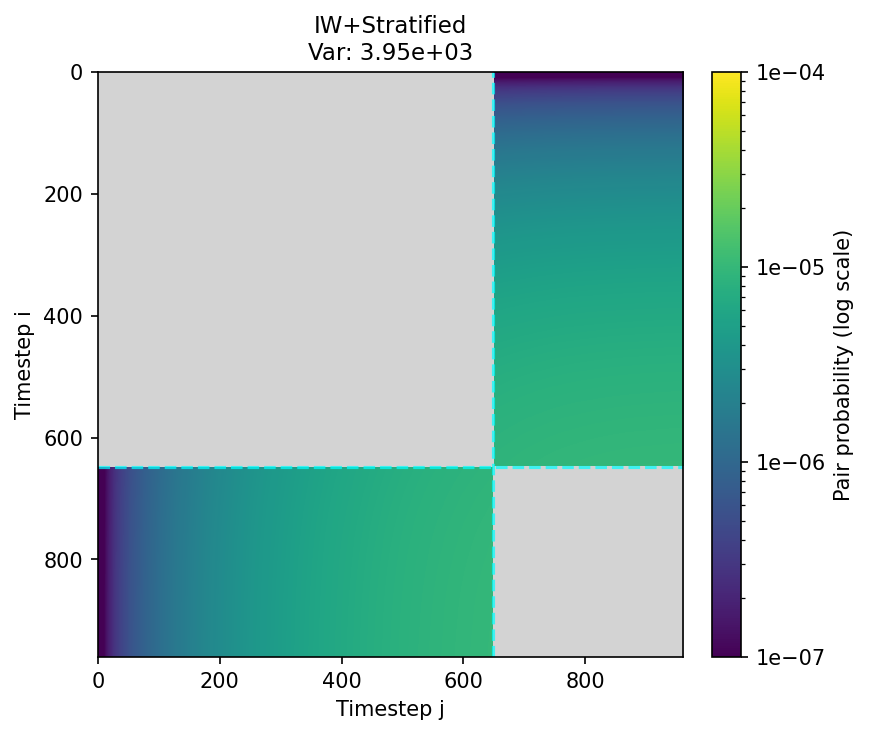}};
                    \node (q5) [right=of q4, xshift=\xsep] {\includegraphics[trim={1.5cm 0.7cm 3.0cm 1.1cm}, clip, width=\imgw]{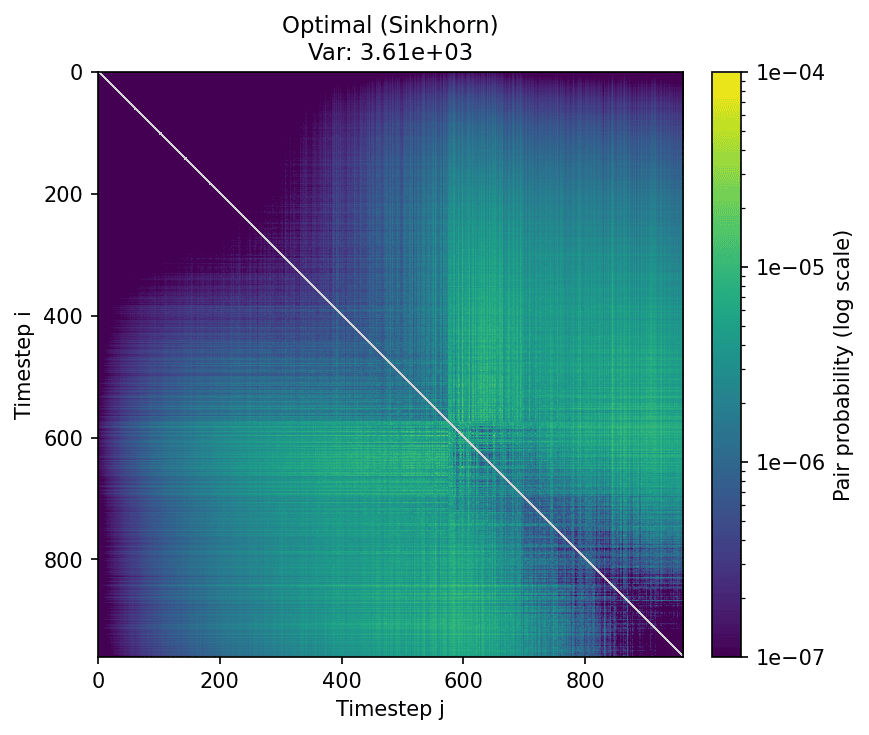}};
                    \node (legend) [right=of q5, xshift=\xsep] {\includegraphics[trim={12.0cm 0.7cm 0.7cm 1.0cm}, clip, width=0.054\linewidth]{images/pair_prob_matrices/ys_pickled_tensor_final_different_eps_Q_optimal.png}};
                    
                    \node[above=of q1, yshift=-1.1cm, font=\scriptsize] {(a) IID};
                    \node[above=of q2, yshift=-1.1cm, font=\scriptsize] {(b) Stratified};
                    \node[above=of q3, yshift=-1.1cm, font=\scriptsize] {(c) IW};
                    \node[above=of q4, yshift=-1.1cm, font=\scriptsize] {(d) IW+Strat};
                    \node[above=of q5, yshift=-1.1cm, font=\scriptsize] {(e) Optimal};
                    \node[right=of legend, yshift=0.9cm, xshift=-0.9cm, rotate=270, font=\scriptsize] {Pair probability};
                    \node[left=of q1, yshift=1.1cm, xshift=0.9cm, rotate=90, font=\scriptsize] {Timestep index $i$};
                    \node[below=of q5, yshift=1.1cm, font=\scriptsize] {Timestep index $j$};
                    \node[left=of q4, yshift=1.1cm, xshift=0.9cm, rotate=90, font=\scriptsize] {Timestep index $i$};
                \end{tikzpicture}
                }
                \vspace{-0.01\textheight}
                \caption{
                    \textbf{Pair probability matrices $\tilde{Q}(i,j)$ for two-sample strategies}, computed on gradient data from a single SDS prompt at the end of training.
                    Each panel shows the probability of selecting pair $(i,j)$ on a log scale (brighter = higher probability, gray = zero).
                    (a) IID places equal mass on all pairs ($1.00\times$, baseline).
                    (b) Index-based stratification concentrates mass in off-diagonal blocks ($1.04\times$).
                    (c) Importance weighting concentrates on high-weight timesteps ($1.22\times$).
                    (d) IW+Stratified combines both via CDF-space stratification ($1.29\times$).
                    (e) Sinkhorn-optimal pairs timesteps with complementary gradient directions ($1.42\times$).
                    Effective compute multipliers (in parentheses) are computed from theoretical variances. IW+Stratified reaches ${\sim}91\%$ of the optimal allocation's efficiency multiplier, validating it as a practical near-optimum.
                }\label{fig:pair_prob_comparison}
            \end{figure}

            \textbf{Pair distributions compared.}
                We compare five pair probability matrices (\Fig~\ref{fig:pair_prob_comparison}):
                \begin{enumerate}[nosep,leftmargin=*]
                    \item \emph{IID (uniform):} $\tilde{Q}(i,j) \propto 1$ for $i \neq j$, corresponding to uniform sampling.
                    \item \emph{Stratified (index):} Partition indices into halves $\stratum_0, \stratum_1$ and sample one from each, giving $\tilde{Q}(i,j) = 1/(|\stratum_0||\stratum_1|)$ for $i \in \stratum_0, j \in \stratum_1$.
                    \item \emph{IW only:} $\tilde{Q}(i,j) \propto \weight(\timevar_i) \weight(\timevar_j)$, with $\weight$ the diffusion objective's timestep weight.
                    \item \emph{IW+Stratified:} Stratify in CDF space of the importance distribution (\Sec~\ref{sec:method-stratified-sampling}), then sample within each stratum proportionally to $\weight(\timevar)$.
                    \item \emph{Optimal (Sinkhorn):} An entropy-regularized approximation of the variance-minimizing $\tilde{Q}$ with target marginals $\kappa_i \propto \|\mathbf{y}_i\|_2$.
                \end{enumerate}
        
            \textbf{Optimal pair distribution via Sinkhorn.}
                The optimal $\tilde{Q}$ minimizes variance subject to marginal constraints $\sum_{j \neq i} \tilde{Q}(i,j) = \kappa_i$ for target inclusion probabilities $\kappa_i \propto \|\mathbf{y}_i\|_2$ (a per-snapshot optimum that treats $\mathbf{y}_i$ as deterministic). We cast this as an entropy-regularized optimal-transport problem~\citep{cuturi2013sinkhorn} with cost matrix $C_{ij} = (\mathbf{y}_i/\kappa_i)^\top (\mathbf{y}_j/\kappa_j)$, penalizing pairs whose scaled contributions are aligned. The Gibbs kernel $K_{ij} = \exp(-\beta C_{ij} / \mathrm{scale})$ is normalized to the 95th percentile of off-diagonal costs. Sinkhorn iteration rescales the kernel to match the target marginals, producing a $\tilde{Q}$ that concentrates mass on pairs with diverse gradient directions while respecting the target marginals.
        
            \textbf{Results.}
                \Fig~\ref{fig:pair_prob_comparison} visualizes the five pair-probability matrices computed from gradient data on a single SDS prompt at the end of training (\Sec~\ref{sec:experiments-sds}). The Sinkhorn distribution has the lowest variance of the five, pairing timesteps with complementary gradient directions. IW+Stratified comes close, reaching ${\sim}91\%$ of its efficiency multiplier without solving the transport problem. The variance ordering (IID $>$ Stratified $>$ IW $>$ IW+Strat $>$ Optimal) and the corresponding effective compute multipliers over IID ($1.04\times$, $1.22\times$, $1.29\times$, $1.42\times$) match our empirical findings, so the tractable estimator captures most of the theoretical benefit.
        
    \subsubsection{Sensitivity to Render vs.\ Denoise Cost Ratio}\label{sec:compute_cost_sensitivity}
        The results in \Fig~\ref{fig:quantifying_variance_hierarchical_cost_aware_iw_strat} use measured wall-clock compute, which conflates rendering, encoding, and denoising costs. To isolate the effect of the render-to-denoise cost ratio, we repeat the analysis (\Fig~\ref{fig:compute_cost_sensitivity}) with simulated cost models $\budget = \alpha \numRenders + \numRenders \numReNoises$, where $\alpha$ controls the relative cost of rendering versus denoising. Setting $\alpha = 0$ simulates the extreme where rendering is free (only denoising contributes), $\alpha = 1$ simulates equal per-operation cost, and $\alpha = 100$ simulates the rendering-dominated regime that occurs in practice for differentiable rendering, latent-diffusion encoders with backpropagation, and physics simulators.

        \textbf{Takeaway.}
            At zero render cost, $\numReNoises\!=\!1$ is best for every method, and IW+Strat stays above the uniform baseline at small $\numReNoises$ because of IW. When the render cost grows, larger $\numReNoises$ becomes increasingly beneficial because the expensive upstream computation amortizes over more cheap denoiser calls. Our wall-clock experiments (\Fig~\ref{fig:quantifying_variance_hierarchical_cost_aware_iw_strat}) reflect the effective scale of these costs after parallelization, and the cost-ratio sweep here shows that the conclusion (IW+Strat beats the uniform baseline) holds across the full range of plausible cost ratios, not only the regime our hardware happens to occupy.
    
                    \begin{figure}[h!]
            \centering
            \resizebox{0.92\linewidth}{!}{
            \begin{tikzpicture}
            \centering
                \node (img11){\includegraphics[trim={1.3cm 1.21cm 0cm 0.8cm}, clip, width=.3\linewidth]{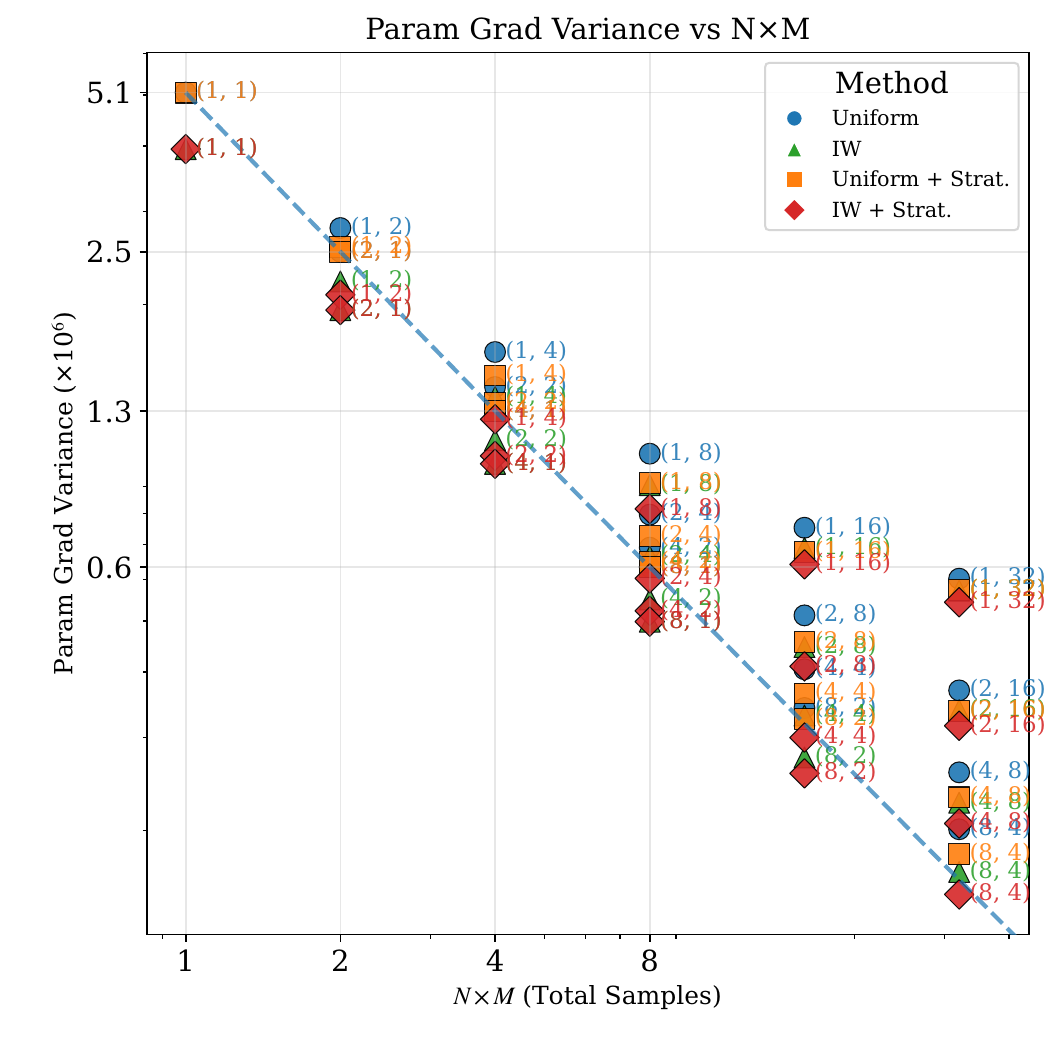}};
                \node [right=of img11, node distance=0cm, xshift=-0.75cm](img12){\includegraphics[trim={1.4cm 1.21cm 0cm 0.8cm}, clip, width=.3\linewidth]{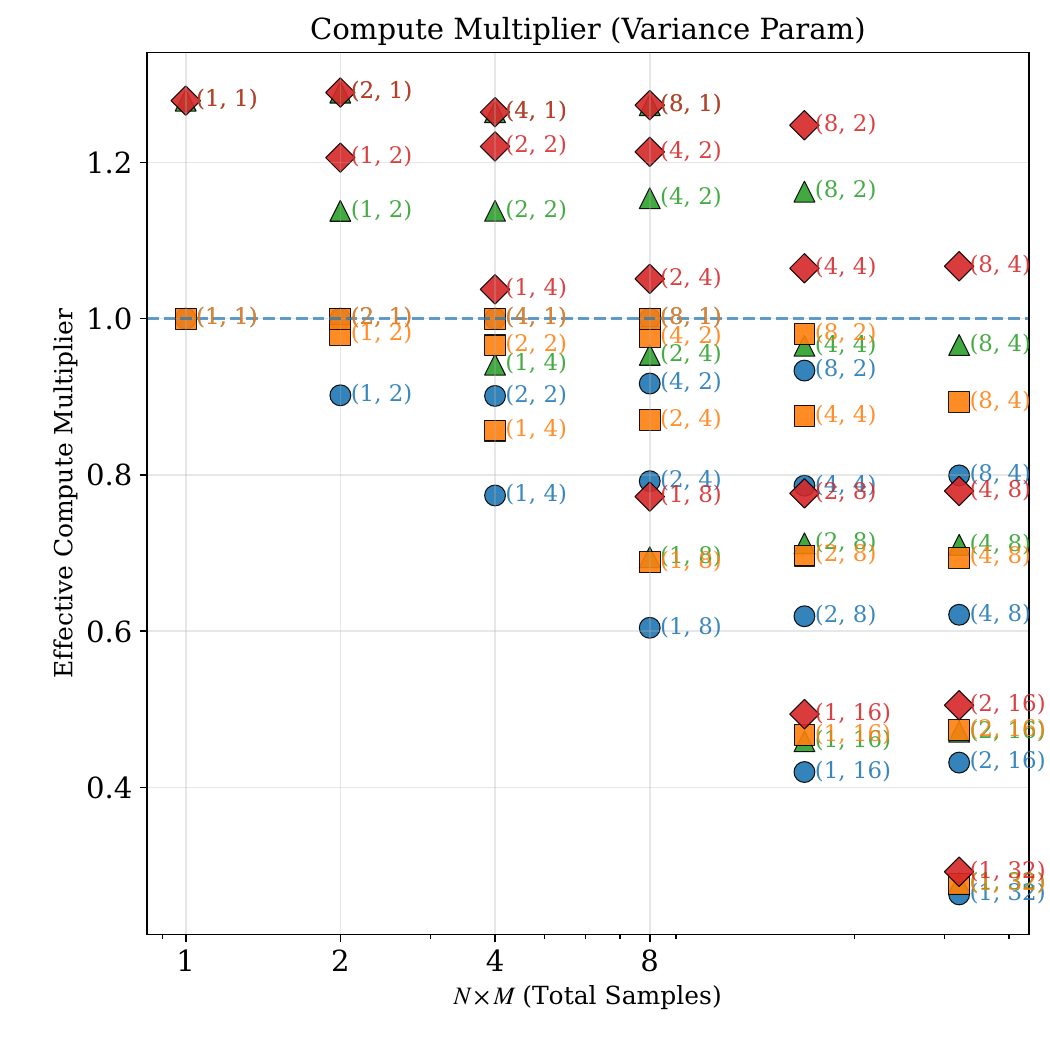}};
                \node [below=of img11, node distance=0cm, yshift=1.2cm](img21){\includegraphics[trim={1.3cm 1.21cm 0cm 0.8cm}, clip, width=.3\linewidth]{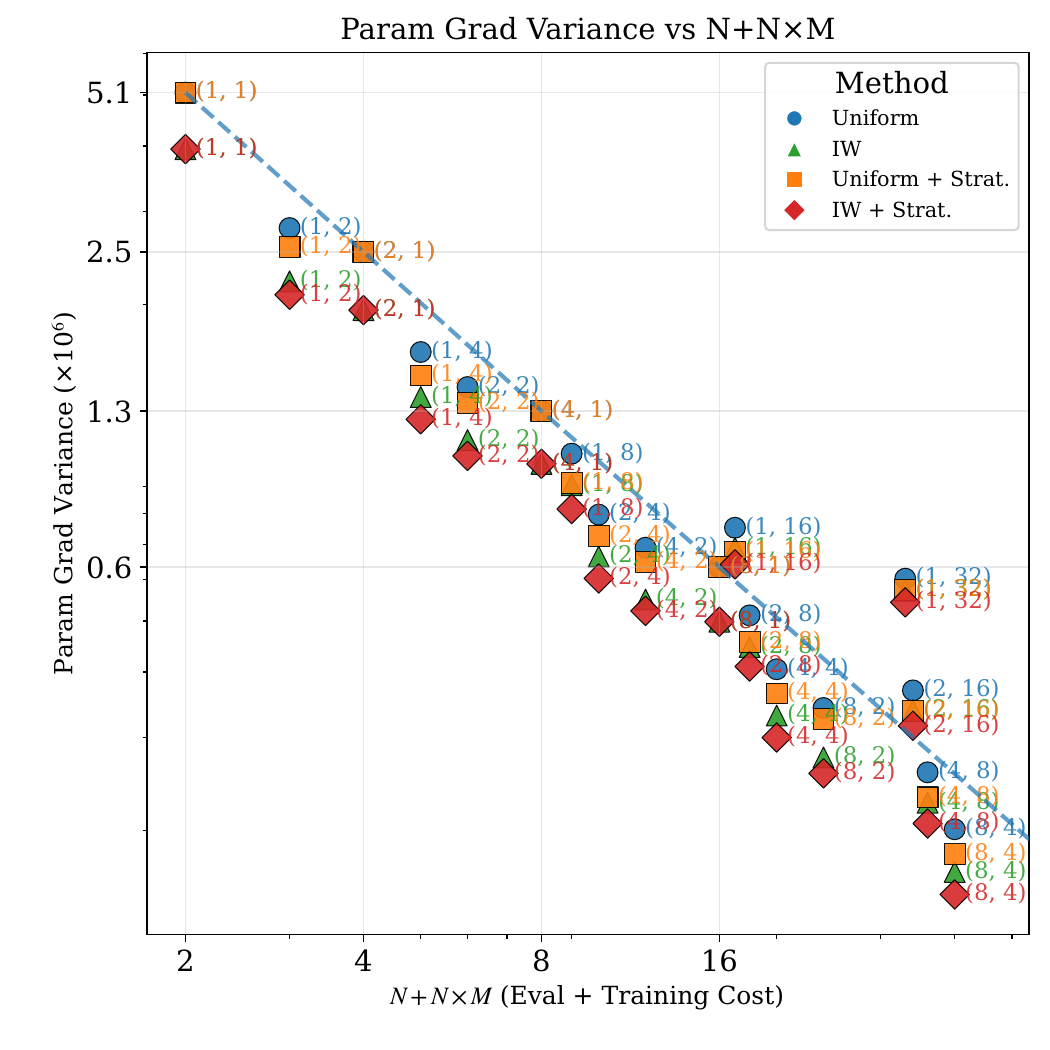}};
                \node [right=of img21, node distance=0cm, xshift=-0.75cm](img22){\includegraphics[trim={1.4cm 1.21cm 0cm 0.8cm}, clip, width=.3\linewidth]{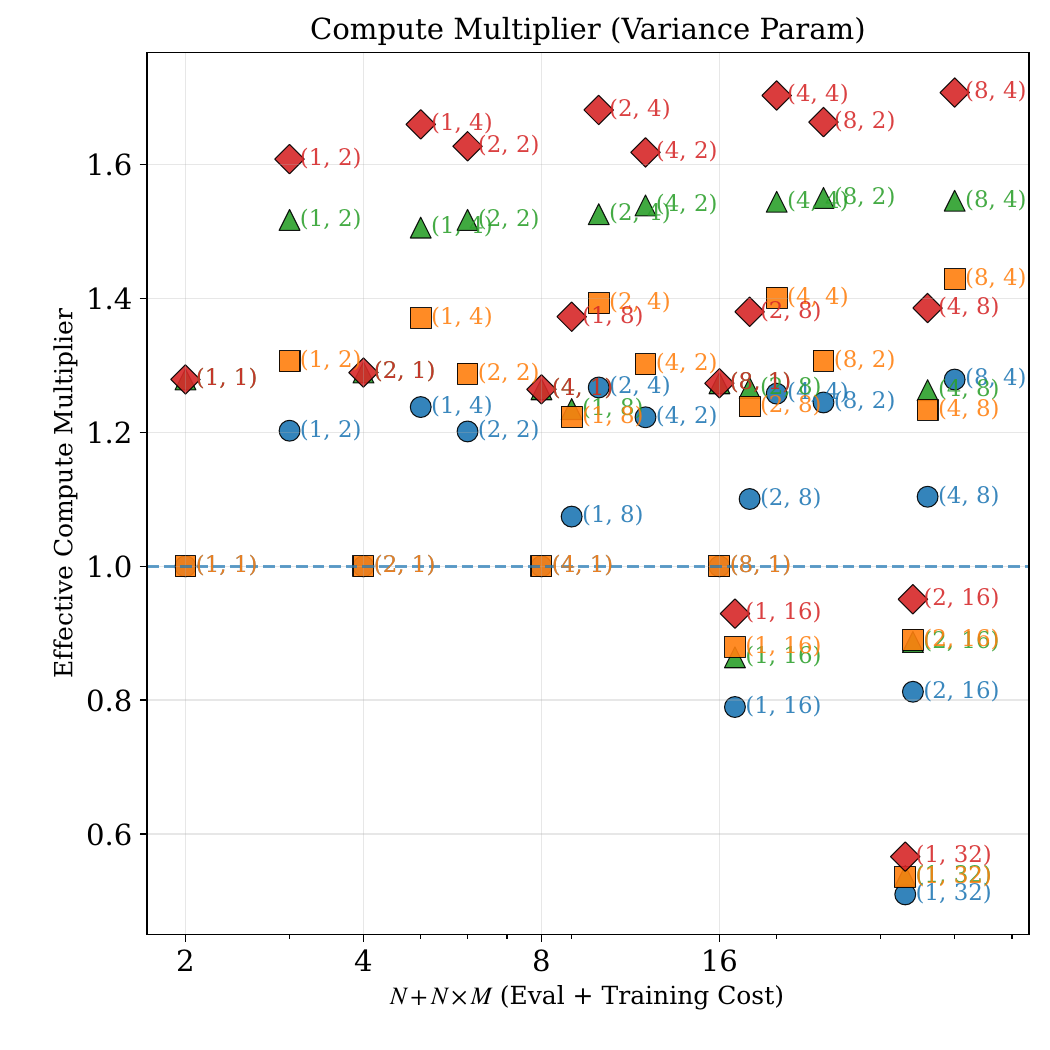}};
                \node [below=of img21, node distance=0cm, yshift=1.2cm](img31){\includegraphics[trim={1.3cm 1.21cm 0cm 0.8cm}, clip, width=.3\linewidth]{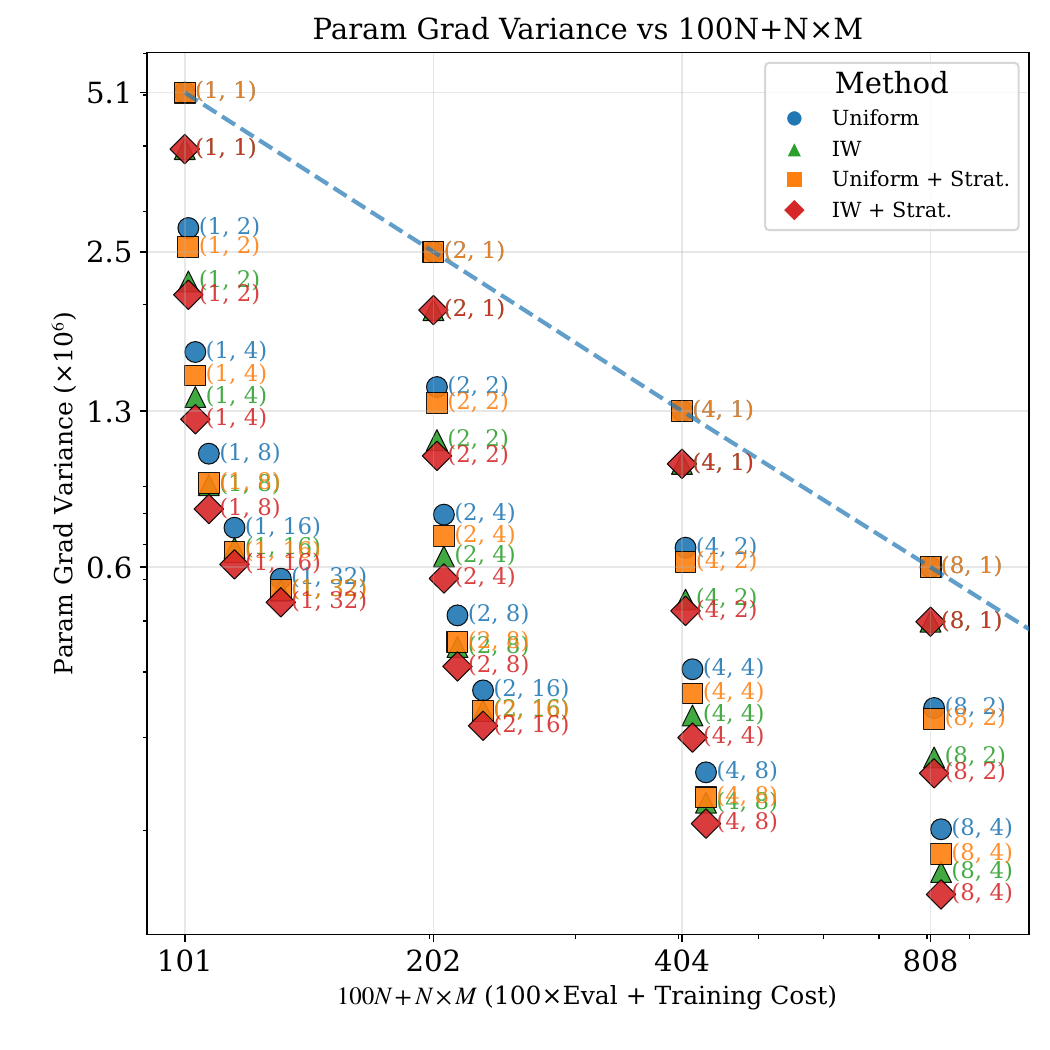}};
                \node [right=of img31, node distance=0cm, xshift=-0.63cm](img32){\includegraphics[trim={1.86cm 1.21cm 0cm 0.8cm}, clip, width=0.2916\linewidth]{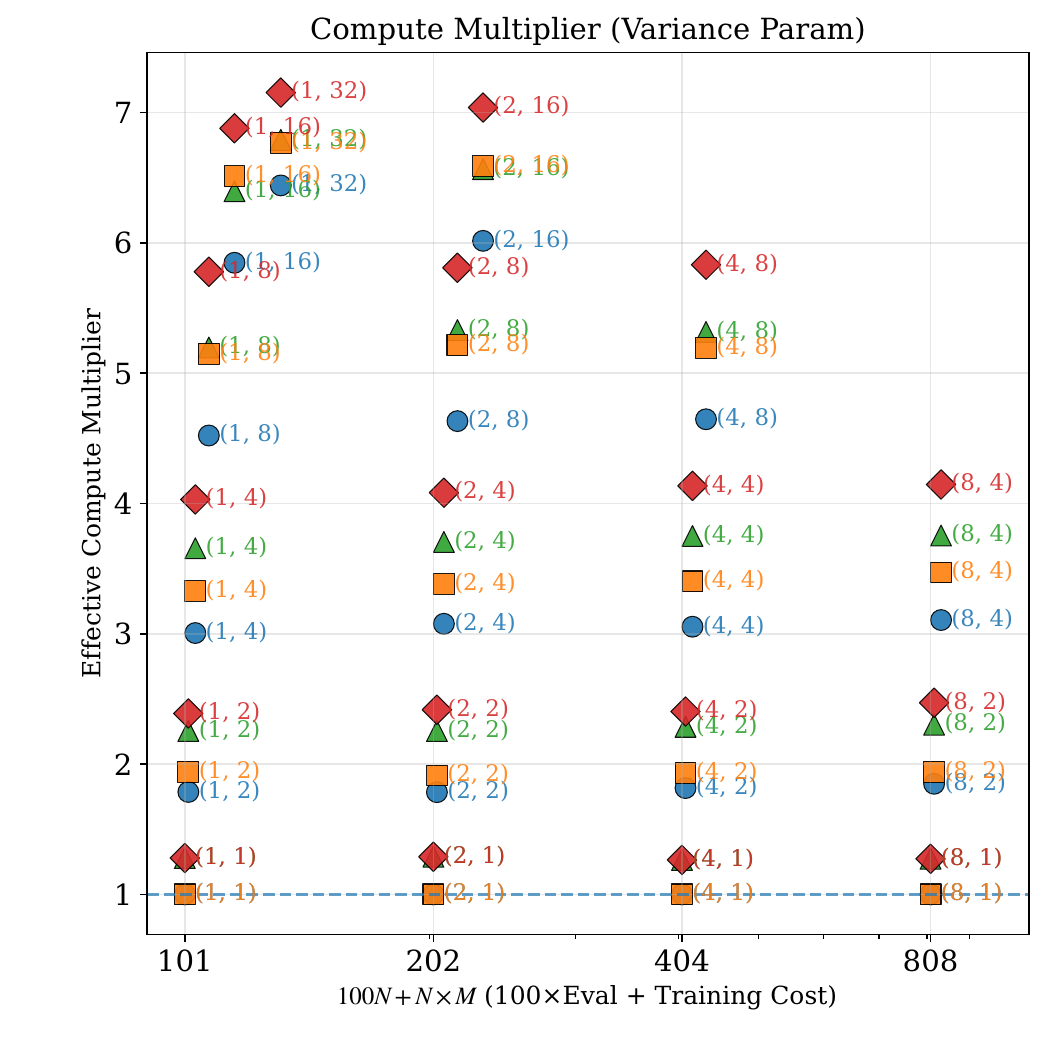}};
                \node[above=of img11, node distance=0cm, xshift=0.0cm, yshift=-1.2cm, font=\fontsize{7.4}{8.9}\selectfont\color{black}]{Variance $(\times10^6)$};
                \node[above=of img12, node distance=0cm, xshift=0.0cm, yshift=-1.2cm, font=\fontsize{7.4}{8.9}\selectfont\color{black}]{Effective Compute Mult. to Baseline};
                \node[left=of img11, node distance=0cm, xshift=1.0cm, yshift=1.5cm, rotate=90, font=\fontsize{7.4}{8.9}\selectfont\color{black}]{Simulated cost scale $\alpha\!=\!0$};
                \node[left=of img21, node distance=0cm, xshift=1.0cm, yshift=1.5cm, rotate=90, font=\fontsize{7.4}{8.9}\selectfont\color{black}]{Simulated cost scale $\alpha\!=\!1$};
                \node[left=of img31, node distance=0cm, xshift=1.0cm, yshift=1.5cm, rotate=90, font=\fontsize{7.4}{8.9}\selectfont\color{black}]{Simulated cost scale $\alpha\!=\!100$};
                \node[below=of img31, node distance=0cm, xshift=0.0cm, yshift=1.2cm, font=\fontsize{7.4}{8.9}\selectfont\color{black}]{Simulated Cost};
                \node[below=of img32, node distance=0cm, xshift=0.0cm, yshift=1.2cm, font=\fontsize{7.4}{8.9}\selectfont\color{black}]{Simulated Cost};
            \end{tikzpicture}
            }
            \caption{
                \textbf{Sensitivity of variance reduction to render-vs-denoise cost ratio.}
                Analysis of \Fig~\ref{fig:quantifying_variance_hierarchical_cost_aware_iw_strat} repeated with simulated cost $\budget = \alpha \numRenders + \numRenders \numReNoises$ to isolate render cost.
                \emph{Top ($\alpha\!=\!0$):} Render free; $\numReNoises\!=\!1$ is best for every method, and IW+Strat stays above the baseline at small $\numReNoises$.
                \emph{Middle ($\alpha\!=\!1$):} Equal render and denoise cost; modest re-noising improves ECM.
                \emph{Bottom ($\alpha\!=\!100$):} Rendering dominates; larger $\numReNoises$ gives larger ECM through render amortization.
                Columns: variance and effective compute multiplier to baseline; the continuation shows relative efficiency to uniform. Colors and annotations follow \Fig~\ref{fig:quantifying_variance_hierarchical_cost_aware_iw_strat}.
            }\label{fig:compute_cost_sensitivity}
            \end{figure}

            \begin{figure}[h!]
            \ContinuedFloat
            \centering
            \resizebox{\linewidth}{!}{
            \begin{tikzpicture}
            \centering
                \node (img13){\includegraphics[trim={1.2cm 1.21cm 0cm 0.8cm}, clip, width=.3\linewidth]{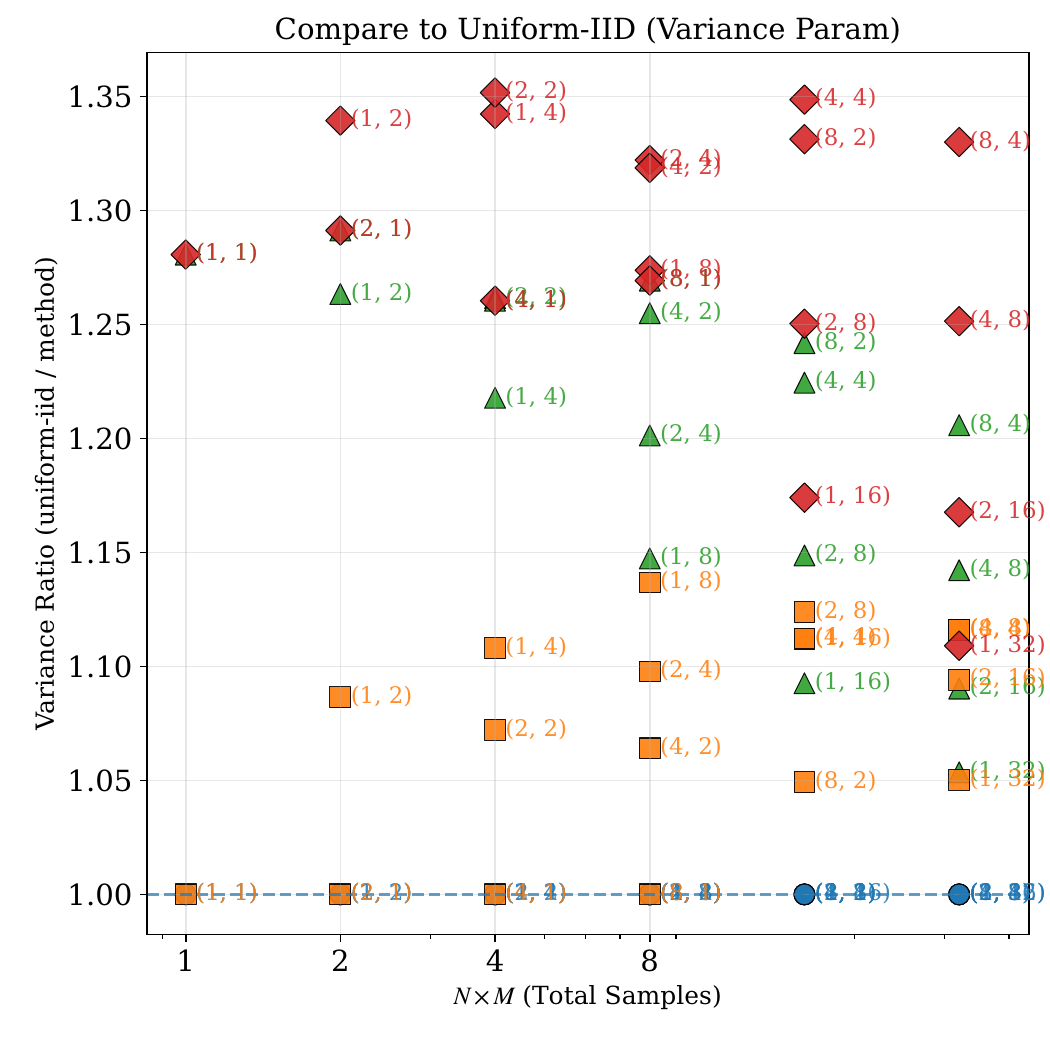}};
                \node [right=of img13, node distance=0cm, xshift=-0.75cm](img23){\includegraphics[trim={1.2cm 1.21cm 0cm 0.8cm}, clip, width=.3\linewidth]{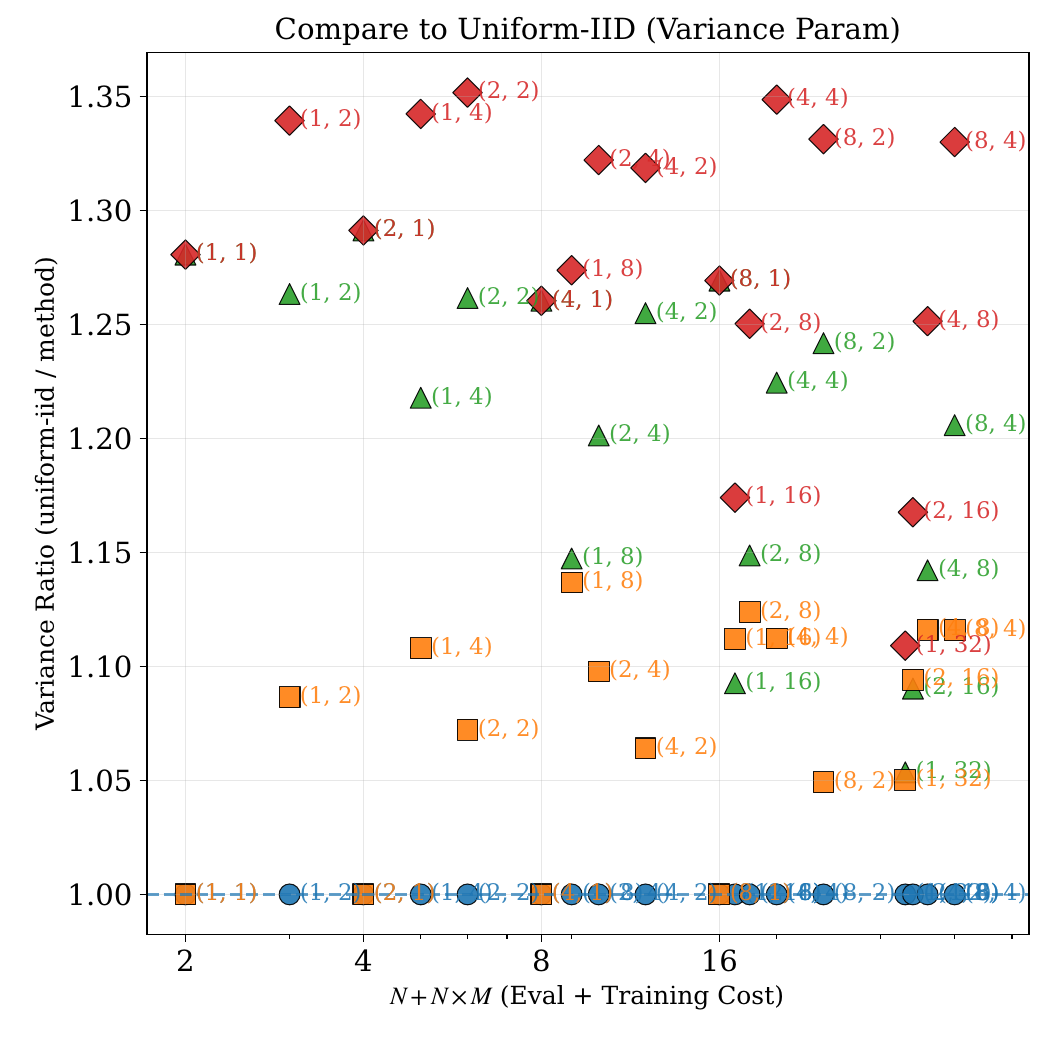}};
                \node [below=of img13, xshift=2.22cm, yshift=0.3cm](img33){\includegraphics[trim={1.2cm 1.21cm 0cm 0.8cm}, clip, width=.3\linewidth]{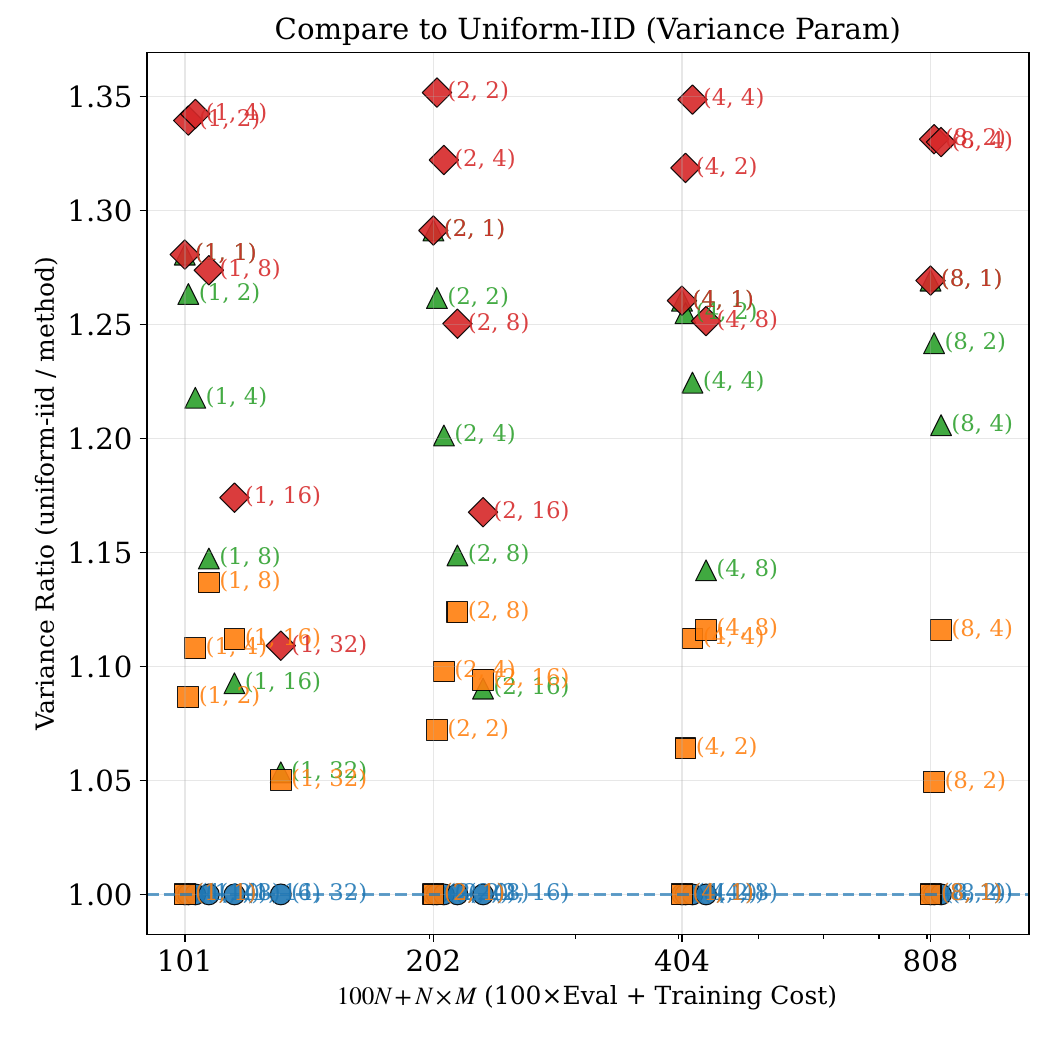}};
                \node[left=of img13, node distance=0cm, rotate=90, xshift=2.1cm, yshift=-.9cm, font=\color{black}]{\scriptsize{Relative Efficiency to Uniform}};
                \node[above=of img13, node distance=0cm, xshift=0.0cm, yshift=-1.2cm, font=\color{black}]{\scriptsize{$\alpha\!=\!0$}};
                \node[below=of img13, node distance=0cm, xshift=0.0cm, yshift=1.15cm, font=\color{black}]{\scriptsize{Simulated Cost}};
                \node[left=of img23, node distance=0cm, rotate=90, xshift=2.1cm, yshift=-.9cm, font=\color{black}]{\scriptsize{Relative Efficiency to Uniform}};
                \node[above=of img23, node distance=0cm, xshift=0.0cm, yshift=-1.2cm, font=\color{black}]{\scriptsize{$\alpha\!=\!1$}};
                \node[below=of img23, node distance=0cm, xshift=0.0cm, yshift=1.15cm, font=\color{black}]{\scriptsize{Simulated Cost}};
                \node[left=of img33, node distance=0cm, rotate=90, xshift=2.1cm, yshift=-.9cm, font=\color{black}]{\scriptsize{Relative Efficiency to Uniform}};
                \node[above=of img33, node distance=0cm, xshift=0.0cm, yshift=-1.2cm, font=\color{black}]{\scriptsize{$\alpha\!=\!100$}};
                \node[below=of img33, node distance=0cm, xshift=0.0cm, yshift=1.15cm, font=\color{black}]{\scriptsize{Simulated Cost}};
            \end{tikzpicture}
            }
            \caption[]{\textbf{(continued)} Relative efficiency to uniform at matched $(\numRenders,\numReNoises)$ for simulated cost scales $\alpha\!=\!0$, $1$, and $100$.}
            \end{figure}
    
   \subsubsection{Importance Sampling: Weight Heuristic vs.\ Oracle}\label{sec:iw_ablation}
           We compare the weight heuristic IW ($\proposalDensity \propto p(\timevar) \sdsWeight(\timevar)$; \Sec~\ref{sec:method-noise-schedules}) against the intractable oracle IW ($\proposalDensity^{\star}$ from \Eq~\ref{eq:opt_proposal}). Here $\gterm$ denotes the full parameter-gradient contribution ($\mathbf{h}$ in \Sec~\ref{sec:method-variance-reduction-strategies}).
            
            \textbf{Why is the oracle intractable?}
                The variance-minimizing proposal $\proposalDensity^{\star}(\timevar) \propto p(\timevar)\sqrt{\E[\|\gterm(\timevar,\randomness)\|_2^2 \mid \timevar]}$ requires estimating the conditional second moment of the gradient contribution at each timestep. For parameter gradients, this requires estimating $\E[\|\nabla_{\genParams}\lossSDS\|_2^2 \mid \timevar]$ across the distribution of renders, camera views, and noise samples. This is prohibitively expensive during training: it requires isolating per-timestep gradient norms, but standard batched backpropagation aggregates contributions across all $\numReNoises$ re-noisings per render (\Sec~\ref{sec:method-compute-reuse}), making per-timestep norms inaccessible without running $\numReNoises$ separate backward passes.

\begin{figure}[h]
                    \centering
                    \fitlinewidth{
                    \begin{tikzpicture}
                    \centering
                        \node (img1){\includegraphics[trim={0.9cm 0.9cm 0cm 0.1cm}, clip, width=.45\linewidth]{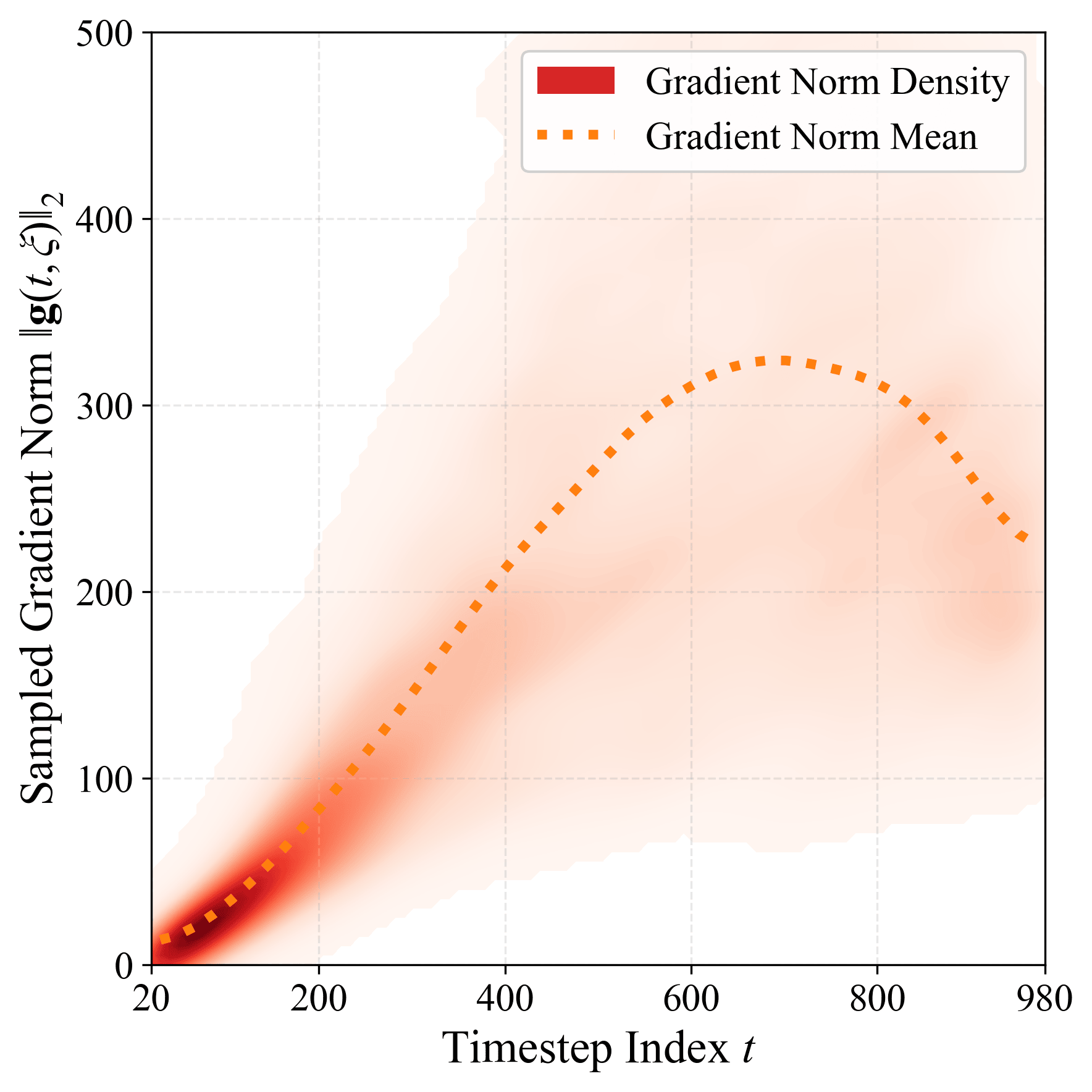}};
                        \node[left=of img1, node distance=0cm, rotate=90, xshift=2.2cm, yshift=-.75cm,  font=\color{black}]{\small{Latent Gradient Norm $\|\sdsWeight(\timevar)\residual\|_2$}};
                        \node[below=of img1, node distance=0cm, xshift=0.0cm, yshift=1.25cm,  font=\color{black}]{\normalsize{Timestep $\timevar$}};
                        
                        \node [right=of img1, xshift=-0.5cm] (img2){\includegraphics[trim={0.9cm 0.9cm 0cm 0.1cm}, clip, width=.45\linewidth]{images/sds/optimal_proposal/sds_param_gradient_norms.png}};
                        \node[left=of img2, node distance=0cm, rotate=90, xshift=2.4cm, yshift=-.8cm,  font=\color{black}]{\small{Parameter Gradient Norm $\|\gterm(\timevar,\randomness)\|_2$}};
                        \node[below=of img2, node distance=0cm, xshift=0.0cm, yshift=1.25cm,  font=\color{black}]{\normalsize{Timestep $\timevar$}};
                    \end{tikzpicture}
                    }
                    \caption{
                        \textbf{Weight function closely tracks gradient magnitude across timesteps.}
                        We visualize empirical gradient norms as a function of timestep $\timevar$ during SDS optimization, alongside the scaled weight function behind the importance-sampling proposal (right panel).
                        \emph{Left:} Latent-space gradient contribution $\|\sdsWeight(\timevar)\residual\|_2$, where $\residual = \noisePred(\noisedData,\timevar,\textCond;\cfgScale)-\noiseVec$ is the noise prediction residual.
                        \emph{Right:} Full parameter gradient norm $\|\gterm(\timevar,\randomness)\|_2$, aggregated over camera views, renders, and noise. The weight-based proposal $\proposalDensity \propto p(\timevar)\sdsWeight(\timevar)$ closely tracks the empirical gradient norm profile, explaining why it attains $94\%$ to $97\%$ of the oracle proposal's relative efficiency (\Tab~\ref{tab:iw_ablation}) without requiring expensive per-timestep norm estimation. The oracle proposal $\proposalDensity^{\star} \propto p(\timevar)\sqrt{\E[\|\gterm(\timevar,\randomness)\|_2^2\mid\timevar]}$ is given for reference.
                    }\label{fig:iwvis_latent_vs_param}
                \end{figure}

            \textbf{How do we evaluate the oracle?}
                Although per-timestep norms are inaccessible when $\numReNoises > 1$, our $\numReNoises = 1$ variance measurement experiments (\Sec~\ref{sec:method-variance-framework-app}) do isolate per-timestep gradient contributions. We construct an empirical oracle by collecting these per-timestep gradient norms, binning $\|\gterm(\timevar, \randomness)\|_2^2$ by timestep, computing the empirical mean within each bin, and using these to define the oracle proposal $\proposalDensity^{\star}$. \Fig~\ref{fig:iwvis_latent_vs_param} visualizes these tracked gradient norms, showing both the weighted latent-space residual $\|\sdsWeight(\timevar)\residual\|_2$ and the full parameter gradient norm $\|\gterm(\timevar, \randomness)\|_2$ as a function of timestep.
            
\begin{figure}[h!]
                    \centering
                    \resizebox{\linewidth}{!}{
                    \begin{tikzpicture}
                    \centering
                        \node (img11){\includegraphics[trim={1.37cm 1.2cm 0cm 0.83cm}, clip, width=.3\linewidth]{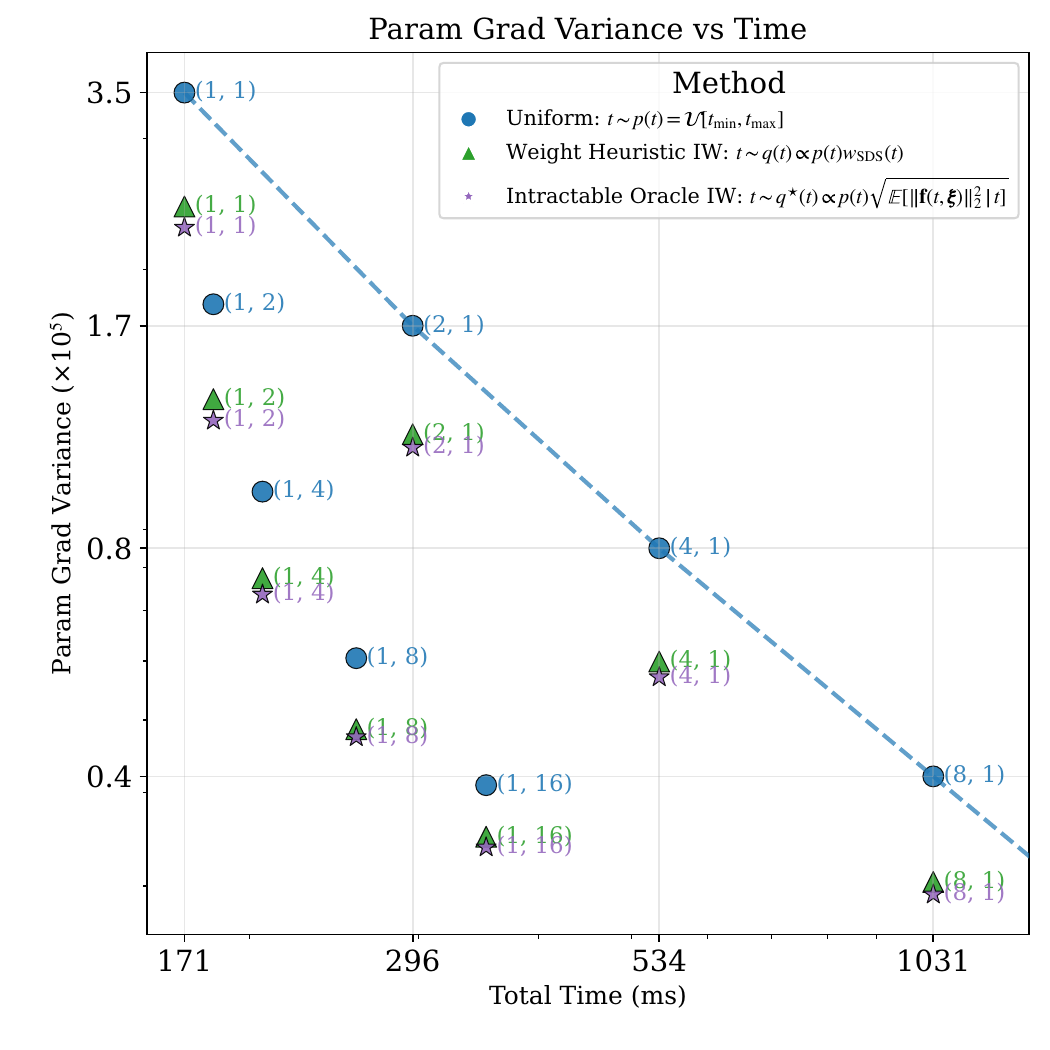}};
                        \node[left=of img11, node distance=0cm, rotate=90, xshift=1.0cm, yshift=-1.0cm,  font=\color{black}]{\scriptsize{Variance $(\times 10^5)$}};
                        \node[below=of img11, node distance=0cm, xshift=-0.0cm, yshift=1.15cm,  font=\color{black}]{\scriptsize{Per-Iteration Compute (ms)}};
        
                        \node [right=of img11, node distance=0cm, xshift=-0.75cm](img21){\includegraphics[trim={1.4cm 1.2cm 0cm 0.83cm}, clip, width=.3\linewidth]{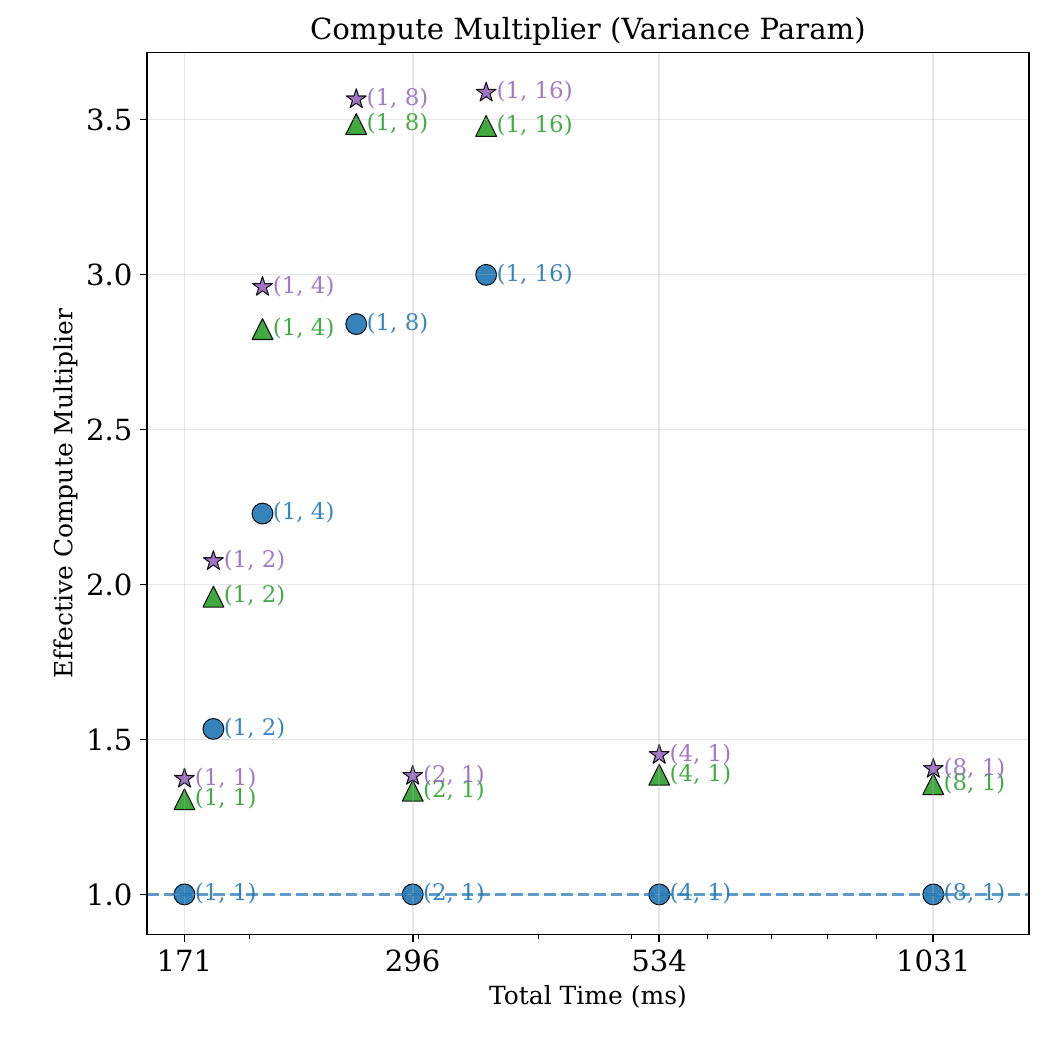}};
                        \node[left=of img21, node distance=0cm, rotate=90, xshift=1.8cm, yshift=-1.0cm,  font=\color{black}]{\scriptsize{Effective Compute Multiplier}};
                        \node[below=of img21, node distance=0cm, xshift=0.0cm, yshift=1.15cm,  font=\color{black}]{\scriptsize{Per-Iteration Compute (ms)}};
        
                        \node [below=of img11, xshift=2.22cm, yshift=0.6cm](img31){\includegraphics[trim={1.41cm 1.2cm 0cm 0.83cm}, clip, width=.3\linewidth]{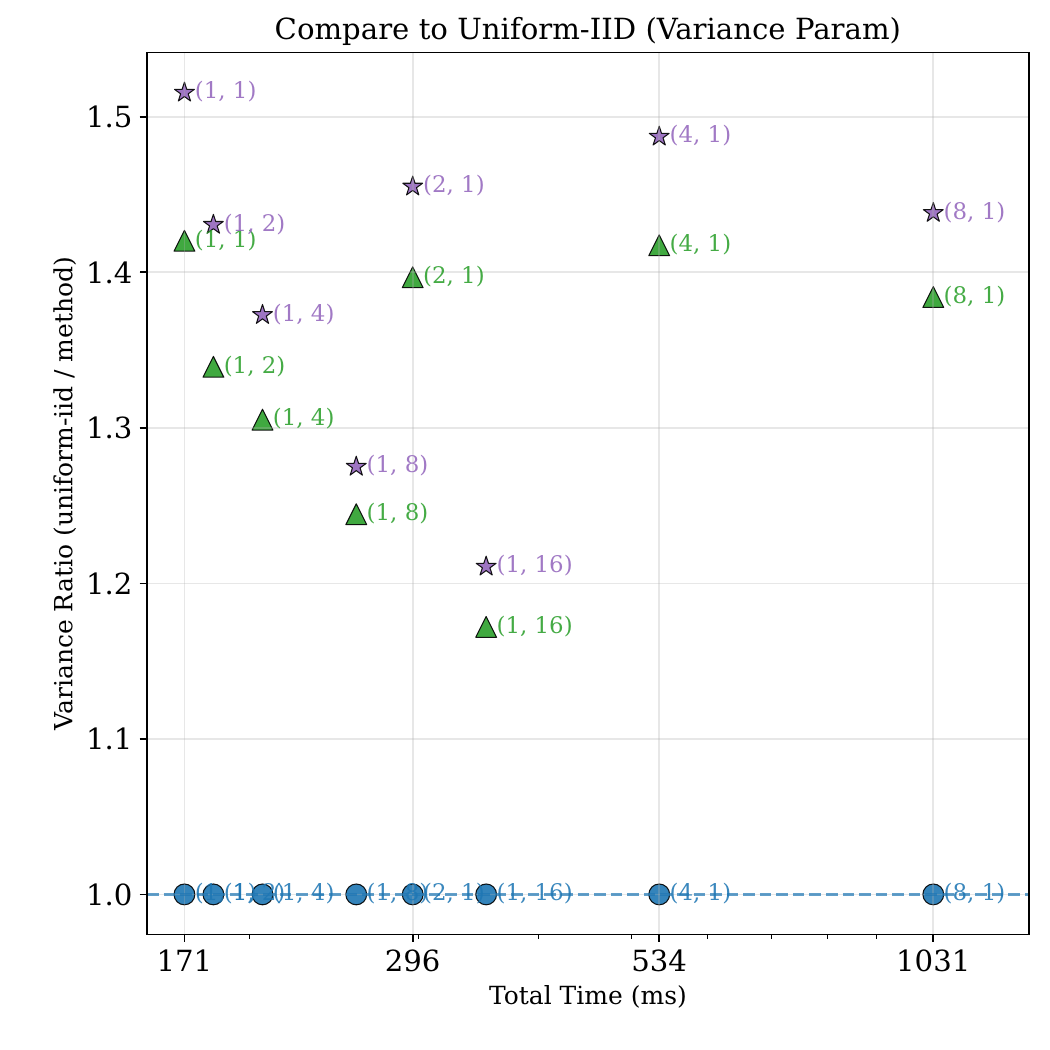}};
                        \node[left=of img31, node distance=0cm, rotate=90, xshift=1.8cm, yshift=-1.0cm,  font=\color{black}]{\scriptsize{Relative Efficiency to Uniform}};
                        \node[below=of img31, node distance=0cm, xshift=0.0cm, yshift=1.15cm,  font=\color{black}]{\scriptsize{Per-Iteration Compute (ms)}};
                    \end{tikzpicture}
                    }
                    \caption{
                        \textbf{Importance Sampling Strategy Comparison: Weight-Based Heuristic versus Oracle.}
                        We compare three importance sampling approaches for parameter gradient estimation: uniform sampling (baseline), our weight-based importance sampling using $\proposalDensity(\timevar) \propto p(\timevar) \sdsWeight(\timevar)$ as described in \Sec~\ref{sec:method-noise-schedules}, and the intractable oracle proposal $\proposalDensity^\star(\timevar) \propto p(\timevar)\sqrt{\E[\|\sdsWeight\jacobian_{\genParams}^\top\residual\|_2^2\mid\timevar]}$ that requires estimating conditional second moments of parameter-gradient contributions.
                        \emph{Top left:} Parameter gradient variance versus compute budget in milliseconds.  Points are annotated by $(\numRenders,\numReNoises)$ configurations.
                        \emph{Top right:} Effective compute multiplier relative to the uniform $\numReNoises\!=\!1$ baseline curve.
                        \emph{Bottom:} Relative efficiency (RE) isolating the gain from importance sampling by comparing to uniform sampling at the same $(\numRenders,\numReNoises)$ configuration. The weight-based heuristic achieves $\sim\!17$--$40\%$ improvements.
                        Our zero-cost, weight-based importance-sampling heuristic performs nearly as well as the intractable oracle across all compute budgets, so $\sdsWeight(\timevar)$ is a practical proxy for gradient magnitude when designing importance proposals.
                    }\label{fig:sds_optimal_importance}
                \end{figure}

            \textbf{Results.}
                \Fig~\ref{fig:sds_optimal_importance} and \Tab~\ref{tab:iw_ablation} summarize results on the low-guidance ($\cfgScale\!=\!25$) runs of \App~\Sec~\ref{sec:low_guidance_analysis} at training step $\num{5000}$, averaged over prompts; the uniform and weight-heuristic points are those of \Fig~\ref{fig:quantifying_variance_low_guidance}. Both IW methods reduce variance more than uniform sampling (\Tab~\ref{tab:iw_ablation}). The weight heuristic attains $94\%$ to $97\%$ of the oracle's relative efficiency across all $\numReNoises$, so $\sdsWeight(\timevar)$ is a practical proxy for the optimal proposal. Because $\sdsWeight$ is deterministic given $\timevar$, the oracle factors as $\proposalDensity^\star(\timevar)\propto p(\timevar)\,\sdsWeight(\timevar)\sqrt{\E[\|\jacobian_{\genParams}^\top\residual\|_2^2\mid\timevar]}$, so $\proposalDensity\propto p\,\sdsWeight$ is exact when that conditional second moment is flat in $\timevar$. The Jacobian stays in the estimator and is dropped only from the proposal; our empirical oracle includes it, so the $94\%$ to $97\%$ relative efficiency measures what dropping it costs. This close agreement is explained by \Fig~\ref{fig:iwvis_latent_vs_param}: the weight function $\sdsWeight(\timevar)$ closely tracks the empirical gradient norm profile, so importance sampling with $\proposalDensity \propto p(\timevar)\sdsWeight(\timevar)$ approximates the oracle without requiring expensive norm estimation.

                \begin{table}[h!]
                    \centering
                    \caption{
                        Importance sampling ablation on the low-guidance ($\cfgScale\!=\!25$) runs of \App~\Sec~\ref{sec:low_guidance_analysis} at training step $\num{5000}$, averaged over prompts. \textbf{(a)} ECM relative to uniform $\numReNoises\!=\!1$. \textbf{(b)} Relative efficiency vs.\ uniform at matched $\numReNoises$. The weight heuristic attains $94\%$ to $97\%$ of the oracle's relative efficiency.
                    }\label{tab:iw_ablation}
                    \vspace{0.5em}
                    \begin{minipage}[t]{0.48\textwidth}
                        \centering
                        \textbf{(a) Effective Compute Multiplier}
                        \vspace{0.3em}
                        \scalebox{0.75}{
                        \begin{tabular}{lrrr}
                        \toprule
                        $\numReNoises$ & \multicolumn{1}{c}{Uniform} & \multicolumn{1}{c}{Weight Heur.\ IW} & \multicolumn{1}{c}{Oracle IW} \\
                        \midrule
                        1 & 1.00$\times$ & 1.31$\times$ & \textbf{1.39}$\times$ \\
                        2 & 1.65$\times$ & 2.12$\times$ & \textbf{2.23}$\times$ \\
                        4 & 2.41$\times$ & 2.98$\times$ & \textbf{3.12}$\times$ \\
                        8 & 3.00$\times$ & 3.55$\times$ & \textbf{3.66}$\times$ \\
                        16 & 3.13$\times$ & 3.54$\times$ & \textbf{3.61}$\times$ \\
                        \bottomrule
                        \end{tabular}
                        }
                    \end{minipage}
                    \hfill
                    \begin{minipage}[t]{0.48\textwidth}
                        \centering
                        \textbf{(b) Relative Efficiency vs.\ Uniform}
                        \vspace{0.3em}
                        \scalebox{0.75}{
                        \begin{tabular}{lrr}
                        \toprule
                        $\numReNoises$ & \multicolumn{1}{c}{Weight Heur.\ IW} & \multicolumn{1}{c}{Oracle IW} \\
                        \midrule
                        1 & 1.40$\times$ & \textbf{1.47}$\times$ \\
                        2 & 1.34$\times$ & \textbf{1.43}$\times$ \\
                        4 & 1.31$\times$ & \textbf{1.37}$\times$ \\
                        8 & 1.24$\times$ & \textbf{1.28}$\times$ \\
                        16 & 1.17$\times$ & \textbf{1.21}$\times$ \\
                        \bottomrule
                        \end{tabular}
                        }
                    \end{minipage}
                \end{table}
        
   \subsubsection{Prompt Ablation}\label{sec:prompt_ablation}
        We evaluate the consistency of variance reduction benefits across five diverse text prompts: emerald beetle, gold mask, mahogany piano, orchid pot, and teddy bear. \Tab~\ref{tab:prompt_ablation} summarizes the results.

        \textbf{Results.}
            Our variance reduction methods behave consistently across prompts. All prompts achieve peak ECM at $\numReNoises=8$ (\Tab~\ref{tab:prompt_ablation}a), with values ranging from $2.73\times$ (gold) to $3.58\times$ (mahogany). The marginal benefit of IW+Strat over uniform peaks at $\numReNoises$ from $2$ to $4$ (\Tab~\ref{tab:prompt_ablation}b), consistent with diminishing returns from importance weighting at higher $\numReNoises$ as the estimator approaches the continuous limit. Method rankings remain stable across all prompts (\Tab~\ref{tab:prompt_ablation}c,d): IW+Strat consistently outperforms both IW-only and Strat-only, which in turn outperform uniform sampling. This confirms that the complementary nature of importance weighting and stratification is not prompt-specific.

            \begin{table}[!htbp]
            \centering
            \caption{
                \textbf{Prompt ablation for variance reduction methods.}
                Five prompts (emerald beetle, gold mask, mahogany piano, orchid pot, teddy bear).
                \textbf{(a)} ECM for IW+Strat by $\numReNoises$; $\numReNoises\!=\!8$ is optimal across prompts.
                \textbf{(b)} RE vs.\ uniform for IW+Strat; peak at $\numReNoises\!=\!2$--$4$.
                \textbf{(c)} ECM for all methods at $\numReNoises\!=\!8$; rankings (IW+Strat $>$ IW $\approx$ Strat $>$ Uniform) are stable.
                \textbf{(d)} RE vs.\ uniform at $\numReNoises\!=\!8$; IW and Strat are complementary across prompts.
            }\label{tab:prompt_ablation}
            \vspace{0.3em}
            
            \begin{minipage}[t]{0.48\textwidth}
                \centering
                \textbf{(a) ECM by $\numReNoises$ (IW+Strat)}
                \vspace{0.3em}
                
                \scalebox{0.62}{
                \begin{tabular}{lrrrrrr}
                \toprule
                Prompt & \multicolumn{1}{c}{$\numReNoises\!=\!1$} & \multicolumn{1}{c}{$\numReNoises\!=\!2$} & \multicolumn{1}{c}{$\numReNoises\!=\!4$} & \multicolumn{1}{c}{$\numReNoises\!=\!8$} & \multicolumn{1}{c}{$\numReNoises\!=\!16$} & \multicolumn{1}{c}{$\numReNoises\!=\!32$} \\
                \midrule
                Emerald  & 1.23$\times$ & 2.05$\times$ & 2.99$\times$ & \textbf{3.41$\times$} & 3.12$\times$ & 2.40$\times$ \\
                Gold     & 1.20$\times$ & 1.95$\times$ & 2.59$\times$ & \textbf{2.73$\times$} & 2.33$\times$ & 1.73$\times$ \\
                Mahogany & 1.28$\times$ & 2.07$\times$ & 3.12$\times$ & \textbf{3.58$\times$} & 3.24$\times$ & 2.32$\times$ \\
                Orchid   & 1.24$\times$ & 2.04$\times$ & 2.98$\times$ & \textbf{3.33$\times$} & 2.97$\times$ & 2.15$\times$ \\
                Teddy    & 1.26$\times$ & 2.13$\times$ & 3.01$\times$ & \textbf{3.42$\times$} & 3.05$\times$ & 2.31$\times$ \\
                \midrule
                Avg      & 1.24$\times$ & 2.05$\times$ & 2.94$\times$ & \textbf{3.29$\times$} & 2.94$\times$ & 2.18$\times$ \\
                \bottomrule
                \end{tabular}
                }
            \end{minipage}
            \hfill
            \begin{minipage}[t]{0.48\textwidth}
                \centering
                \textbf{(b) RE by $\numReNoises$ (IW+Strat)}
                \vspace{0.3em}
                
                \scalebox{0.62}{
                \begin{tabular}{lrrrrrr}
                \toprule
                Prompt & \multicolumn{1}{c}{$\numReNoises\!=\!1$} & \multicolumn{1}{c}{$\numReNoises\!=\!2$} & \multicolumn{1}{c}{$\numReNoises\!=\!4$} & \multicolumn{1}{c}{$\numReNoises\!=\!8$} & \multicolumn{1}{c}{$\numReNoises\!=\!16$} & \multicolumn{1}{c}{$\numReNoises\!=\!32$} \\
                \midrule
                Emerald  & 1.23$\times$ & \textbf{1.29$\times$} & 1.28$\times$ & 1.21$\times$ & 1.14$\times$ & 1.07$\times$ \\
                Gold     & 1.20$\times$ & \textbf{1.27$\times$} & 1.25$\times$ & 1.18$\times$ & 1.12$\times$ & 1.12$\times$ \\
                Mahogany & 1.28$\times$ & 1.32$\times$ & \textbf{1.35$\times$} & 1.29$\times$ & 1.22$\times$ & 1.11$\times$ \\
                Orchid   & 1.24$\times$ & 1.31$\times$ & \textbf{1.35$\times$} & 1.28$\times$ & 1.19$\times$ & 1.09$\times$ \\
                Teddy    & 1.26$\times$ & \textbf{1.33$\times$} & 1.33$\times$ & 1.28$\times$ & 1.17$\times$ & 1.11$\times$ \\
                \midrule
                Avg      & 1.24$\times$ & 1.30$\times$ & \textbf{1.31$\times$} & 1.25$\times$ & 1.17$\times$ & 1.10$\times$ \\
                \bottomrule
                \end{tabular}
                }
            \end{minipage}
            
            \vspace{1em}
            
            \begin{minipage}[t]{0.48\textwidth}
                \centering
                \textbf{(c) ECM at $\numReNoises=8$ (all methods)}
                \vspace{0.3em}
                
                \scalebox{0.72}{
                \begin{tabular}{lrrrr}
                \toprule
                Prompt & \multicolumn{1}{c}{Uniform} & \multicolumn{1}{c}{IW} & \multicolumn{1}{c}{Strat.} & \multicolumn{1}{c}{IW+Strat.} \\
                \midrule
                Emerald  & 2.80$\times$ & 3.17$\times$ & 3.15$\times$ & \textbf{3.41$\times$} \\
                Gold     & 2.32$\times$ & 2.55$\times$ & 2.55$\times$ & \textbf{2.73$\times$} \\
                Mahogany & 2.78$\times$ & 3.22$\times$ & 3.07$\times$ & \textbf{3.58$\times$} \\
                Orchid   & 2.60$\times$ & 2.98$\times$ & 2.98$\times$ & \textbf{3.33$\times$} \\
                Teddy    & 2.67$\times$ & 3.13$\times$ & 3.04$\times$ & \textbf{3.42$\times$} \\
                \midrule
                Avg      & 2.63$\times$ & 3.01$\times$ & 2.96$\times$ & \textbf{3.29$\times$} \\
                \bottomrule
                \end{tabular}
                }
            \end{minipage}
            \hfill
            \begin{minipage}[t]{0.48\textwidth}
                \centering
                \textbf{(d) RE at $\numReNoises=8$ (all methods)}                \vspace{0.3em}
                
                \scalebox{0.72}{
                \begin{tabular}{lrrrr}
                \toprule
                Prompt & \multicolumn{1}{c}{Uniform} & \multicolumn{1}{c}{IW} & \multicolumn{1}{c}{Strat.} & \multicolumn{1}{c}{IW+Strat.} \\
                \midrule
                Emerald  & 1.00$\times$ & 1.13$\times$ & 1.12$\times$ & \textbf{1.21$\times$} \\
                Gold     & 1.00$\times$ & 1.10$\times$ & 1.10$\times$ & \textbf{1.18$\times$} \\
                Mahogany & 1.00$\times$ & 1.16$\times$ & 1.10$\times$ & \textbf{1.29$\times$} \\
                Orchid   & 1.00$\times$ & 1.15$\times$ & 1.15$\times$ & \textbf{1.28$\times$} \\
                Teddy    & 1.00$\times$ & 1.17$\times$ & 1.14$\times$ & \textbf{1.28$\times$} \\
                \midrule
                Avg      & 1.00$\times$ & 1.14$\times$ & 1.12$\times$ & \textbf{1.25$\times$} \\
                \bottomrule
                \end{tabular}
                }
            \end{minipage}
            
            \end{table}

    \FloatBarrier
    \subsection{Single-Step Diffusion Distillation}\label{sec:experiments-dmd-app}

        \subsubsection{Details}\label{sec:experiments-dmd-app-details}

            \textbf{Hyperparameters:}
                For all of our diffusion distillation experiments, we use the public DiT-XL/2 weights published alongside the DiT source code~\citep{peebles2023scalable}. We used the ImageNet-256 dataset~\citep{deng2009imagenet}, encoded with the pretrained Stable Diffusion encoder~\citep{rombach2022high}, per the original teacher model.
            
                We train one-step generators using the DMD2 algorithm~\citep{yin2024improved}. This replaces the regression loss with an additional learned-feature discriminator and an alternating update schedule for the fake score and the generator network. Unless otherwise specified, we use a learning rate of $10^{-5}$ for all experiments and perform 5 fake-score/discriminator updates for each student update.

            \textbf{Compute Usage:}
                NVIDIA A100 GPUs. Batch $8$ baseline $(8,1)$: $\sim\!0.38$\,s/iter; $8\times$ resampling $\sim\!0.47$\,s; $16\times$ resampling $\sim\!0.59$\,s. Batch $48$: $\sim\!1.0$\,s/iter. We evaluate FID at checkpoints, averaging over $5$ seeds.

        \subsubsection{Results}\label{sec:experiments-dmd-app-results}

            \begin{figure}
                \centering
                \scalebox{1.0}{
                \begin{tikzpicture}
                \centering
                    \node (img11){\includegraphics[trim={0.8cm 0.8cm 0.0cm 0.8cm}, clip, width=.7\linewidth]{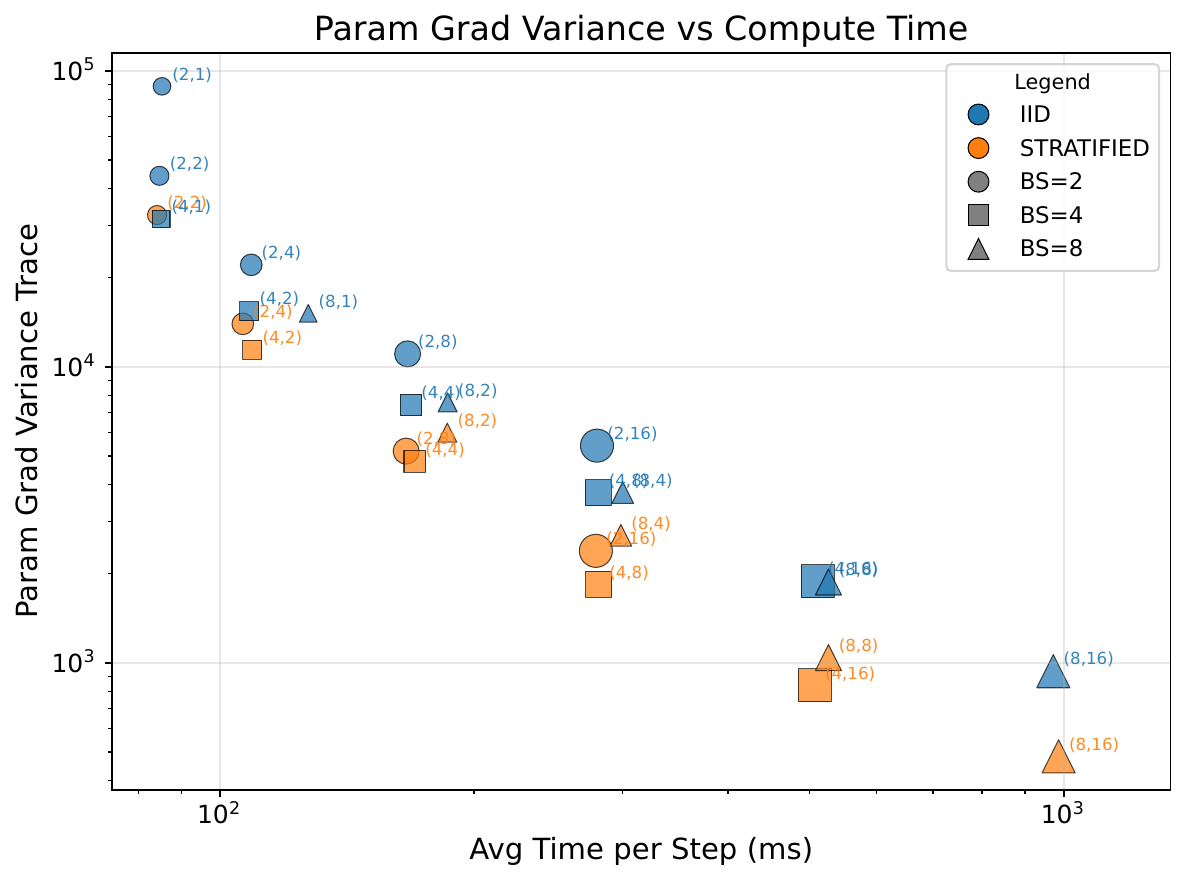}};
                    \node[left=of img11, node distance=0cm, rotate=90, xshift=1.5cm, yshift=-.8cm,  font=\color{black}]{\normalsize{Gradient Variance}};
                    \node[below=of img11, node distance=0cm, xshift=0.0cm, yshift=1.15cm,  font=\color{black}]{\normalsize{Per-Iteration Compute (ms)}};
                \end{tikzpicture}
                }
                \caption{
                    \textbf{Quantifying variance reduction against compute cost for one-step distillation.} Variance of the parameter-gradient estimator (\Eq~\ref{eq:two-level}) versus per-step compute in the variance runs; absolute values are on a different scale from \Tab~\ref{tab:dmd_variance} and the training timings of \App~\Sec~\ref{sec:experiments-dmd-app-details}, but variance ratios between configurations agree with \Tab~\ref{tab:dmd_variance} (${\sim}16\times$ from $(8,1)$ to $(8,16)$ IID). Points are labeled with $(\numRenders, \numReNoises)$; marker shape encodes $\numRenders$ (legend ``BS''). Resampling and stratification both reduce variance, with combined methods achieving the lowest variance per unit compute. However, these variance reductions do not translate to improved FID scores at matched wall-clock time (\Fig~\ref{fig:student_fid_combined}).
                } \label{fig:dmd2_frontier}
            \end{figure}

            We follow our variance computation framework to evaluate the variance of the Monte Carlo estimator of \Eq~\ref{eq:dmd_mc_estimator}. We run the online variance estimator for at least $\num{1000}$ steps, stopping when the parameter gradient variance remains within $0.1\%$ for $50$ consecutive steps. We compute the variance estimates after training a one-step generator for $\num{20000}$ steps using a batch size of $8$. The $(8,1)$ to $(8,16)$ IID parameter-gradient ratio in \Tab~\ref{tab:dmd_variance}, $982/59.9\!\approx\!16.4$, slightly exceeds the factor of $16$ that \Eq~\ref{eq:two-level} allows for $16$ IID re-noisings, so these ratios carry estimation noise. We also investigated the effectiveness of variance reduction at initialization\footnote{At initialization, the VSD loss gradient is zero as the fake-score model is initialized from the teacher weights. But we found that the teacher SDS loss exhibited an $\sim\!4\times$ variance reduction.} and earlier checkpoints, finding a similar overall pattern.
    
            \textbf{Investigating the effect on FID.}
                 These experiments ask why lower gradient variance did not translate into better FID in diffusion distillation.

            \begin{figure}[t]
                \centering
                \scalebox{1.0}{
                \begin{tikzpicture}
                \centering
                    \node (img11){\includegraphics[trim={0.8cm 1.3cm 0cm 0.8cm}, clip, width=.46\linewidth]{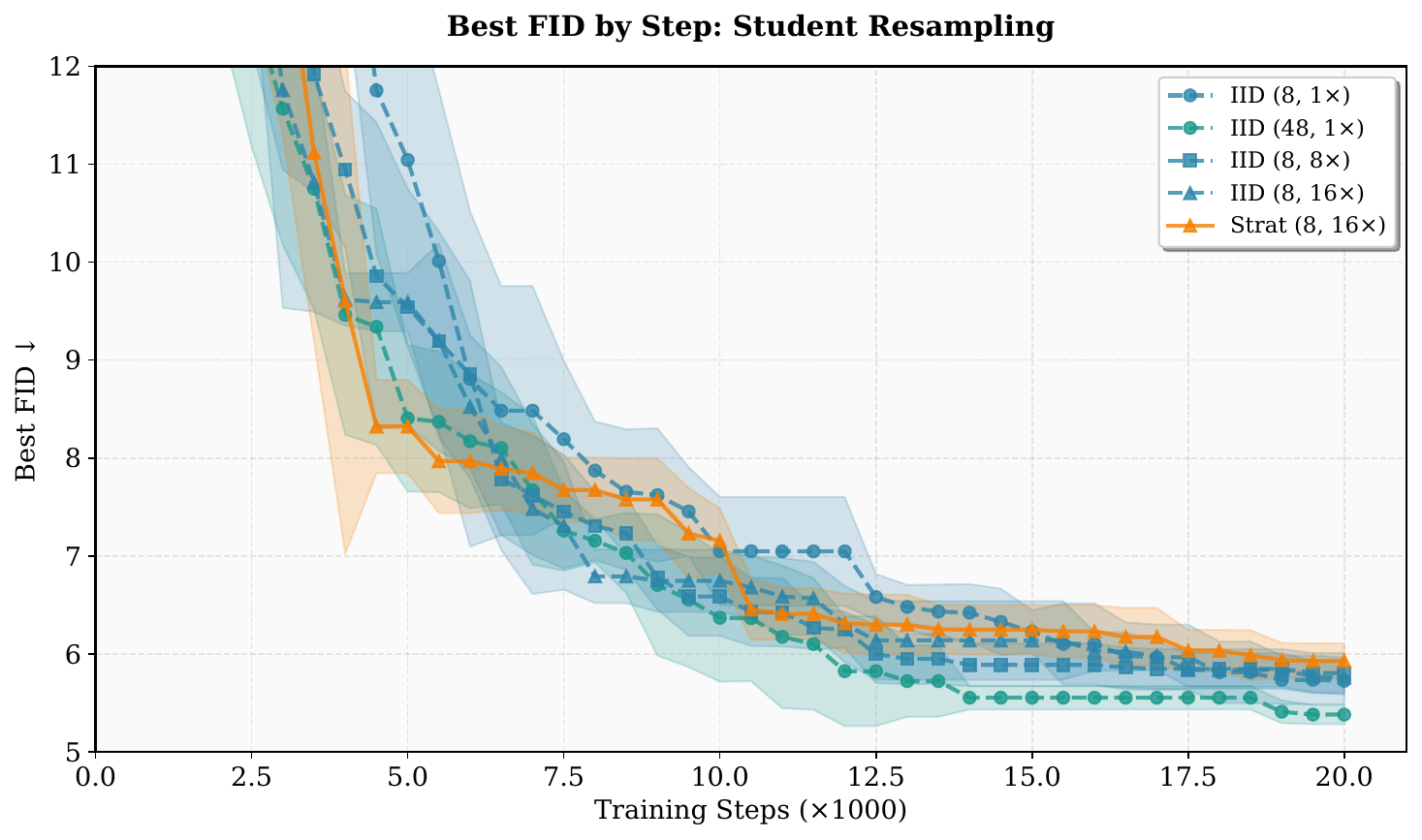}};
                    \node[left=of img11, node distance=0cm, rotate=90, xshift=1.25cm, yshift=-.8cm,  font=\color{black}]{\normalsize{Best FID $\downarrow$}};
                    
                    \node [right=of img11, node distance=0cm, xshift=-1.0cm](img12){\includegraphics[trim={1.5cm 1.3cm 0cm 0.8cm}, clip, width=.45\linewidth]{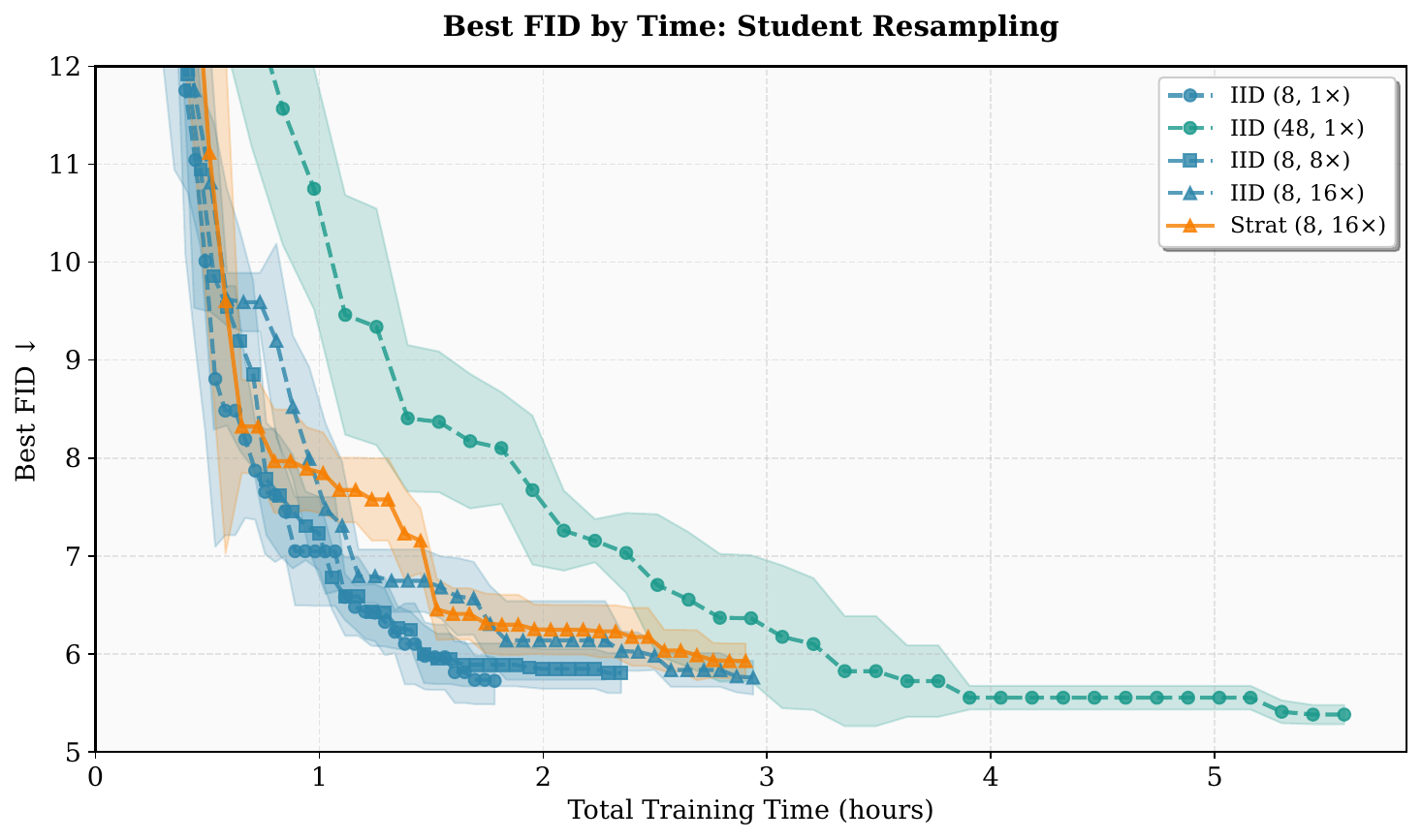}};
                    
                    \node [below=of img11, node distance=0cm, yshift=1.1cm](img21){\includegraphics[trim={0.8cm 0.8cm 0cm 0.8cm}, clip, width=.46\linewidth]{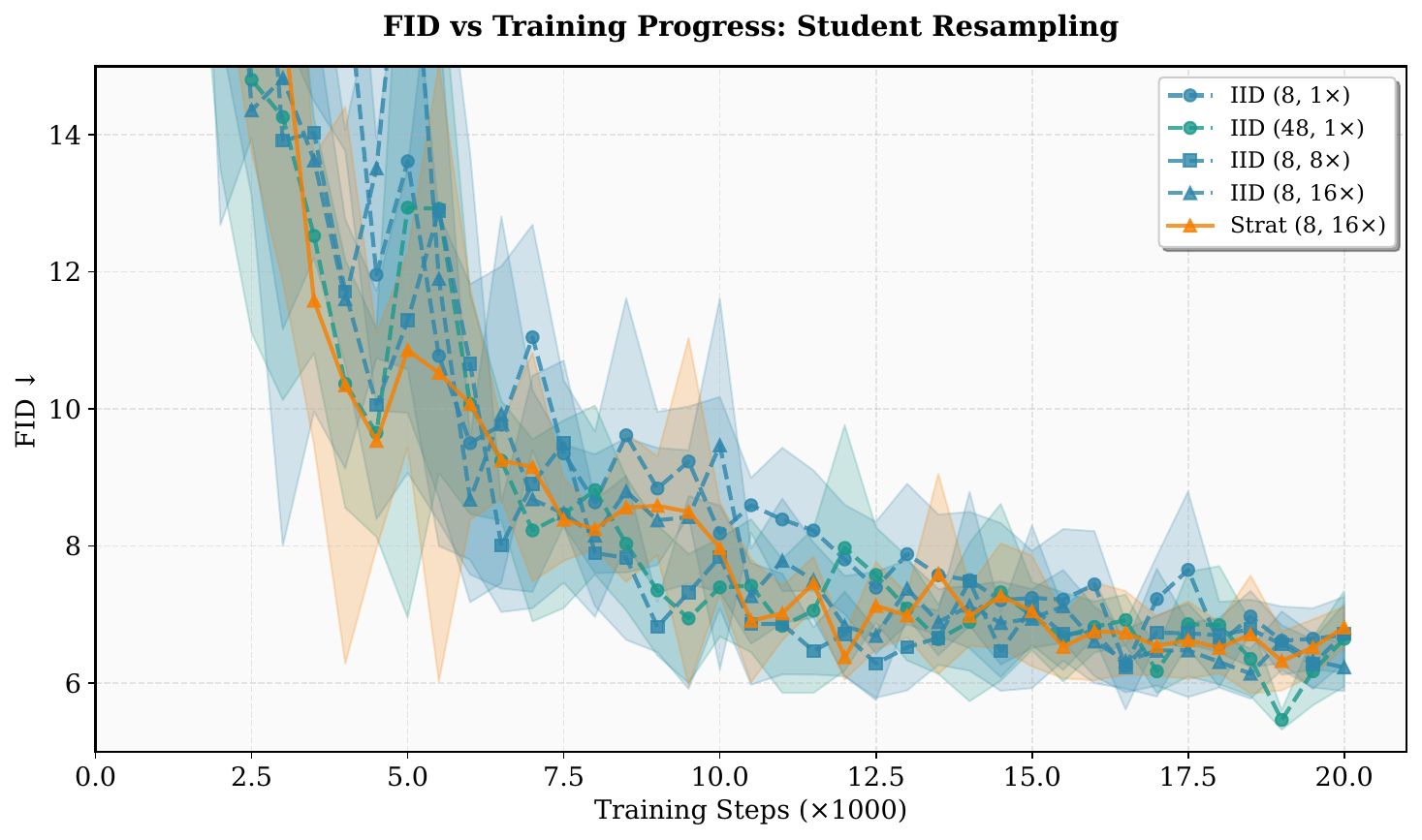}};
                    \node[left=of img21, node distance=0cm, rotate=90, xshift=0.8cm, yshift=-.8cm,  font=\color{black}]{\normalsize{FID $\downarrow$}};
                    \node[below=of img21, node distance=0cm, xshift=0.0cm, yshift=1.15cm,  font=\color{black}]{\normalsize{Training Steps ($\times 1000$)}};
                    
                    \node [right=of img21, node distance=0cm, xshift=-1.0cm](img22){\includegraphics[trim={1.5cm 0.8cm 0cm 0.8cm}, clip, width=.45\linewidth]{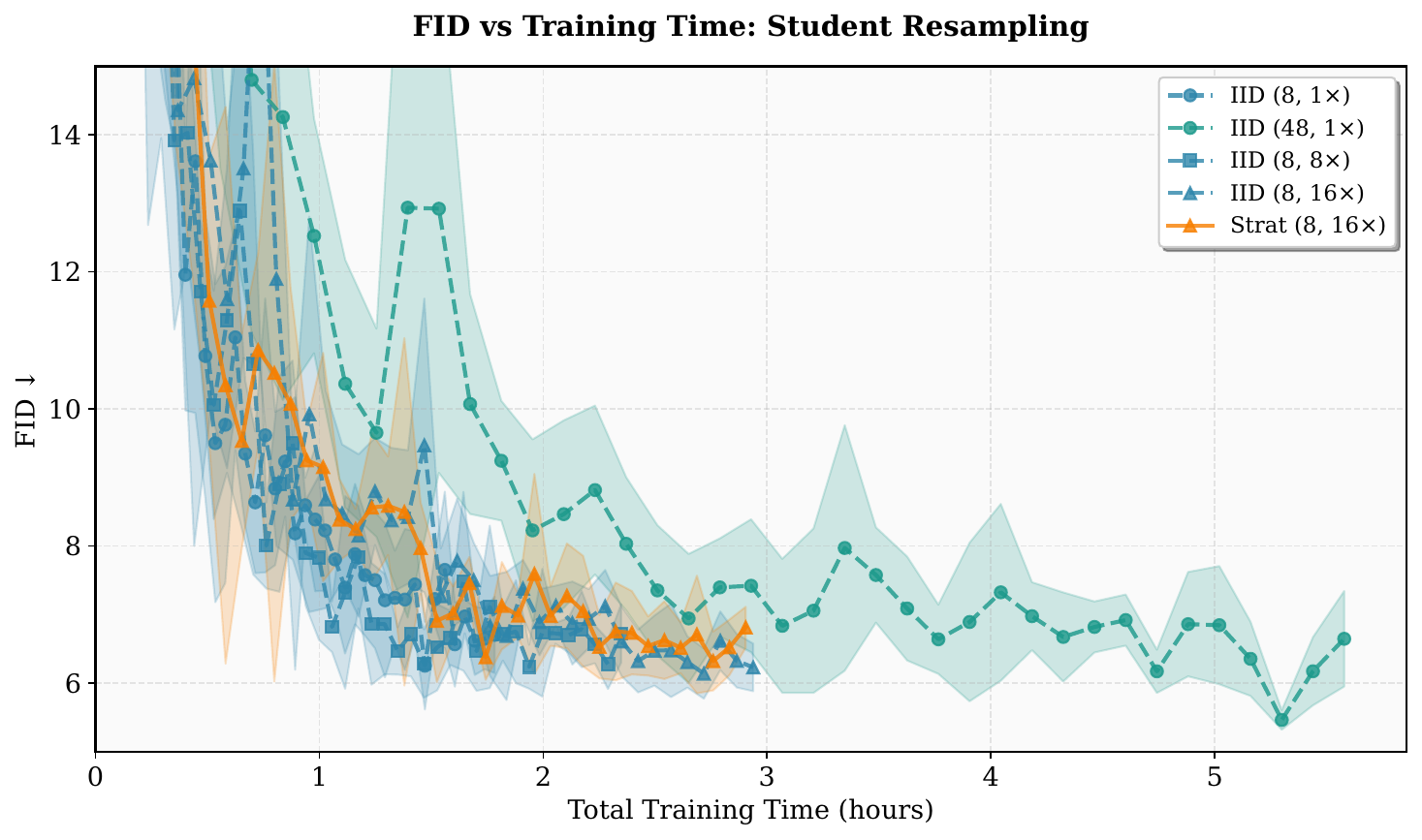}};
                    \node[below=of img22, node distance=0cm, xshift=0.0cm, yshift=1.15cm,  font=\color{black}]{\normalsize{Total Training Time (hours)}};
                \end{tikzpicture}
                }
                \caption{
                    \textbf{FID convergence during DMD training for student-step resampling.}
                    \emph{Top:} best FID vs.\ steps (left) and wall-clock (right). \emph{Bottom:} per-iteration FID vs.\ steps (left) and wall-clock (right). Shaded $\pm1\sigma$ over $5$ seeds.
                    Batch $48$ reaches the lowest final FID but is the slowest in wall-clock time at $1.0$\,s/iter. Resampling $(8,8)$/$(8,16)$ matches baseline $(8,1)$ wall-clock convergence despite $3.4$--$16\times$ variance reduction (\Tab~\ref{tab:dmd_variance}); raw FID is noisier than best-so-far.
                }
                \label{fig:student_fid_combined}
            \end{figure}
            
            We explore student-timestep resampling at batch $8$, which gives large variance reduction at low cost. Per-iter timings: batch $48$ $\sim\!1$\,s; batch $8$ $\sim\!0.38$\,s; $8\times$ resampling $\sim\!0.47$\,s; $16\times$ $\sim\!0.59$\,s.
        
            We show FID over the course of optimization. We show the convergence rate as a function of the optimization step and also rescale the $x$-axis based on the compute time spent. Because the raw FID values at each checkpoint exhibit high variance, we compute the average over $5$ random training seeds and display a shaded band indicating one standard deviation. In \Fig~\ref{fig:student_fid_combined}, top, we plot the best FID achieved by iteration $i$ while the raw (much noisier) FID values per iteration are shown in \Fig~\ref{fig:student_fid_combined}, bottom, for this setting only.
        
            Overall, we find that optimization steps with variance reduction yield convergence rates and final values that are similar to or better than those of the baseline. However, when accounting for the additional compute cost, the baseline method remains comparable.
        
            \begin{figure}[t]
                \centering
                \scalebox{1.0}{
                \begin{tikzpicture}
                \centering
                    \node (img11){\includegraphics[trim={0.8cm 0.8cm 0cm 0.8cm}, clip, width=.46\linewidth]{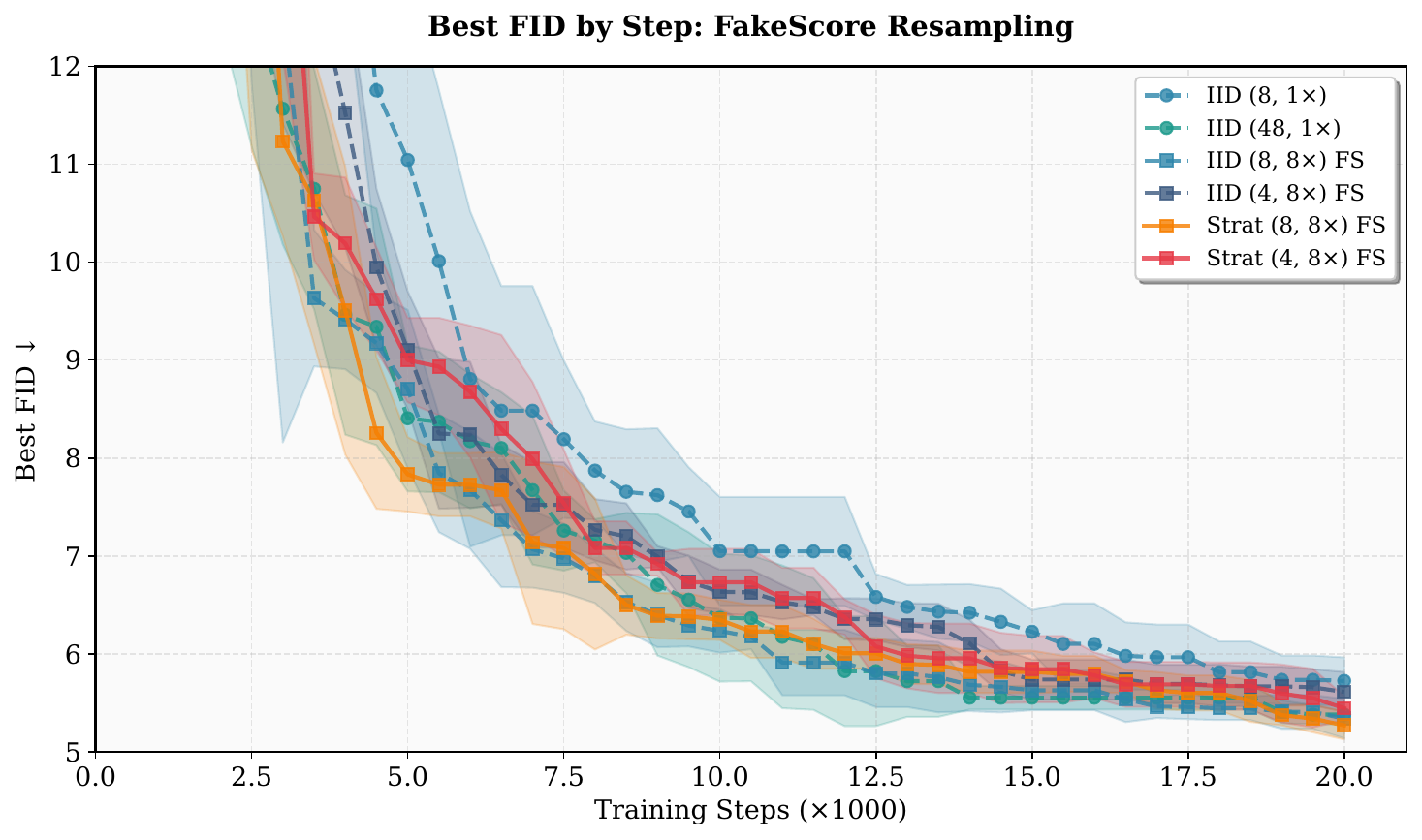}};
                    \node[left=of img11, node distance=0cm, rotate=90, xshift=1.25cm, yshift=-.8cm,  font=\color{black}]{\normalsize{Best FID $\downarrow$}};
                    \node[below=of img11, node distance=0cm, xshift=0.0cm, yshift=1.15cm,  font=\color{black}]{\normalsize{Training Steps ($\times 1000$)}};
                    
                    \node [right=of img11, node distance=0cm, xshift=-1.0cm](img21){\includegraphics[trim={1.5cm 0.8cm 0cm 0.8cm}, clip, width=.45\linewidth]{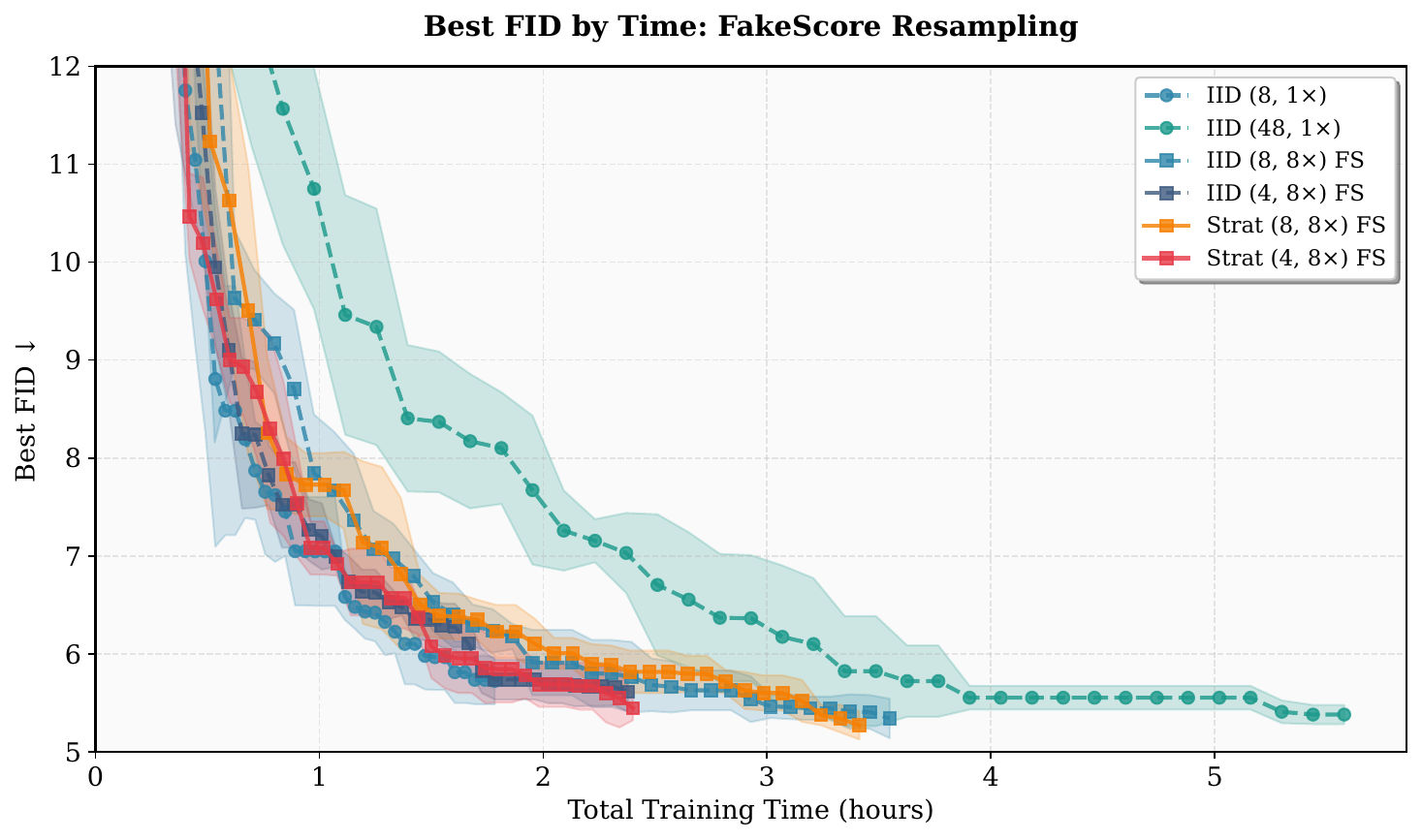}};
                    \node[below=of img21, node distance=0cm, xshift=0.0cm, yshift=1.15cm,  font=\color{black}]{\normalsize{Total Training Time (hours)}};
                \end{tikzpicture}
                }
                \caption{
                    \textbf{Best FID during training with resampling and stratification in the fake-score step.}
                    \emph{Left:} Best FID versus training iteration. \emph{Right:} Best FID versus wall-clock time. Shaded regions show $\pm 1$ standard deviation across $5$ seeds. Resampling and stratifying the fake-score update (runs marked FS) improves per-iteration convergence and reaches a final FID similar to batch size $48$ at a lower per-iteration cost, but the batch-$8$ baseline remains the most compute-efficient overall.
                }\label{fig:fake_score_best_fid}
            \end{figure}
            
            We also applied resampling and stratification to the fake-score update step (\Fig~\ref{fig:fake_score_best_fid}). This costs more compute per step but gives a larger per-step convergence gain, in line with the larger batch size; adjusted for compute, the benefit is less apparent.

        \subsubsection{Understanding the DMD Variance-Metric Disconnect}\label{sec:experiments-dmd-disconnect}
            The methods in \Tab~\ref{tab:dmd_variance} cut parameter-gradient variance by up to ${\sim}32\times$ vs.\ the $(8,1)$ baseline (with $(8,16)$ resampling + stratification) and score-difference variance by ${\sim}5.9\times$, but \Fig~\ref{fig:student_fid_combined} shows no best-FID gain vs.\ wall-clock. We triage four candidate explanations and discuss them in numerical order below: H4 is the only one ruled out; H2 is most consistent with the data; H1 is plausible but not isolated; and H3 is untested by our experiments.

            \textbf{Positive control: VR machinery acts on the teacher signal.}
                At initialization, $\meanFake$ equals the teacher, and the score-difference term is zero. In this regime, we observe ${\sim}4\times$ variance reduction in the teacher-side SDS-like signal using the same machinery, confirming the pipeline runs end-to-end on the DMD codebase (consistent with \Sec~\ref{sec:experiments-sds}). This does not validate VR on the trained score-difference gradient at intermediate iterations where $\meanFake$ has trained but is not at equilibrium; the strongest closure would be measuring score-difference VR at iter $\sim\!\num{5000}$--$\num{10000}$, which we leave to future work. So, FID flatness is a downstream insensitivity rather than a measurement artifact. Approximation error in $\meanFake$ is a plausible contributor that our unbiased-estimator analysis does not isolate.

            \textbf{Hypothesis 1 (auxiliary objectives stabilize training): plausible, not isolated.}
                At fixed $\numRenders\!=\!8$ and $1.55\times$ per-iteration wall-clock cost, score-difference variance shrinks $\sim\!5.94\times$ while parameter-gradient variance shrinks $\sim\!32\times$ (\Tab~\ref{tab:dmd_variance}, $(8,16)$ Strat vs $(8,1)$ IID). The two ratios need not agree, because \Eq~\ref{eq:dmd_mc_estimator} maps the score difference through $\partial\generator/\partial\genParams$, which can amplify the reduction. Both measurements are of the distribution-matching gradient alone: the GAN-feature discriminator's generator gradient is not included, and $\lossDenoise$ trains the fake score on stop-gradient generator outputs, so these data do not isolate the auxiliary objectives. They remain a plausible reason that the variance reduction does not reach FID; with the positive control above, the reduction itself is real.

            \textbf{Hypothesis 2 (generator-input diversity is the bottleneck): consistent with the data.}
                \Fig~\ref{fig:student_fid_combined}: batch $48$ with $\numReNoises{=}1$ ($1.0$\,s/iter, $48$ generator-input draws) reaches the lowest final FID, while batch $8$ with $\numReNoises{=}16$ ($0.59$\,s/iter, $8$ draws) does not, despite $16\times$ more $(\timevar,\noiseVec')$ per render. The comparison is consistent with H2 but not strictly compute-matched ($1.0$\,s vs $0.59$\,s/iter means the $(48,1)$ run also gets more total $\noiseVec$ draws per wall-clock); a clean $(R,K)$ sweep at fixed wall-clock is future work. The qualitative pattern matches what reverse-KL matching predicts: per-input mass-shifting moves FID, while per-input timestep variance has a smaller marginal effect.

            \textbf{Hypothesis 3 (data-dependent weighting): untested by our data.}
                The DMD weight in \Eq~\ref{eq:dmd_mc_estimator} has a data-dependent normalizer ($\|\meanBase(\noisedData,\timevar)-\encodedData\|_1$), so $\proposalDensity^\star\!\propto\!p\sqrt{\E[\|\gterm\|_2^2\mid\timevar]}$ is data-dependent and the SDS weight proxy of \Sec~\ref{sec:method-noise-schedules} need not approximate it. Our ratios use stratification with a uniform $\timevar$; whether a DMD-tuned proposal helps is left for future work.

            \textbf{Hypothesis 4 (compute scaling): ruled out.}
                From the timings in \App~\Sec~\ref{sec:experiments-dmd-app-details}, scaling $\numReNoises\!=\!1{\to}16$ raises per-iter cost $0.38$\,s$\to\!0.59$\,s ($0.21$\,s for $15$ extra denoising calls), giving $\costDenoise\!\approx\!0.014$\,s, $\costRender\!\approx\!0.37$\,s and $\costRender/\costDenoise\!\approx\!26$. DMD sits in the \emph{render-dominated} regime ($\alpha\!\approx\!26$ on \App~\Fig~\ref{fig:compute_cost_sensitivity}, between $\alpha\!=\!1$ and $\alpha\!=\!100$), where re-noising amortizes its largest gains. The $(8,16)$ Strat vs $(8,1)$ IID parameter-gradient variance reduction of ${\sim}32\times$ at $1.55\times$ wall-clock gives ${\sim}20\times$ variance reduction per unit wall-clock, ruling out H4 (``no compute headroom''); FID flatness must come from H1, H2, or H3.

            \textbf{Implications for practitioners.}
                VR in DMD is most likely to translate to FID gains when the practitioner: (1) raises generator-input diversity $\noiseVec$ jointly with timestep VR, not holding $\numRenders$ fixed; (2) lowers the GAN discriminator and denoising auxiliary weights, isolating the distribution-matching gradient as binding; (3) builds a DMD-specific proposal that handles the data-dependent normalization in $\weight(\timevar)$. Otherwise, our methods still yield stable, lower-variance gradient estimates, but should not be expected to move FID on their own.

    \FloatBarrier
    \subsection{Data Attribution}\label{sec:experiments-motive-app}

        \subsubsection{Details}\label{sec:experiments-motive-app-details}
            \textbf{Hyperparameter Choices:} 
                 We follow MOTIVE~\cite{wu2026motion} settings for \texttt{Wan2.1-T2V-1.3B}~\citep{wan2025wan}. We fix random seeds for the Gaussian noise and the TRAK projector. Per-sample gradients are projected to $512$ dimensions. Gradients are computed without classifier-free guidance. For influence scores, we use $11$ video samples from \texttt{VIDGEN-1M}~\citep{tan2024vidgen} with leave-one-out evaluation: each sample serves as query, and the remaining $10$ form the candidate set, averaging influence scores across all $11$ query-candidate assignments. Influence scores average the cosine similarities of normalized projected gradients over shared $(\timevar, \noiseVec)$ draws, as in \Eq~\ref{eq:diffusion_attrib_cosine}.

            \textbf{Compute Usage:}
                NVIDIA A100. Per-video gradients $\sim\!54$\,s (DiT forward-backward); $512$-dim TRAK $\sim\!2$\,s/sample. Influence scores $\sim\!46$\,ms/pair. Sampling-strategy runs ($\num{1000}$ MC trials, $768$ timesteps) finish in $<\!4$\,min on one GPU with cached scores.

        \subsubsection{Results}\label{sec:experiments-motive-app-results}
\begin{figure}[h!]
                \centering
                \vspace{-0.004\textheight}
                \fitlinewidth{
                \begin{tikzpicture}
                \centering
                    \node (img21){\includegraphics[trim={0.3cm 0.0cm 0.0cm 0.25cm}, clip, width=.45\linewidth]{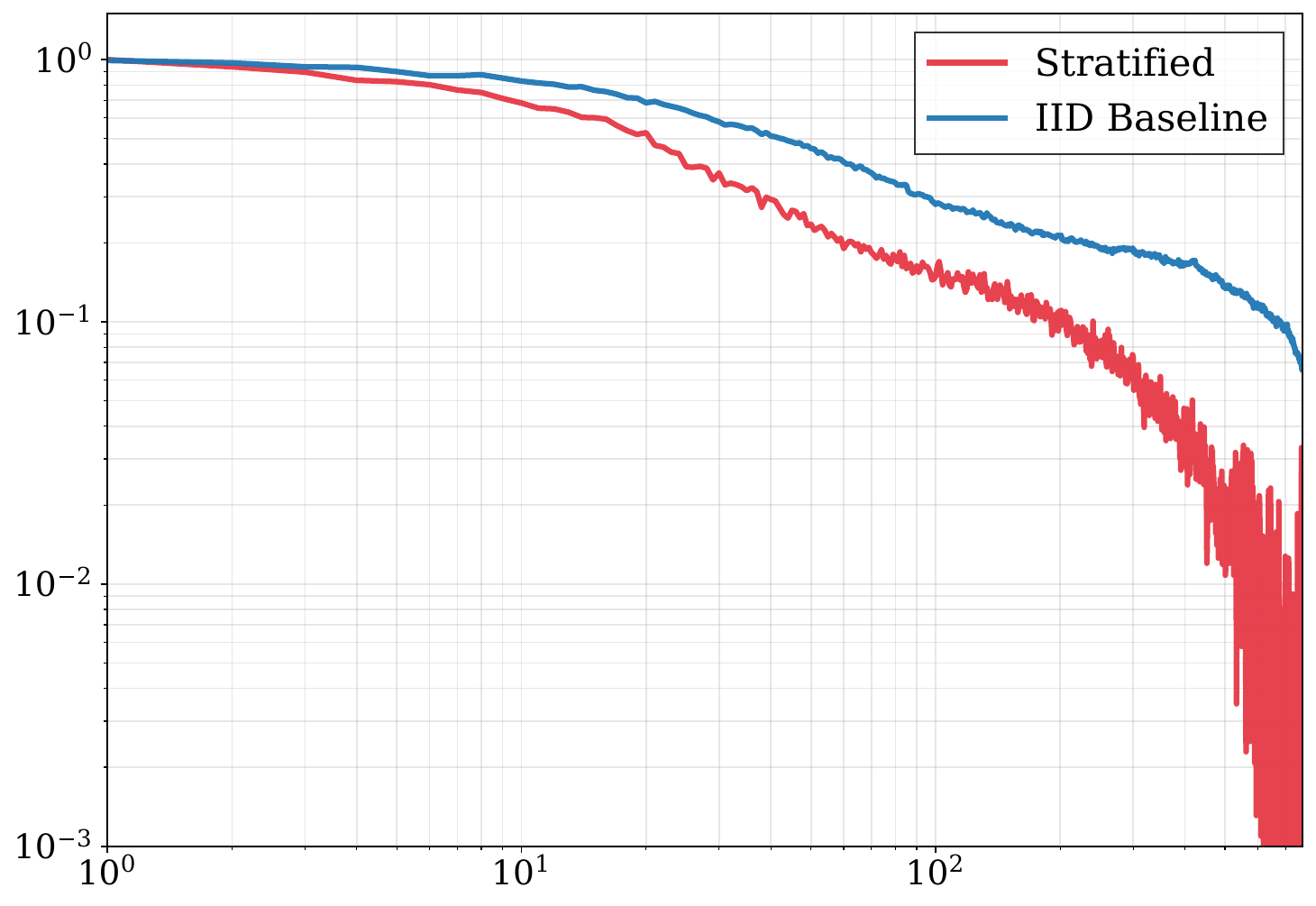}};
                    \node[left=of img21, node distance=0cm, rotate=90, xshift=1.6cm, yshift=-.9cm,  font=\color{black}]{\small{$1-{}$Mean Correlation}};
                    
                    \node [right=of img21, node distance=0cm, yshift=0.0cm, xshift=-0.2cm] (img31){\includegraphics[trim={0.0cm 0.0cm 0.0cm 0.25cm}, clip, width=.45\linewidth]{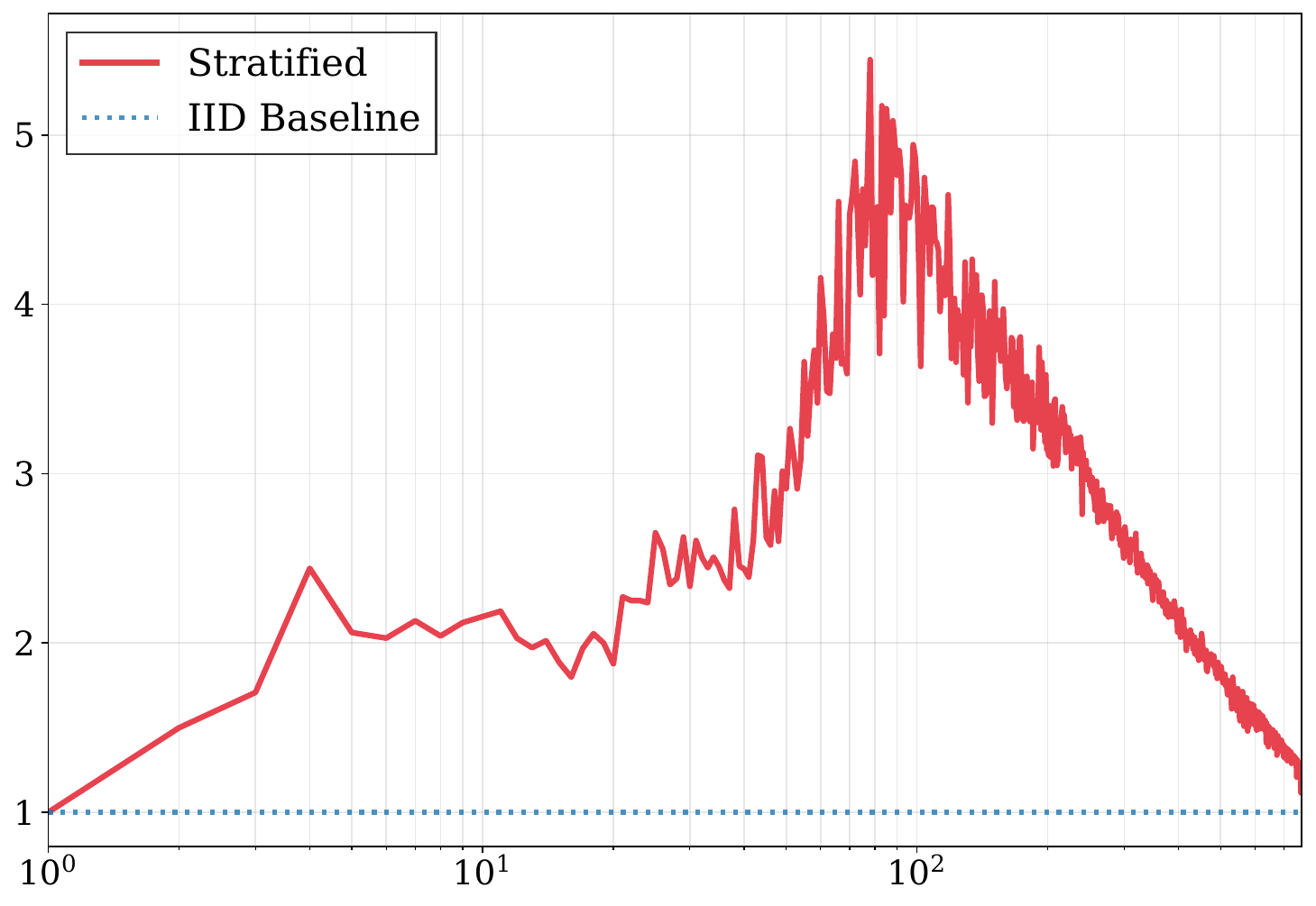}};
                    \node[left=of img31, node distance=0cm, rotate=90, xshift=1.15cm, yshift=-.8cm,  font=\color{black}]{\small{Correlation ECM}};
    
                    \node [below=of img21, node distance=0cm, yshift=1.0cm, xshift=0.0cm] (img22){\includegraphics[trim={0.0cm 0.0cm 0.0cm 0.25cm}, clip, width=.45\linewidth]{images/data_attribution/v4_gradient_convergence.pdf}};
                    \node[left=of img22, node distance=0cm, rotate=90, xshift=1.35cm, yshift=-.9cm,  font=\color{black}]{\small{Gradient Variance}};
                    \node[below=of img22, node distance=0cm, xshift=0.0cm, yshift=1.15cm,  font=\color{black}]{\normalsize{Gradient Samples per Data Point}};
    
                    \node [right=of img22, node distance=0cm, yshift=0.0cm, xshift=-0.2cm] (img32){\includegraphics[trim={0.0cm 0.0cm 0.0cm 0.25cm}, clip, width=.45\linewidth]{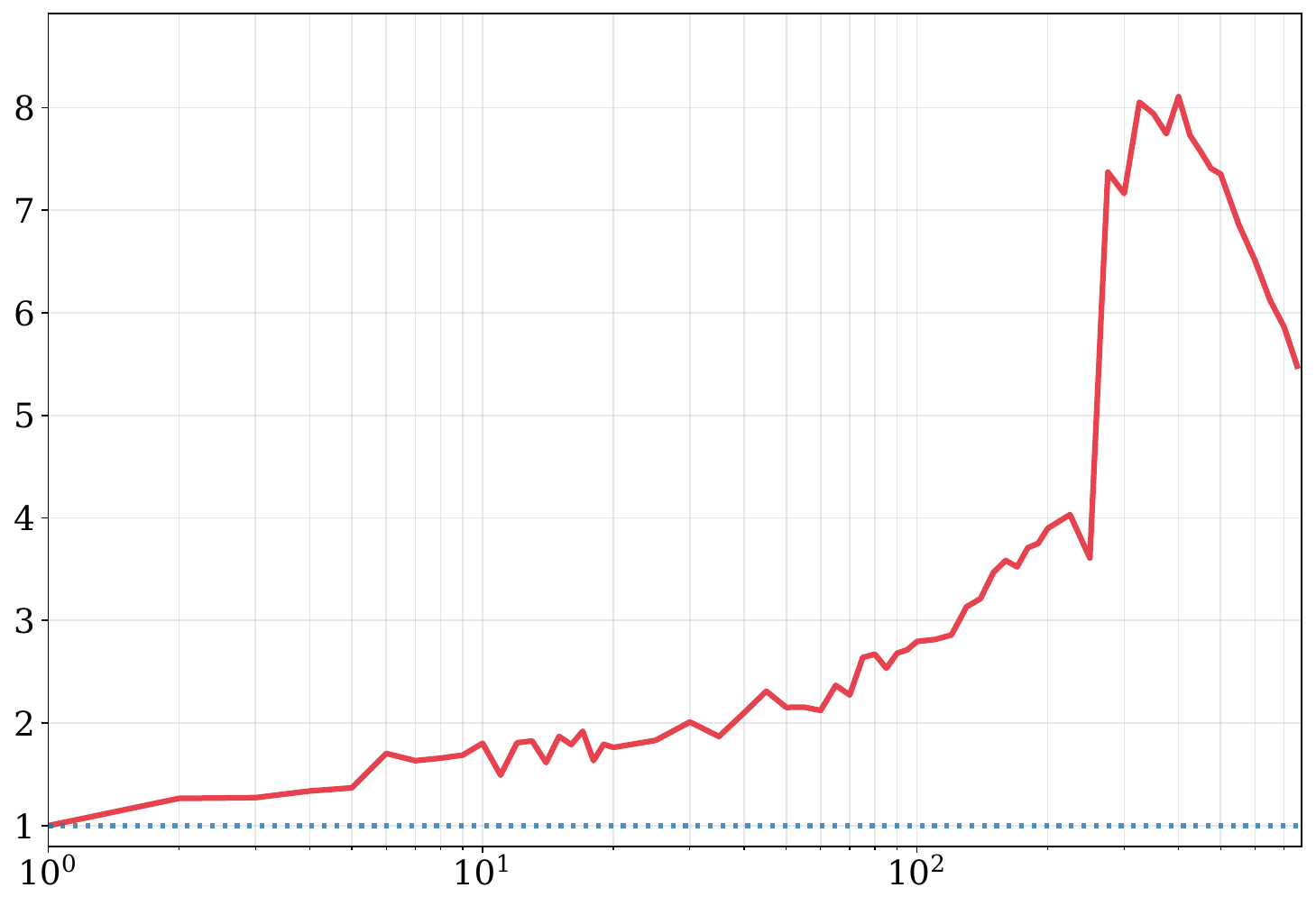}};
                    \node[left=of img32, node distance=0cm, rotate=90, xshift=1.0cm, yshift=-.8cm,  font=\color{black}]{\small{Variance ECM}};
                    \node[below=of img32, node distance=0cm, xshift=0.0cm, yshift=1.15cm,  font=\color{black}]{\normalsize{Gradient Samples per Data Point}};
                \end{tikzpicture}
                }
                \vspace{-0.02\textheight}
                \caption{
                    \textbf{(Extended) Quantifying Changes in Data Attribution:}
                    Extended version of \Fig~\ref{fig:quantifying_attribution} showing convergence (left) and effective compute multiplier (right) for both correlation and variance metrics.
                    \emph{Top row:} Correlation between rankings computed with limited evaluations and full-budget rankings. Left: $1-\text{correlation}$ versus number of evaluations, showing convergence to ground truth. Right: effective compute multiplier, quantifying how much more compute uniform sampling requires to match stratified sampling's correlation.
                    \emph{Bottom row:} Gradient variance convergence. Left: variance versus number of evaluations, showing the expected decrease with more samples. Right: variance effective compute multiplier, quantifying stratified sampling's variance reduction relative to uniform sampling.
                    Stratified sampling consistently outperforms uniform sampling, with compute multipliers above $1.5\times$ in correlation and above $1.25\times$ in variance for $2$ to $500$ samples.
                }\label{fig:quantifying_attribution_full}
                \vspace{-0.01\textheight}
            \end{figure}

            Our experiments demonstrate that stratified sampling substantially improves the correlation in influence rankings (\Tab~\ref{tab:motive_gradient_estimation_summary}). While we do not validate downstream fine-tuning performance, the metric measures agreement with the full-budget attribution, and correlation-based evaluation is standard for attribution methods \cite{park2023trak}. At $768$ timesteps, the full budget, the stratified estimate reproduces this reference (correlation $1.000$; \App~\Fig~\ref{fig:quantifying_attribution_full}, top left), so the last row of \Tab~\ref{tab:motive_gradient_estimation_summary} is not a like-for-like comparison. Direct validation via fine-tuning on variance-reduced rankings remains future work.
            
            \begin{figure}[h!]
                \centering
                \scalebox{1.0}{
                \begin{tikzpicture}
                \centering
                    \node (img11){\includegraphics[trim={0.9cm 0.9cm 0cm 0.2cm}, clip, width=.4\linewidth]{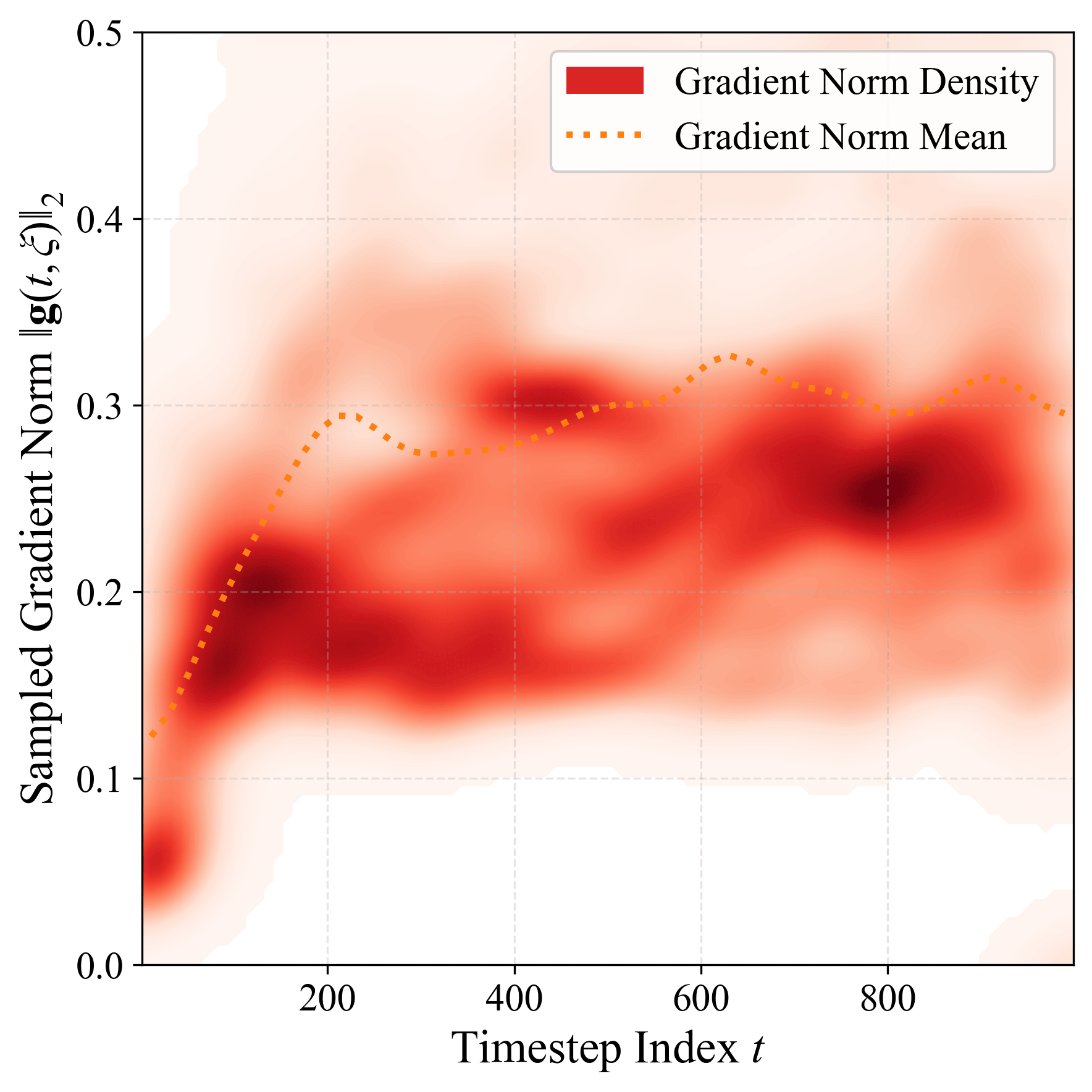}};
                    \node[left=of img11, node distance=0cm, rotate=90, xshift=2.8cm, yshift=-.8cm,  font=\color{black}]{\normalsize{Sampled Gradient Norm $\|\gterm(\timevar, \randomness)\|_2$}};
                    \node[below=of img11, node distance=0cm, xshift=0.0cm, yshift=1.15cm,  font=\color{black}]{\normalsize{Timestep Index $\timevar$}};
                \end{tikzpicture}
                }
                \caption{
                    \textbf{Is there an improvement from importance sampling for data attribution?}
                    We show the sampled gradient norm $\|\gterm(\timevar,\randomness)\|_2$, where $\gterm(\timevar,\randomness) = \nabla_{\denParams}\costDiffusion(\encoder(\dataSample),\textCond,\timevar,\noiseVec,\denParams)$ and $\randomness$ denotes all other sources of randomness, including the data and conditioning $(\dataSample, \textCond)$ and the Gaussian noise $\noiseVec$.
                    \emph{Takeaway:} By \Eq~\ref{eq:opt_proposal} we have $\proposalDensity^\star(\timevar) \propto p(\timevar)\sqrt{\E[\|\gterm(\timevar,\randomness)\|_2^2\mid\timevar]}$, and $\E[\|\gterm(\timevar,\randomness)\|_2^2\mid\timevar]$ is already roughly constant in $\timevar$, except near $0$ where the norm contribution is low, so the current sampling may not see large improvements.
                }\label{fig:optimal_importance_motive}
            \end{figure}
    
            \begin{figure}[h!]
                \centering
                \scalebox{1.0}{
                \begin{tikzpicture}
                \centering
                    \node (img11){\includegraphics[trim={0.8cm 6.2cm 0.0cm 0.25cm}, clip, width=.45\linewidth]{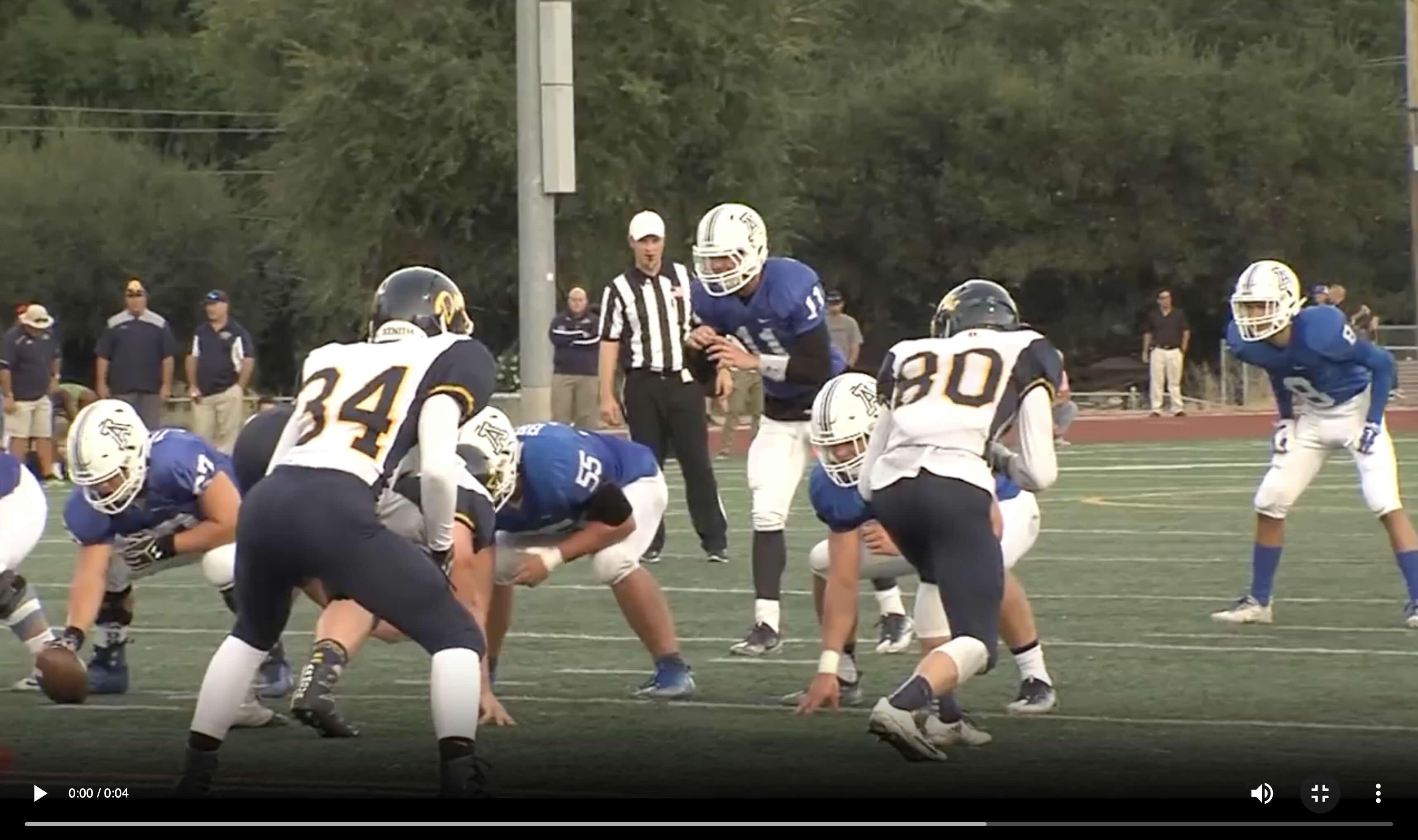}};
                    \node [right=of img11, xshift=-1cm] (img21){\includegraphics[trim={0.3cm 6.2cm 0.0cm 0.25cm}, clip, width=.45\linewidth]{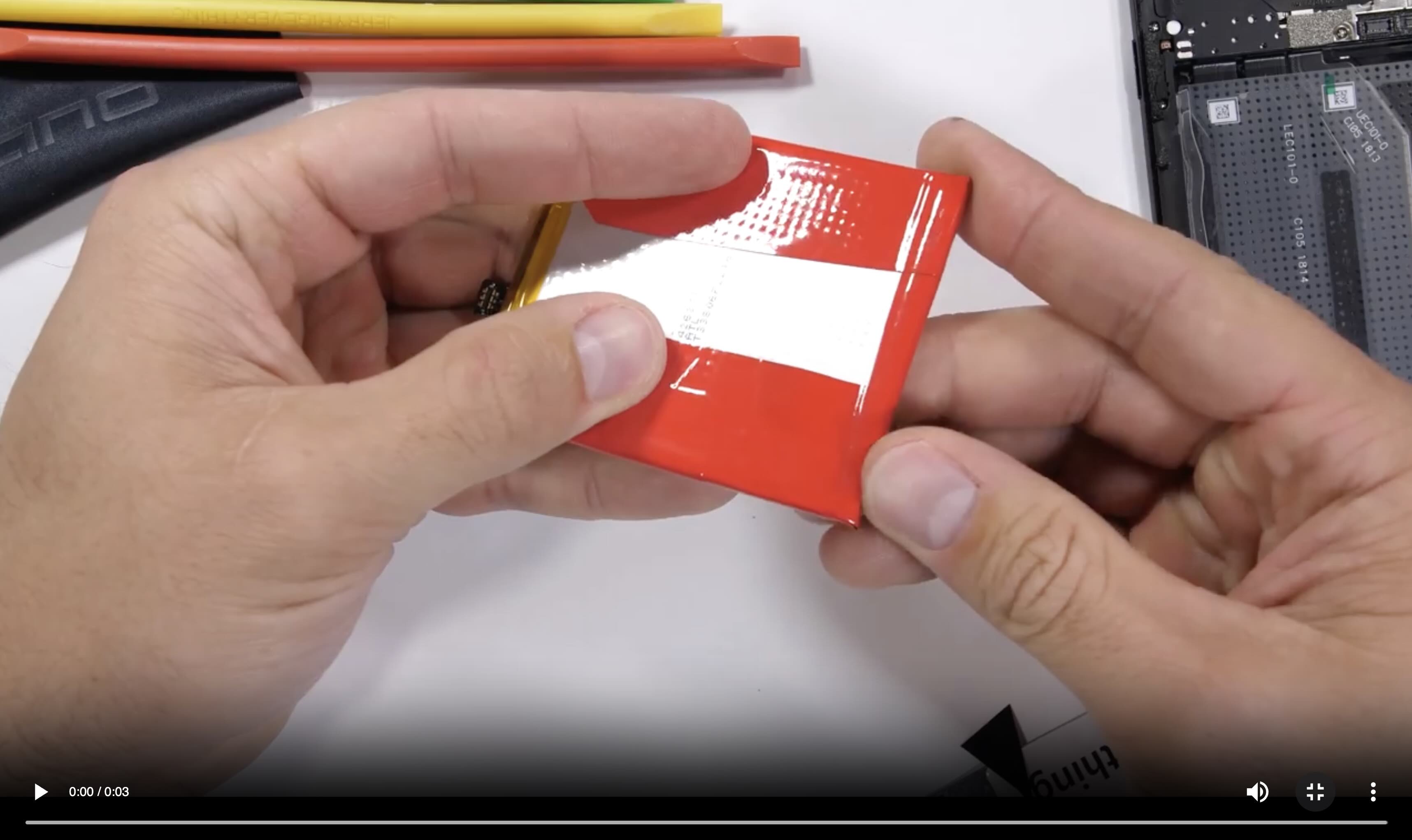}};
                    \node [below=of img11, node distance=0cm, yshift=1cm, xshift=0.0cm] (img12){\includegraphics[trim={0.0cm 6.2cm 0.0cm 0.25cm}, clip, width=.45\linewidth]{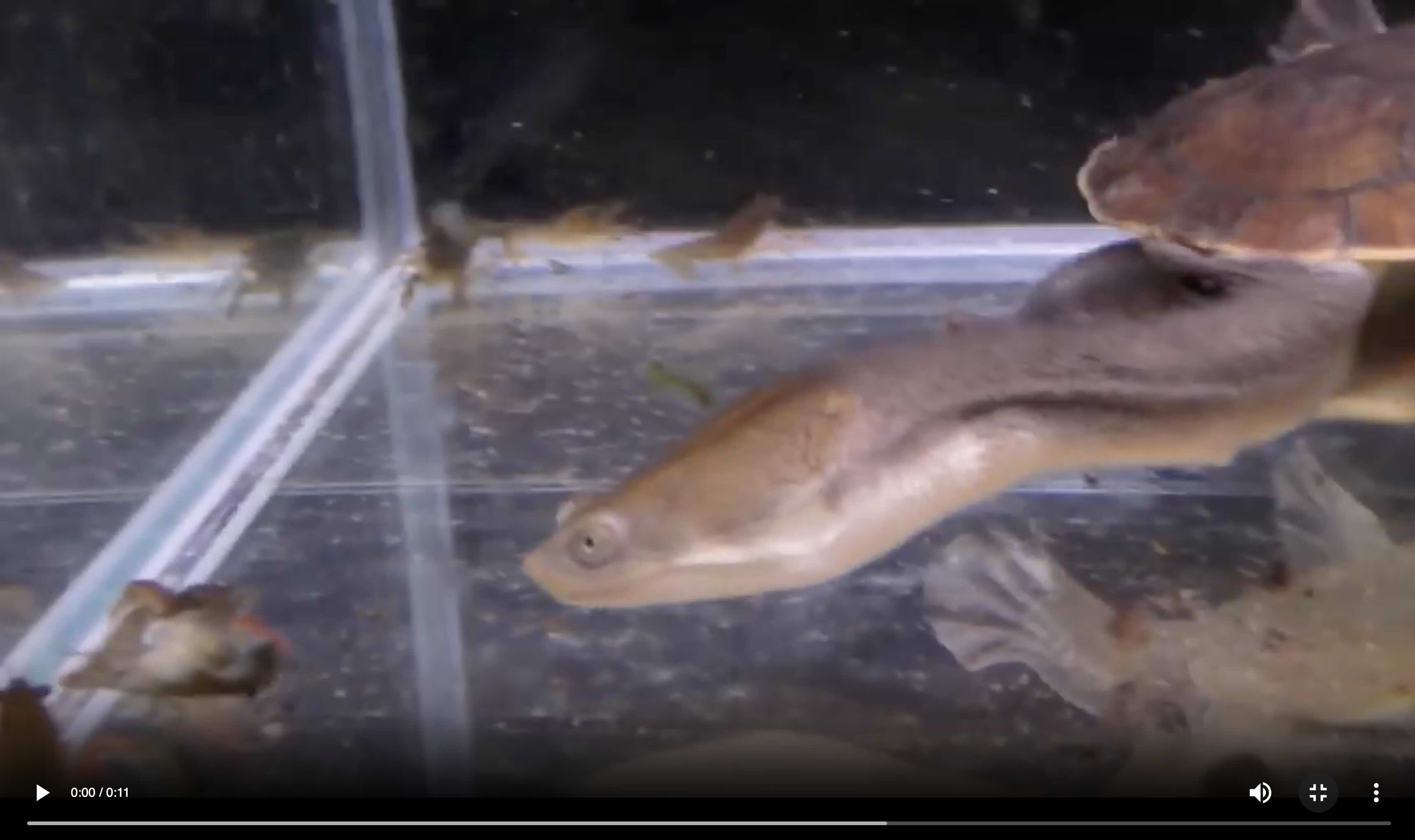}};
                    \node [right=of img12, xshift=-1cm] (img2){\includegraphics[trim={0.3cm 6.2cm 0.0cm 0.25cm}, clip, width=.45\linewidth]{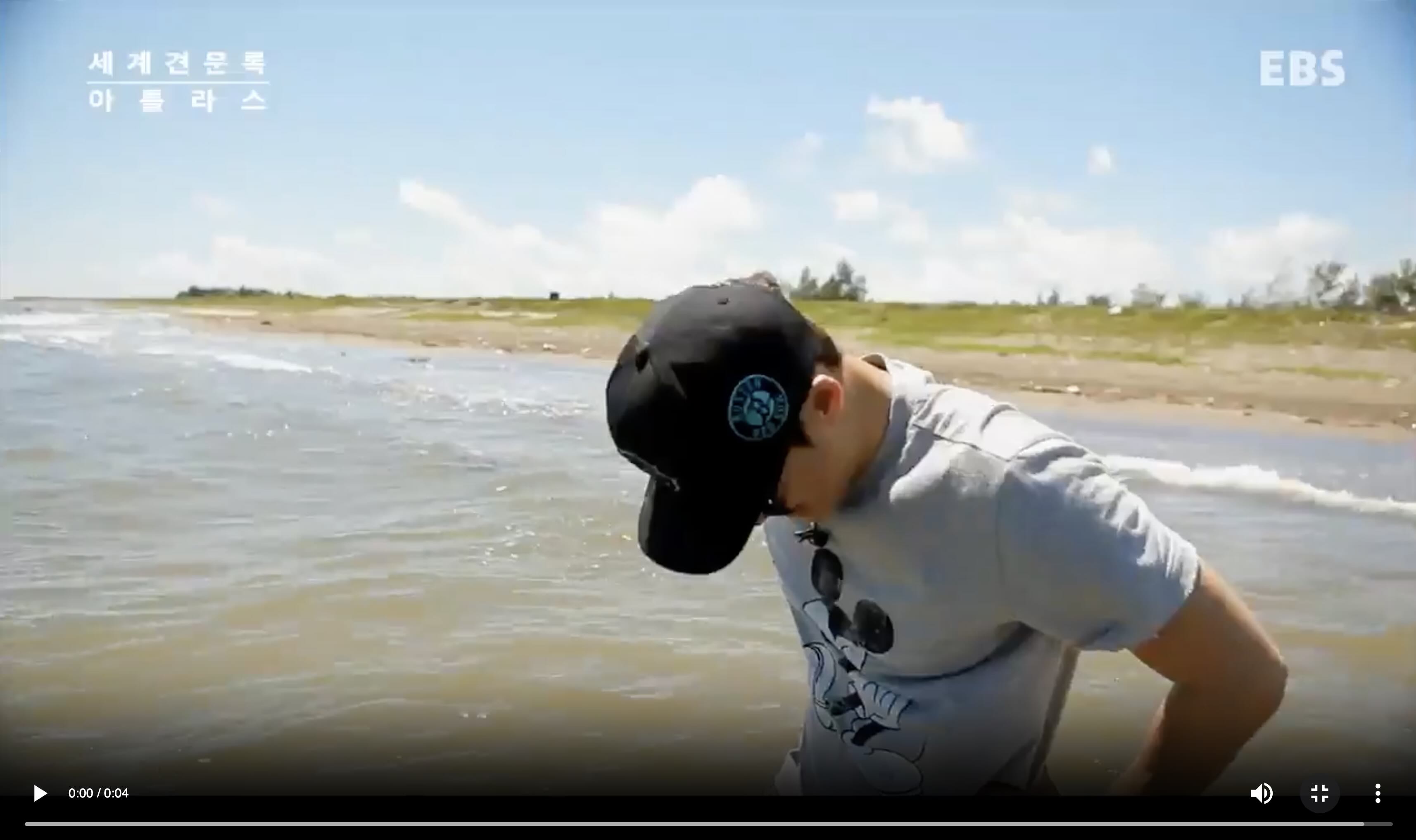}};
                \end{tikzpicture}
                }
                \vspace{-0.00\textheight}
                \caption{
                    \textbf{Example Videos for Attribution:}
                    We show assorted clips from \texttt{VIDGEN-1M}~\citep{tan2024vidgen} used for our video data attribution experiments, where the influence is being calculated for \texttt{Wan2.1-T2V-1.3B}~\citep{wan2025wan}.
                }\label{fig:motive_qualitative}
            \end{figure}

    \FloatBarrier
\section{Related Work}\label{app:related}
    We now cover related works on latent diffusion models in \Sec~\ref{sec:related-diffusion-background}, followed by related methods to reduce estimator variance in \Sec~\ref{sec:related-diffusion-variance}, and finally an exploration of variance-bounded diffusion model applications of interest in \Sec~\ref{sec:related-diffusion-applications}.

    \subsection{Latent Diffusion Models}\label{sec:related-diffusion-background}
        Diffusion models dominate generative modeling across images, video, audio, and 3D. Early work established denoising diffusion and score-based formulations \citep{sohl2015deep, ho2020denoising, song2019generative, song2020score}. Latent diffusion \citep{rombach2022high} compresses diffusion into a learned latent via an autoencoder, cutting compute while preserving quality, and underpins Stable Diffusion \citep{rombach2022high}, Stable Audio Open \citep{evans2025stable}, and video generators such as Sora \citep{videoworldsimulators2024}, CogVideoX \citep{yang2024cogvideox}, and Wan \citep{wan2025wan}. Diffusion transformers (DiT) \citep{peebles2023scalable} and related architectures scaled these models with transformer backbones. We treat pretrained teachers as given and target gradient-estimator variance for downstream optimization and distillation.

    \subsection{Reducing Estimator Variance in Diffusion Models}\label{sec:related-diffusion-variance}
        Monte Carlo variance reduction is a standard topic in simulation and optimization. Standard techniques include importance sampling, stratified sampling, control variates, and antithetic variates \citep{rubinstein2016simulation, owen2013monte}. In diffusion model training, variance reduction primarily focuses on noise schedule design, loss reweighting, and improvements to training dynamics. Early noise schedules used simple linear or cosine interpolations \citep{ho2020denoising, nichol2021improved}; later work parametrizes diffusion training in terms of signal-to-noise ratio and learns noise schedules that minimize estimator variance \citep{kingma2023variational} or shifts them with image resolution \citep{hoogeboom2023simple}. Loss reweighting adjusts per-timestep contributions to balance gradient magnitudes and prioritize perceptually important noise levels \citep{salimans2022progressive, karras2022elucidating, choi2022perception, hang2023minsnr}. Complementary training-side techniques randomize the supervised time interval to avoid a fixed low-noise truncation \citep{kim2022soft} and stabilize training dynamics through normalization and architecture choices \citep{karras2024analyzing}.
        
        Downstream users of frozen diffusion teachers typically inherit the teacher's training schedule without revisiting the timestep distribution. Recent work has begun exploring variance reduction here: Variational Score Distillation \citep{wang2023prolificdreamer} introduces a particle-based variational objective that, as later analyses note~\citep[e.g.,][]{wang2025steindreamer}, lowers SDS gradient noise at extra memory and compute. Concurrent few-step distillation work, such as moment matching \citep{salimans2024multistep}, does not measure estimator variance or use unbiased IS. We give a principled compute-aware framework and show that simple unbiased techniques (IS, stratification, reuse) deliver large variance reduction across diffusion-teacher tasks while preserving the objective.

    \subsection{Diffusion Model Applications}\label{sec:related-diffusion-applications}
        We now cover diffusion tasks related to our setup, including optimizing parametrized models with diffusion priors (\Sec~\ref{sec:related-diffusion-applications-sds}), single-step diffusion model distillation (\Sec~\ref{sec:related-diffusion-applications-single-step}), data attribution (\Sec~\ref{sec:related-diffusion-applications-data-attribution}), and beyond (\Sec~\ref{sec:related-diffusion-applications-other}).

        \subsubsection{Diffusion Priors for Optimization}\label{sec:related-diffusion-applications-sds}
            Diffusion models are widely used as frozen teachers that provide gradients for optimizing parametrized generators, beginning with diffusion-teacher-guided text-to-3D optimization and its refinements for fidelity, stability, and efficiency \citep{poole2022dreamfusion, lin2023magic3d, chen2023fantasia3d, zhu2023hifa, wang2023score, wang2023prolificdreamer, ma2024scaledreamer, lukoianov2024score, wang2025steindreamer, yu2024csd, mcallister2024rethinking, shi2024mvdream}. This teacher-gradient template has since been extended to amortized text-to-3D \citep{lorraine2023att3d, xie2024latte3d}, dynamic 4D scenes \citep{ling2024align, ren2023dreamgaussian4d, bahmani20244d}, physics and simulation parameters \citep{liu2024physics3d, zhang2024physdreamer}, and materials and textures \citep{deng2024flashtex}, alongside other routes to 3D content such as language-model mesh generation \citep{wang2024llamamesh} and multimodal generative 3D \citep{cheng20253d}. It also appears outside vision, for example in diffusion-guided audio optimization \citep{richter2025audiosds}. We keep these SDS-style objectives fixed and reduce the variance of the Monte Carlo gradients they share, using unbiased, compute-aware sampling.

        \subsubsection{Single-Step Distillation}\label{sec:related-diffusion-applications-single-step}
            Single-step and few-step distillation compress a pretrained diffusion teacher into a fast student sampler by training the student to match the teacher's induced sample distribution or score field. The training objective is an expectation over diffusion timesteps and noise, so optimization relies on Monte Carlo gradient estimators whose variance and cost depend on the timestep and noise-sampling designs. A prominent family matches student and teacher distributions via score-difference objectives, including distribution-matching distillation and follow-ups \citep{yin2024one, yin2024improved, luo2024diff, zhou2024score, nguyen2024swiftbrush, zhou2024long}; related methods match moments \citep{salimans2024multistep} or use EM-based maximum likelihood \citep{xie2024distillation}. A complementary adversarial branch combines distillation with a discriminator on student outputs \citep{sauer2024add}. Variants include multi-student distillation \citep{song2024multi} and flow-alignment formulations that bridge diffusion and flow-based distillation views \citep{sabour2025align}. Consistency models \citep{song2023consistency, song2024improved} provide an alternative distillation path by enforcing self-consistency along ODE trajectories. Our variance-reduction methods apply naturally to these score-difference and consistency objectives, reducing the cost of gradient estimation without modifying the distillation loss.

        \subsubsection{Data Attribution}\label{sec:related-diffusion-applications-data-attribution}
            Gradient-based data attribution estimates how training examples affect model outputs via influence functions \cite{koh2017understanding} or approximations such as TracIn~\citep{pruthi2020estimating}, TRAK~\citep{park2023trak}, and approximate-unrolled-differentiation methods \citep{bae2024training}. Recent work adapts these to diffusion and LoRA-tuned models, proposing attribution methods for image and video generation \citep{georgiev2023journey, zhengintriguing, brokman2024montrage, lin2024diffusion, mlodozeniecinfluence, kwon2024datainf}. MOTIVE~\citep{wu2026motion} specializes in attribution for motion in video diffusion models. These estimators rely on Monte Carlo averages over diffusion timesteps and noise, so reducing variance improves attribution quality at fixed compute. Our variance-reduction techniques apply naturally and yield more stable influence~rankings.

        \subsubsection{Other Diffusion-Guided Tasks}\label{sec:related-diffusion-applications-other}
            Beyond the three applications we evaluate, diffusion-teacher gradients appear in many other settings where variance reduction could provide similar benefits. Diffusion guidance has been applied to image editing and inpainting \citep{meng2021sdedit, lugmayr2022repaint}, controllable generation via spatial or semantic constraints \citep{zhang2023adding, mou2024t2i}, and inverse problems such as super-resolution and deblurring \citep{song2021solving, kawar2022denoising}. In robotics and reinforcement learning, diffusion models serve as planners and policies \citep{janner2022planning, chi2023diffusion}. Our variance-accounting framework (\Sec~\ref{sec:method-variance-framework}) and drop-in variance-reduction techniques (\Sec~\ref{sec:method-variance-reduction-strategies}) extend naturally to these tasks whenever the computational bottleneck lies in Monte Carlo estimation over timesteps and~noise.

    \subsection{Comparison to Variational Score Distillation (VSD)}\label{app:comparison-vsd}
        VSD \cite{wang2023prolificdreamer} keeps the frozen teacher and replaces SDS's injected-noise term with a LoRA-adapted score of the rendered images, turning SDS into a variational objective that jointly optimizes the scene and that score; the learned score gives more consistent gradients than the injected noise.
        
        \textbf{Relationship to our work.}
            Our re-noising, importance-sampling, and stratification construction can be applied to the VSD gradient with an appropriate timestep proposal, though we have not tested this combination. A head-to-head comparison of SDS with our estimator against VSD would confound estimator changes with VSD's learned-score objective and the additional joint-tuning budget \citep{maclaurin2015gradient, pedregosa2016hyperparameter, lorraine2024scalable}.
        
        \textbf{Other related methods:}
            Beyond VSD, many techniques improve SDS-based text-to-3D generation through domain-specific modifications: other 3D representations such as Instant-NGP hash grids \cite{muller2022instant}, amortization across prompts \citep{lorraine2023att3d,xie2024latte3d}, multimodal 3D generation \citep{cheng20253d}, trajectory reparametrizations \citep{lukoianov2024score}, regularization terms, camera pose scheduling, and guidance-scale annealing. Language-model mesh generation \citep{wang2024llamamesh} is a separate route to 3D that does not use SDS. These approaches are orthogonal to ours in two ways. First, they are specific to 3D optimization and do not generalize to other diffusion-teacher applications (distillation, data attribution), whereas our methods apply to any Monte Carlo gradient from a frozen teacher. Second, many approaches introduce bias or alter the target objective (e.g., via regularization or modified guidance), whereas we focus exclusively on changes to unbiased estimators that preserve the original objective. These 3D-specific techniques could be combined with our variance-reduction strategies to further improve text-to-3D~tasks.

    \subsection{Comparison to SteinDreamer}\label{app:comparison-stein}
        SteinDreamer~\citep{wang2025steindreamer} reduces SDS variance via a Stein-identity control variate, instantiated with a frozen MiDaS depth (or normal) estimator and a learnable scaling $\boldsymbol{\mu}$ for the Stein term. The control variate is zero-mean in expectation and operates on the noise-direction estimator at fixed $\timevar$; our methods operate at the timestep-sampling and compute-amortization levels. The two attack different randomness axes of the same estimator: SteinDreamer targets the noise-direction component, while \ourMethod{} targets the joint $(\timevar,\noiseVec)$ allocation and the rendering-vs-denoising compute split.

        \textbf{Relationship to our work.}
            Our hierarchical Monte Carlo construction (\Algo~\ref{alg:combined}) can be applied to SteinDreamer's control-variate-corrected estimator, though the combined gains remain untested. A head-to-head substitution would confound noise-direction control variates with changes to timestep allocation and compute reuse, whereas a stacked ablation would isolate their contributions. SteinDreamer's implementation has not been released at the time of writing (\url{https://github.com/Vita-Group/SteinDreamer} contains a README, a teaser, and project-page assets).
        
        \textbf{Why CLIP rather than FID for our setting.}
            FID requires a reference distribution. Single-prompt text-to-3D produces one scene per prompt; rendered views of that scene are not drawn from a meaningful distribution, so FID measures distance between an ad hoc reference set and a single object's renders rather than fidelity-and-diversity. We therefore follow the standard SDS literature in using CLIP score for prompt alignment and qualitative visualizations for fidelity (\Fig~\ref{fig:qualitative_sds}, \Fig~\ref{fig:qualitative_sds_trajectory}); SteinDreamer reports CLIP distance for their convergence study while also reporting FID in their Tab.~1, consistent with this choice for convergence-curve comparisons.

        \textbf{Independent/concurrent variance-reduction directions for SDS.}
            Beyond control variates and our timestep-allocation approach, several recent directions change the SDS objective or its gradient: particle-based learned-score reformulations \citep{wang2023prolificdreamer}, multi-step trajectory reparametrizations \citep{lukoianov2024score}, and mean-shift formulations of the distillation gradient \citep{thamizharasan2026meanshift}. Others change the training setup: multi-student distillation \citep{song2024multi} and amortization across prompts \citep{lorraine2023att3d, xie2024latte3d}. Our work is the only direction we are aware of that preserves the SDS objective \emph{and} provides compute-aware, estimator-level accounting. Amortized text-to-3D \citep{lorraine2023att3d, xie2024latte3d} also trains with SDS-style losses, so it can use the same~estimator.

    \subsection{Novelty vs. Prior Work}\label{app:novelty-discussion}
        Importance sampling, stratified sampling, and compute reuse are standard Monte Carlo variance-reduction techniques; adapting them to frozen diffusion teachers requires care:
        
        \textbf{What is standard:}
            The frameworks for importance sampling (\Sec~\ref{sec:background-reducing-variance-importance_sampling}) and stratified sampling (\Sec~\ref{sec:background-reducing-variance-stratified}) are classical, as are the two-level variance decomposition (\Eq~\ref{eq:two-level}), the cost-optimal allocation between levels, the composition of stratification with subsampling, and stratified inverse-transform sampling \citep{cochran1977sampling, thompson2012sampling, owen2013monte}. Reusing expensive computation while resampling cheap randomness is a general Monte Carlo principle.
        
        \textbf{What is novel:}
        \begin{enumerate}[leftmargin=*]
            \item \textbf{The identification:} frozen-teacher gradient estimation is an instance of the classical two-level sampling problem with asymmetric per-level costs, with renders as primary sampling units and diffusion draws as subunits (\Sec~\ref{sec:method-variance-reduction-strategies}). The items below operationalize it.
            
            \item \textbf{Weight-based importance sampling proxy:} We identify that the explicit weight functions $\weight(\timevar)$ already present in diffusion objectives (e.g., $\sdsWeight(\timevar)$ in SDS) can serve as effective proxies for the variance-optimal proposal without requiring gradient-norm estimation. Prior frozen-teacher work in the cited SDS, DMD, and attribution lines typically uses uniform or loss-based timestep sampling; $\sdsWeight(\timevar)$ itself, motivated by its empirical tracking of parameter gradient norms (\Fig~\ref{fig:toyIWVisualization}), is the proxy we add.
            
            \item \textbf{Stratified inverse-CDF timestep allocation (an adaptation):} Stratification and inverse-transform sampling are each standard, and the equal-mass quantile construction is the standard way to combine them. What we add is its application to frozen-teacher timestep allocation, composed with per-render stratification; we are not aware of prior frozen-teacher pipelines applying it to timestep allocation.
            
            \item \textbf{Compute-aware variance accounting:} Our framework (\Sec~\ref{sec:method-variance-framework}) separates rendering/encoding from denoising costs and measures variance per unit compute via effective compute multipliers (ECM) and relative efficiency (RE). Prior work in this setting counts gradient evaluations or wall-clock time without decomposing costs, masking the asymmetric-cost regime in which reuse pays off.
            
            \item \textbf{Systematic empirical measurement of parameter-gradient variance:} Prior SDS work typically reports latent-space update variance or scalar losses. We measure \emph{parameter-gradient estimator variance} (\Eq~\ref{eq:two-level}) across three applications, and use the resulting measurements to expose a regime (DMD, \Sec~\ref{sec:experiments-dmd}) where variance reduction does not improve downstream metrics.
            
            \item \textbf{Timestep stratification in this context:} To our knowledge, no published work in the cited frozen-teacher SDS, DMD, or attribution lines applies stratified sampling over diffusion timesteps. Timestep stratification has been used in diffusion model \emph{training} (e.g., the low-discrepancy timestep sampler of VDM \citep[App.~I.1]{kingma2023variational}), but not in the downstream frozen-teacher pipelines we evaluate, where the computational hierarchy and gradient structure differ.
        \end{enumerate}
        
        \textbf{Other estimator-level baselines.}
            Adaptive importance sampling over $\timevar$ is bounded by the gradient-norm oracle of \App~\Sec~\ref{sec:iw_ablation}, and \App~\Sec~\ref{app:optimal_pair_distributions} computes a Sinkhorn-optimal two-sample design; IW$+$Strat reaches ${\sim}91\%$ of its efficiency. Both serve as reference points rather than competing heuristics. Antithetic sampling and control variates act on the innermost $\noiseVec$ level, conditional on render and timestep, whereas \Eq~\ref{eq:two-level} covers the two levels that carry the cost asymmetry; applying the same accounting to a third level is a natural extension.
        
        \textbf{Related variance reduction in diffusion and graphics.}
            The most closely related prior work is on noise-schedule design for diffusion model training. Variational diffusion training \cite{kingma2023variational} parametrizes diffusion training in terms of SNR and learns a noise schedule that minimizes the variance of the training-objective estimator, and Min-SNR weighting \citep{hang2023minsnr} reweights training losses by signal-to-noise ratio to balance contributions across $\timevar$. Both, however, target training the teacher itself, where the gradient is with respect to denoiser parameters $\denParams$; we focus on downstream use of frozen teachers, where gradients flow through generators, encoders, or renderers, and the computational hierarchy is fundamentally different. Methods like Variational Score Distillation \cite{wang2023prolificdreamer} reduce SDS variance by changing the objective (replacing the injected noise with a learned score while keeping the frozen teacher); we preserve the objective and work at the estimator level. Compute reuse has a long pedigree in graphics: spatiotemporal resampled importance sampling (ReSTIR) \citep{bitterli2020spatiotemporal} and its formal generalized basis (GRIS) \citep{lin2022generalized} amortize expensive scene queries across reused samples, the same principle our hierarchical estimator exploits in the diffusion-teacher setting.

    \subsection{Adjacent Tools and Methods}\label{app:adjacent-tools}
        Several adjacent research threads outside the immediate diffusion-teacher setting provide tools and theory that complement our framework.

        \textbf{Bilevel and nested optimization.}
            DMD's learned-score formulation and VSD (\App~\Sec~\ref{sec:experiments-dmd-disconnect}, \App~\Sec~\ref{app:comparison-vsd}) are bilevel (inner: auxiliary score; outer: generator/scene). Gradient-based bilevel optimization \citep{maclaurin2015gradient, pedregosa2016hyperparameter, liu2019darts, lorraine2024scalable} provides implicit differentiation, structured best responses, and scalable nested optimization, useful when gradient-variance reduction is muted by auxiliary stabilizers. Hyperparameter optimization that uses checkpointed model weights \citep{mehta2024improving} could speed up tuning of the extra hyperparameters (LoRA rank, $\beta$, and the $\proposalDensity$ shape). The optimization-in-games view \citep{lorraine2022complex, lorraine2022lyapunov} captures DMD's alternating-update dynamics.
        
        \textbf{Structured Jacobians and architecture-aware tooling.}
            Our analysis of the SDS gradient $\sdsWeight\,\jacobian_{\genParams}^{\!\top}\!\residual$ depends on the structure of $\jacobian_{\genParams}$ through encoder-renderer chains. Structured-Jacobian networks \citep{lorraine2019jacnet, richterpowell2021input} provide tools for analyzing or learning such Jacobians directly, and AutoML task selection \citep{lorraine2022task}, which picks representative tasks for evaluating system changes, could choose the prompts and modalities on which to test estimator settings.

        \textbf{Distillation and 3D-generation pipelines that benefit from estimator-level VR.}
            Multi-student distillation \citep{song2024multi} trains several specialized one-step students with distillation losses that are Monte Carlo estimates, so the same estimators apply. In 3D generation, amortized text-to-3D \citep{lorraine2023att3d, xie2024latte3d} and multimodal 3D generation \citep{cheng20253d} train with SDS-style losses, where our IW $+$ Strat $+$ Reuse estimator applies unchanged; language-model mesh generators \citep{wang2024llamamesh} take a route that does not use SDS. Score distillation for audio \citep{richter2025audiosds} is a direct application in a modality we do not evaluate, and our video attribution experiments build on MOTIVE \citep{wu2026motion}.

\section{Additional Discussion}\label{app:sec_discussion}

    \subsection{Limitations}\label{sec:limitations}
        \textbf{Variance reduction does not always translate to downstream gains.}
            In DMD (\Sec~\ref{sec:experiments-dmd}, \App~\Sec~\ref{sec:experiments-dmd-app}), resampling reduces gradient variance by $3.4$--$16\times$ without improving FID at matched wall-clock time. Generator-input diversity is most consistent with the data, while auxiliary-objective effects are plausible but not isolated (\App~\Sec~\ref{sec:experiments-dmd-disconnect}). We treat this as a deliberate negative result that maps the boundary of applicability rather than a failure of the method.
        
        \textbf{IS proxy depends on gradient structure.}
            Our weight-based IS proposal $\proposalDensity(\timevar)\propto p(\timevar)\sdsWeight(\timevar)$ assumes the explicit weight is correlated with the per-timestep gradient norm. For tasks where gradient contributions are roughly uniform across timesteps (e.g., data attribution, \App~\Fig~\ref{fig:optimal_importance_motive}), IS provides minimal benefit, and stratification alone suffices. When $\weight(\timevar)$ is unavailable or miscalibrated, adaptive binned proposals can substitute but require periodic recomputation. We did not test SDS variants that change the form of the per-timestep gradient, such as VSD \citep{wang2023prolificdreamer}.

        \textbf{Compute reuse depends on cost structure.}
            Re-noising (\Sec~\ref{sec:method-compute-reuse}) is most effective when upstream costs (rendering, encoding, generator forwards) exceed denoising and within-render variance is timestep/noise-driven. When input variability dominates or $\costRender/\costDenoise$ is small, marginal gains shrink (\App~\Fig~\ref{fig:compute_cost_sensitivity}).

        \textbf{Variance-measurement overhead.}
            The framework (\Sec~\ref{sec:method-variance-framework}) runs each estimator to convergence, an upfront cost for method comparison. Its cost grows with the number of configurations compared and the precision required, so practitioners optimizing wall-clock time may prefer to validate on downstream metrics directly and use our framework as design guidance.

        \textbf{Frozen-teacher assumption.}
            We assume the teacher is frozen downstream. Settings where the teacher is fine-tuned or co-adapted (e.g., joint distillation) would require accounting for teacher-parameter drift and its interaction with timestep-sampling strategies.

        \textbf{SDS evaluation stack vintage.}
            Our SDS experiments use Stable Diffusion 2.1 as the teacher \citep{rombach2022high}, NeRF / Instant-NGP \citep{muller2022instant} as the 3D representation, and the threestudio \citep{threestudio2023} framework. These were SOTA at the time of the experimental sweep, but more recent stacks (FLUX or SDXL teachers, MVDream-style multi-view conditioning \citep{shi2024mvdream}, 3D Gaussian Splatting renderers) would change the absolute compute budget and the precise $\costRender/\costDenoise$ ratio. Our framework is teacher- and renderer-agnostic by construction (\Sec~\ref{sec:method-variance-framework}), so the qualitative wins should transfer; precise quantitative ECMs in those stacks would require re-running the variance sweep with the new~pipeline.

    \subsection{Future Directions}\label{sec:future_directions}
        Several promising directions extend our framework beyond the tasks and methods explored here.
    
        \textbf{Extension to diverse diffusion-guided tasks.}
            Our evaluation covers text-to-3D optimization, one-step distillation, and data attribution, but the teacher-gradient pattern appears across many other settings. Natural extensions include 4D scene optimization, physics-informed diffusion guidance, material and texture synthesis, audio generation, and video editing pipelines. Our framework provides a systematic way to quantify variance-cost trade-offs in these domains, so practitioners can identify efficient sampling strategies without exhaustive tuning.
    
        \textbf{Alternative prediction parameterizations and teacher architectures.}
            Our framework applies unchanged across noise-prediction, $\dataSample$-prediction, and $v$-prediction parameterizations: each corresponds to a different per-timestep weight $\weight(\timevar)$, and the IS proxy $\proposalDensity\!\propto\!p\,\weight$ uses whichever weight the teacher exposes. Stratification and compute reuse are parameterization-agnostic. Our data-attribution experiments (\Sec~\ref{sec:experiments-motive}) already cover the flow-matching case (Wan2.1), where the velocity field carries an analogous time-dependent weight. Investigating how optimal $(\numRenders,\numReNoises)$ shifts across these parameterizations on a fixed task is a natural next step.
    
        \textbf{Biased variance reduction and trade-offs.}
            Our methods preserve unbiasedness. Many practical pipelines introduce bias: they clip gradients, truncate guidance, or clamp timesteps to improve stability or perceptual quality. We leave two questions open. How do stratification and importance sampling interact with these biased techniques? Can we trade a slight bias for further variance reduction? Control variates that use inexpensive auxiliary estimates could complement our approach and reduce variance further.
    
        \textbf{Non-frozen teachers and co-adaptation.}
            Our framework assumes a frozen teacher, but some pipelines fine-tune or distill the teacher alongside downstream optimization. In such settings, the optimal timestep distribution may shift as the teacher adapts, and variance-reduction strategies may need to account for the joint dynamics of the teacher and downstream parameters. Extending our methods to these coupled settings could improve both training efficiency and final performance.
    
        \textbf{Connecting estimator design to downstream performance.}
            While variance reduction improves gradient quality, the relationship between gradient variance and downstream metrics (e.g., CLIP score, FID, influence ranking correlation) is task-dependent and not fully understood. Developing theory or empirical principles that predict when variance reduction will translate into metric improvements and which variance sources matter most for a given task would help practitioners allocate compute more effectively and design better estimators.

    \subsection{Detailed Practitioner Guidance}\label{app:practitioner-guide}
        We provide a decision tree to help practitioners choose variance-reduction methods based on their application's computational structure and objectives. \Sec~\ref{sec:method-variance-framework} gives the procedure in brief; this section adds thresholds and the full decision tree.
        
        \textbf{Step 1: Assess cost structure}
        \begin{itemize}[leftmargin=*]
            \item Measure rendering/encoding cost $\costRender$ vs. denoising cost $\costDenoise$.
            \item As a rule of thumb, $\costRender > 10 \times \costDenoise$ motivates testing compute reuse; the cost ratio alone is not sufficient.
            \item If $\costRender \approx \costDenoise$: stratification may be more effective (data attribution).
            \item If $\costRender < \costDenoise$: focus on importance sampling and stratification with small $\numReNoises$.
            \item To set $\numReNoises$, estimate $\sigma_A^2$ and $\sigma_W^2$ of \Eq~\ref{eq:two-level} from a short pilot using Welford's algorithm with cached-render re-noising (\Sec~\ref{sec:method-variance-framework}, \App~\Sec~\ref{app:caching}), time $\costRender$ and $\costDenoise$, and use $\numReNoises^\star\approx\sqrt{(\sigma_W^2/\sigma_A^2)(\costRender/\costDenoise)}$ (\Sec~\ref{sec:method-compute-reuse}). For IID subunits, compare the feasible floor and ceiling (at least $1$) using \Eq~\ref{eq:budget-variance}; $\numReNoises=2$ beats $\numReNoises=1$ exactly when $\numReNoises^\star>\sqrt{2}$. With stratification, validate the final allocation by its measured ECM.
        \end{itemize}
        
        \textbf{Step 2: Identify gradient structure}
        \begin{itemize}[leftmargin=*]
            \item If a known timestep weight dominates how gradient norms vary with $\timevar$ (e.g., $\sdsWeight(\timevar)$ in SDS): use weight-based importance sampling with the likelihood-ratio correction.
            \item If gradient norms vary significantly across timesteps: use adaptive importance sampling.
            \item If gradient norms are approximately constant: skip importance sampling.
        \end{itemize}
        
        \textbf{Step 3: Check for auxiliary objectives}
        \begin{itemize}[leftmargin=*]
            \item If training uses strong auxiliary losses (DMD): variance reduction may not improve final metrics, but can stabilize training.
            \item If using only Monte Carlo gradients (SDS, data attribution): variance reduction will likely help.
        \end{itemize}
        
        \textbf{Step 4: Choose stratification design}
        \begin{itemize}[leftmargin=*]
            \item Per-render stratification: Use when rendering dominates, and you sample multiple timesteps per render (SDS default).
            \item Global stratification: Use when encoding is moderate relative to denoising (data attribution).
            \item Number of strata: $\numStrata = \numReNoises$ per render, or $\numStrata = \numRenders \times \numReNoises$ for global stratification; use equal-mass quantile strata for a non-uniform proposal.
        \end{itemize}
        
        \textbf{Step 5: Combine methods}
        \begin{itemize}[leftmargin=*]
            \item Importance + stratification: Use stratified inverse-CDF.
            \item Stratification + reuse: Compatible, combine for complementary benefits.
            \item All three: Best results in SDS experiments.
        \end{itemize}
        
        \textbf{Expected effective compute multipliers (envelope across our experiments):}
        \begin{itemize}[leftmargin=*]
            \item Importance sampling alone: $1.05$--$1.24\times$ (RE vs uniform; \Tab~\ref{tab:relative_improvement_by_m_param}), up to $1.40\times$ at $\cfgScale\!=\!25$ (\Tab~\ref{tab:iw_ablation})
            \item Stratification plus reuse: $\sim\!1.0$--$3.0\times$ ECM in SDS (\Tab~\ref{tab:absolute_ecm_by_m_param}); stratification's isolated gain is reported as matched-configuration RE in \Tab~\ref{tab:relative_improvement_by_m_param}
            \item Compute reuse alone (high $\costRender/\costDenoise$): $\sim\!1.6$--$2.6\times$ on SDS (\Tab~\ref{tab:absolute_ecm_by_m_param})
            \item Combined IW+Strat+Reuse: $\sim\!2$--$3.3\times$ (SDS averages over $\numRenders$, peaking at $\numReNoises\!=\!8$; \Tab~\ref{tab:absolute_ecm_by_m_param})
        \end{itemize}

\clearpage
\raggedbottom
\section{Glossary and Notation}\label{app_sec_notation}
    \begin{table}[H]
    \caption{Glossary and notation (Part I: Fundamentals)}
        \begin{center}
        \begin{tabular}{c l}
        \toprule
        \multicolumn{2}{c}{\textbf{Acronyms and abbreviations}} \\
        \midrule
        IID & Independent and identically distributed \\
        IS & Importance sampling \\
        MC & Monte Carlo \\
        SDS & Score Distillation Sampling \\
        DMD & Distribution Matching Distillation \\
        VSD & Variational Score Distillation \citep{wang2023prolificdreamer} \\
        CFG & Classifier-free guidance \\
        ECM & Effective compute multiplier (\Sec~\ref{sec:method-variance-framework}) \\
        RE & Relative efficiency (\Sec~\ref{sec:method-variance-framework}) \\
        HT & Horvitz-Thompson (estimator; \App~\Sec~\ref{app:optimal_pair_distributions}) \\
        FID & Fr\'echet inception distance \\
        NeRF & Neural radiance field \\
        VAE & Variational autoencoder \\
        DiT & Diffusion transformer \citep{peebles2023scalable} \\
        \midrule
        \multicolumn{2}{c}{\textbf{Core mathematical notation}} \\
        \midrule
        $\Eop$ & Expectation operator \\
        $\KL$ & Kullback-Leibler divergence \\
        $\Normal$ & Gaussian/normal distribution \\
        $\Uniform$ & Uniform distribution \\
        $\standardNormal$ & Standard normal $\Normal(\mathbf{0},\identity)$ \\
        $\identity$ & Identity matrix \\
        $\diffd$ & Differential (use as $\diffd x / \diffd y$) \\
        $\defeq$ & Defined to be equal to \\
        $\jacobian$ & Jacobian matrix \\
        \midrule
        \multicolumn{2}{c}{\textbf{Diffusion model time and schedules}} \\
        \midrule
        $\timevar$ & Continuous time / noise level in $[0,1]$ \\
        $\maxTimevar$ & Maximum time / total timesteps \\
        $\logsnr$ & Log signal-to-noise ratio \\
        $\logsnrSchedule$ & Schedule mapping $\timevar \to \logsnr$ \\
        $\logsnrMin,\logsnrMax$ & Schedule endpoints \\
        $\signalcoef,\noisecoeff$ & Signal and noise coefficients at time $\timevar$ \\
        \bottomrule
        \end{tabular}
        \end{center}
        \label{tab:TableOfNotationPart1}
    \end{table}
    
    \begin{table}[H]
    \caption{Glossary and notation (Part II: Data and Models)}
        \begin{center}
        \begin{tabular}{c l}
        \toprule
        \multicolumn{2}{c}{\textbf{Data and latent representations}} \\
        \midrule
        $\dataSample$ & Clean data/image; generic upstream draw in \Sec~\ref{sec:method-variance-reduction-strategies} \\
        $\encodedData$ & Latent-space representation (VAE-encoded; generator output in DMD) \\
        $\noisedData$ & Forward-noised latent: $\noisedData=\signalcoef\encodedData+\noisecoeff\noiseVec$ \\
        $\noiseVec$ & Gaussian noise input $\sim\standardNormal$ \\
        $\noisePred$ & Predicted noise from denoiser network \\
        $\residual$ & Denoising residual $\noisePred-\noiseVec$ \\
        $\encoder$ & Encoder to latent space (e.g., VAE encoder) \\
        \midrule
        \multicolumn{2}{c}{\textbf{Model components and parameters}} \\
        \midrule
        $\genParams$ & Generator or renderer parameters (e.g., NeRF weights) \\
        $\denParams$ & Denoiser or score network parameters \\
        $\generator$ & One-step generator mapping noise to samples \\
        $\render$ & Encoded rendering map $\encoder\circ\prerender$ \\
        $\prerender$ & Differentiable renderer before encoding (e.g., NeRF) \\
        $\textCond$ & Conditioning signal (e.g., text prompt) \\
        $\cameraSample$ & Camera or rendering condition sample \\
        \midrule
        \multicolumn{2}{c}{\textbf{Weighting and guidance}} \\
        \midrule
        $\weight$ & Timestep weighting function \\
        $\sdsWeight$ & SDS-specific timestep weight \\
        $\cfgScale$ & Guidance scale for CFG \\
        \bottomrule
        \end{tabular}
        \end{center}
        \label{tab:TableOfNotationPart2}
    \end{table}
    
    \begin{table}[H]
    \caption{Glossary and notation (Part III: Losses and Distributions)}
        \begin{center}
        \begin{tabular}{c l}
        \toprule
        \multicolumn{2}{c}{\textbf{Loss functions and objectives}} \\
        \midrule
        $\loss$ & Generic loss function \\
        $\cost$ & Per-sample cost function \\
        $\lossDiffusion$ & Diffusion training loss \\
        $\lossWeighted$ & Weighted diffusion objective \\
        $\costDiffusion$ & Per-sample diffusion cost \\
        $\lossSDS$ & SDS objective (parameter-space) \\
        $\lossDenoise$ & Auxiliary denoising loss for fake model \\
        $\lossReg$ & Regression loss \\

        $\update$ & Update vector (gradient-style) \\
        $\sdsupdate$ & SDS update direction \\
        $\sdsupdatehat$ & Stochastic estimator of $\sdsupdate$ \\
        \midrule
        \multicolumn{2}{c}{\textbf{Distributions and density models}} \\
        \midrule
        $\realDensity$ & Real data distribution \\
        $\fakeDensity$ & Generated/fake distribution \\
        $\proposalDensity$ & Importance sampling proposal distribution \\
        $\importanceWeight$ & Importance weight $p(\timevar)/\proposalDensity(\timevar)$ \\
        $\scoreReal,\scoreFake$ & Real and fake score functions \\
        $\mean$ & Target mean (the expectation being estimated) \\
        $\estimatedMean$ & Estimated mean (the realized estimator) \\
        $\hat\mean_{\mathrm{GT}}$ & High-sample ground-truth mean reference (\App~\Sec~\ref{sec:method-variance-framework-app}) \\
        $\boldsymbol{\mu}_{\mathbf{y}}$ & Pair-prob target sum $\sum_i \mathbf{y}_i$ (\App~\Sec~\ref{app:optimal_pair_distributions}) \\
        $\meanBase$ & Denoised mean from base diffusion model \\
        $\meanFake$ & Denoised mean from learned fake model \\
        $\labelBase,\labelFake,\labelReal$ & Labels \texttt{base}, \texttt{fake}, \texttt{real} \\
        \bottomrule
        \end{tabular}
        \end{center}
        \label{tab:TableOfNotationPart3}
    \end{table}
    
    \begin{table}[H]
    \caption{Glossary and notation (Part IV: Sampling and Variance Reduction)}
        \begin{center}
        \begin{tabular}{c l}
        \toprule
        \multicolumn{2}{c}{\textbf{Monte Carlo sampling}} \\
        \midrule
        $\numSamples$ & Number of samples (also $\numSamples_{\mathrm{GT}}$ for the ground-truth reference) \\
        $\sampleIndex$ & Sample index $\in\{1,\dots,\numSamples\}$ \\
        $\sampleStep,\numSampleStep$ & Sampler index and step count (diffusion sampling only) \\
        $\testFunc$ & Test function in expectations: $\testFunc(\timevar)\!=\!\E_{\randomness}[\gterm(\timevar,\randomness)\mid\timevar]$ \\
        $\randomness$ & Generic randomness variable (data, conditioning, noise) \\
        $\randomness^{(\renderIndex)}_{\stratumIndex}$ & Uniform $(0,1)$ jitter for stratum $\stratumIndex$ of render $\renderIndex$ \\
        \midrule
        \multicolumn{2}{c}{\textbf{Stratified sampling}} \\
        \midrule
        $\numStrata$ & Number of strata for stratified sampling \\
        $\stratum_{\stratumIndex}$ & The $\stratumIndex$-th stratum \\
        $\stratumIndex$ & Stratum index \\
        $\samplePerStratum$ & Number of samples in stratum $\stratumIndex$ \\
        $\sampleIndexInStratum$ & Index within a stratum \\
        $\quantile$ & Quantile level for inverse-CDF sampling \\
        \midrule
        \multicolumn{2}{c}{\textbf{Compute reuse and efficiency}} \\
        \midrule
        $\numRenders$ & Number of distinct renders per step \\
        $\numReNoises$ & Number of re-noisings per render (compute reuse) \\
        $\renderIndex$ & Render index \\
        $\renoiseIndex$ & Re-noise index \\
        $\costRender$ & Cost per primary unit (render or generator pass and encode; \Sec~\ref{sec:method-variance-reduction-strategies}) \\
        $\costDenoise$ & Cost per subunit (one noising and denoising; \Sec~\ref{sec:method-variance-reduction-strategies}) \\
        $\budget$ & Compute budget \\
        $\generalCost$ & Generic per-operation cost \\
        $\gterm$ & Per-$(\timevar,\randomness)$ contribution vector (e.g., $\sdsWeight\,\residual$ in latent space; \Sec~\ref{sec:method-variance-reduction-strategies}) \\
        $\mathbf{h}$ & Full gradient contribution, including the Jacobian \\
        $\hat{\mathbf{h}}^{(\renderIndex)}$ & Mean of $\numReNoises$ contributions for primary unit $\renderIndex$ \\
        $\sigma_A^2$ & Across-unit variance of conditional means (\Eq~\ref{eq:two-level}) \\
        $\sigma_W^2$ & Mean within-unit variance (\Eq~\ref{eq:two-level}) \\
        $\numReNoises^\star$ & Continuous cost-optimal allocation for IID subunits \\
        \bottomrule
        \end{tabular}
        \end{center}
        \label{tab:TableOfNotationPart4}
    \end{table}
    
    \begin{table}[H]
    \caption{Glossary and notation (Part V: Attribution and Auxiliary)}
        \begin{center}
        \begin{tabular}{c l}
        \toprule
        \multicolumn{2}{c}{\textbf{Data attribution}} \\
        \midrule
        $\influence$ & Influence score between training and query examples \\
        $\motiveGrad$ & Per-example gradient for attribution \\
        $\query$ & Query example label \\
        $\ntrain$ & Training example index \\
        $\sampleSet$ & Set of shared $(\timevar,\noiseVec)$ samples \\
        $\dataset$ & Training dataset \\
        $\hessian$ & Hessian matrix \\
        \midrule
        \multicolumn{2}{c}{\textbf{Auxiliary notation}} \\
        \midrule
        $\numChannels,\numSpatial$ & Number of channels and spatial locations \\
        \bottomrule
        \end{tabular}
        \end{center}
        \label{tab:TableOfNotationPart5}
    \end{table}

\ifarxiv\else
\clearpage
\fi

\end{document}